\documentclass[11pt]{article}

\usepackage[final]{acl}

\usepackage{times}
\usepackage{latexsym}

\usepackage[T1]{fontenc}

\usepackage[utf8]{inputenc}

\usepackage{inconsolata}

\usepackage{graphicx}

\usepackage{hyperref}
\usepackage{url}
\usepackage{xspace}
\usepackage{subcaption}
\usepackage{booktabs}
\usepackage{circledsteps}
\usepackage{xcolor}
\usepackage[table]{xcolor}
\usepackage{algorithm}
\usepackage{algpseudocode}
\usepackage{pifont} 
\usepackage{multirow}
\usepackage{fvextra}
\usepackage{amsmath}
\usepackage[normalem]{ulem}
\usepackage{tikz}

\newcommand{\darkcirc}[1]{%
\tikz[baseline=(char.base)]{
  \node[circle, fill=black, inner sep=1.2pt] (char)
  {\textcolor{white}{\footnotesize #1}};
}}

\newcommand{\review}[1]{\textcolor{black}{#1}}

\newcommand{\Clare}{\textsc{CLaRE}\xspace}
\title{\Clare-ty Amid Chaos: Quantifying Representational Entanglement to Predict Ripple Effects in LLM Editing}

\author{Manit Baser$^1$, Alperen Yildiz$^2$, Dinil Mon Divakaran$^{1,2}$, Mohan Gurusamy$^1$ \\
  $^1$National University of Singapore, Singapore \\
  $^2$A*STAR Institute for Infocomm Research (A*STAR I$^2$R), Singapore}

\begin{document}
\maketitle
\begin{abstract}
The static knowledge representations of large language models (LLMs) inevitably become outdated or incorrect over time. While model-editing techniques offer a promising solution by modifying a model's factual associations, they often produce unpredictable ripple effects, which are unintended behavioral changes that propagate even to the hidden space. In this work, we introduce \Clare, a lightweight representation-level technique to identify where these ripple effects may occur. Unlike prior gradient-based methods, \Clare quantifies entanglement between facts using forward activations from a single intermediate layer, avoiding costly backward passes. To enable systematic study, we prepare and analyse a corpus of 11,427 facts drawn from three existing datasets. Using \Clare, we compute large-scale entanglement graphs of this corpus for multiple models, capturing how local edits propagate through representational space. These graphs enable stronger preservation sets for model editing, audit trails, efficient red-teaming, and scalable post-edit evaluation. In comparison to baselines, \Clare achieves an average of 62.2\% improvement in Spearman correlation with ripple effects while being 2.74× faster, and using 2.85× less peak GPU memory. Besides, \Clare requires only a fraction of the storage needed by the baselines to compute and preserve fact representations. Our entanglement graphs and corpus are available at \href{https://github.com/manitbaser/CLaRE}{https://github.com/manitbaser/CLaRE}.

\end{abstract}

\begin{figure}
    \centering
    \includegraphics[width=\linewidth]{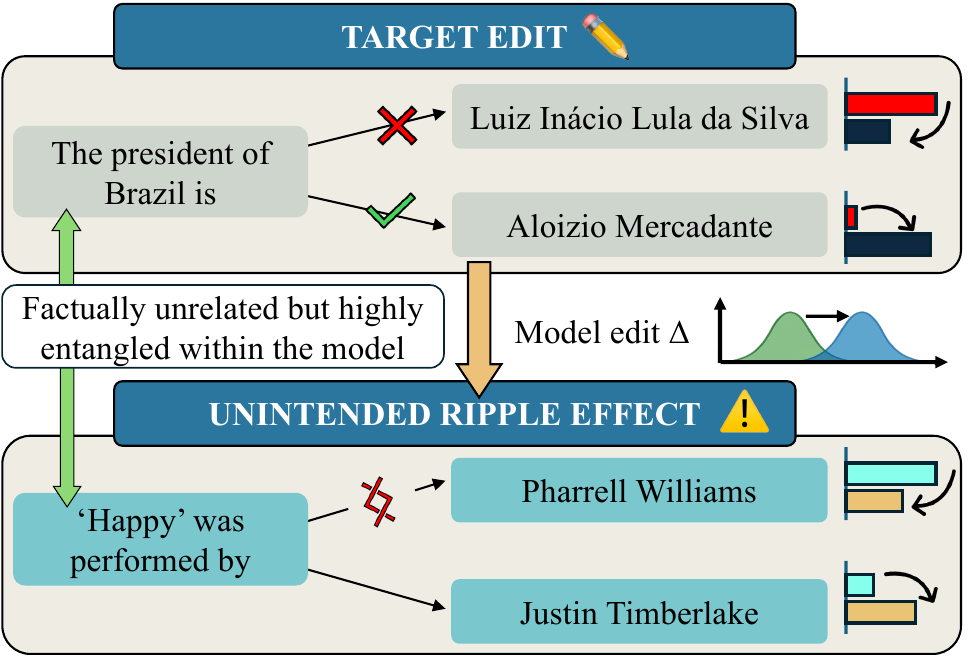}
    \caption{A targeted update to a political fact may inadvertently alter the model's prediction for an unrelated musical fact, despite no semantic connection. This demonstrates how edits can trigger ripple effects far beyond the intended factual neighborhood.}
    \label{fig:example}
\end{figure}

\section{Introduction}

Large language models (LLMs) are increasingly used in applications such as search~\cite{zhao2023survey}, dialogue systems~\cite{raza2025industrial}, education~\cite{yigci2024large}, and decision support~\cite{yang2023foundationmodels}. However, their encoded knowledge needs updating, as they may contain misinformation or outdated content~\cite{sallami2024deception, lin-etal-2025-investigating}. The fine-tuning process updates model parameters but scales poorly for large models~\cite{Fang2025_AlphaEdit}. Retrieval-augmented generation (RAG) supplements the LLM's input with external context, but may fail when it misinterprets the retrieved content or when the retrieved content itself is inaccurate~\cite{10852457}. As a complementary strategy, model editing updates specific factual associations in model weights, but may have its own trade-offs.

Parameter-modifying model-editing techniques~\cite{meng2022mass,meng2022locating} localize and alter weights for targeted factual associations. Although they offer significant improvements in computational efficiency compared to fine-tuning, they also lead to ripple effects. These are unintended behavioral changes at other, semantically related (or even unrelated) model outputs~\cite{cohen2024rippleedits}. While some ripple effects may enhance generalization, many introduce new hallucinations, distort factual consistency, or degrade model performance. Prior efforts such as GradSim~\cite{qin2024does} and the RippleEdits benchmark~\cite{cohen2024rippleedits} have attempted to quantify how these effects propagate to semantically neighboring facts, typically within one or two hops of the edited entity in a knowledge graph. They reveal that LLM representations are highly interconnected: altering one fact may influence others that share representational subspaces. However, ripple effects can also extend to hidden spaces, producing distortions even where no direct factual relationship exists~\cite{missingPiece2025hiddenDamage}, as illustrated in Figure \ref{fig:example}. These {\em inter-domain} ripple effects may degrade a model's reliability and deteriorate its performance. Furthermore, although GradSim estimates entanglement between two facts, we show in Section~\ref{experiments} that such gradient similarity poorly correlates with inter-domain ripple effects, and it is computationally inefficient as it calculates the entire gradient for each fact.

These cross-domain distortions due to model editing remain poorly understood and largely unmeasured in current evaluation frameworks. For comprehensive analysis, we curate 11,427 facts spanning diverse domains, drawn from 3 existing datasets~\cite{zhong2023mquake, cohen2024rippleedits, liu-etal-2025-know}. This corpus, containing 212 unique prompt formats and 6,140 unique subjects, enables analysis of how model edits propagate globally through a model's knowledge base and how these dynamics vary across model architectures~(Section~\ref{experiments}). We introduce \textbf{\Clare} (Critical Layer Representation Entanglement), a lightweight representation-level technique for identifying where ripple effects are most likely to occur~(Section~\ref{Clare}). Unlike computationally heavy gradient-based approaches like GradSim, \Clare quantifies entanglement directly from forward activations from a single intermediate layer, without requiring backward passes or gradient computation. Our experiments show that in comparison to GradSim, \Clare achieves an average of 62.2\% improvement in Spearman correlation with observed ripple effects across multiple editing techniques and models~(Section~\ref{rq1}). Beyond predictive accuracy, \Clare is 2.74× faster and uses 2.85× lower peak GPU memory, while requiring only a fraction of storage needed to preserve fact representations~(Section~\ref{rq2}).



We release entanglement graphs computed with \Clare over the corpus of 11,427 facts across multiple models, providing a basis for safer, more interpretable, and cost-effective model editing with practical utility for both research and deployment. These graphs enable safer editing by enabling stronger preservation-set construction and facilitate audit trails and post-edit explainability. They further support downstream applications such as budget-constrained red-teaming of editing techniques by identifying high-risk regions within the model~(Section~\ref{rq3}). Together, these capabilities help in assessing ripple effects and advance reliable, auditable and interpretable model editing.


\textbf{Contributions.} \darkcirc{1} We introduce \Clare, a lightweight and scalable technique for identifying where ripple effects are most likely to occur. \Clare offers 2.74× speedup, and 2.85× lower peak GPU memory usage as compared to gradient-based methods.
\darkcirc{2} We curate 11,427 facts across diverse domains, to systematically study how edits propagate globally across a model's knowledge base. \Clare achieves up to 0.31 higher Spearman correlation with observed ripple effects, averaging a 62.2\% improvement over gradient-based methods in predictive alignment.
\darkcirc{3} We release large-scale entanglement graphs for these 11,427 facts computed with \Clare for multiple models, enabling construction of stronger preservation sets for model editing, and support cost-effective red-teaming and post-edit evaluation at scale.

\section{Related Work}

\textbf{Model-editing techniques.}
Parameter-modifying and parameter-preserving approaches are two primary paradigms for model editing in LLMs. Parameter-modifying techniques directly alter model weights through targeted optimization. Approaches like MEND use gradient decomposition for fast, local edits~\cite{mitchell2021fast}. Recently, locate-then-edit techniques have emerged which target specific knowledge, yet they are also subject to ripple effects~\cite{meng2022mass, meng2022locating, Fang2025_AlphaEdit, gu2401model, ma2024perturbation}. Parameter-preserving techniques augment models with external modules instead of modifying weights~\cite{mitchell2022memory, zheng2023can, huang2023transformer, hartvigsen2023aging, chen-etal-2024-lifelong}. Emerging techniques integrate knowledge graphs and chain-of-thought reasoning with parameter modification~\cite{zhang2024knowledge, dong2025chainedit, zhao2024ripplecot}. We focus on parameter-modifying techniques, as they update the model parameters.

\textbf{Model-editing evaluation techniques and benchmarks.} CounterFact~\cite{meng2022locating} and CounterFactPlus~\cite{hoelscher2023detecting} benchmark specificity through counterfactual edits. MQuAKE~\cite{zhong2023mquake} and ThinkEval~\cite{baserthinkeval} evaluate multi-hop and multi-step reasoning, while KnowEdit~\cite{zhang2024comprehensive} provides a unified framework across six datasets. LEME~\cite{rosati2024long} and Evedit~\cite{liu-etal-2024-evedit} target long-form and event-based editing. UnKEBench~\cite{deng2024everything} tests editing over noisy, unstructured data, and ZsRE~\cite{levy-etal-2017-zero} focuses on QA-based factual samples. Know-MRI~\cite{liu-etal-2025-know} offers a toolkit integrating 13 datasets to analyse LLM knowledge mechanisms. RippleEdits~\cite{cohen2024rippleedits} is a benchmark of factual edits meant to capture ripple effects caused by editing techniques. While RippleEdits focuses on semantically or graph-neighboring facts, our goal is to detect ripple effects in the hidden space, including across unrelated or cross-domain facts. Hence, we curate a corpus of 11,427 facts from MQuAKE, RippleEdits and Know-MRI, enabling broader domain coverage and fine-grained analysis.

\textbf{Ripple Effects and Entanglement.} DiKE~\cite{zhang2025disentangling} is an editing technique which uses a disentanglement module to preserve other facts about the same entity, but does not consider damage brought to the hidden space. SIR~\cite{missingPiece2025hiddenDamage} is an editing technique which addresses ripple effects in hidden space, but it only detects them after editing. It requires evaluation after editing followed by re-editing to repair the damage, making it reactive rather than preventive, and unsuitable for mapping entanglement pre-edit or at scale. GradSim~\cite{qin2024does} uses gradient similarity to estimate entanglement, but computing full gradients for each fact remains inefficient for large corpora. These limitations motivate the need for alternatives like \Clare, which directly estimates entanglement using forward activations from a single intermediate layer. Through our experiments, we show that \Clare achieves higher predictive alignment with observed ripple effects while scaling efficiently to thousands of facts, enabling corpus-wide entanglement mapping.


\begin{figure*}
\centering

\begin{subfigure}{\linewidth}
    \centering
    \includegraphics[width=\linewidth]{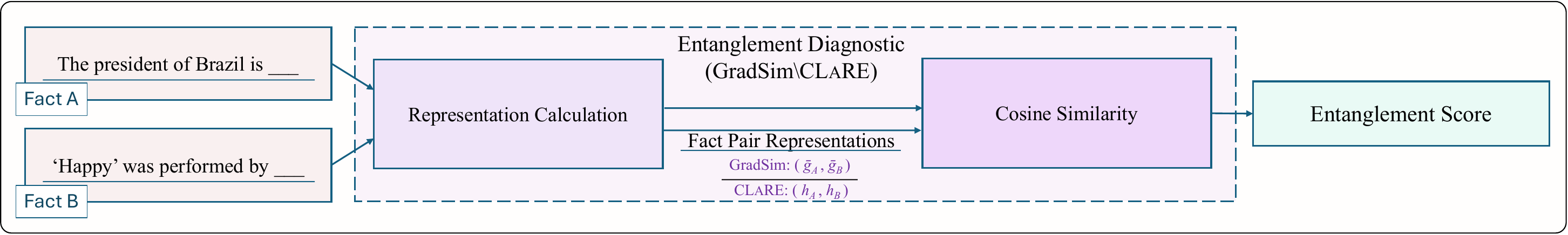}
    \caption{\review{Entanglement computation pipeline. Representations of fact pairs are calculated and processed using cosine similarity to produce the entanglement score.}}
    \label{fig:pipeline}
\end{subfigure}

\vspace{6pt}

\begin{subfigure}{\linewidth}
    \centering
    \includegraphics[width=\linewidth]{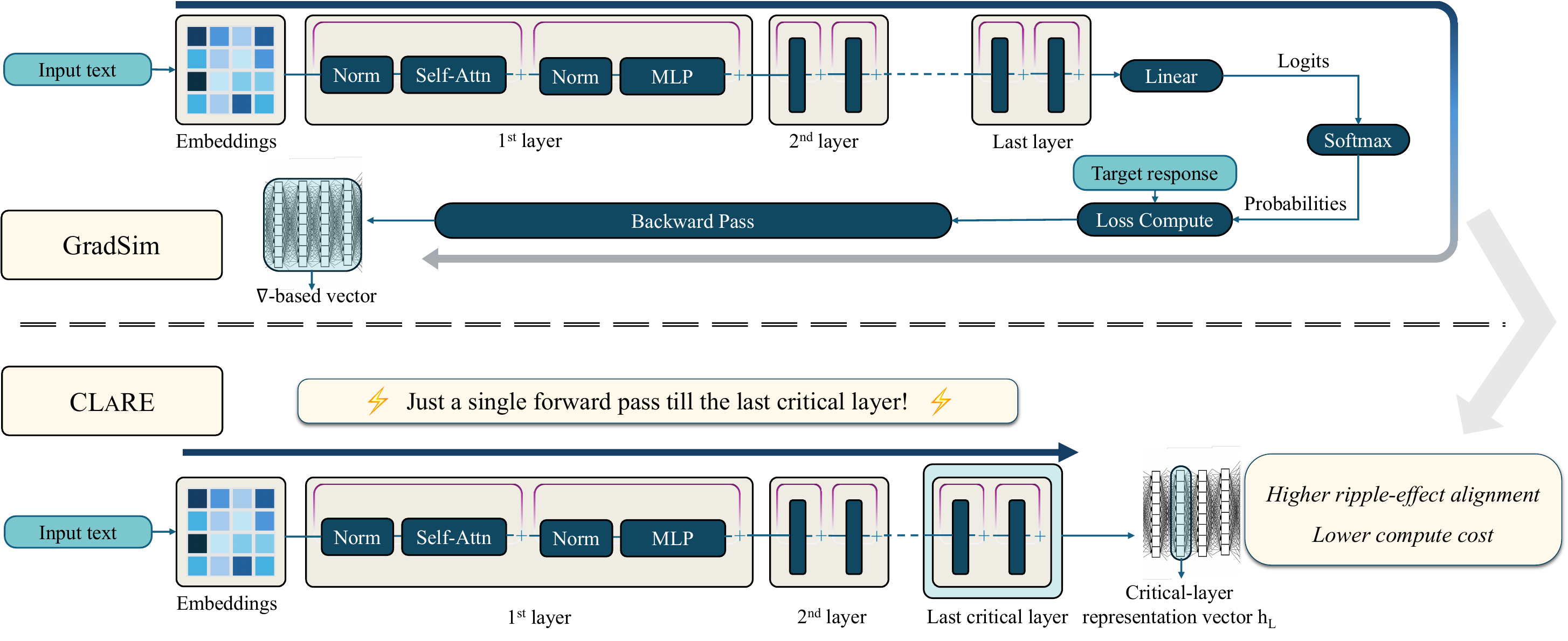}
    \caption{\review{Comparison between GradSim and \Clare.}}
    \label{fig:arch}
\end{subfigure}

\caption{\review{For each fact, GradSim computes the entire gradient, while \Clare uses a single forward pass up till the last critical layer, enabling faster and scalable entanglement mapping.}}
\label{fig:clare_pipeline}
\end{figure*}

\section{Preliminary}

\review{For simplicity, Eq.~\ref{eq:residual} follows the parallel attention-MLP formulation.}
As an input token $x_{[t]}$ moves through hidden layer $l$ of a transformer, its representation is updated by following components:

\begin{align}
h^{l}_{[t]}(x) &= h^{l-1}_{[t]}(x) + a^{l}_{[t]}(x) + m^{l}_{[t]}(x),
\label{eq:residual}
\end{align}
where $h^{0}_{[t]}(x)$ denotes the input embedding, $a^{l}_{[t]}$ is the attention output, and $m^{l}_{[t]}$ is the MLP (feedforward) contribution:
\begin{align}
a^{l}_{[t]} &= \mathrm{attn}\big(h^{l-1}_{[1]},\dots,h^{l-1}_{[t]}\big),\\
m^{l}_{[t]} &= W^{l}_{\mathrm{out}}\,\sigma\!\big(W^{l}_{\mathrm{in}}\,\gamma(h^{l-1}_{[t]})\big).
\end{align}

Here, $\gamma$ denotes layer normalization and $\sigma$ a nonlinear activation. The residual stream $h^{l}_{[t]}$ aggregates attention and nonlinear feedforward signals, and this layer-wise buildup enables the LLM to compose and propagate factual associations to its final prediction. Prior works find that factual associations are often localized in a small subset of intermediate MLP layers, termed the critical MLP layers $\mathcal{R}$~\cite{meng2022locating,meng2022mass}. Intuitively, these layers perform an associative ``key$\to$value'' computation: subject tokens elicit key-like activation patterns, and MLPs generate values that later shape output logits. Interventions at these layers may yield the largest changes in answers, identifying them as major causal sites for storing facts.

A standard procedure to identify critical layers is causal tracing~\cite{meng2022locating,meng2022mass}. Causal tracing holds the model computation fixed up to a candidate layer and then swaps or ablates subsequent activations from a reference input, measuring the downstream effect on the model's prediction. Layers whose interventions produce the largest change receive the highest causal scores and are labeled as the critical mediators for that fact. This per-layer causal ranking provides a principled basis to select layers for targeted editing. These mechanistic insights have motivated a family of parameter-modifying editing techniques, such as ROME~\cite{meng2022locating}, MEMIT~\cite{meng2022mass} and EMMET~\cite{gupta2024unifiedModelEditing}. \cite{guptalinearly} characterize approximate linear associative structure in factual storage, \cite{yao2024knowledgecircuits} describe distributed ``knowledge circuits'' that implement higher-level computation , and \cite{guerin2025qualifying} analyse neuron-level distinctions between language-structural and semantic storage.

\subsection{Ripple Effects in Model Editing}

\review{Parameter updates due to editing may unintentionally affect a model's predictions on other facts. These unintended changes are referred to as ripple effects \cite{cohen2024rippleedits}. Let $F$ denote the original set of facts represented by the model and $\Delta F$ denote the intended edits. After editing, the model's knowledge becomes}

\review{\begin{equation}
F' = F + \Delta F + R(\Delta F),
\end{equation}}

\review{where $R(\Delta F)$ denotes the ripple effects introduced by the editing operation. \cite{missingPiece2025hiddenDamage} distinguishes ripple effects in primarily two types:}

\review{\textbf{Ripple Effect in the Same Entity ($R_E$).}
Edits may influence other facts associated with the same entity as the edited fact. For example, modifying the birthplace of an individual may unintentionally affect other attributes related to that entity.}

\review{\textbf{Ripple Effect in Hidden Space ($R_H$).}
Edits may also propagate to seemingly unrelated facts due to proximity in the model's latent representation space. These effects arise from representational entanglement between facts and are often not captured by entity-based locality evaluations.}

\section{\Clare: A lightweight and scalable technique for identifying ripple effects}
\label{Clare}
We represent facts as triplets (subject $s$, relation $r$, object $o$), and the input prompt as $p_{i} = p(s_{i}, r_{i})$. Let the memory to be updated in the model be $(s_{i}, r_{i}, o_{i})$. To simplify notation, following Eq.~\ref{eq:residual}, let $h_{i}^{l} = h_{[S]}^{l}(p_{i})$, where $S$ is the last subject token of $p_{i}$. Let $\mathcal{L} = \max(\mathcal{R})$ be the last critical layer of the model. If a fact's causal influence is concentrated in a contiguous band of layers $\mathcal{R}$, then $\mathcal{L}$ tends to be the deepest point where the fact is still clearly represented before downstream mixing by later attention and MLPs. Using this layer as a probe gives a representation that contains the associative signal produced by earlier critical layers. As later layers can combine or diffuse information, a representation taken at $\mathcal{L}$ is more stable for cross-fact comparisons than either earlier activations or the final output logits that also encode decoding constraints. Thus, for a fact $(s_{i}, r_{i}, o_{i})$:

\begin{equation}\label{eq:6}
h_{i}^{\mathcal{L}} = h_{i}^{0} + \sum_{l=1}^{\mathcal{L}} a_{i}^{l} + \sum_{l=1}^{\mathcal{L}} m_{i}^{l}.
\end{equation}

$h_{i}^{\mathcal{L}}$ is compact and informative, and relatively local snapshot of where the model ``stores'' the fact as compared to entire model gradients. To be specific, we do not perform weight edits. We use $h_i^\mathcal{L}$ purely as a probe. We quantify entanglement, i.e., whether two facts $(s_{i}, r_{i}, o_{i})$ and $(s_{j}, r_{j}, o_{j})$, overlap in the way model encodes them prior to any editing. This overlap serves as a proxy to identify which facts may get impacted by ripple effects.

\begin{multline}
\mathrm{\Clare}(i, j)
= \cos\!\big(h_{\mathrm{i}}^{\mathcal{L}},\, h_{j}^{\mathcal{L}}\big) \\
= \frac{\langle h_{\mathrm{i}}^{\mathcal{L}}, h_{j}^{\mathcal{L}}\rangle + \varepsilon}
         {\|h_{\mathrm{i}}^{\mathcal{L}}\|\,\|h_{j}^{\mathcal{L}}\| + \varepsilon}
\label{eq:entanglement}
\end{multline}


A high \Clare entanglement score indicates that the model stores both facts in similar subspaces, increasing the likelihood that an edit to one will influence the other. Each fact requires only a single forward pass up to $\mathcal{L}$ (and not the last layer) to obtain its hidden representation at $\mathcal{L}$. As illustrated in Figure~\ref{fig:arch}, \Clare requires no gradient computations or backward passes, making it faster and lightweight than existing approaches, enabling large-scale entanglement mapping.

While both \Clare and GradSim have a time complexity of \( \mathcal{O}(L \cdot d^2) \), where \( d \) is the hidden dimension size and \( L \) is the number of MLP layers in an LLM, \Clare is substantially faster in practice as it avoids loss computation and backward passes. Moreover, GradSim requires \( \mathcal{O}(L \cdot d^2) \) storage to retain full gradients for a fact, whereas \Clare only needs \( \mathcal{O}(d) \), which is to store a single hidden state vector at layer \( \mathcal{L} \). These compact representations allow \Clare to scale to thousands of facts efficiently. A detailed computational and storage complexity analysis is provided in Appendix~\ref{appendix:complexity}, and runtime cost analysis in Appendix~\ref{appendix:cost_breakdown}.

\begin{figure}[t]
    \centering
    \begin{subfigure}[b]{0.7\columnwidth}
        \includegraphics[width=\textwidth]{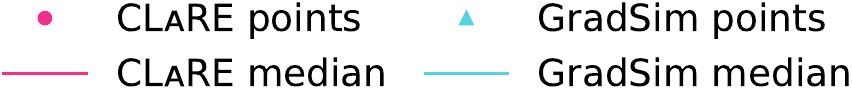}
    \end{subfigure}
    \par\medskip
    GPT2-XL \\[2pt]
    \begin{subfigure}[b]{0.49\columnwidth}
        \includegraphics[width=\textwidth]{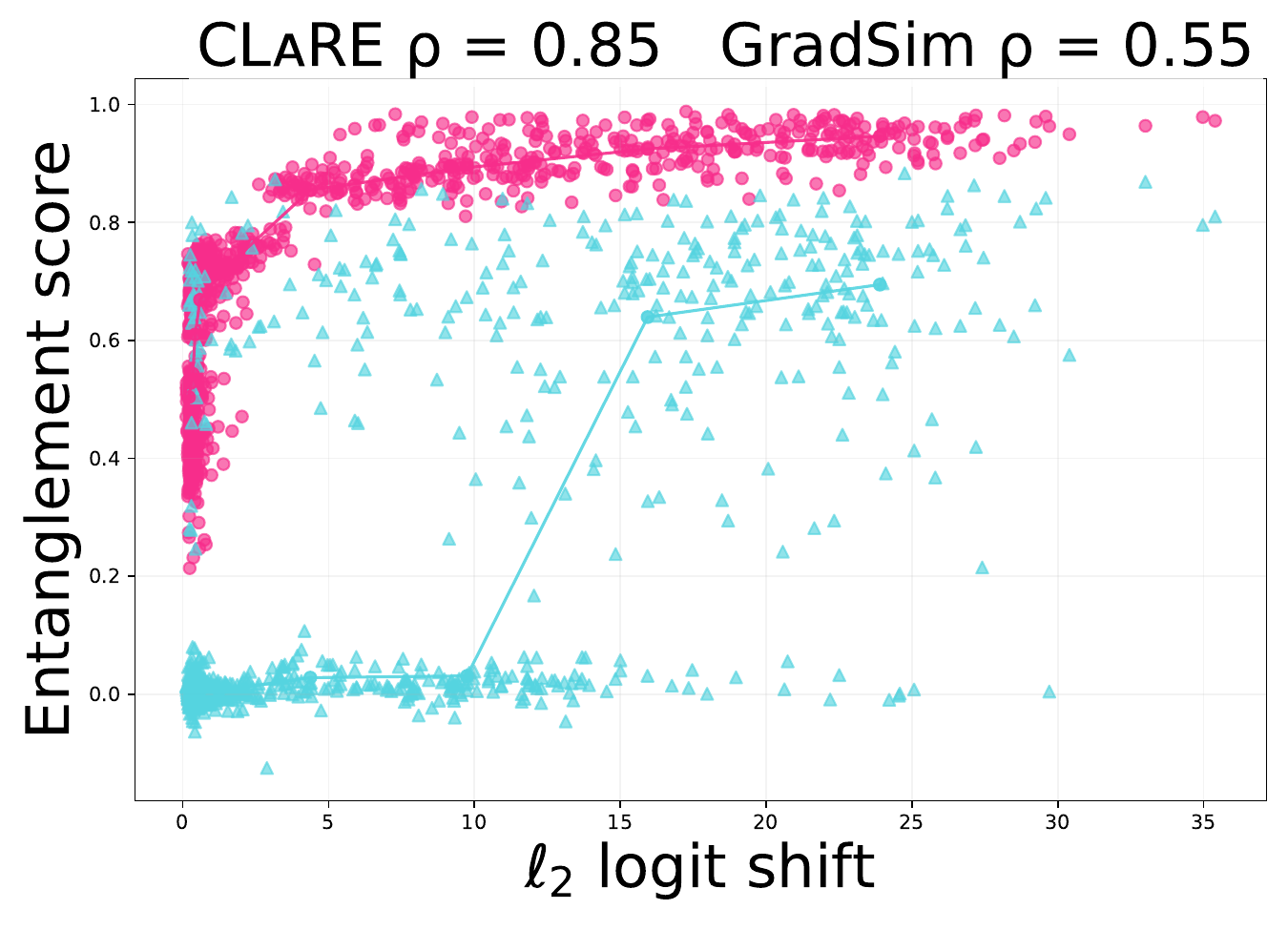}
    \end{subfigure}
    \begin{subfigure}[b]{0.49\columnwidth}
        \includegraphics[width=\textwidth]{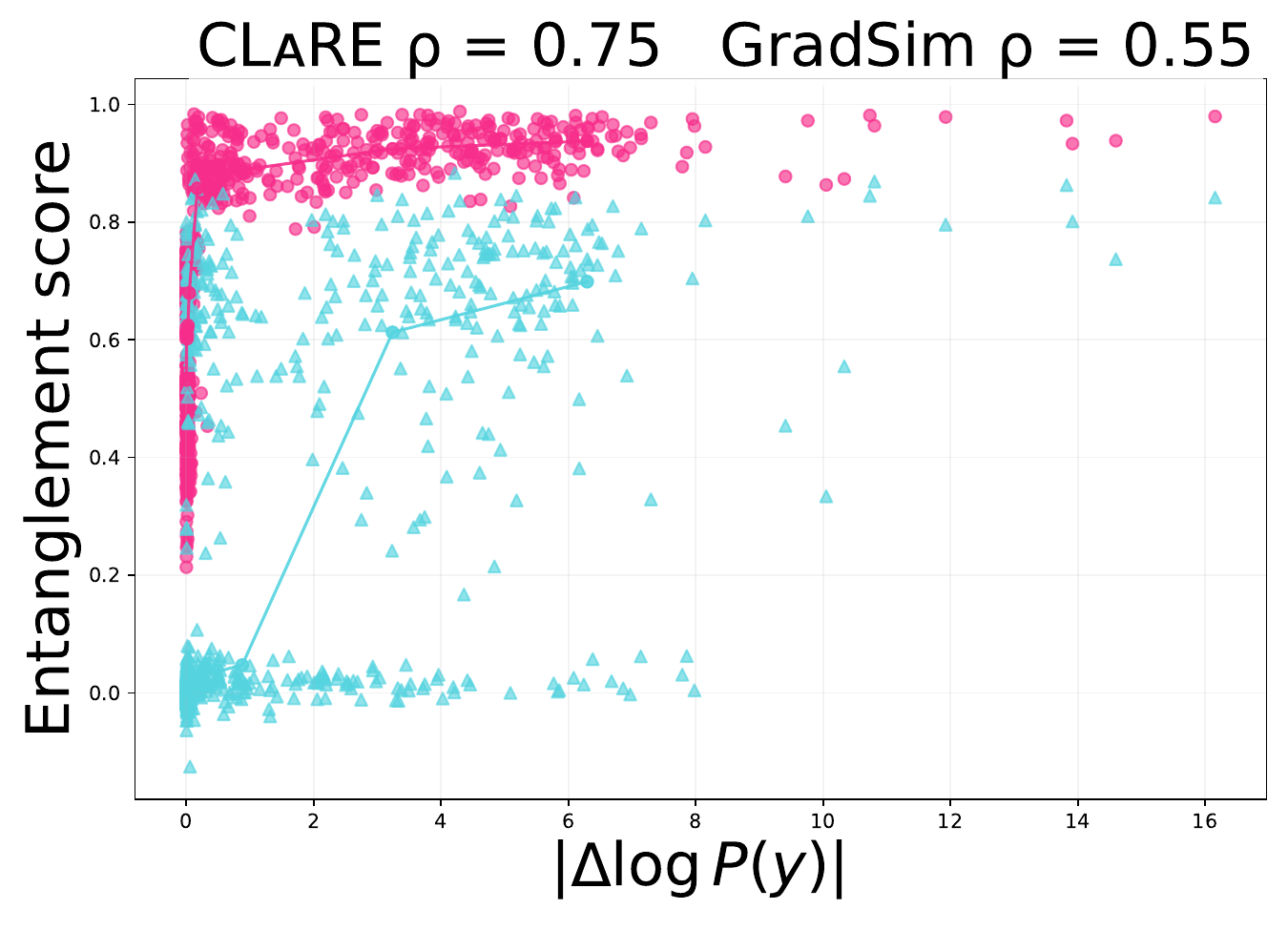}
    \end{subfigure}

    Llama3 \\[2pt]
    \begin{subfigure}[b]{0.49\columnwidth}
        \includegraphics[width=\textwidth]{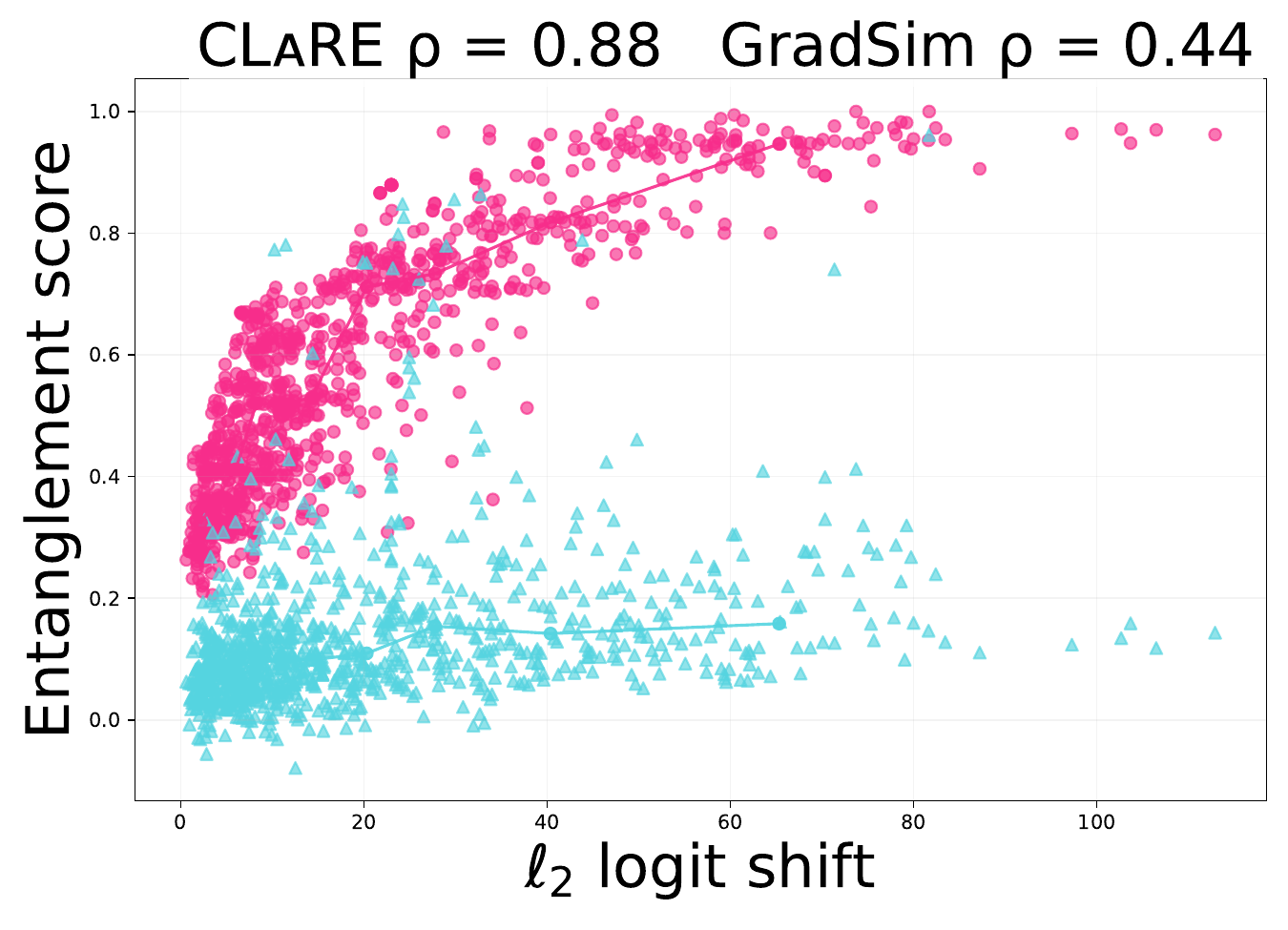}
    \end{subfigure}
    \begin{subfigure}[b]{0.49\columnwidth}
        \includegraphics[width=\textwidth]{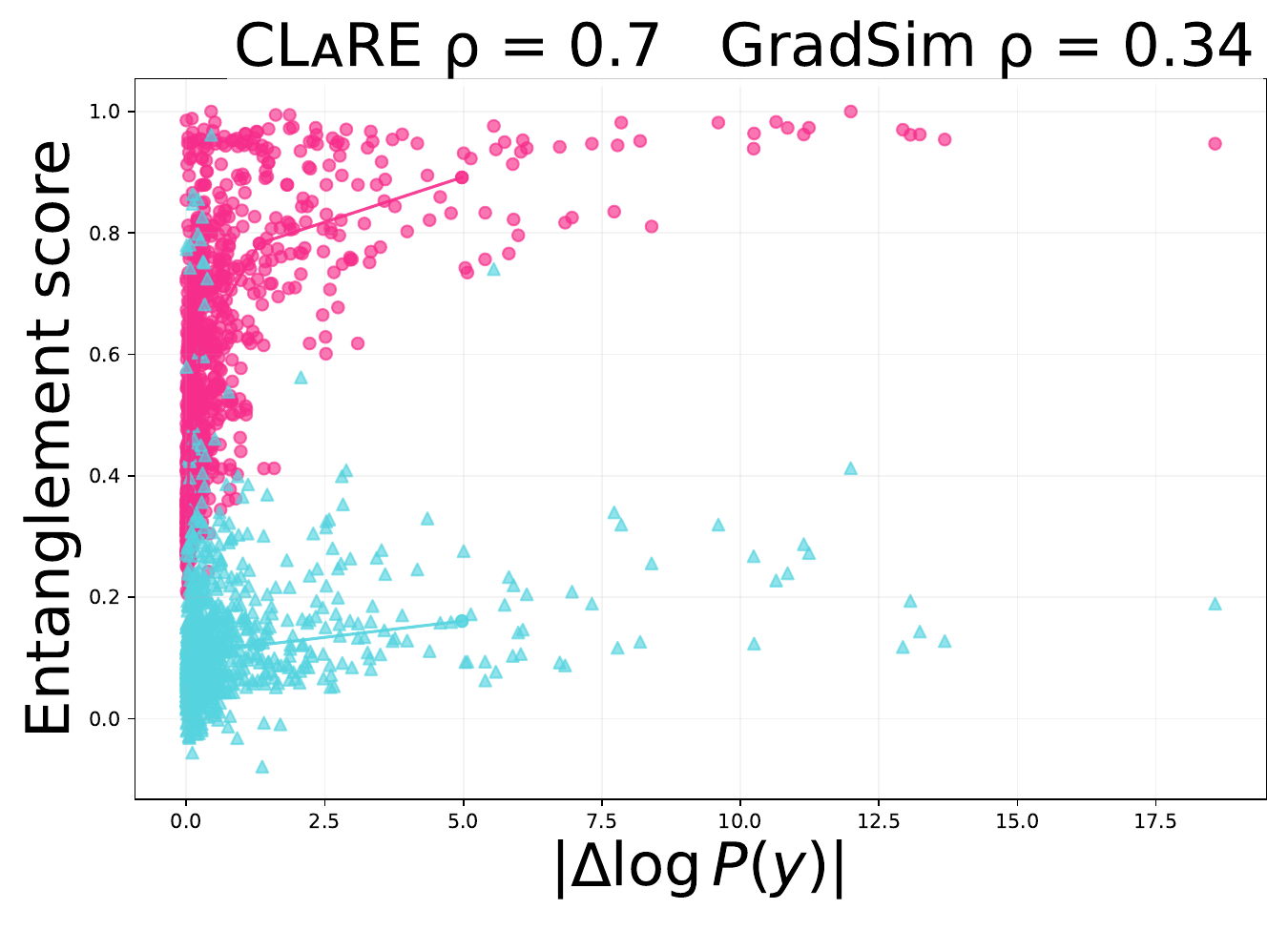}
    \end{subfigure}

    GPT-J \\[2pt]
    \begin{subfigure}[b]{0.49\columnwidth}
        \includegraphics[width=\textwidth]{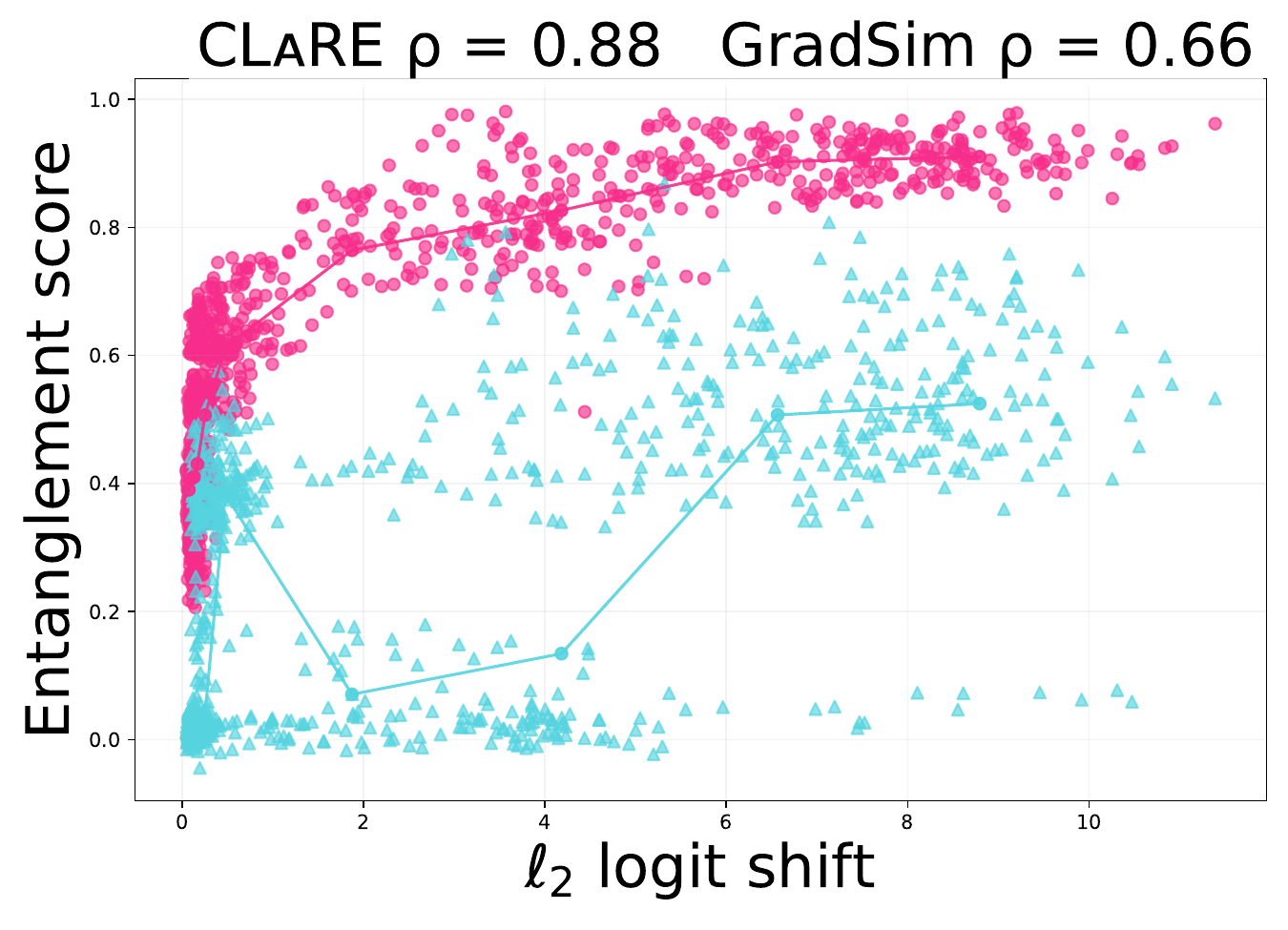}
    \end{subfigure}
    \begin{subfigure}[b]{0.49\columnwidth}
        \includegraphics[width=\textwidth]{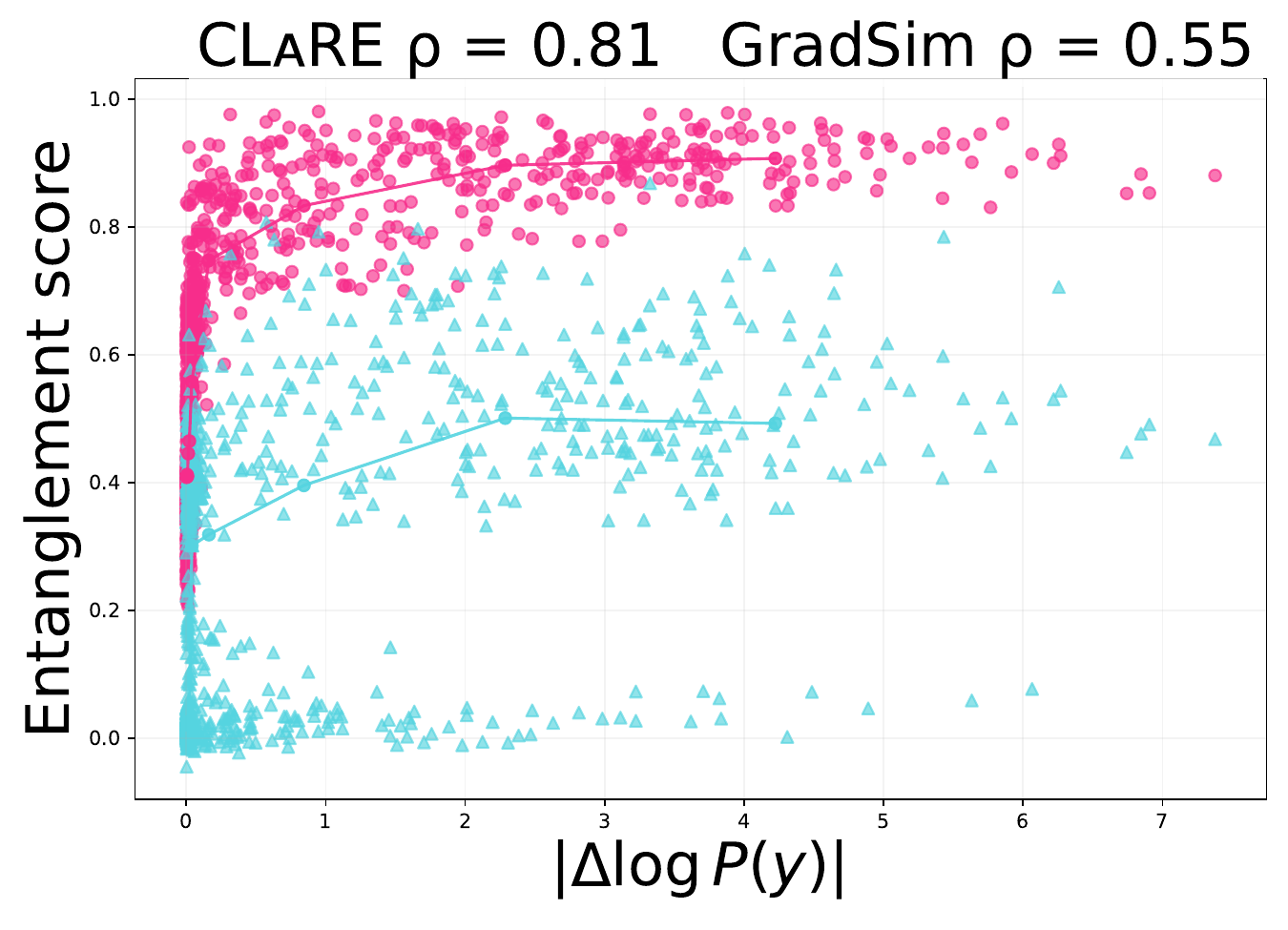}
    \end{subfigure}

    \caption{Correlation patterns for AlphaEdit: entanglement vs.\ $\ell_2$ logit shift (left) and $|\Delta \log P(y)|$ (right).}
    \label{fig:alphaedit_combined}
\end{figure}

\section{Experiments}
\label{experiments}

Our curated corpus of 11,427 facts spans 212 unique prompt formats and 6,140 unique subjects, enabling a systematic analysis of how ripple effects propagate globally across a model's knowledge base and how these dynamics vary across model architectures. All of our experiments are conducted on an NVIDIA H200 141GB GPU. We adopt the hyperparameter configurations outlined in the EasyEdit framework~\cite{wang-etal-2024-easyedit}. We address three research questions here, and discuss more research questions in Appendix~\ref{appendix:more_rq}.

\textbf{\underline{RQ \ding{172}}}: How is \Clare's predictive accuracy and computational efficiency compared to gradient-based methods like GradSim?

\textbf{\underline{RQ \ding{173}}}: How does the choice of layer affect \Clare's entanglement estimation quality?

\textbf{\underline{RQ \ding{174}}}: How effectively can \Clare compute large-scale entanglement graphs to enable downstream tasks like budget-constrained red-teaming and building model-editing preservation sets?

\begin{table*} \centering 
\caption{Spearman Correlation Scores for \Clare and GradSim across $|\Delta \log P(y)|$ and $\ell_2$ logit shift.} 
\label{tab:models-clare-gradsim-combined} 
\resizebox{0.99\textwidth}{!}{ 
\renewcommand{\arraystretch}{1.5} 
\begin{tabular}{c|cc|ccccc} \toprule 
\multirow{2}{*}{\rotatebox{90}{\textbf{Model}}} & \multirow{2}{*}{\textbf{\shortstack{Metric\\($\rho_{s}$)}}} & \multirow{2}{*}{\textbf{\shortstack{Method}}} & \multicolumn{5}{c}{\textbf{Editing Techniques}} \\ \cmidrule(lr){4-8} 
& & & \textbf{MEMIT} & \textbf{RECT} & \textbf{ROME} & \textbf{PRUNE} & \textbf{AlphaEdit} \\ \midrule 
\multirow{4}{*}{\rotatebox{90}{GPT-J}} & \multirow{2}{*}{$|\Delta \log P(y)|$} & GradSim & 0.582 & 0.549 & 0.509 & 0.597 & 0.547 \\ 
& & \cellcolor{gray!15}\textbf{\Clare} & \cellcolor{gray!15}\textbf{0.838} {\color{blue!70!black}(+44.0\%)} & \cellcolor{gray!15}\textbf{0.805} {\color{blue!70!black}(+46.6\%)} & \cellcolor{gray!15}\textbf{0.797} {\color{blue!70!black}(+56.6\%)} & \cellcolor{gray!15}\textbf{0.854} {\color{blue!70!black}(+43.0\%)} & \cellcolor{gray!15}\textbf{0.805} {\color{blue!70!black}(+47.2\%)} \\ \cmidrule(lr){2-8} 
& \multirow{2}{*}{$\ell_2$ logit shift} & GradSim & 0.677 & 0.646 & 0.634 & 0.691 & 0.663 \\ 
& & \cellcolor{gray!15}\textbf{\Clare} & \cellcolor{gray!15}\textbf{0.913} {\color{blue!70!black}(+34.9\%)} & \cellcolor{gray!15}\textbf{0.878} {\color{blue!70!black}(+35.9\%)} & \cellcolor{gray!15}\textbf{0.857} {\color{blue!70!black}(+35.2\%)} & \cellcolor{gray!15}\textbf{0.917} {\color{blue!70!black}(+32.7\%)} & \cellcolor{gray!15}\textbf{0.875} {\color{blue!70!black}(+32.0\%)} \\ \midrule 
\multirow{4}{*}{\rotatebox{90}{GPT2-XL}} & \multirow{2}{*}{$|\Delta \log P(y)|$} & GradSim & 0.545 & 0.531 & 0.544 & 0.558 & 0.550 \\ 
& & \cellcolor{gray!15}\textbf{\Clare} & \cellcolor{gray!15}\textbf{0.770} {\color{blue!70!black}(+41.3\%)} & \cellcolor{gray!15}\textbf{0.774} {\color{blue!70!black}(+45.8\%)} & \cellcolor{gray!15}\textbf{0.785} {\color{blue!70!black}(+44.3\%)} & \cellcolor{gray!15}\textbf{0.822} {\color{blue!70!black}(+47.3\%)} & \cellcolor{gray!15}\textbf{0.753} {\color{blue!70!black}(+36.9\%)} \\ \cmidrule(lr){2-8} 
& \multirow{2}{*}{$\ell_2$ logit shift} & GradSim & 0.543 & 0.548 & 0.558 & 0.561 & 0.554 \\ 
& & \cellcolor{gray!15}\textbf{\Clare} & \cellcolor{gray!15}\textbf{0.915} {\color{blue!70!black}(+68.5\%)} & \cellcolor{gray!15}\textbf{0.911} {\color{blue!70!black}(+66.2\%)} & \cellcolor{gray!15}\textbf{0.909} {\color{blue!70!black}(+62.9\%)} & \cellcolor{gray!15}\textbf{0.918} {\color{blue!70!black}(+63.6\%)} & \cellcolor{gray!15}\textbf{0.854} {\color{blue!70!black}(+54.2\%)} \\ \midrule 
\multirow{4}{*}{\rotatebox{90}{Llama3}} & \multirow{2}{*}{$|\Delta \log P(y)|$} & GradSim & 0.471 & 0.432 & 0.308 & 0.428 & 0.339 \\ 
& & \cellcolor{gray!15}\textbf{\Clare} & \cellcolor{gray!15}\textbf{0.772} {\color{blue!70!black}(+63.9\%)} & \cellcolor{gray!15}\textbf{0.747} {\color{blue!70!black}(+72.9\%)} & \cellcolor{gray!15}\textbf{0.706} {\color{blue!70!black}(+129.2\%)} & \cellcolor{gray!15}\textbf{0.798} {\color{blue!70!black}(+86.4\%)} & \cellcolor{gray!15}\textbf{0.697} {\color{blue!70!black}(+105.6\%)} \\ \cmidrule(lr){2-8} 
& \multirow{2}{*}{$\ell_2$ logit shift} & GradSim & 0.442 & 0.468 & 0.389 & 0.460 & 0.442 \\ 
& & \cellcolor{gray!15}\textbf{\Clare} & \cellcolor{gray!15}\textbf{0.848} {\color{blue!70!black}(+91.9\%)} & \cellcolor{gray!15}\textbf{0.861} {\color{blue!70!black}(+84.0\%)} & \cellcolor{gray!15}\textbf{0.801} {\color{blue!70!black}(+105.9\%)} & \cellcolor{gray!15}\textbf{0.871} {\color{blue!70!black}(+89.3\%)} & \cellcolor{gray!15}\textbf{0.876} {\color{blue!70!black}(+98.2\%)} \\ \bottomrule 
\end{tabular} 
} 
\end{table*}




\subsection{Performance on entanglement estimation and computational efficiency (RQ \ding{172})}
\label{rq1}
To quantify the magnitude of ripple effects induced by a model edit, we measure how predictions on control facts—facts not directly edited—change before and after editing. Let $\theta$ and $\theta'$ be the parameters of the original and edited models, respectively. For control fact $y$ with query $q_y$ and original answer $a_y$, we define two ripple-effect magnitudes that capture both geometric and probabilistic shifts in model behavior:

\paragraph{(1) $\ell_2$ logit shift} captures the overall geometric shift in the LLM's output distribution for $y$:
\begin{equation}
\mathcal{L}_2(y) = 
\big\| f_{\theta}(q_y) - f_{\theta'}(q_y) \big\|_2
\end{equation}
where $f_{\theta}(q_y)$ and $f_{\theta'}(q_y)$ are the output logit vectors of the original and edited models for the input prompt $q_y$. A larger $\mathcal{L}_2(y)$ indicates stronger ripple effect on the model's internal representations.

\paragraph{(2) Original answer log-probability shift} measures the probabilistic change of the answer $a_y$:
\begin{multline}
|\Delta \log P(y)| = \\
|\log P_{\theta'}(a_y \mid q_y) - 
\log P_{\theta}(a_y \mid q_y)|
\end{multline}
A larger $|\Delta \log P(y)|$ signifies that the LLM's belief about the control fact is altered (similar to Absolute Likelihood Gain in~\cite{qin2024messyRippleEffects}).

For each model, we sampled 1,000 edit–control pairs from our corpus, ensuring coverage across a broad spectrum of $\ell_2$ logit shift magnitudes resulting from a single trial when edited by AlphaEdit~\cite{Fang2025_AlphaEdit}. Each edit–control pair consists of an edit fact and a control fact. For each LLM and editing technique, we compute \Clare~(Eq.~\ref{eq:entanglement}) and GradSim entanglement scores for control facts with their corresponding edit facts, and compare them against both the ripple-effect magnitudes. We compute the Spearman rank correlation coefficient ($\rho_{s}$) between the entanglement scores and the observed ripple magnitudes. A higher $\rho_{s}$ indicates that the method more accurately captures ripple effects. We evaluate five editing techniques (ROME~\cite{meng2022locating}, MEMIT~\cite{meng2022mass}, PRUNE~\cite{ma2024perturbation}, RECT~\cite{gu2401model}, and AlphaEdit~\cite{Fang2025_AlphaEdit}) over three LLMs of varying sizes: GPT2-XL (1.5B), GPT-J (6B), and Llama3 (8B, Instruct). Figure~\ref{fig:alphaedit_combined} shows correlation patterns of these methods when edited by AlphaEdit. We present correlation patterns for other techniques and further analysis in Appendix~\ref{appendix:exp1_diagrams}.

\begin{figure}
    \centering
    \begin{subfigure}{\linewidth}
    \includegraphics[width=\textwidth]{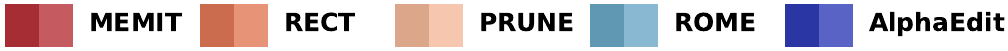}
    \end{subfigure}
    \begin{subfigure}{\linewidth}
    \includegraphics[width=\textwidth]{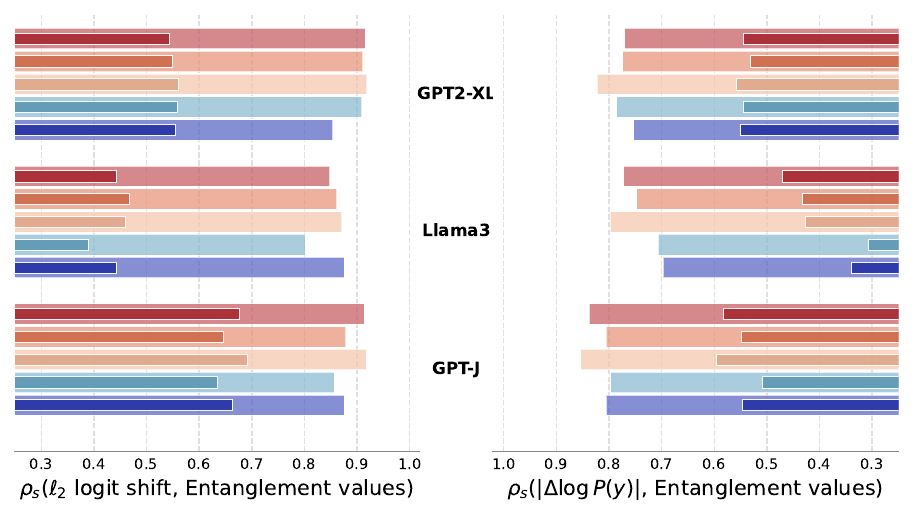}
    \end{subfigure}
    \caption{\review{Performance comparison between \Clare and GradSim in terms of Spearman correlation ($\rho_s$). The left panel shows $\rho_s$ between entanglement values and $\ell_2$ logit shift, and right panel shows $\rho_s$ between entanglement values and $|\Delta \log P(y)|$. \Clare (wider, transparent bars) consistently achieves higher $\rho_s$ than GradSim (narrower, solid bars).}}
    \label{fig:wide-figure}
\end{figure}

Table~\ref{tab:models-clare-gradsim-combined} reports $\rho_{s}$ across all LLMs and editing techniques. While GradSim provides moderate correlation ($\rho_{s} \approx 0.45$–$0.68$), \Clare achieves consistently strong alignment ($\rho_{s} \approx 0.75$–$0.92$), with an average improvement of 62.2\% over GradSim. For GPT2-XL, \Clare achieves an average improvement of 40.8\% over GradSim. GPT-J exhibits similar strong predictive alignment, with mean gain of 53.1\%. As shown in Figure~\ref{fig:wide-figure}, improvements are even more pronounced for Llama3, with \Clare achieving a mean 92.7\% higher correlation than GradSim, indicating that \Clare generalizes well to even larger models.


To evaluate \Clare's computational cost, we track total processing time, peak GPU memory use, and the size of per-fact representations during entanglement computation. We benchmark these across seven LLMs: GPT2-XL (1.5B), Llama3.2 (3B), GPT-J (6B), Qwen2 (7B), Qwen2.5 (7B), Mistral (7B, v0.1), and Llama3 (8B). For all quantities, a lower value indicates better performance.

\begin{table*}
\centering
\caption{Performance comparison between \Clare and GradSim. Lower values for all quantities indicate better performance. The ratio rows (e.g., Speed-up) show how many times \Clare is faster or more memory-efficient than GradSim. For example, a speed-up of \textbf{3.0×} means \Clare completes the task 3.0 times faster than GradSim.}
\label{tab:entanglement_results}
\resizebox{0.86\textwidth}{!}{
\begin{tabular}{cccccccc}
\hline
\textbf{Method} & \textbf{GPT2-XL} & \textbf{Qwen2.5} & \textbf{Qwen2} &
\textbf{Llama3} & \textbf{Llama3.2} & \textbf{GPT-J} & \textbf{Mistral} \\
\hline
\hline
\multicolumn{8}{c}{Total Processing Time (seconds)}\\
\hline
GradSim           & 0.796   & 0.915   & 0.868   & 0.917   & 0.774   & 0.763   & 0.833 \\
\Clare & \textbf{0.305} & \textbf{0.305} & \textbf{0.307} & \textbf{0.305} & \textbf{0.307} & \textbf{0.305} & \textbf{0.306} \\
\rowcolor{gray!15}Speed-up & \textcolor{blue!70!black}{2.610×} & \textcolor{blue!70!black}{3.000×} & \textcolor{blue!70!black}{2.827×} & \textcolor{blue!70!black}{3.007×} & \textcolor{blue!70!black}{2.521×} & \textcolor{blue!70!black}{2.502×} & \textcolor{blue!70!black}{2.722×} \\
\hline
\hline
\multicolumn{8}{c}{Size of representation used for computation (bytes)}\\
\hline
GradSim           & 6.23e9    & 3.05e10   & 3.05e10   & 3.21e10   & 1.29e10   & 2.42e10   & 2.90e10 \\
\Clare & \textbf{6400.00} & \textbf{14340.00} & \textbf{14340.00} & \textbf{16380.00} & \textbf{12290.00} & \textbf{16380.00} & \textbf{16380.00} \\
\rowcolor{gray!15}Compression & \textcolor{blue!70!black}{9.73e5×} & \textcolor{blue!70!black}{2.13e6×} & \textcolor{blue!70!black}{2.13e6×} & \textcolor{blue!70!black}{1.96e6×} & \textcolor{blue!70!black}{1.05e6×} & \textcolor{blue!70!black}{1.48e6×} & \textcolor{blue!70!black}{1.77e6×} \\
\hline
\hline
\multicolumn{8}{c}{Peak Utilised GPU Memory (GB)}\\
\hline
GradSim           & 18.212  & 87.513  & 87.513  & 91.765  & 37.445  & 68.474  & 81.483 \\
\Clare & \textbf{7.016} & \textbf{29.981} & \textbf{29.981} & \textbf{31.539} & \textbf{12.995} & \textbf{23.905} & \textbf{28.478} \\
\rowcolor{gray!15}Reduction & \textcolor{blue!70!black}{2.596×} & \textcolor{blue!70!black}{2.919×} & \textcolor{blue!70!black}{2.919×} & \textcolor{blue!70!black}{2.910×} & \textcolor{blue!70!black}{2.881×} & \textcolor{blue!70!black}{2.864×} & \textcolor{blue!70!black}{2.861×} \\
\hline
\end{tabular}
}
\end{table*}

\begin{figure*}[ht]
    \centering
    \begin{subfigure}[b]{0.3\textwidth}
        \includegraphics[width=\linewidth]{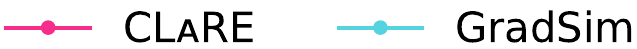}
    \end{subfigure}
    \par\medskip
    \begin{subfigure}[b]{0.31\textwidth}
        \includegraphics[width=\linewidth]{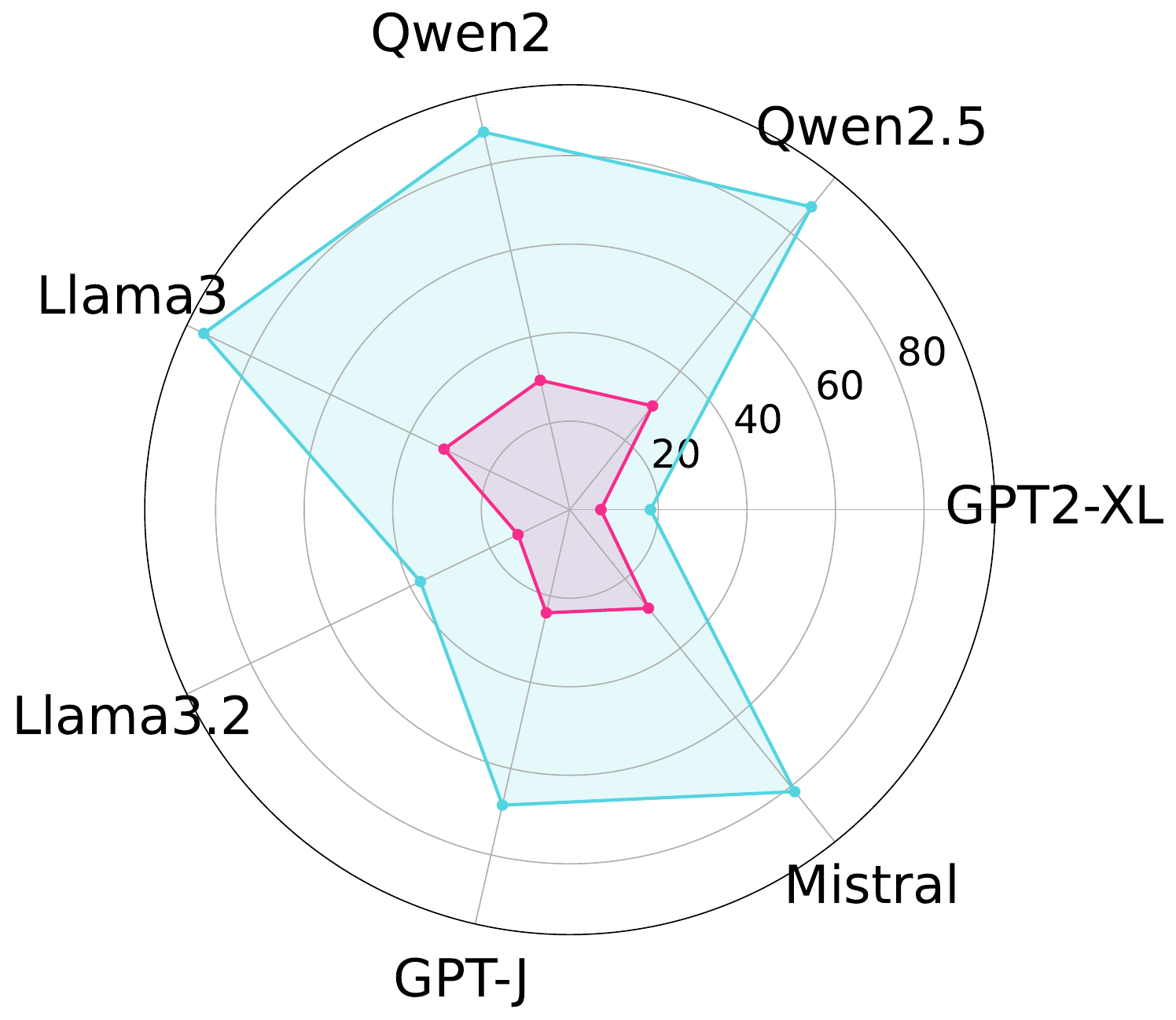}
        \caption{Peak Utilised GPU Usage (GB)}
    \end{subfigure}
    \begin{subfigure}[b]{0.31\textwidth}
        \includegraphics[width=\linewidth]{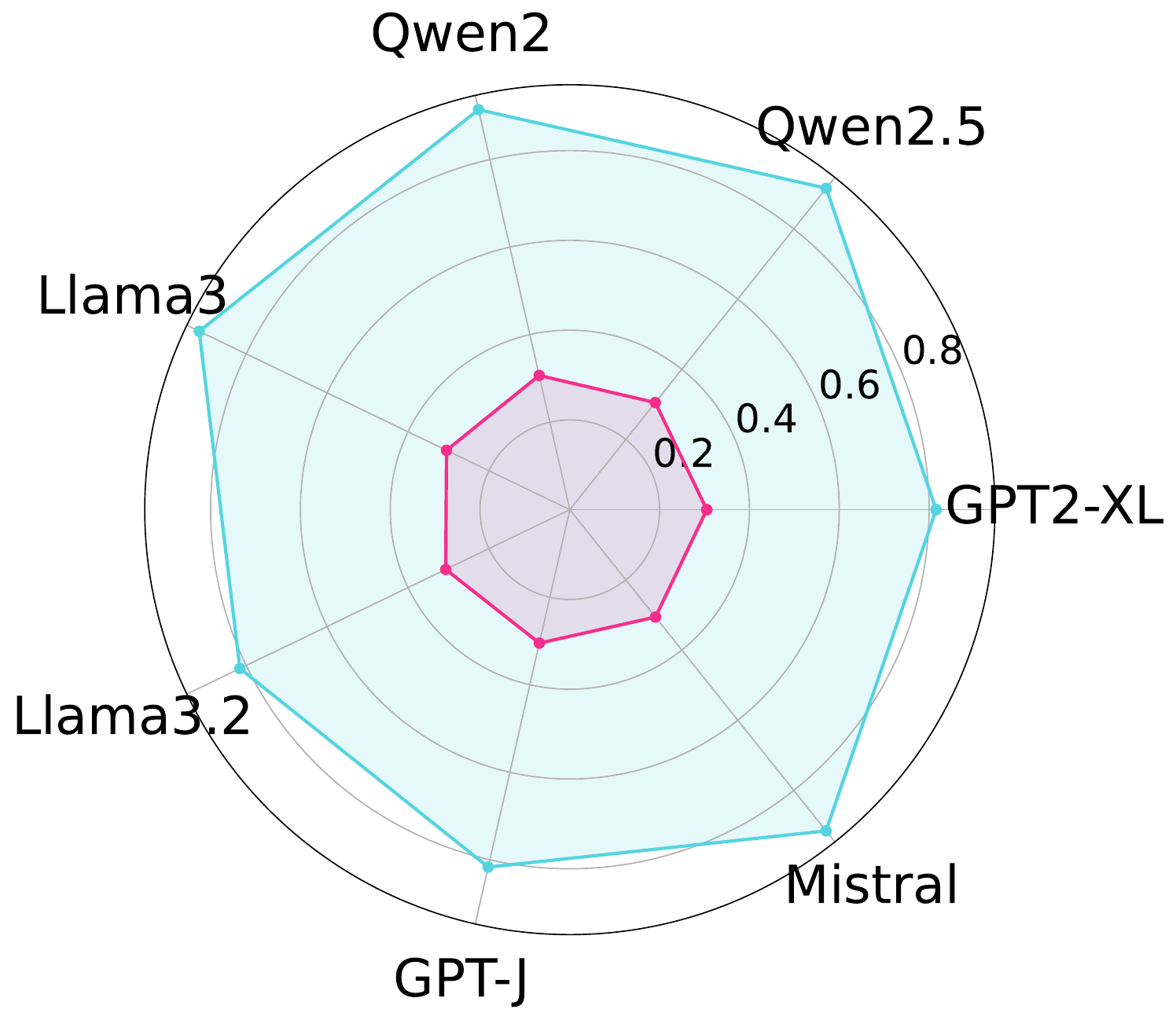}
        \caption{Total Processing Time (seconds)}
    \end{subfigure}
    \begin{subfigure}[b]{0.31\textwidth}
        \includegraphics[width=\linewidth]{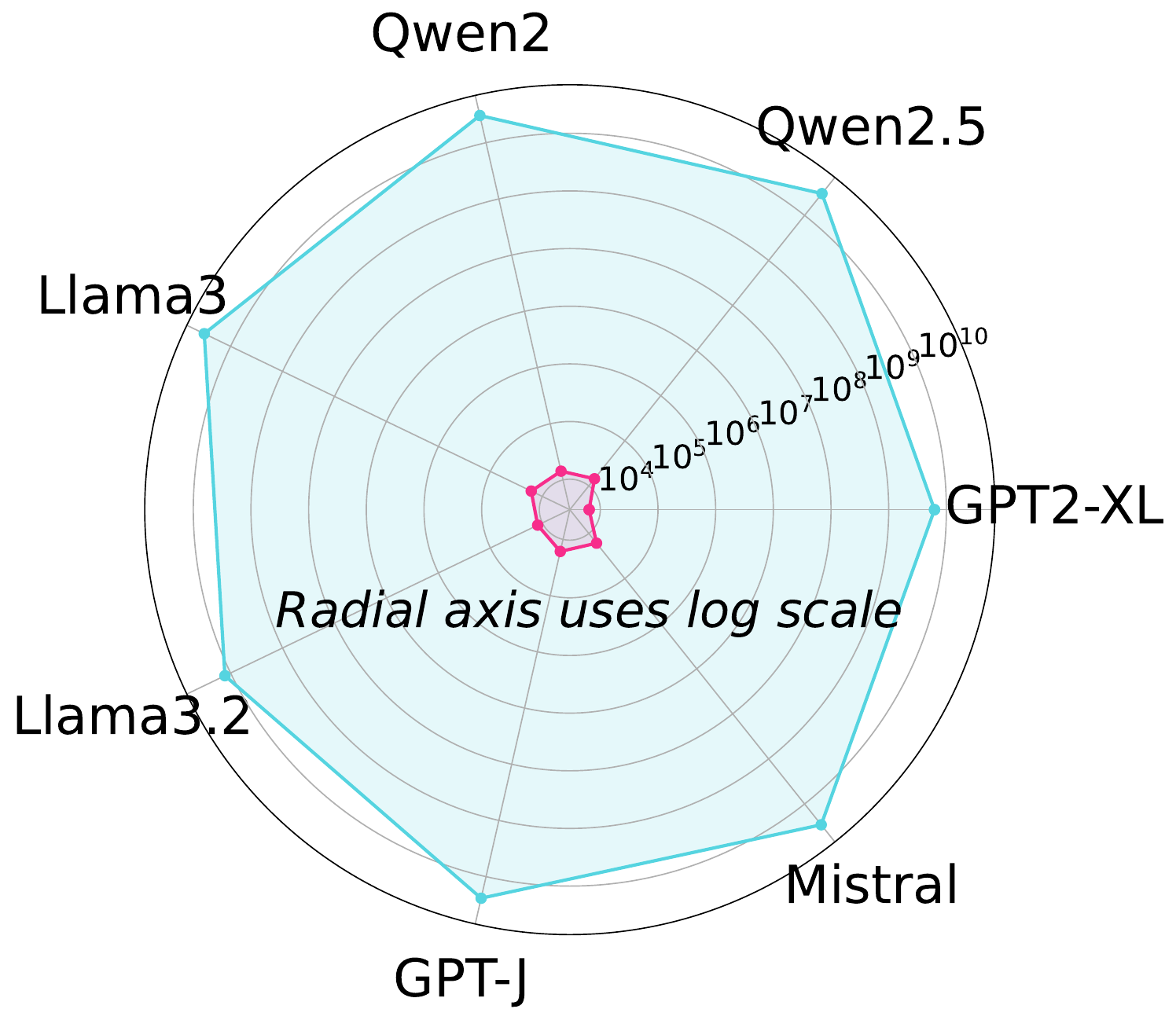}
        \caption{Size of representation (bytes)}
    \end{subfigure}
    \caption{Computational efficiency comparison. Closer to center => better performance.}
    \label{fig:main}
\end{figure*}

As shown in Table~\ref{tab:entanglement_results}, \Clare achieves substantial speed improvements, achieving an average of \textbf{2.74×} speed-up over GradSim. \Clare's factual representations are extremely compact, being measured in kilobytes. On average, \Clare achieves over a \textbf{1.64-million-fold compression} relative to GradSim. Peak GPU memory use in \Clare is \textbf{~2.85× lower} than GradSim. As observed in Figure~\ref{fig:main}, these results show that factual entanglement at the LLM's last critical layer reflects vulnerability to ripple propagation, and a lightweight, forward-only technique can outperform computationally intensive gradient methods.

\subsection{Layer-wise Correlation Analysis (RQ \ding{173})}
\label{rq2}
To identify the most suitable layer to capture factual entanglement, we conduct a layerwise correlation study for the same five editing techniques, three models and edit-control pairs as in RQ \ding{172}. For each control fact, instead of only $\mathcal{L}$, we take each transformer layer $l$'s hidden representation $h_{i}^{l}$, and find entanglement values using Eq.~\ref{eq:entanglement}. For each $l$, we measure how well its scores predict ripple magnitude in control facts following edits, in terms of $\ell_2$ logit shift. We compute the Spearman correlation for each $l$'s entanglement scores and $\ell_2$ logit shift, yielding a per-layer correlation profile.

From figure~\ref{fig:all_layers}, we observe a clear and consistent plateau around the last critical layer in correlation across all editing techniques, corresponding closely to the region previously identified to be storing factual associations by causal-tracing analyses~\cite{meng2022locating, meng2022mass}. Quantitatively, Table~\ref{tab:spearman_depth_final_layers_grouped_peak} shows that for all three models and five editing techniques, the correlation computed at the final critical layer lies within \textbf{0–1 percentage points} of the global maximum observed across all layers. This indicates that ripple effects can most effectively be predicted when analysed using hidden representations of the deepest layer of the critical region, $\mathcal{L}$. We discuss $\mathcal{L}$-approximation in Appendix~\ref{appendix:critical_layer_approximation} in case if prior causal analysis of identifying critical layers is unavailable.

\begin{table*}
\caption{$\rho_{s}$ between \Clare and ripple effect ($\ell_2$ logit shift) in control facts observed at $\mathcal{L}$, compared to maximum value possible at any layer. "Difference from peak" reports the absolute deviation in percentage points (p.p.).}
\label{tab:spearman_depth_final_layers_grouped_peak}
\centering
\resizebox{0.8\textwidth}{!}{
\begin{tabular}{l l c c c c c}
\hline
\textbf{Model} & \textbf{Layer Type} &
\textbf{AlphaEdit} & \textbf{MEMIT} & \textbf{PRUNE} & \textbf{RECT} & \textbf{ROME} \\
\hline
\multirow{3}{*}{Llama3}
 & $\mathcal{L} = 9$      & 65.16\% & 63.76\% & 64.35\% & 61.73\% & 70.54\% \\
 & Maximum $\rho_{s}$ (any layer) & 65.16\% & 64.16\% & 64.84\% & 61.73\% & 70.54\% \\
 & Difference from peak (p.p.)   & 0.00 & 0.40 & 0.49 & 0.00 & 0.00 \\
\hline
\multirow{3}{*}{GPT-J}
 & $\mathcal{L} = 9$      & 89.30\% & 80.13\% & 92.72\% & 90.26\% & 88.46\% \\
 & Maximum $\rho_{s}$ (any layer) & 89.72\% & 80.13\% & 92.97\% & 90.26\% & 88.58\% \\
 & Difference from peak (p.p.)   & 0.42 & 0.00 & 0.25 & 0.00 & 0.12 \\
\hline
\multirow{3}{*}{GPT2-XL}
 & $\mathcal{L} = 18$     & 76.61\% & 87.61\% & 87.98\% & 86.33\% & 85.10\% \\
 & Maximum $\rho_{s}$ (any layer) & 77.49\% & 87.67\% & 87.99\% & 86.42\% & 85.26\% \\
 & Difference from peak (p.p.)   & 0.88 & 0.06 & 0.01 & 0.09 & 0.16 \\
\hline
\end{tabular}
}
\end{table*}

\subsection{Performance on scalability and downstream applications support (RQ \ding{174})}
\label{rq3}

To demonstrate scalability, we release entanglement graphs for our entire corpus of 11,427 facts for GPT2-XL, GPT-J and Llama3. This is made possible by \Clare's compact vector representation, which occupies only a fraction of storage required by GradSim~(see Table~\ref{tab:entanglement_results}). These lightweight vectors are efficiently stored and quickly loaded to the GPU for pairwise cosine similarity computation for thousands of fact pairs. In contrast, GradSim requires storing the full gradient for each fact, which is equal in size to the model itself, making it infeasible for large-scale analysis. \Clare reduces this storage burden in orders of $10^{6}\times$, enabling corpus-wide analysis at scale. We analyse distribution of the corpus's facts and their entanglement patterns in Appendix~\ref{appendix:distribution}.

\begin{figure}[t]
\centering
\begin{Verbatim}[breaklines=true,breakanywhere=true,fontsize=\small,commandchars=\\\{\}]
1. The name of the country of citizenship of \textbf{Kate Winslet} is \underline{UK} (affects 1257 facts)
2. The place of burial of \textbf{Audrey Hepburn} is \underline{Cemetery of Tolochenaz} (affects 1233 facts)
3. The place of birth of \textbf{Audrey Hepburn} is \underline{Rue Keyenveld} (affects 1222 facts)
4. The place of death of \textbf{Audrey Hepburn} is \underline{Tolochenaz} (affects 1218 facts)
5. The name of the country of citizenship of \textbf{Mark Wahlberg} is \underline{USA} (affects 1204 facts)
\end{Verbatim}
\caption{Top-5 entangled facts in GPT-J in our corpus.}
\label{fig:gptj_hub_1_main}

\end{figure}

To support comprehensive evaluation of editing techniques under a constrained evaluation budget, we identify highly entangled facts in our corpus. These high-density facts are critical pressure points in LLMs, as editing them may trigger widespread ripple effects. Figure~\ref{fig:gptj_hub_1_main} shows the top 5 most entangled facts in GPT-J, ranked by the number of other facts they are connected to with an entanglement score above 0.7. As shown in Appendix~\ref{appendix:rq4}, facts exceeding this threshold are highly ripple-prone, serving as ideal adversarial candidates. Thus, by surfacing these highly connected regions, \Clare enables fact prioritization for red-teaming. We present more clusters for various models in Appendix~\ref{appendix:redteaming}.

\begin{figure}
\centering
    \includegraphics[width=0.86\linewidth]{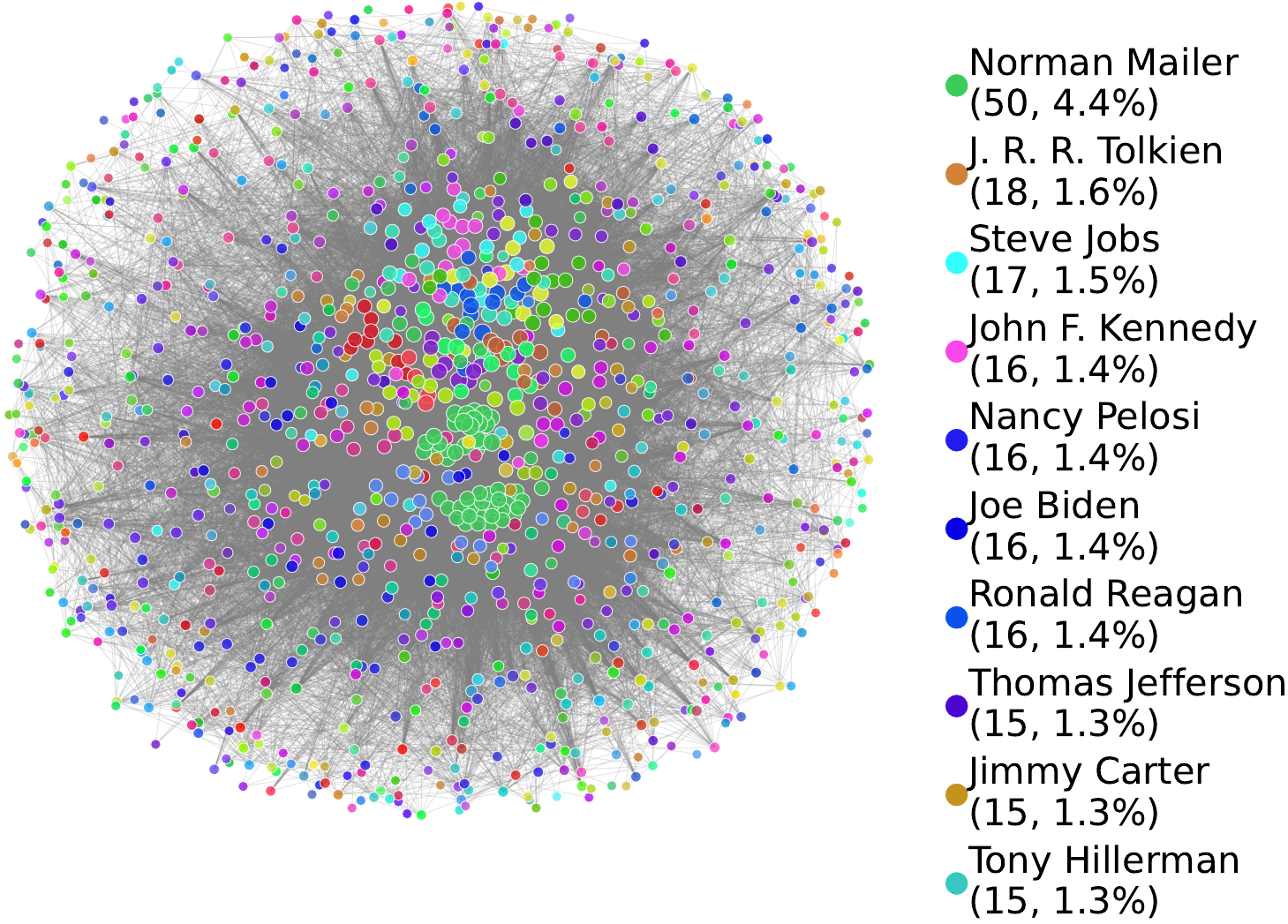}
    \caption{An entanglement cluster in GPT-J with its ten most entangled subjects listed with their number and percentage of facts. It consists of 1,149 facts and 397 subjects, with 87.9\% of edges between inter-subject facts. Node color indicates subject identity.}
    \label{fig:gptj_c1_main}
\end{figure}

We cluster the entanglement graphs using the Louvain community detection algorithm, where nodes represent facts and edges connect pairs with high entanglement scores (cosine similarity > 0.7). Figure~\ref{fig:gptj_c1_main} shows one such cluster in GPT-J, highlighting the top 10 subjects whose facts have the highest connectivity and thus the strongest ripple potential. These clusters span multiple subjects, revealing substantial cross-entity interference. They are especially useful for constructing preservation sets: when editing a fact about subject A, constraints must be applied to other facts within the same cluster to minimize collateral degradation. This extends the capabilities of DiKE~\cite{zhang2025disentangling}, enabling it to preserve facts in hidden space as well. We present more clusters for various LLMs in Appendix~\ref{appendix:clusters}.

\section{Conclusion}
We presented \Clare, a lightweight and scalable technique to estimate factual entanglement within LLMs. \Clare uses forward activations from a single intermediate layer without requiring model gradients. Through extensive model-editing experiments, we show that \Clare achieves stronger alignment with observed ripple effects than gradient-based baselines, with an average 62.2\% improvement in Spearman correlation, while offering 2.74× speedups, 2.85× lower peak GPU memory usage, and requiring only a fraction of the storage needed to preserve fact representations. \review{Beyond improving ripple-effect diagnostics, \Clare enables several capabilities that can support the development of future model-editing techniques. It allows for entanglement-aware fact prioritization for budget-constrained red-teaming, facilitates the construction of preservation sets that extend beyond predefined semantic neighborhoods, and provides a scalable mechanism for auditing editing safety in LLMs.}

\section{Limitations}
While \Clare provides a scalable and empirically strong proxy for predicting ripple effects through activation-based entanglement scores, our analysis remains correlational in nature. That is, although we observe high alignment between \Clare scores and observed behavioral interference post-editing, we do not establish a formal causal mechanism linking representational entanglement to model generalization or degradation. Investigating the causal pathways by which knowledge is stored, shared, and disrupted during pretraining and editing remains an open and important direction for future work. A second limitation is that \Clare currently functions as a diagnostic tool rather than an editing strategy. While \Clare enables the construction of entanglement-aware preservation sets, we do not integrate these constraints into existing editing techniques. Future research could build on \Clare's preservation sets to create entanglement-aware editing techniques that reduce collateral degradation during updates.

\section{Ethical considerations}

From an ethical standpoint, the ability to map these regions allows developers to be more accountable for the ``collateral degradation'' caused by their updates. By building preservation sets based on entanglement clusters, researchers can actively prevent the erosion of model integrity. However, the paper acknowledges that \Clare is currently a diagnostic tool, and the ethical responsibility remains with the practitioners to integrate these constraints into the actual editing process to ensure a ``preventive'' rather than ``reactive'' approach to model safety. We emphasize that our release of entanglement graphs is intended for research and safety applications only, and we encourage future work to adopt responsible deployment practices. Incorporating entanglement-aware auditing may play a key role in developing accountable model-editing pipelines in high-stakes or sensitive domains. All entanglement graphs and data samples are released under the MIT License.

\section*{Acknowledgments}
This research/project is supported by the National Research Foundation, Singapore, and the Cyber Security Agency of Singapore under the National Cybersecurity R\&D Programme (NCR25-CRPO-0001) and the CyberSG R\&D Programme Office (Award CRPO-GC5-ASTAR-001).

Any opinions, findings, conclusions, or recommendations expressed in these materials are those of the author(s) and do not reflect the views of the National Research Foundation, Singapore, the Cyber Security Agency of Singapore, or the CyberSG R\&D Programme Office.

This work is also supported by the National University of Singapore Research Scholarship.

\bibliography{custom}

\appendix

\section{Computational complexity analysis}
\label{appendix:complexity}
We analyze the time and space complexity of GradSim and \Clare to compute entanglement between two facts. For GradSim, the process can be divided into primarily into four components: forward pass, loss computation, backward pass (for both facts), and then cosine similarity computation over the parameter gradients. \Clare requires only a forward pass up to the last critical layer $\mathcal{L}$ for both facts, followed by cosine similarity computation for the hidden representation at $\mathcal{L}$. We provide the details for each step in GradSim and \Clare in Table~\ref{tab:complexity_comparison}. GradSim requires both a forward and backward pass per factual triplet, while \Clare achieves comparable entanglement estimation using only a single forward pass.


\begin{figure*}
    \centering
    \begin{subfigure}{0.7\textwidth}
        \centering
        \includegraphics[width=\linewidth]{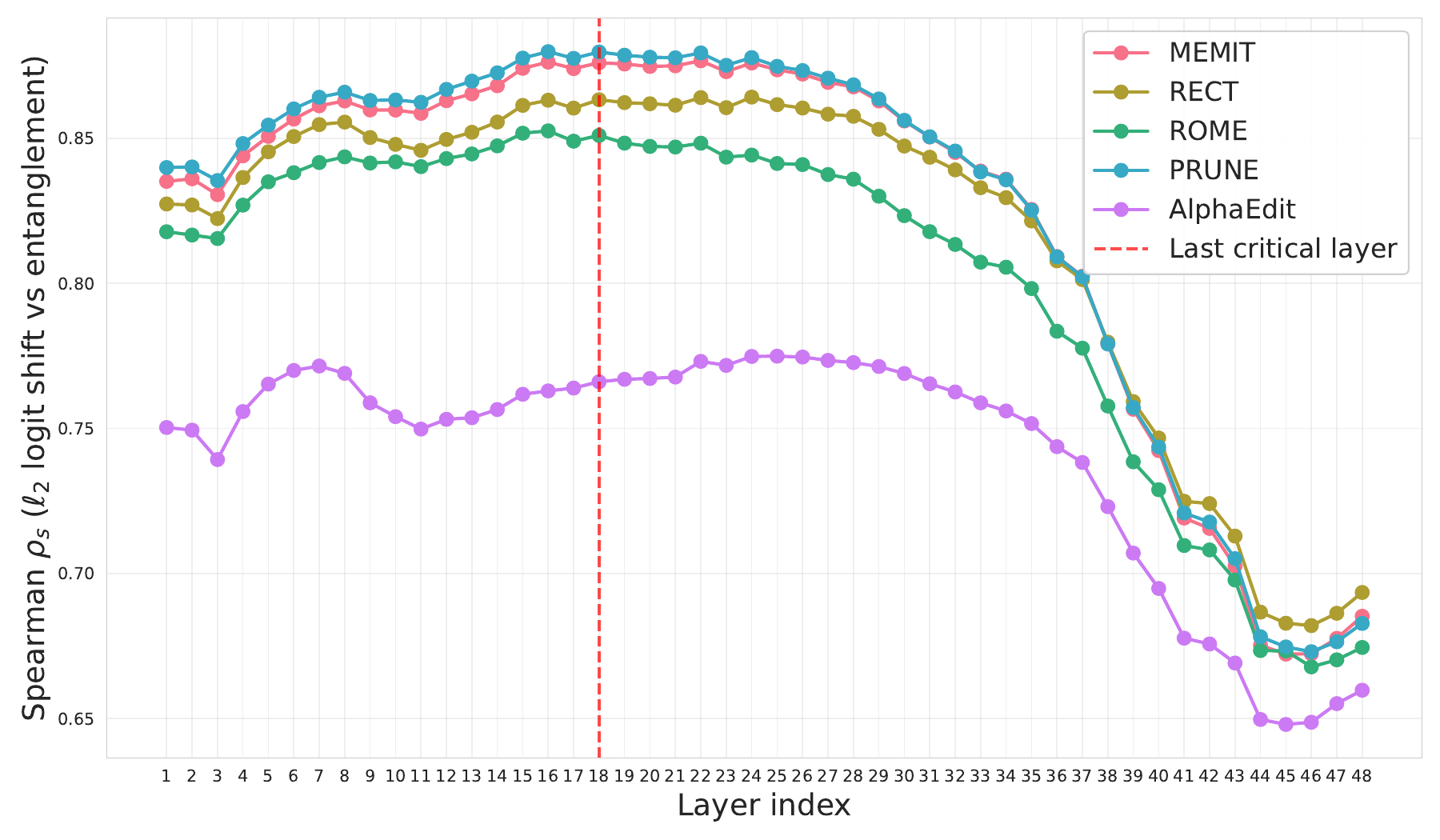}
        \caption{GPT2-XL}
        \label{fig:top1}
    \end{subfigure}
    \begin{subfigure}{0.7\textwidth}
        \centering
        \includegraphics[width=\linewidth]{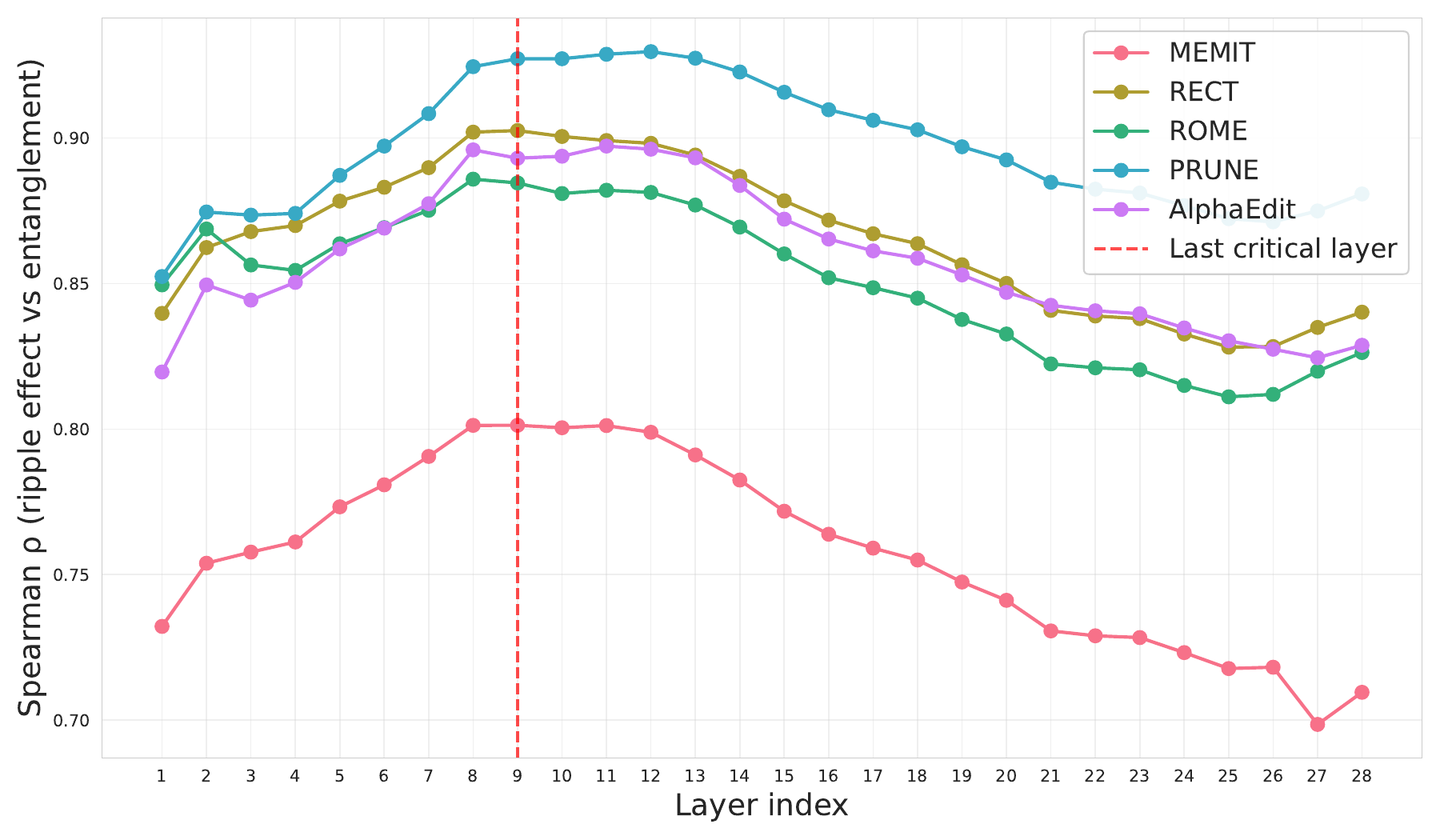}
        \caption{GPT-J}
        \label{fig:top2}
    \end{subfigure}
    \begin{subfigure}{0.7\textwidth}
        \centering
         \includegraphics[width=\linewidth]{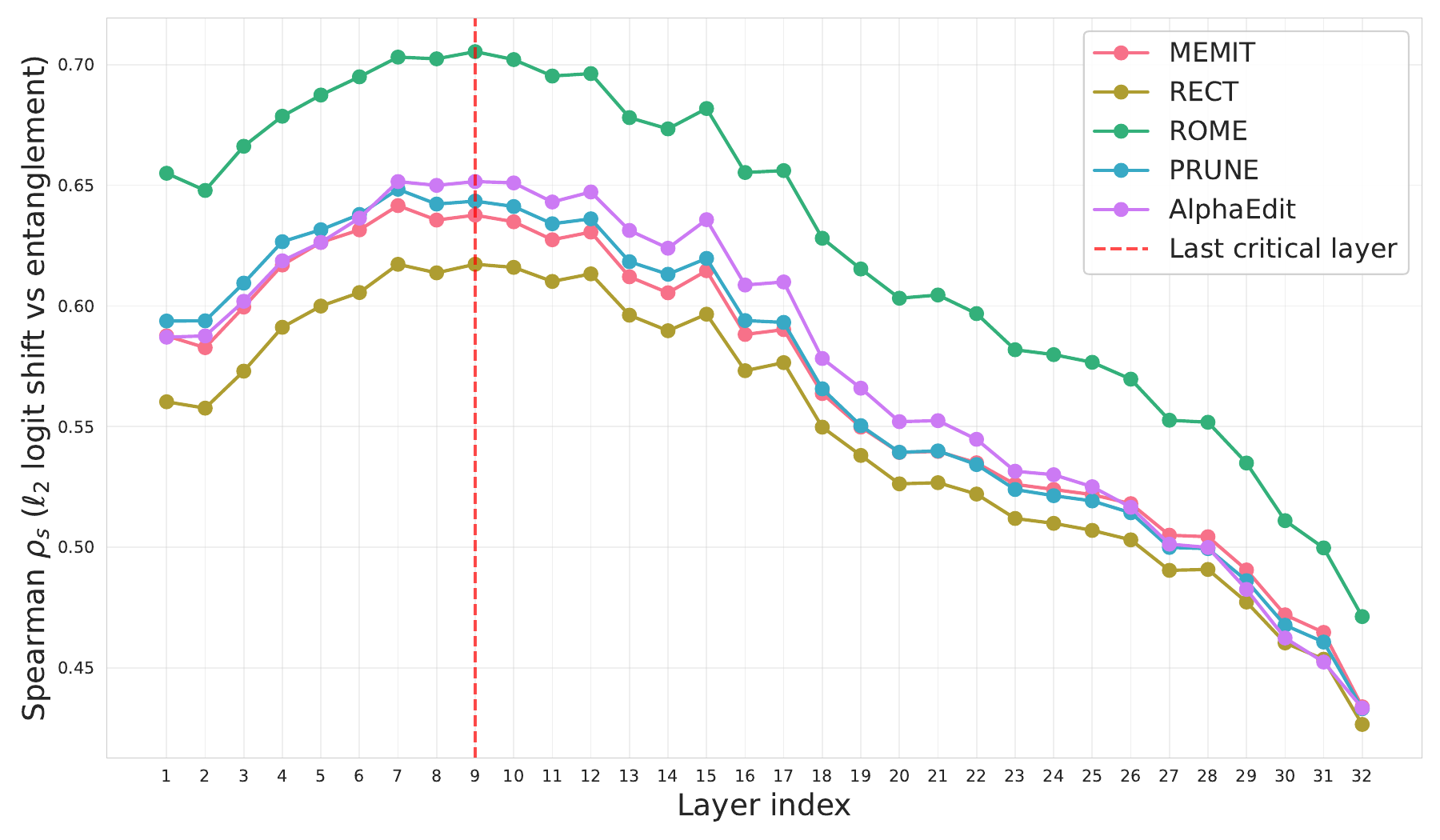}
        \caption{Llama3}
        \label{fig:top3}
    \end{subfigure}
    \caption{Layerwise Spearman correlation ($\rho_{s}$) between \Clare and observed ripple magnitudes. Correlation peaks around the last critical layer indicate that it is most informative about entanglement estimation.}
    \label{fig:all_layers}
\end{figure*}

\begin{table*}
\centering
\caption{Step-by-step comparison of time and space complexity between GradSim and \Clare for computing entanglement between two facts.}
\label{tab:complexity_comparison}
\begin{tabular}{p{5.7cm}cc}
\toprule
\textbf{Component} & \textbf{GradSim} & \textbf{\Clare} \\
\midrule
\textbf{Forward pass (per fact)} & 
$\mathcal{O}(L \cdot d^2 + N^2 \cdot L \cdot d)$ &
$\mathcal{O}(L \cdot d^2 + N^2 \cdot L \cdot d)$ \\

\textbf{Loss computation (per fact)} &
$\mathcal{O}(d \cdot v)$ &
--- \\

\textbf{Backward pass (per fact)} &
$\mathcal{O}(L \cdot d^2)$ &
--- \\

\textbf{Cosine similarity computation} &
$\mathcal{O}(L \cdot d^2)$ &
$\mathcal{O}(d)$ \\

\textbf{Total Time Complexity} &
$\sim \mathcal{O}(L \cdot d^2)$ &
$\sim \mathcal{O}(L \cdot d^2)$ \\
\midrule

\textbf{Gradient storage (per fact)} &
$\mathcal{O}(L \cdot d^2)$ &
--- \\

\textbf{Activation storage (per fact)} &
$\mathcal{O}(N \cdot L \cdot d)$ &
$\mathcal{O}(d)$ \\

\textbf{Total Space Complexity (per fact)} &
$\sim \mathcal{O}(L \cdot d^2)$ &
$\mathcal{O}(d)$ \\

\bottomrule
\end{tabular}
\end{table*}

\subparagraph{Time Complexity.}
For GradSim, the forward pass processes $N$ input tokens through $L$ transformer layers, with each layer usually having the same hidden dimension size  $d$. Each layer consists of multi-head attention with complexity $\mathcal{O}(N^2 \cdot d)$ and feed-forward MLPs with complexity $\mathcal{O}(d^2)$, yielding a total forward pass complexity of $\mathcal{O}(L \cdot d^2 + N^2 \cdot L \cdot d)$. For typical model-editing prompts where $N \approx 10$--$20$ tokens and $d$ is in the orders of $10^{3}$, (4096 in the case of Llama3-8B), the MLP term dominates since $d^2 \gg N^2 \cdot d$. Thus, the forward pass simplifies to $\mathcal{O}(L \cdot d^2)$. The loss computation involves a final projection from hidden dimension $d$ to vocabulary size $v$, requiring $\mathcal{O}(d \cdot v)$ operations. This is also typically negligible compared to the $\mathcal{O}(L \cdot d^2)$ forward pass cost. The backward pass has the same computational complexity as the forward pass, $\mathcal{O}(L \cdot d^2)$, as it traverses in reverse to compute gradients with respect to all parameters. Finally, computing the cosine similarity between gradient vectors requires $\mathcal{O}(L \cdot d^2)$ operations since we must compute the dot product and norms over all $\mathcal{O}(L \cdot d^2)$ model parameters. The total time complexity per fact pair is thus $\mathcal{O}(L \cdot d^2)$, dominated by the forward and backward passes.

For \Clare, the forward pass up to layer $\mathcal{L}$ has complexity $\mathcal{O}(L \cdot d^2 + N^2 \cdot L \cdot d) \approx \mathcal{O}(L \cdot d^2)$. No backward pass is required, as \Clare operates purely on forward activations. The cosine similarity computation operates on $d$-dimensional hidden state vectors (one per fact), requiring only $\mathcal{O}(d)$ operations to compute the dot product $\langle h_i^{\mathcal{L}}, h_j^{\mathcal{L}} \rangle$ and vector norms $\|h_i^{\mathcal{L}}\|$ and $\|h_j^{\mathcal{L}}\|$. While \Clare and GradSim have the same time complexity, \Clare is $\sim$2.8× faster as it eliminates the need for backward passes.

\subparagraph{Space Complexity.}
GradSim requires $\mathcal{O}(L \cdot d^2)$ space to store gradients for all model parameters (weight matrices of dimension $d \times d$ across $L$ layers), plus $\mathcal{O}(N \cdot L \cdot d)$ space for caching intermediate activations needed during backpropagation. In practice, gradient storage dominates, yielding $\mathcal{O}(L \cdot d^2)$ total space complexity. In contrast, \Clare requires only $\mathcal{O}(d)$ space to store a single $d$-dimensional hidden state vector per fact at layer $\mathcal{L}$. This efficiency enables \Clare to scale to analyzing thousands of fact pairs, as each fact requires only a compact $d$-dimensional representation rather than the full gradient footprint, and thus \Clare achieves an average of $1.64$-million-fold compression in storing fact representations.

\section{Runtime cost analysis}
\label{appendix:cost_breakdown}

We empirically benchmark the component-wise computational costs of \Clare and GradSim across seven language models. We measure both execution time and peak GPU memory usage for each step involved in computing entanglement between a pair of facts.

\paragraph{Runtime Analysis.} Table~\ref{tab:time_breakdown} reports the time (in seconds) required by each step in \Clare and GradSim. \Clare involves two forward passes (one for each fact) followed by cosine similarity computation over $d$-dimensional activation vectors. GradSim, in contrast, requires forward, loss, and backward passes for both facts, followed by cosine similarity computation over full gradients. Across all models, \Clare achieves substantial speedups. This efficiency gain stems primarily from the elimination of loss calculation and backward passes.

\paragraph{Memory Utilization.} Table~\ref{tab:gpu_breakdown} presents the peak GPU memory usage (in GB) at each stage of computation. For \Clare, peak memory remains below 32 GB even for 7B and 8B models, enabling full execution on a single H200 GPU. In contrast, GradSim exhibits significantly higher memory demands, peaking near or beyond 90 GB for large models due to the need to retain gradients and activations for all parameters.

As shown in Table~\ref{tab:entanglement_results}, \Clare stores representations of size $4 \times d$, where $d$ is the hidden dimension of the LLM. In contrast, GradSim stores the full gradient for each fact, with a storage cost of $4 \times P$, where $P$ is the total number of model parameters. All storage measurements are computed using FP32 precision.

\begin{table*}
\centering
\caption{Detailed time breakdown (seconds) for \Clare and GradSim. \Clare follows: Forward Pass ($\times 2$) $\oplus$ Cosine similarity (Sim.) of two vectors (Vec). GradSim follows: Forward ($\times 2$) $\oplus$ Loss ($\times 2$) $\oplus$ Backward ($\times 2$) $\oplus$ Cosine similarity of gradients (Grad).}
\label{tab:time_breakdown}
\resizebox{\textwidth}{!}{%
\begin{tabular}{l|cc|c|cccc|c}
\toprule
\multirow{2}{*}{\textbf{Model}} & \multicolumn{3}{c|}{\textbf{Method 1: \Clare}} & \multicolumn{5}{c}{\textbf{Method 2: GradSim}} \\ \cmidrule(lr){2-4} \cmidrule(lr){5-9}
 & \begin{tabular}[c]{@{}c@{}}Forward\\ ($\times 2$ Vec)\end{tabular} & Sim. & \textbf{Total} & \begin{tabular}[c]{@{}c@{}}Forward\\ ($\times 2$ Grad)\end{tabular} & \begin{tabular}[c]{@{}c@{}}Loss\\ ($\times 2$ Grad)\end{tabular} & \begin{tabular}[c]{@{}c@{}}Backward\\ ($\times 2$ Grad)\end{tabular} & Sim. & \textbf{Total} \\ \midrule
GPT-J & 0.204 & 0.101 & \textbf{0.305} & 0.251 & 0.205 & 0.205 & 0.102 & \textbf{0.763} \\
GPT2-XL & 0.204 & 0.101 & \textbf{0.305} & 0.284 & 0.205 & 0.205 & 0.102 & \textbf{0.796} \\
Llama3.2 & 0.206 & 0.101 & \textbf{0.307} & 0.262 & 0.204 & 0.205 & 0.102 & \textbf{0.773} \\
Llama3 & 0.204 & 0.101 & \textbf{0.305} & 0.305 & 0.205 & 0.205 & 0.202 & \textbf{0.917} \\
Mistral & 0.205 & 0.101 & \textbf{0.306} & 0.304 & 0.205 & 0.223 & 0.101 & \textbf{0.833} \\
Qwen2 & 0.206 & 0.101 & \textbf{0.307} & 0.255 & 0.204 & 0.206 & 0.202 & \textbf{0.867} \\
Qwen2.5 & 0.204 & 0.101 & \textbf{0.305} & 0.303 & 0.205 & 0.205 & 0.202 & \textbf{0.915} \\ \bottomrule
\end{tabular}%
}
\end{table*}

\section{More research questions}
\label{appendix:more_rq}

We address the following research questions here:

\textbf{\underline{RQ \ding{175}}}: Can \Clare serve as a diagnostic for evaluating the disruptive impact of model-editing techniques and comparing them?

\textbf{\underline{RQ \ding{176}}}: How are the entanglement patterns across different language models?

\subsection{Evaluating the Disruptive Impact of Model-Editing Techniques (RQ\ding{175})}
\label{appendix:rq4}

Using \Clare, we qualitatively compare editing techniques by their propensity to induce ripple effects. Appendix~\ref{appendix:exp1_diagrams} shows \Clare's entanglement correlation patterns for a single edited fact and many control facts, offering a finer-grained view of ripple susceptibility. They show a consistent threshold across models and techniques: entanglement scores below 0.7 yield minimal change, while scores more than 0.7 lead to sharply increasing ripple effects. This nonlinear transition shows that highly entangled facts are more prone to ripple effects, a risk \Clare can detect prior to editing. When comparing editing techniques across GPT2-XL, GPT-J and Llama3, we observe clear differences in ripple severity. ROME and PRUNE often induce stronger and more widespread ripple effects as control facts become more entangled, whereas RECT and MEMIT generally yield more localized, controlled updates. AlphaEdit exhibits intermediate behavior. Thus, \Clare both anticipates where ripple effects will occur and provides a model- and technique-agnostic way to assess robustness and disruptiveness in editing, supporting informed deployment and design decisions.

\subsection{Entanglement score Distribution (RQ \ding{176})}
\label{appendix:distribution}

\begin{figure}
    \centering
    \includegraphics[width=\linewidth]{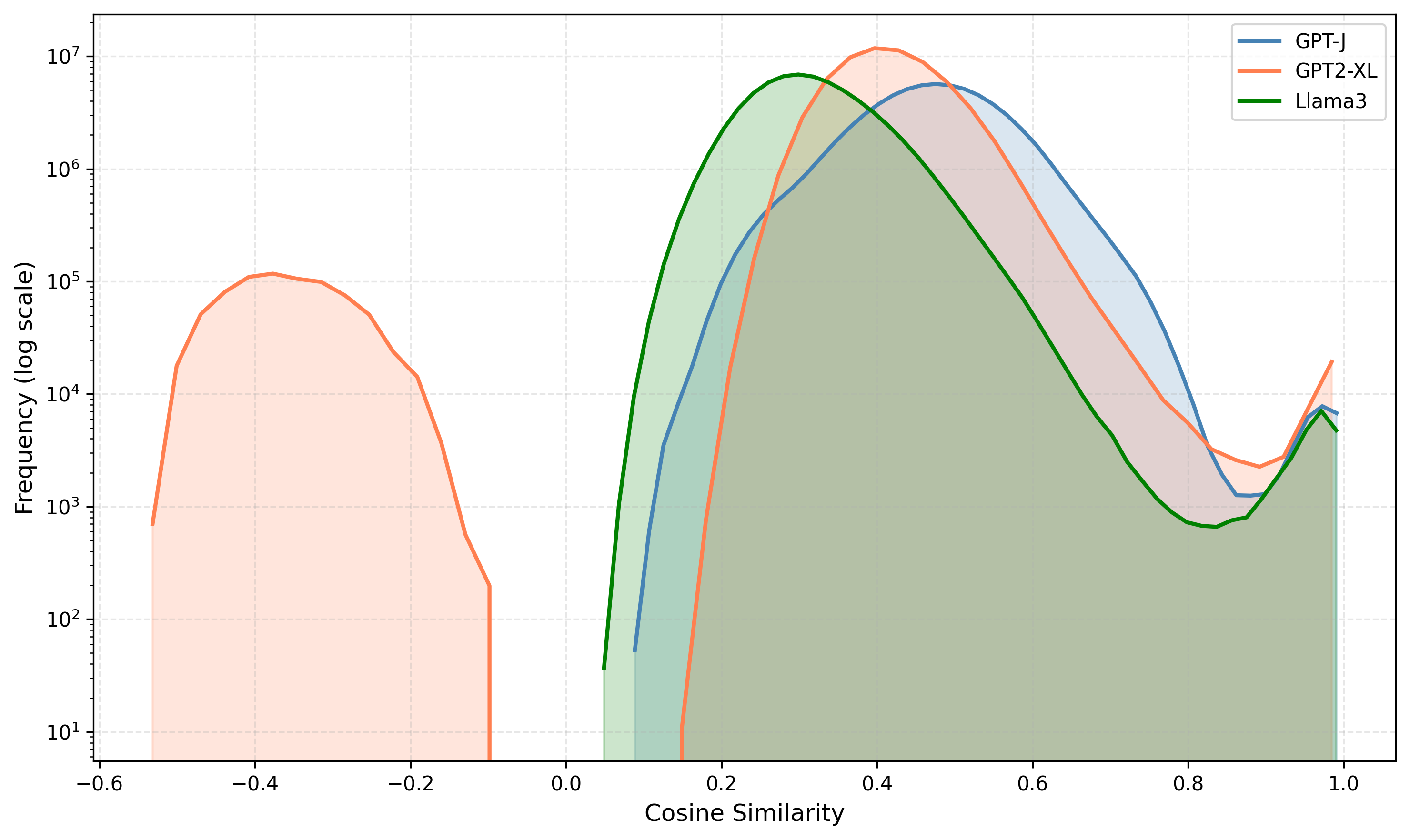}
    \caption{Distribution of cosine similarities between \Clare representations across 11,427 facts.}
    \label{fig:distribution}
\end{figure}

Figure~\ref{fig:distribution} shows cosine similarity distribution between \Clare representations across all 11,427 factual triplets for GPT-J, GPT2-XL and Llama3. Most entanglement scores lie between 0.2 and 0.6, indicating moderate representational overlap between most fact pairs. Notably, a non-trivial mass exists near the upper extreme (0.95-1.0), suggesting a high-density zone of strongly entangled facts that may be especially susceptible to ripple effects. GPT2-XL also shows a distinct secondary cluster with negative cosine similarities (centered near -0.35), not present in the other models, pointing to architectural differences in how subject representations are organized. The long-tailed shape and multimodal structure of the distribution underscore the need for selective entanglement-aware editing strategies.

\section{Performance of other editing techniques in RQ \ding{172}}
\label{appendix:exp1_diagrams}
\begin{figure*}[t]
    \centering
    
    \begin{subfigure}[b]{0.32\textwidth}
        \includegraphics[width=\textwidth]{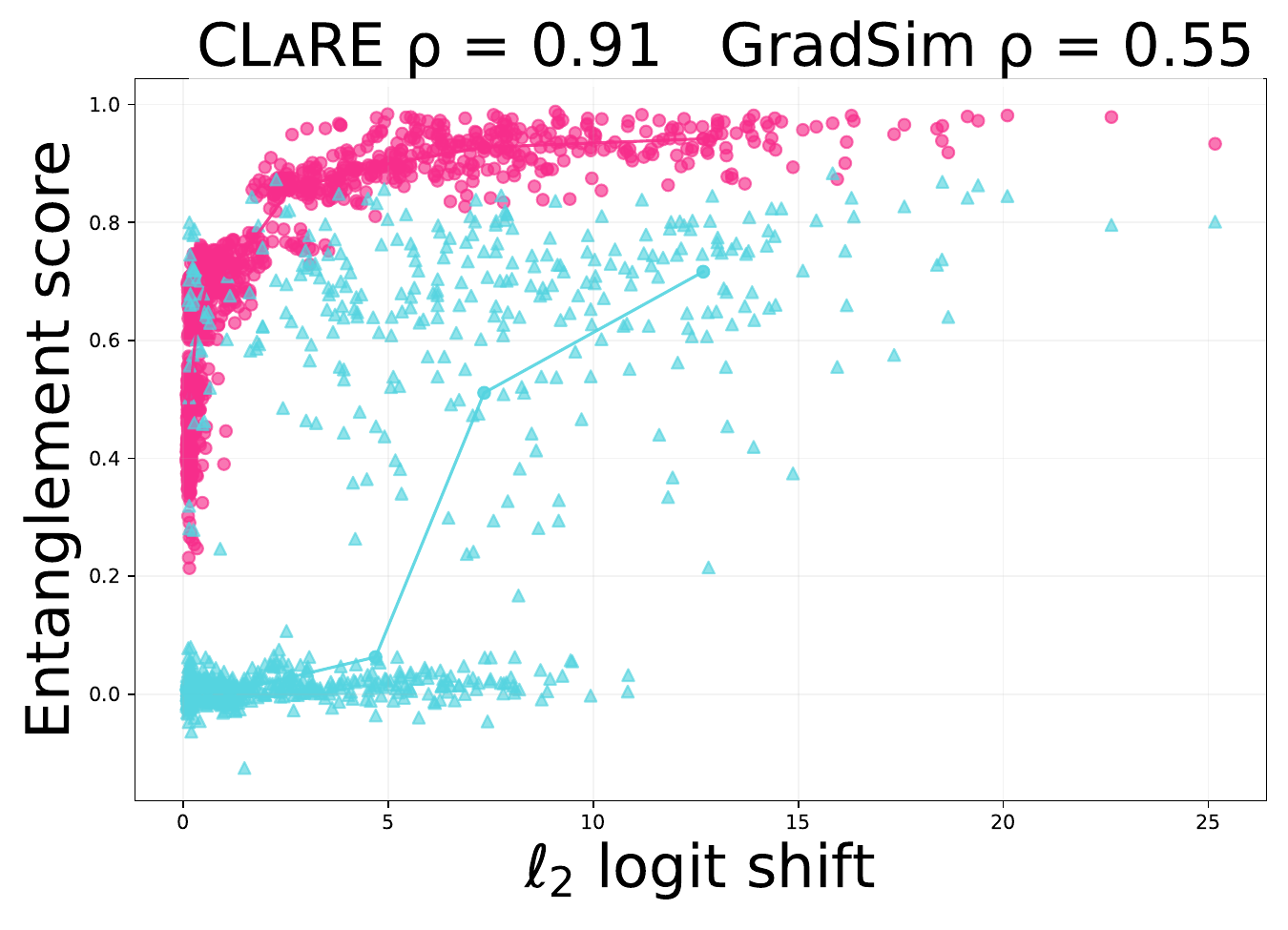}
        \caption{GPT2-XL}
    \end{subfigure}
    \hfill
    \begin{subfigure}[b]{0.32\textwidth}
        \includegraphics[width=\textwidth]{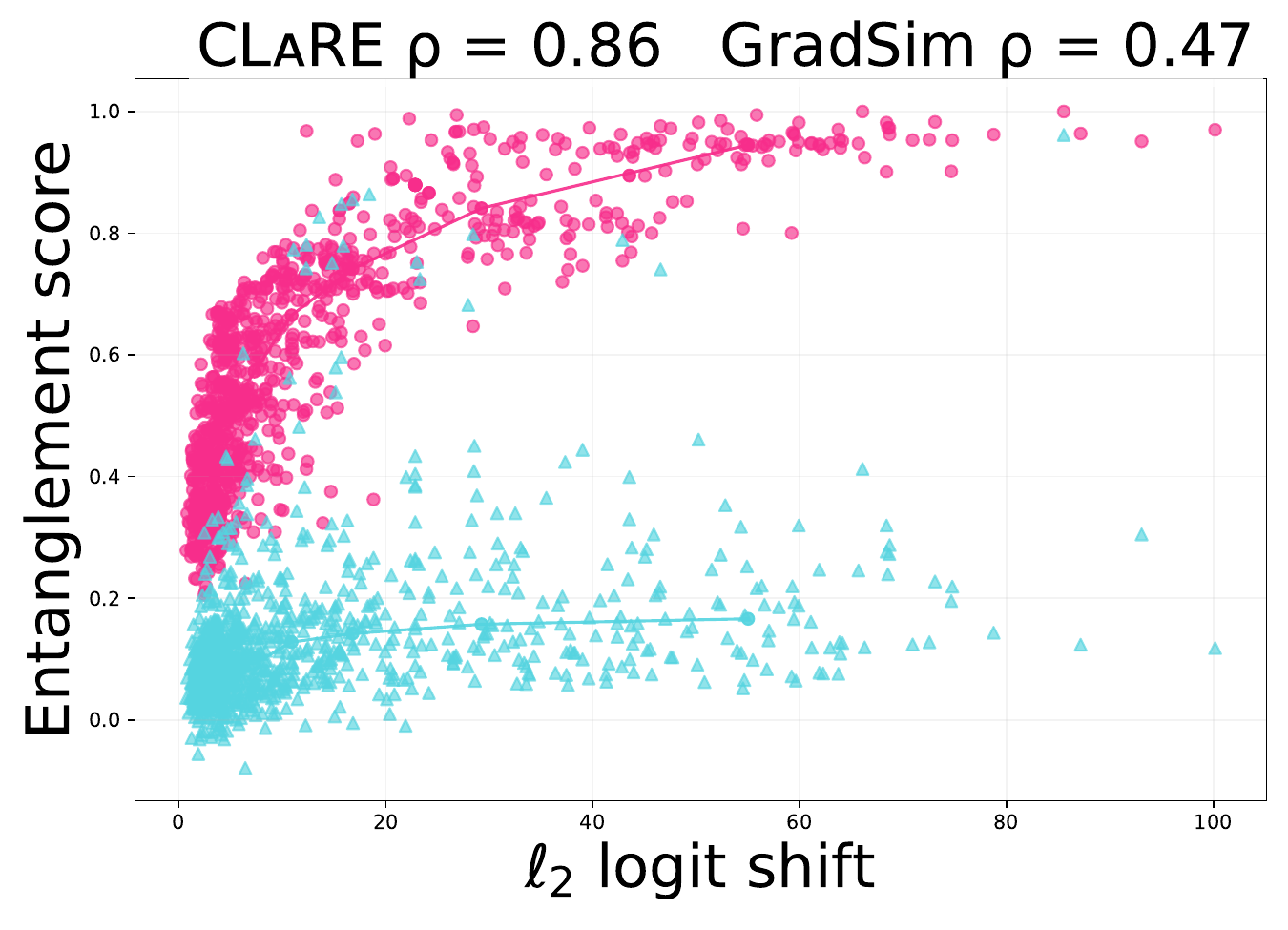}
        \caption{Llama3}
    \end{subfigure}
    \hfill
    \begin{subfigure}[b]{0.32\textwidth}
        \includegraphics[width=\textwidth]{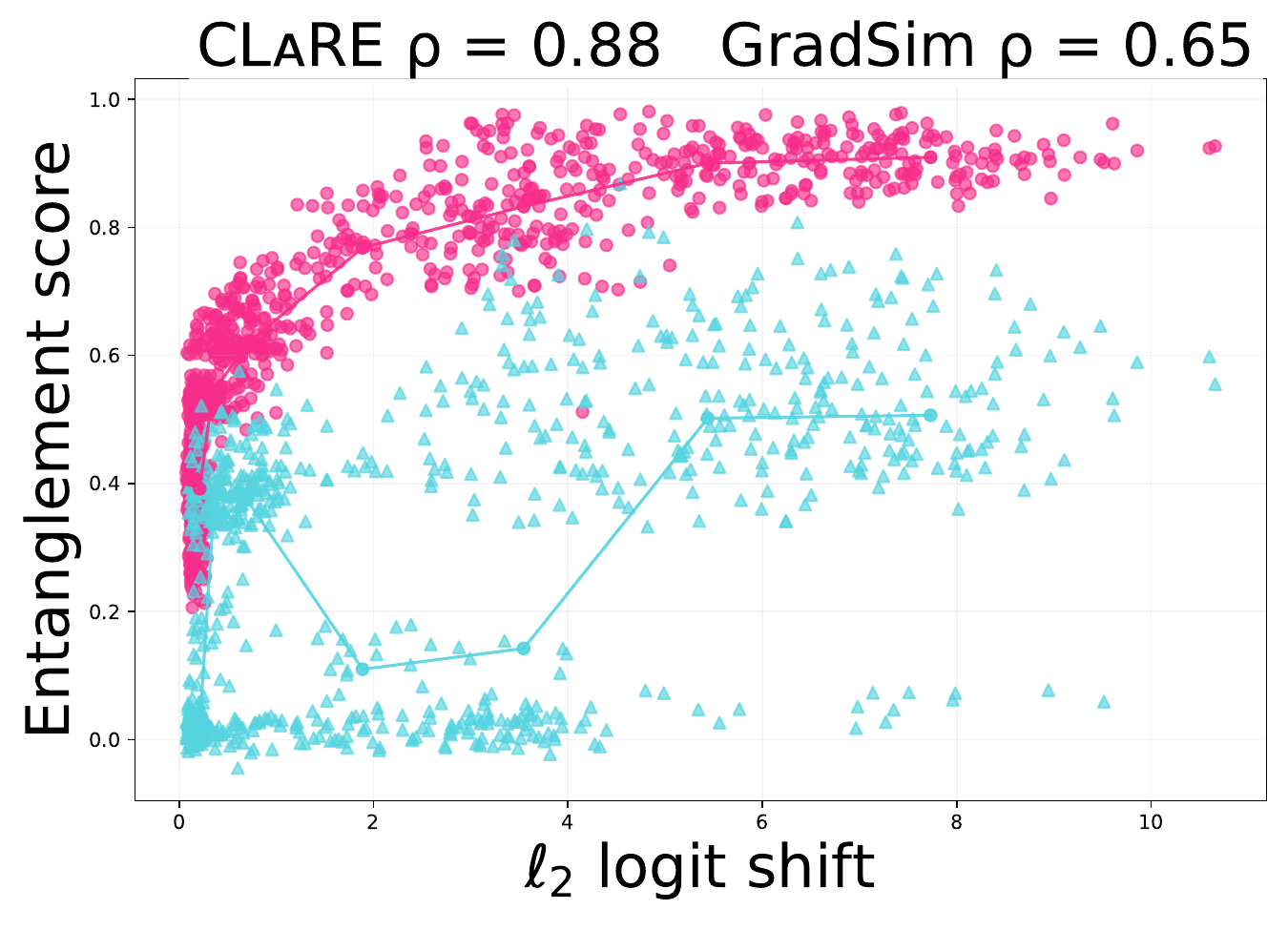}
        \caption{GPT-J}
    \end{subfigure}
    
    \caption{Correlation patterns for RECT across different models for entanglement vs $\ell_2$ logit shift.}
    \label{fig:rect_l2}
\end{figure*}

\begin{figure*}[t]
    \centering
    
    \begin{subfigure}[b]{0.32\textwidth}
        \includegraphics[width=\textwidth]{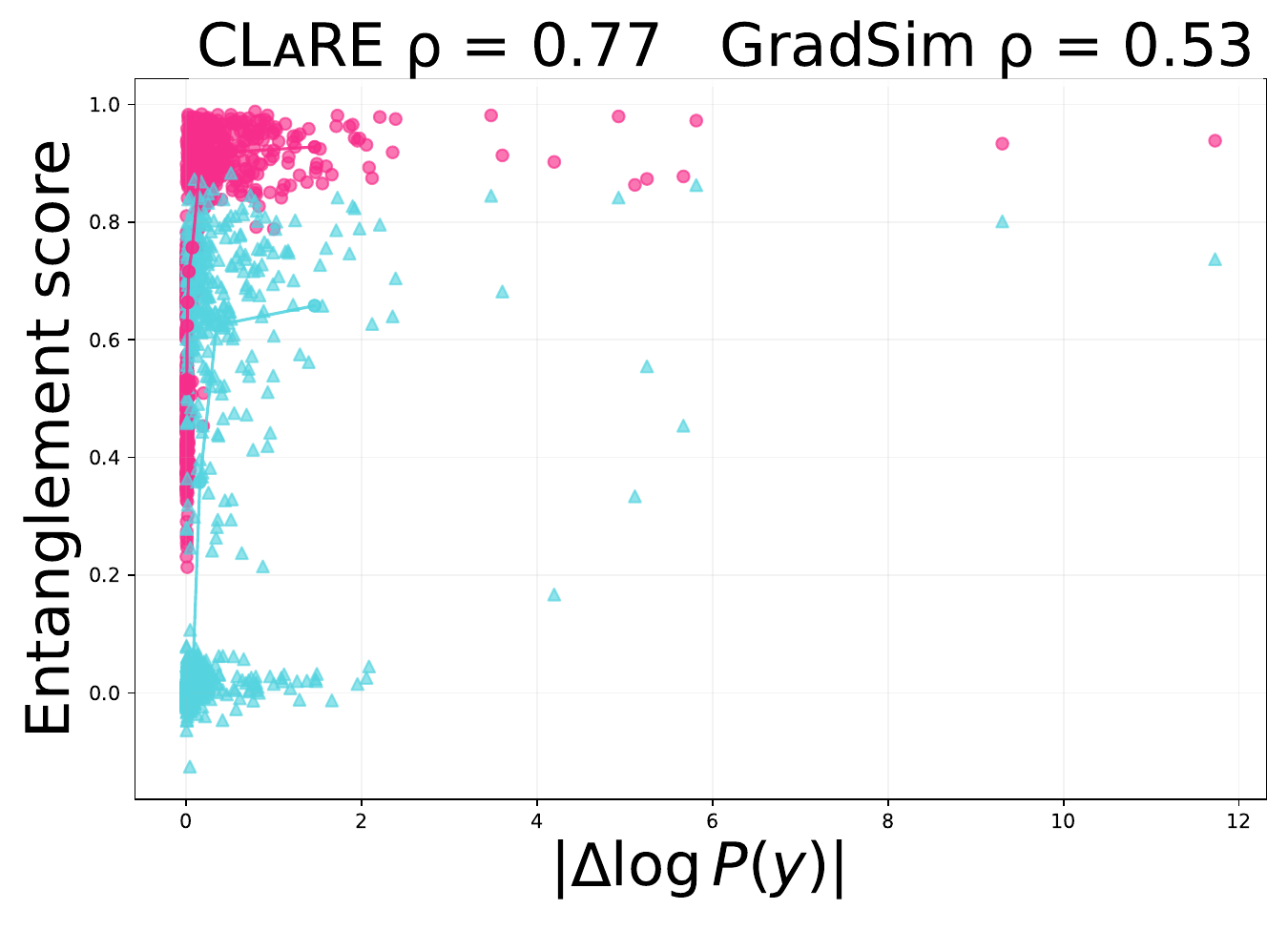}
        \caption{GPT2-XL}
    \end{subfigure}
    \hfill
    \begin{subfigure}[b]{0.32\textwidth}
        \includegraphics[width=\textwidth]{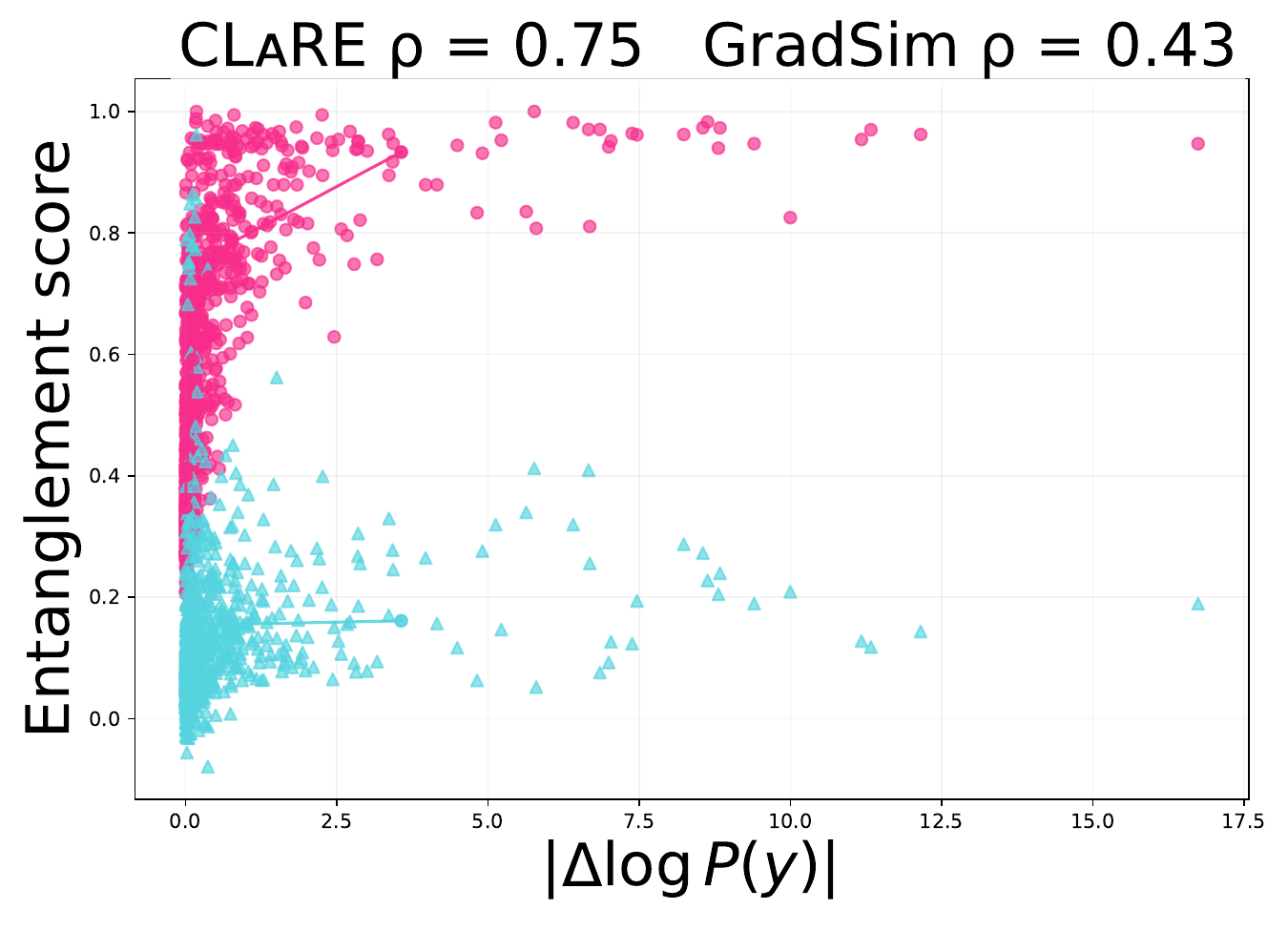}
        \caption{Llama3}
    \end{subfigure}
    \hfill
    \begin{subfigure}[b]{0.32\textwidth}
        \includegraphics[width=\textwidth]{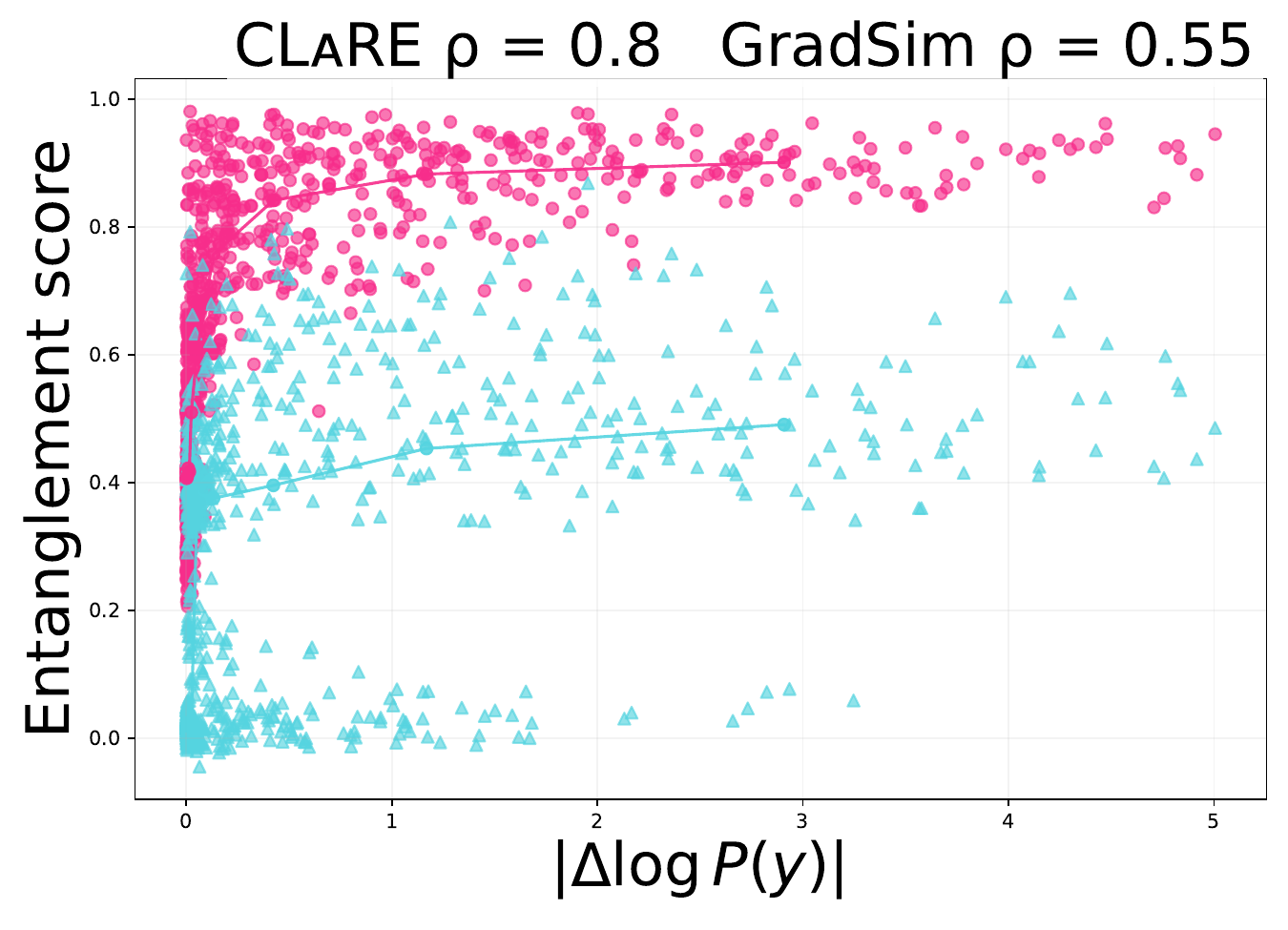}
        \caption{GPT-J}
    \end{subfigure}
    
    \caption{Correlation patterns for RECT across different models for entanglement vs $|\Delta \log P(y)|$.}
    \label{fig:rect_delta}
\end{figure*}

\begin{figure*}[t]
    \centering
    
    \begin{subfigure}[b]{0.32\textwidth}
        \includegraphics[width=\textwidth]{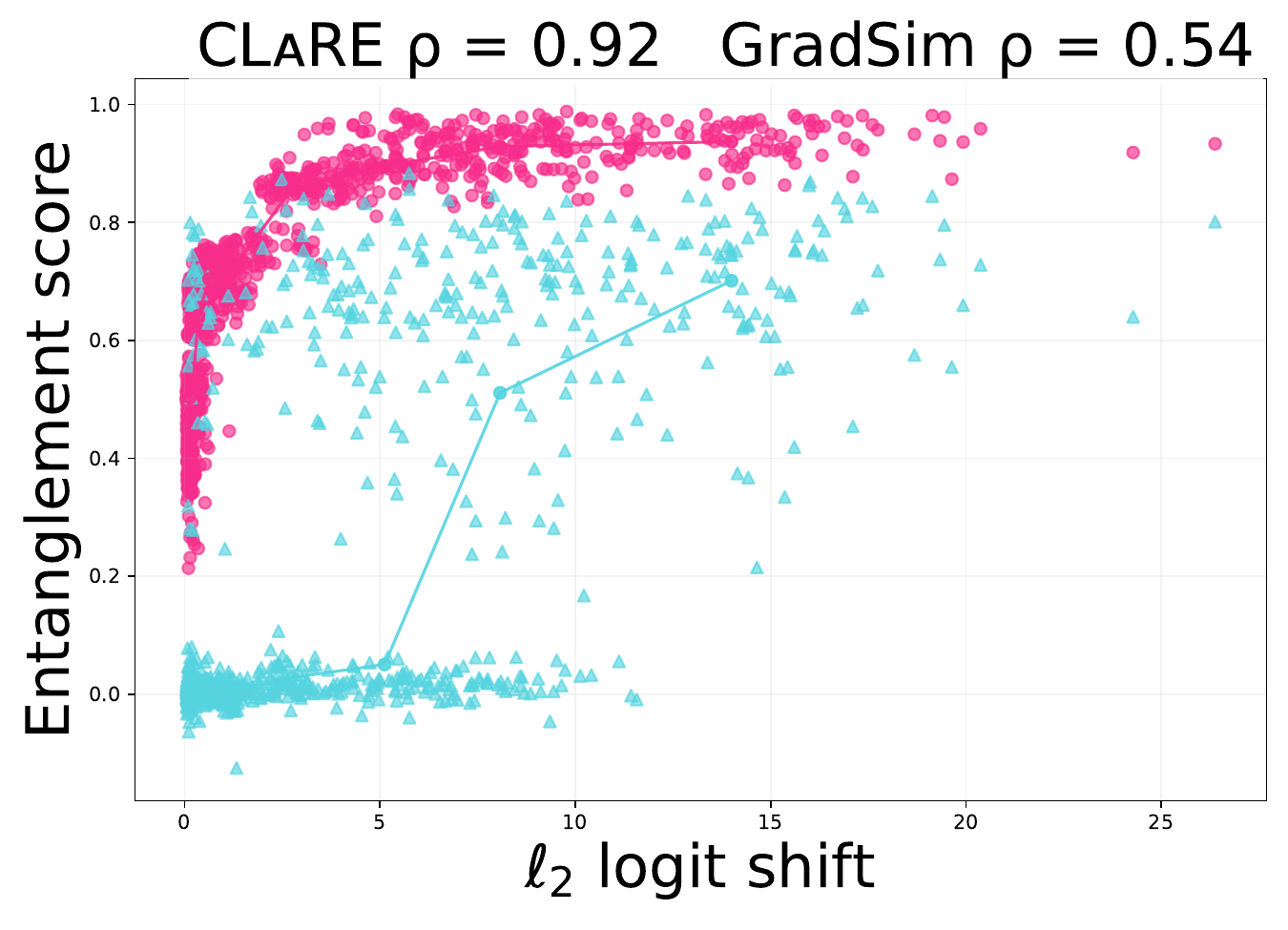}
        \caption{GPT2-XL}
    \end{subfigure}
    \hfill
    \begin{subfigure}[b]{0.32\textwidth}
        \includegraphics[width=\textwidth]{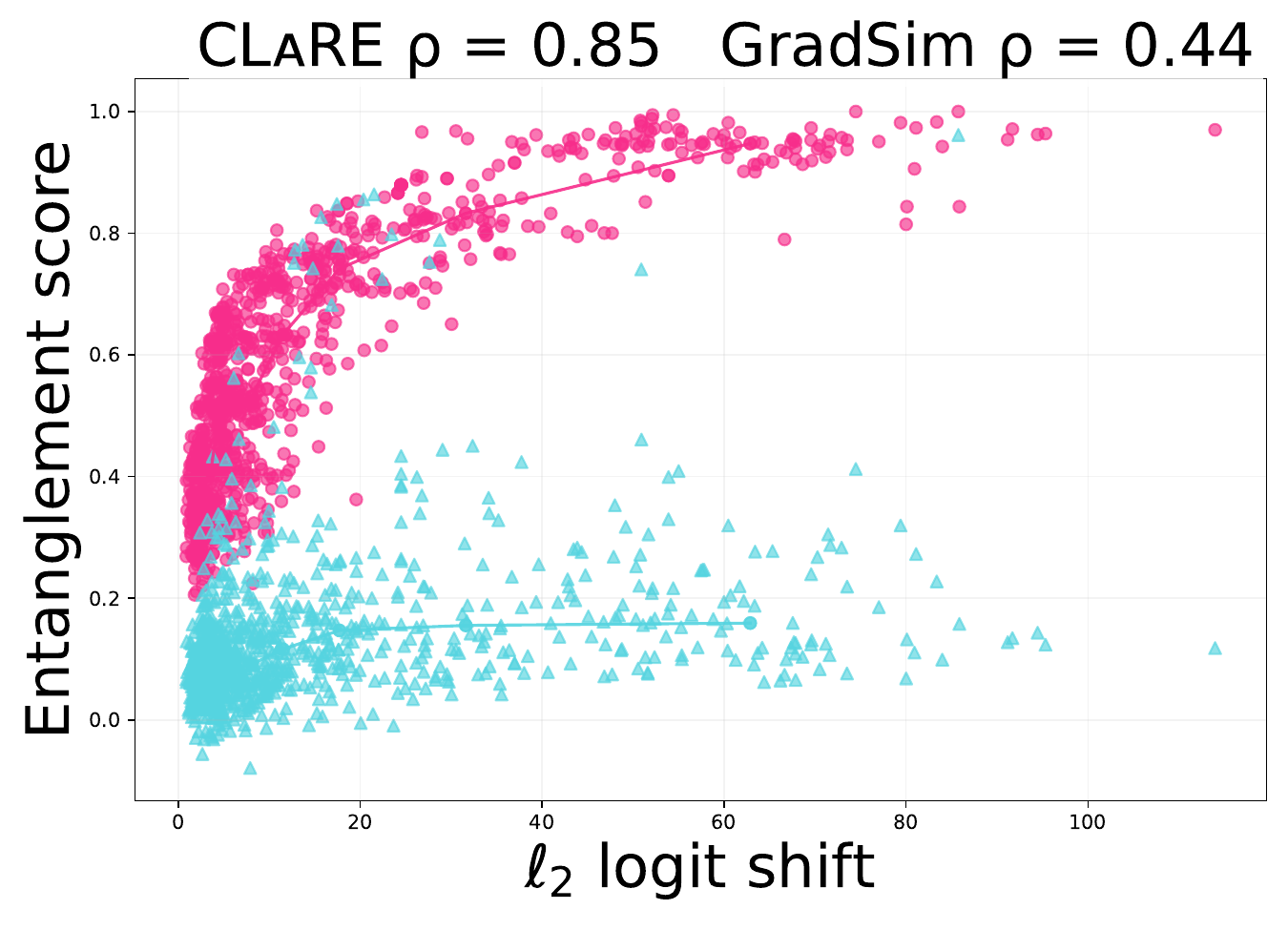}
        \caption{Llama3}
    \end{subfigure}
    \hfill
    \begin{subfigure}[b]{0.32\textwidth}
        \includegraphics[width=\textwidth]{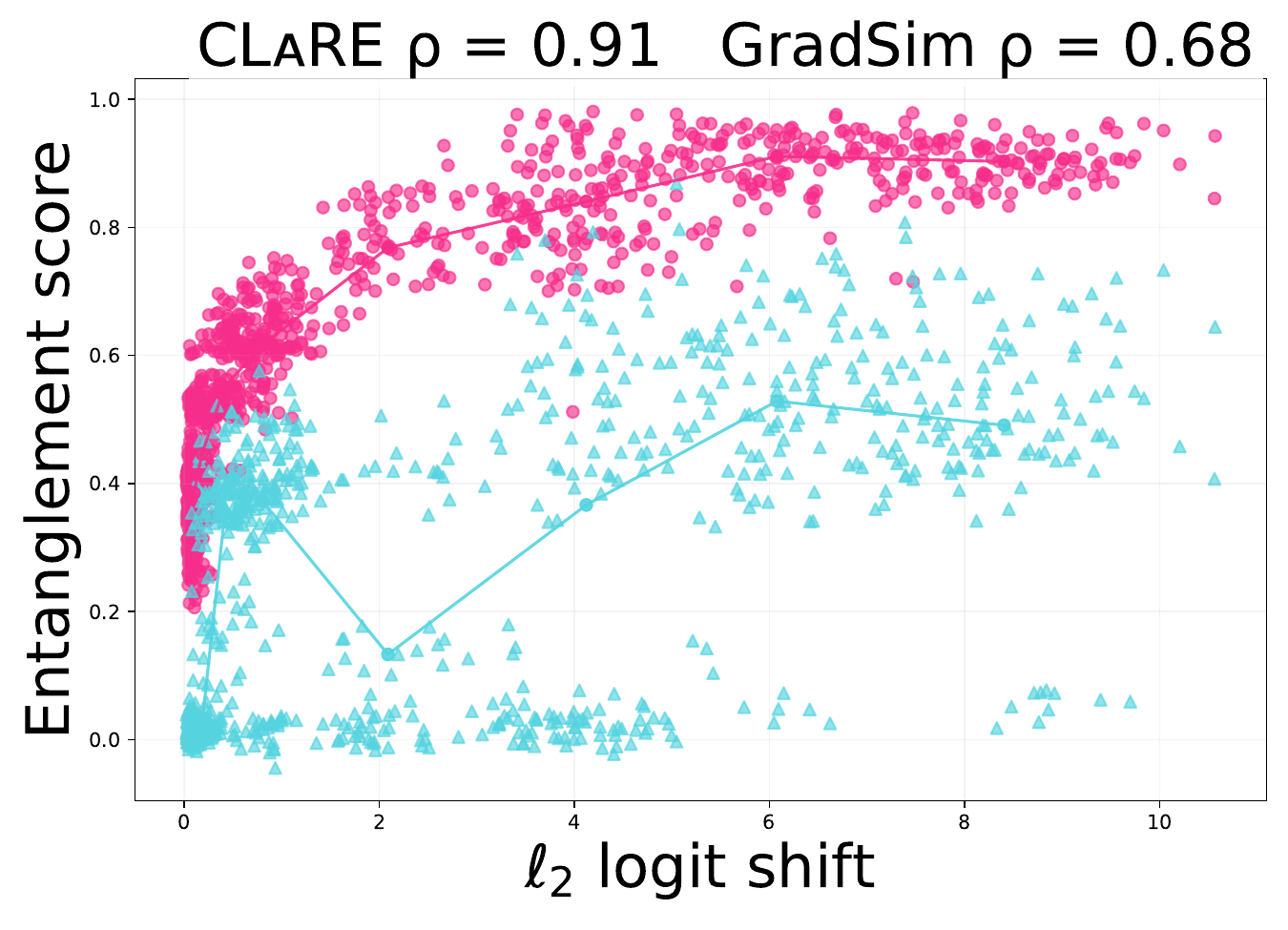}
        \caption{GPT-J}
    \end{subfigure}
    
    \caption{Correlation patterns for MEMIT across different models for entanglement vs $\ell_2$ logit shift.}
    \label{fig:memit_l2}
\end{figure*}

\begin{figure*}[t]
    \centering
    
    \begin{subfigure}[b]{0.32\textwidth}
        \includegraphics[width=\textwidth]{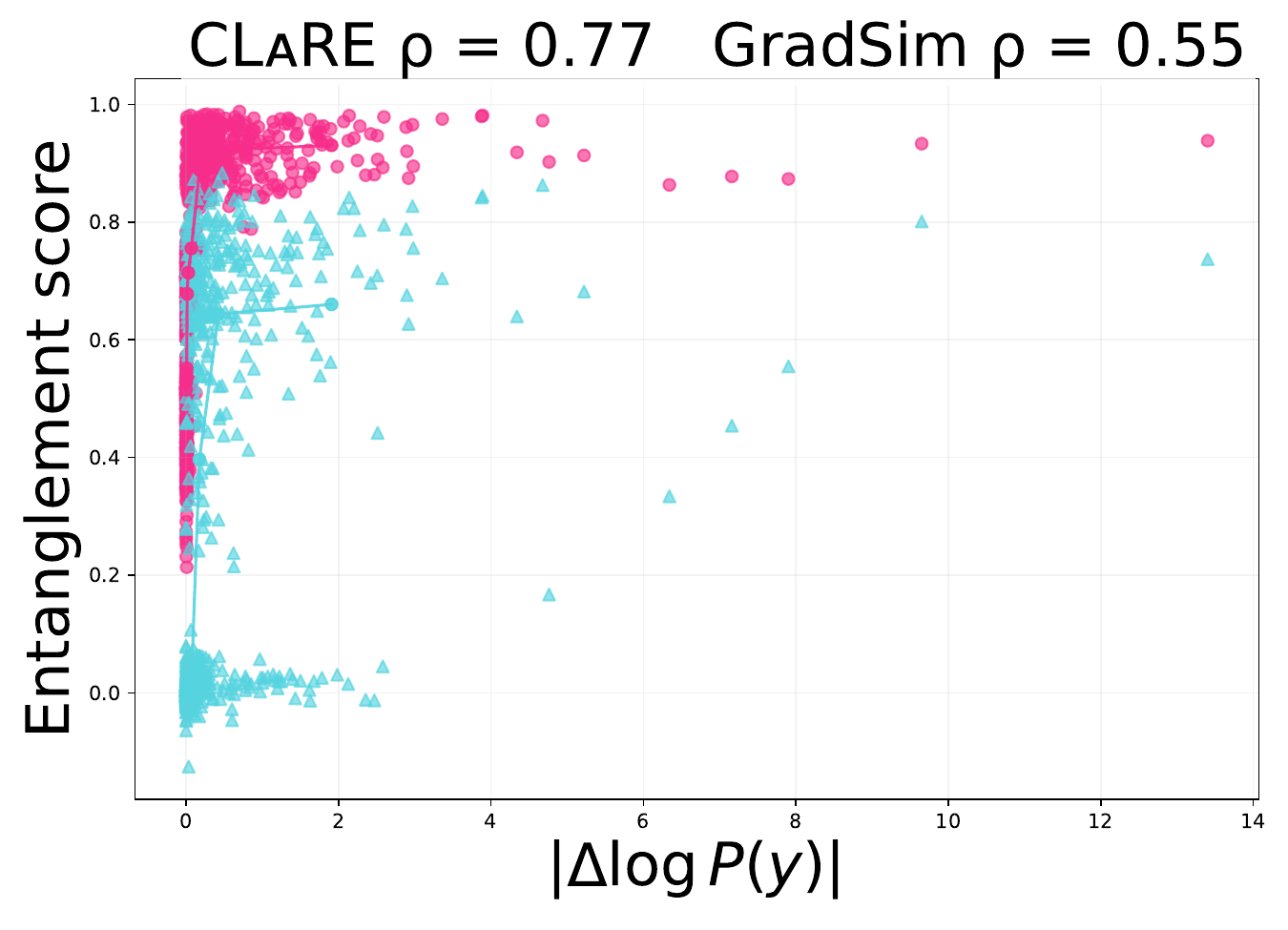}
        \caption{GPT2-XL}
    \end{subfigure}
    \hfill
    \begin{subfigure}[b]{0.32\textwidth}
        \includegraphics[width=\textwidth]{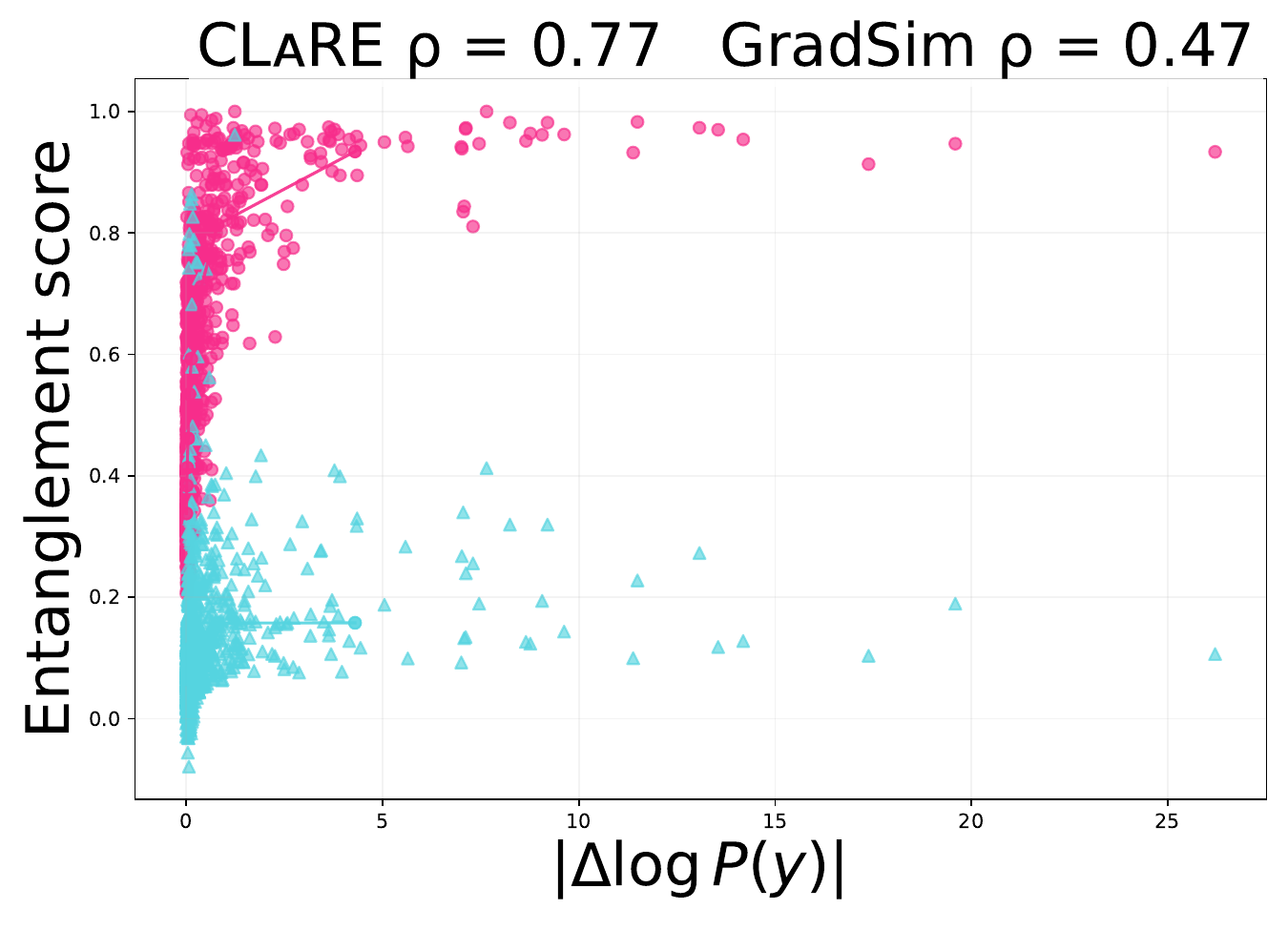}
        \caption{Llama3}
    \end{subfigure}
    \hfill
    \begin{subfigure}[b]{0.32\textwidth}
        \includegraphics[width=\textwidth]{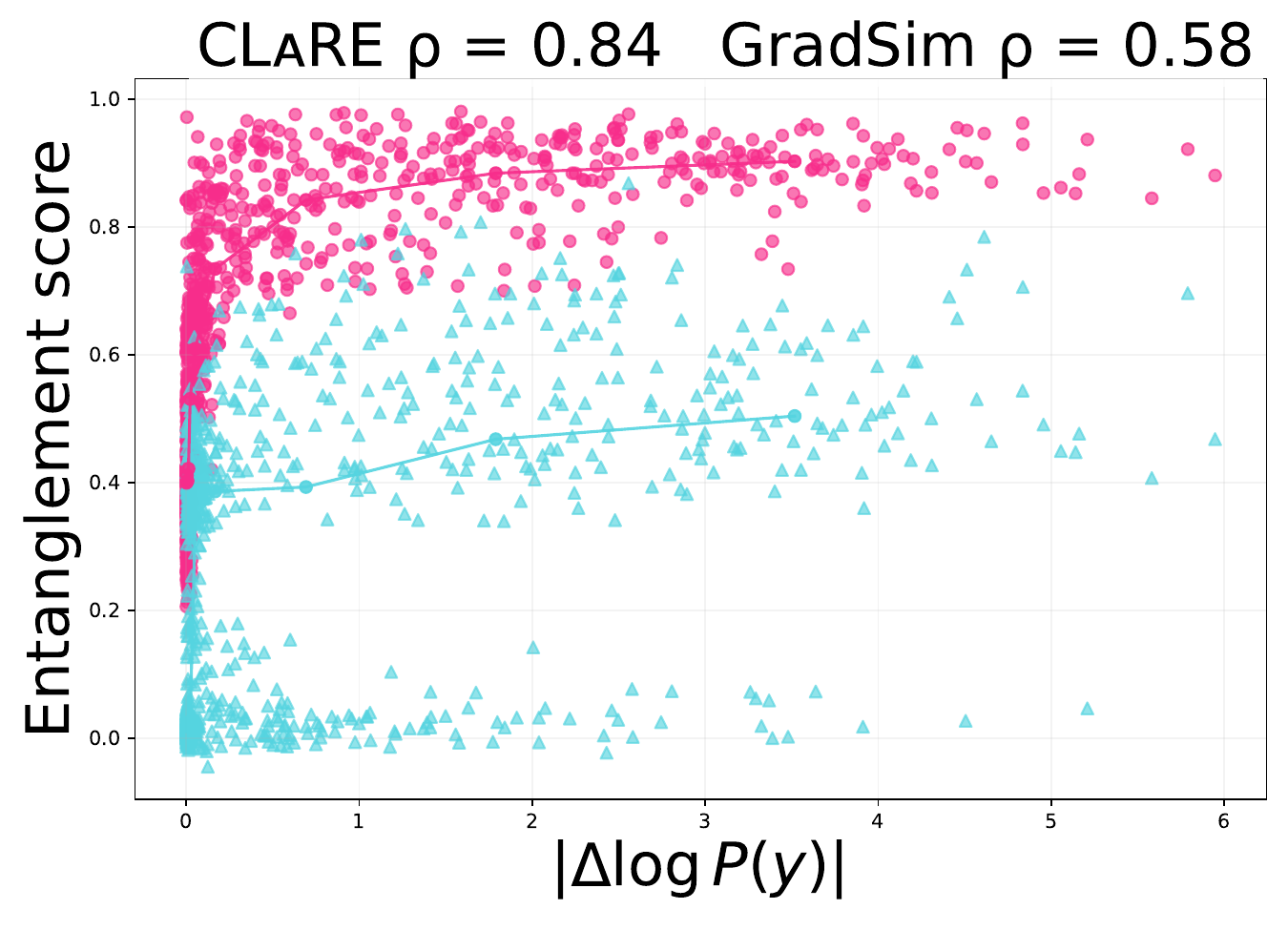}
        \caption{GPT-J}
    \end{subfigure}
    
    \caption{Correlation patterns for MEMIT across different models for entanglement vs $|\Delta \log P(y)|$.}
    \label{fig:memit_delta}
\end{figure*}

\begin{figure*}[t]
    \centering
    
    \begin{subfigure}[b]{0.32\textwidth}
        \includegraphics[width=\textwidth]{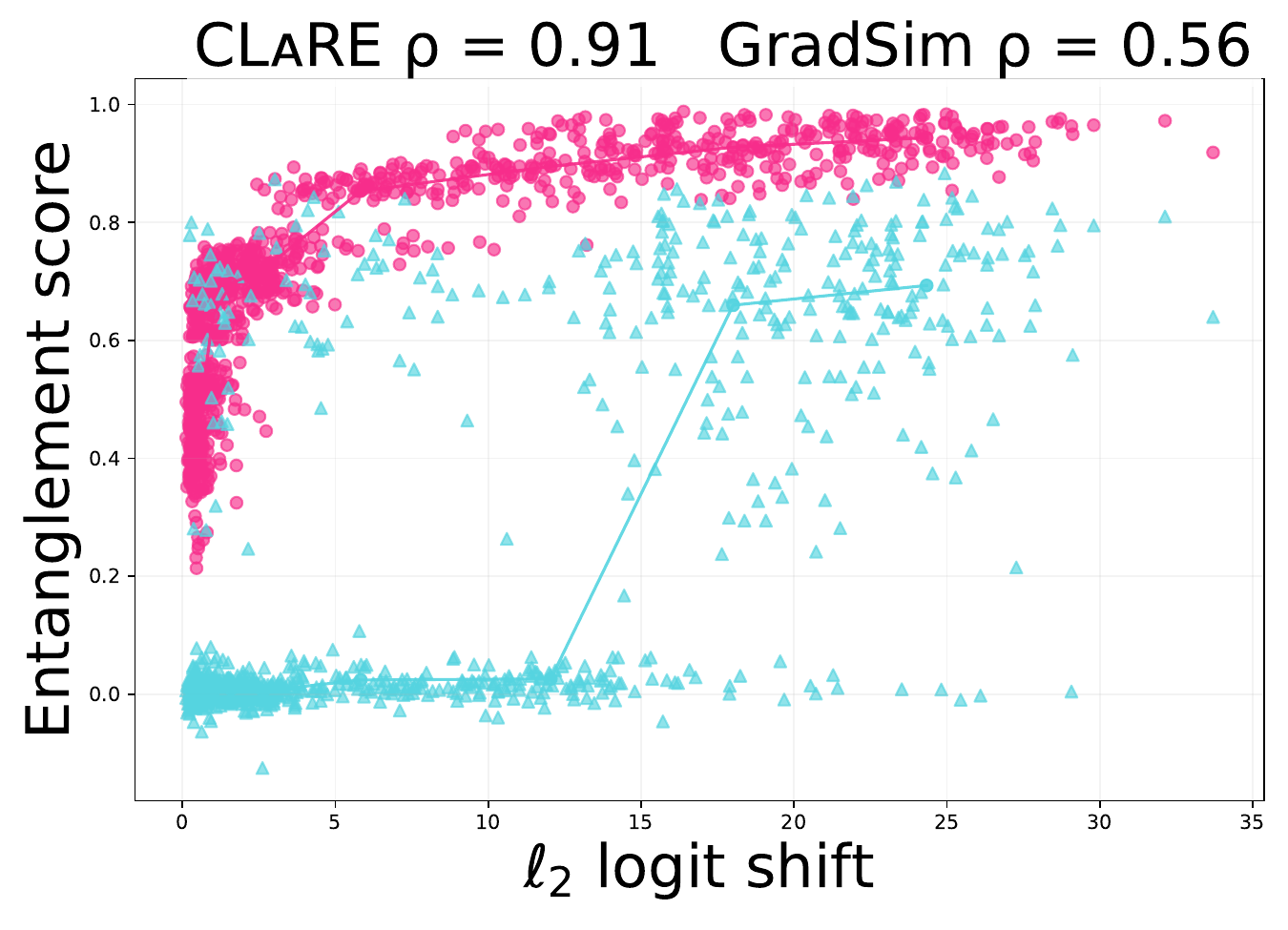}
        \caption{GPT2-XL}
    \end{subfigure}
    \hfill
    \begin{subfigure}[b]{0.32\textwidth}
        \includegraphics[width=\textwidth]{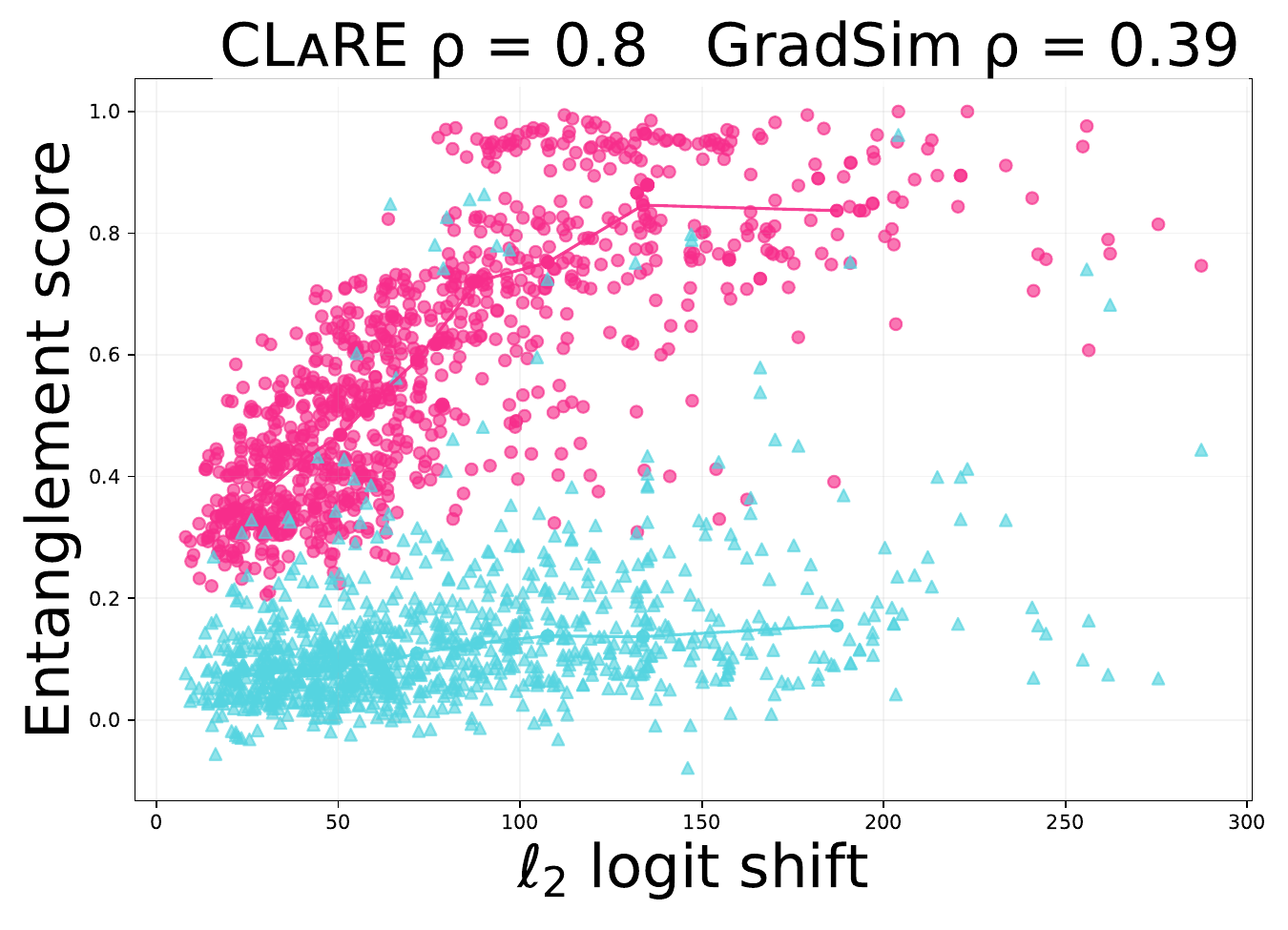}
        \caption{Llama3}
    \end{subfigure}
    \hfill
    \begin{subfigure}[b]{0.32\textwidth}
        \includegraphics[width=\textwidth]{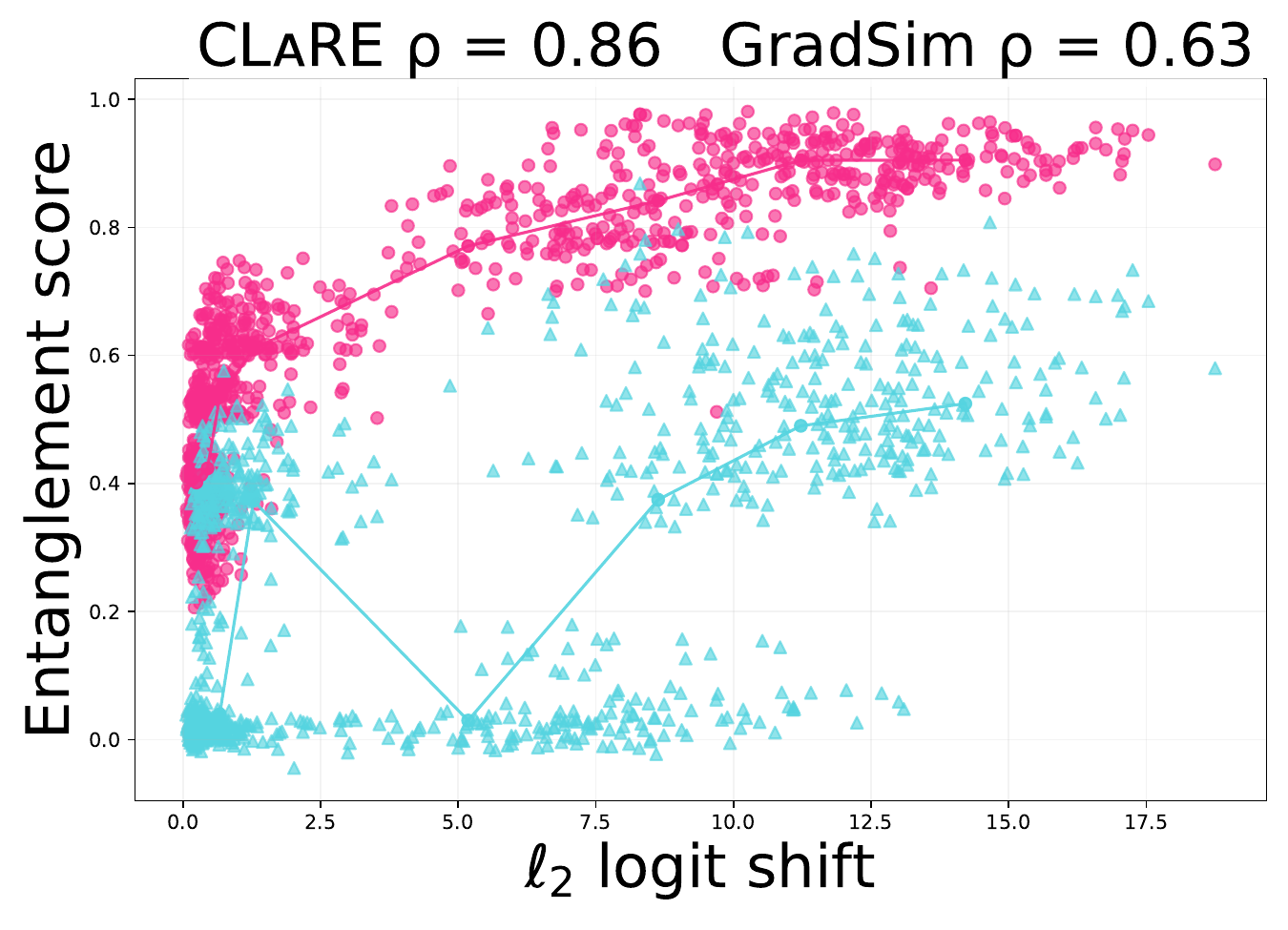}
        \caption{GPT-J}
    \end{subfigure}
    
    \caption{Correlation patterns for ROME across different models for entanglement vs $\ell_2$ logit shift.}
    \label{fig:rome_l2}
\end{figure*}

\begin{figure*}[t]
    \centering
    
    \begin{subfigure}[b]{0.32\textwidth}
        \includegraphics[width=\textwidth]{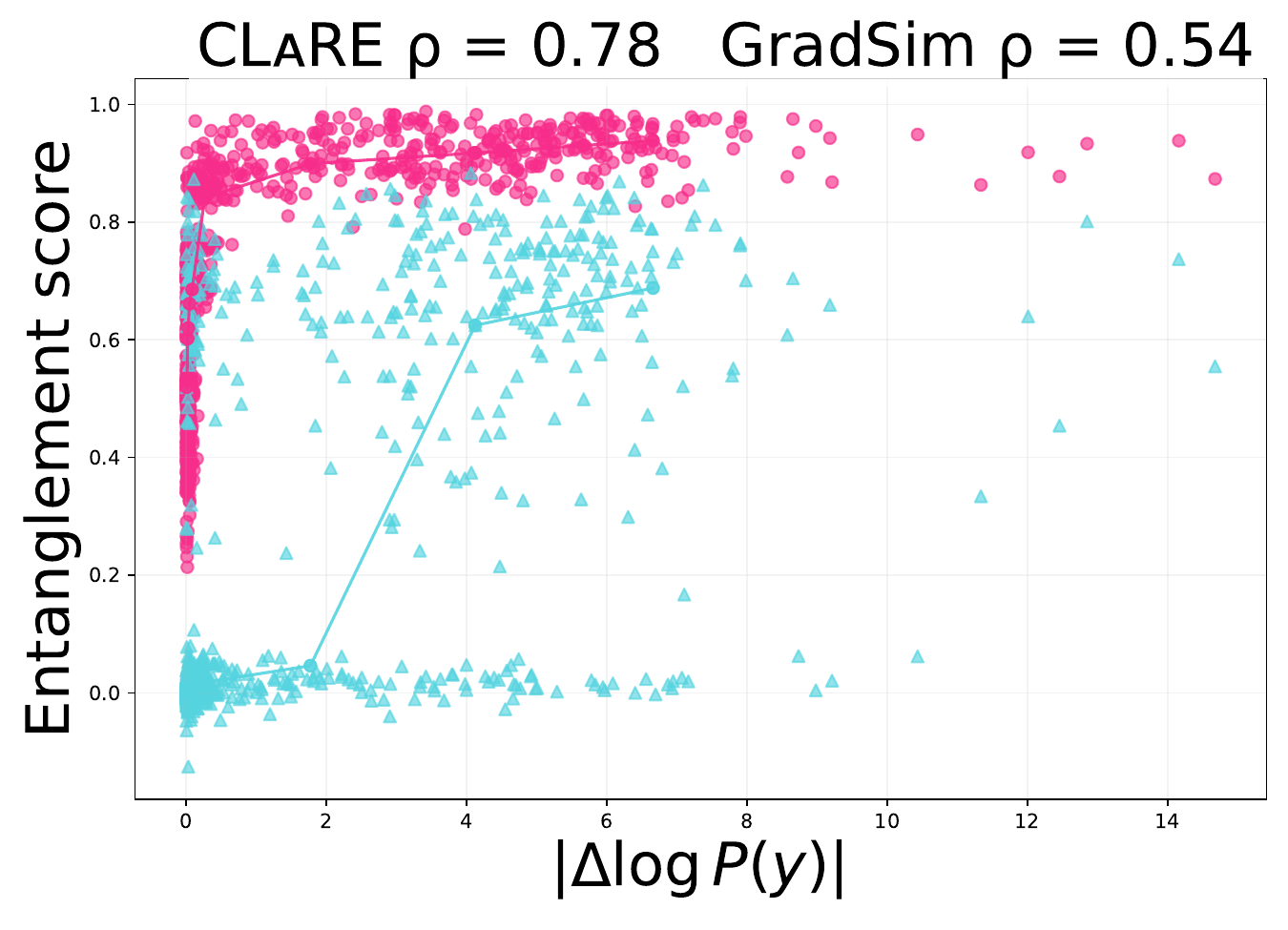}
        \caption{GPT2-XL}
    \end{subfigure}
    \hfill
    \begin{subfigure}[b]{0.32\textwidth}
        \includegraphics[width=\textwidth]{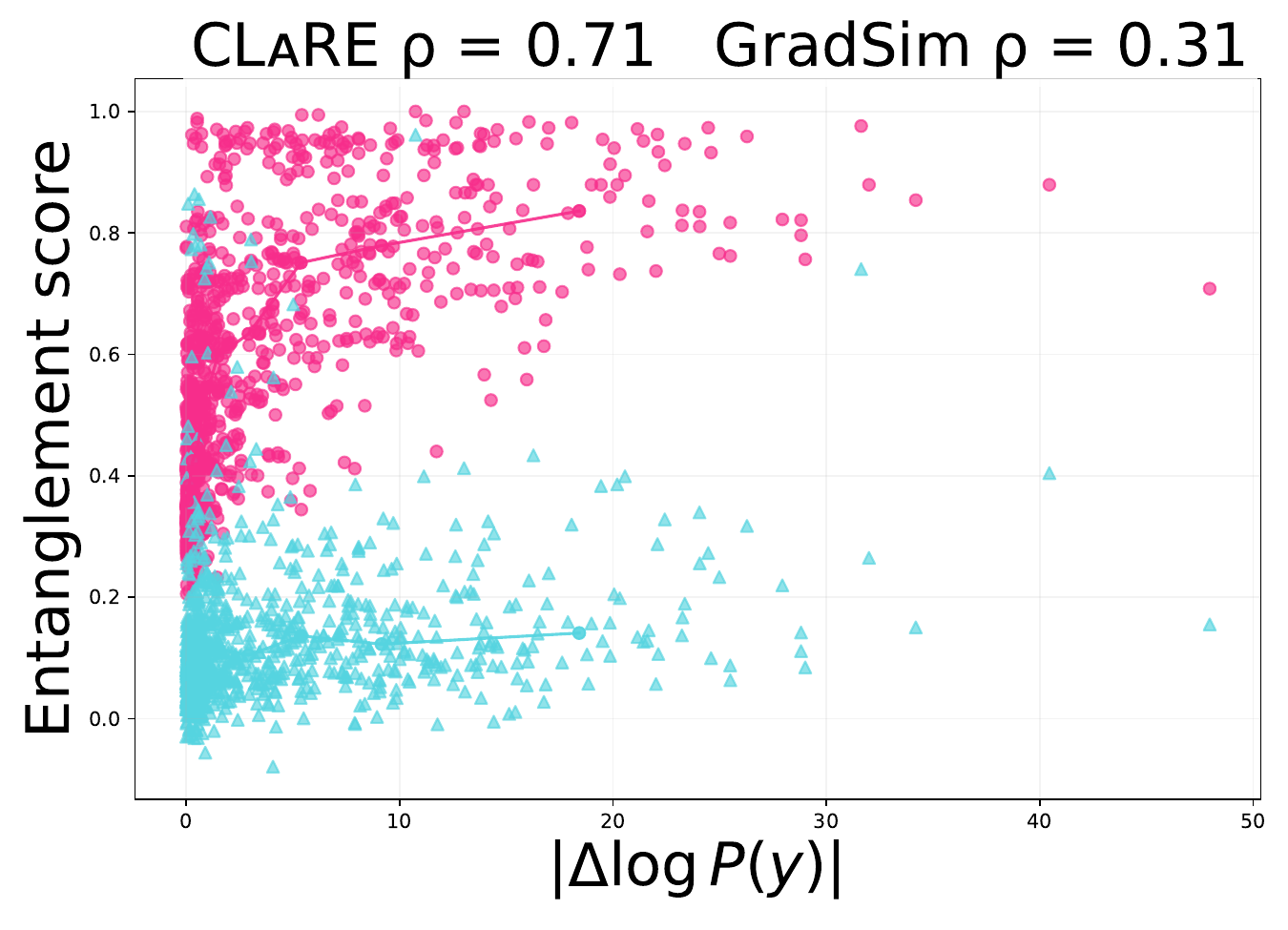}
        \caption{Llama3}
    \end{subfigure}
    \hfill
    \begin{subfigure}[b]{0.32\textwidth}
        \includegraphics[width=\textwidth]{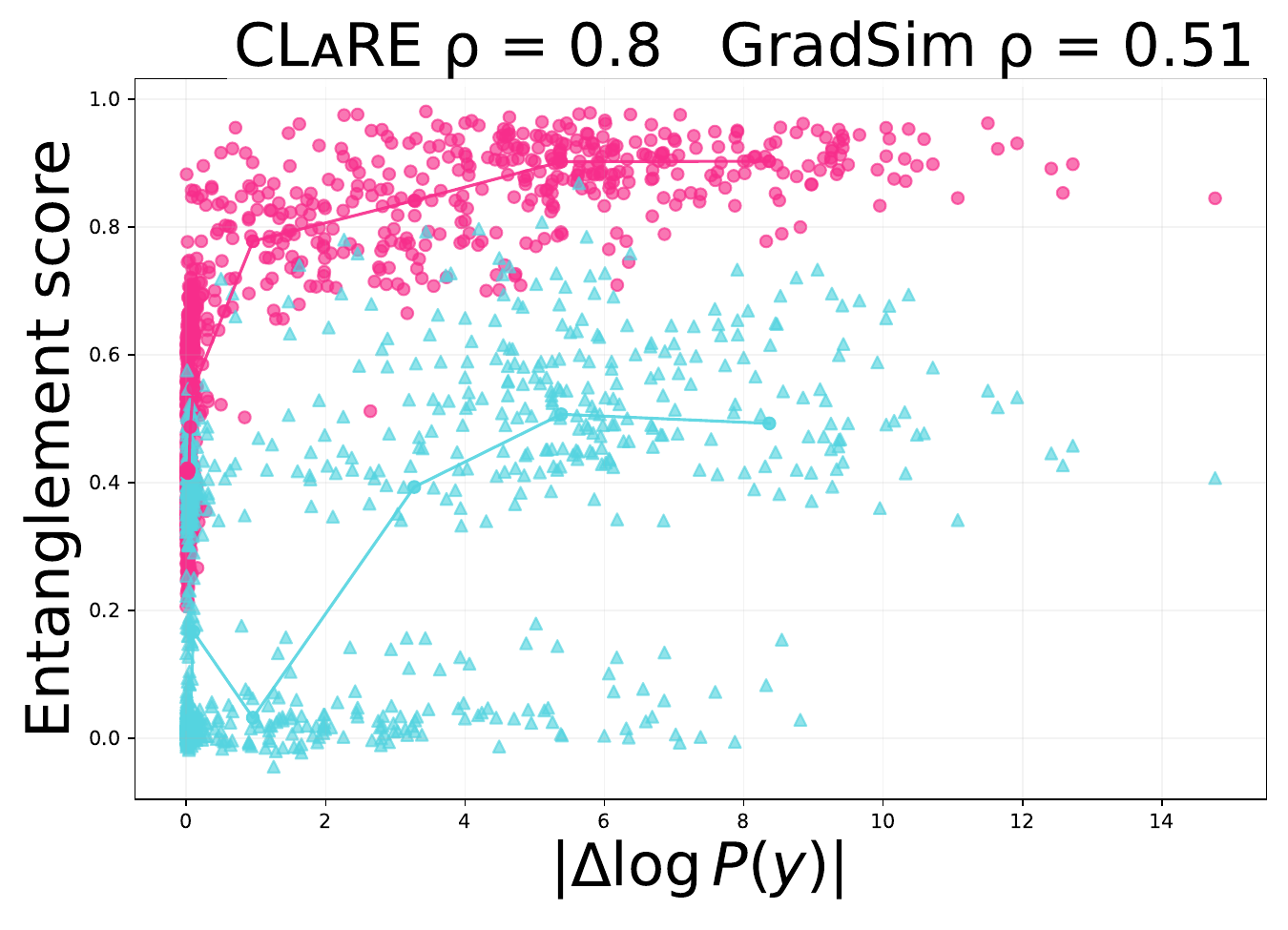}
        \caption{GPT-J}
    \end{subfigure}
    
    \caption{Correlation patterns for ROME across different models for entanglement vs $|\Delta \log P(y)|$.}
    \label{fig:rome_delta}
\end{figure*}

\begin{figure*}[t]
    \centering
    
    \begin{subfigure}[b]{0.32\textwidth}
        \includegraphics[width=\textwidth]{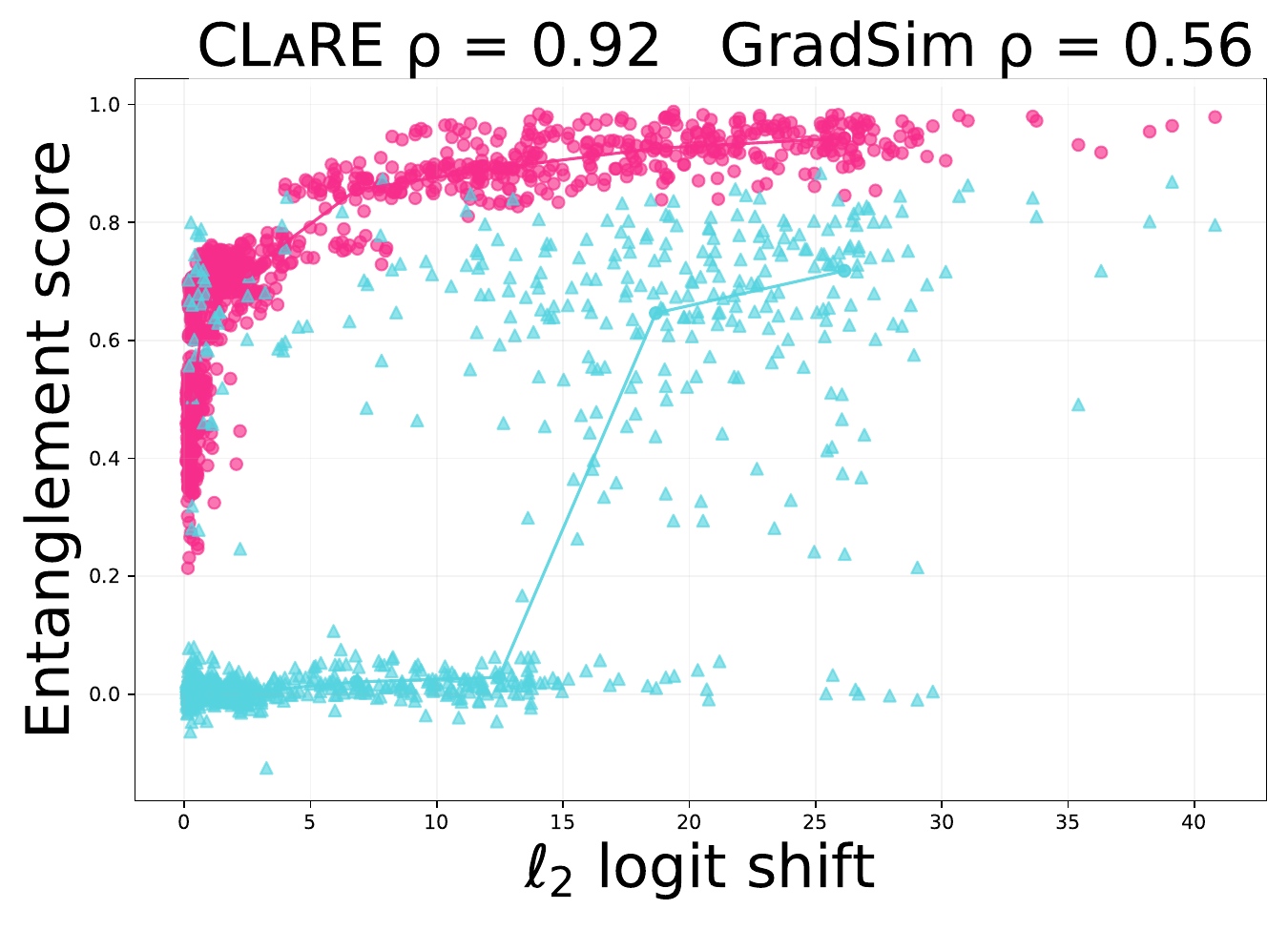}
        \caption{GPT2-XL}
    \end{subfigure}
    \hfill
    \begin{subfigure}[b]{0.32\textwidth}
        \includegraphics[width=\textwidth]{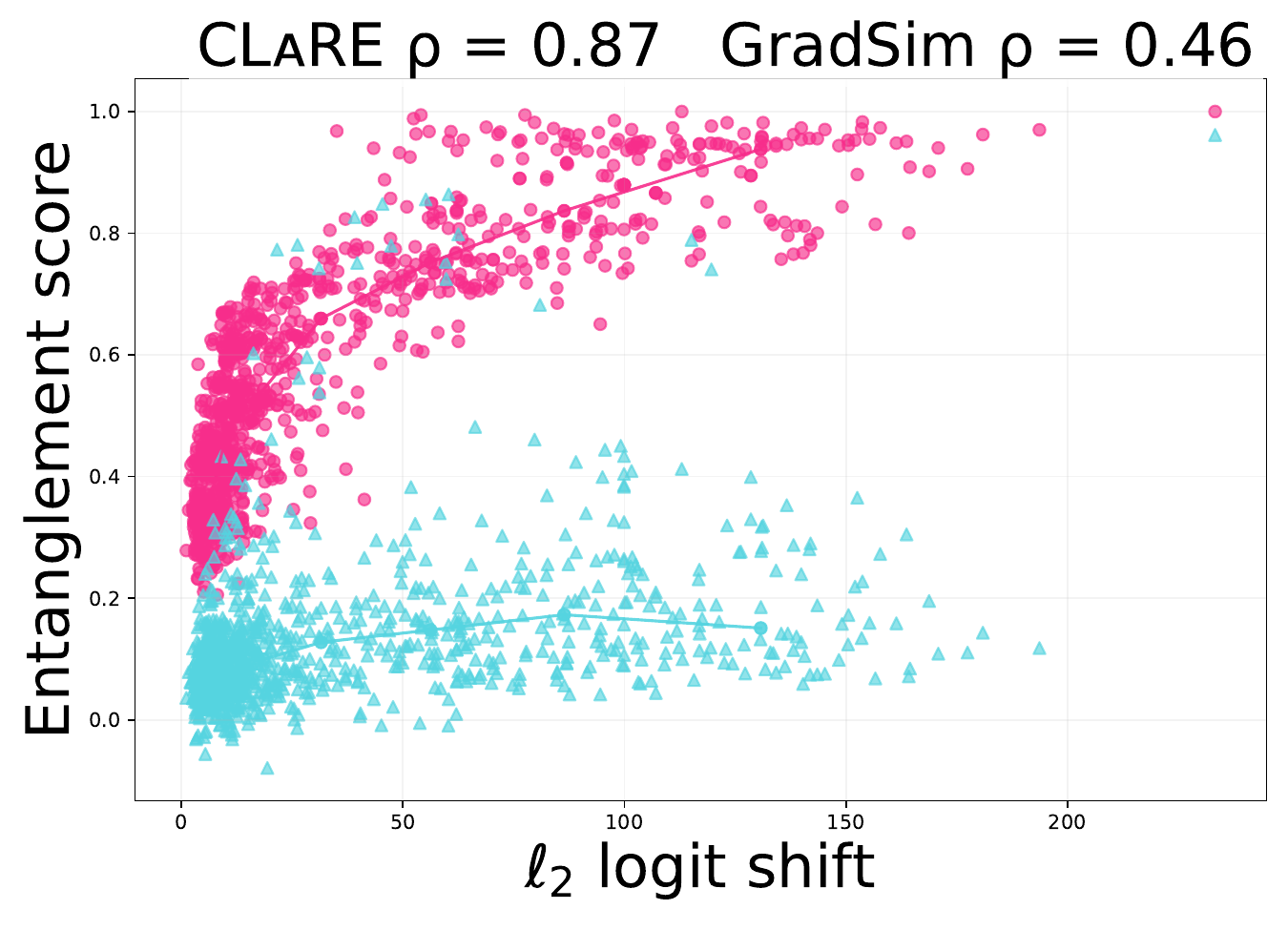}
        \caption{Llama3}
    \end{subfigure}
    \hfill
    \begin{subfigure}[b]{0.32\textwidth}
        \includegraphics[width=\textwidth]{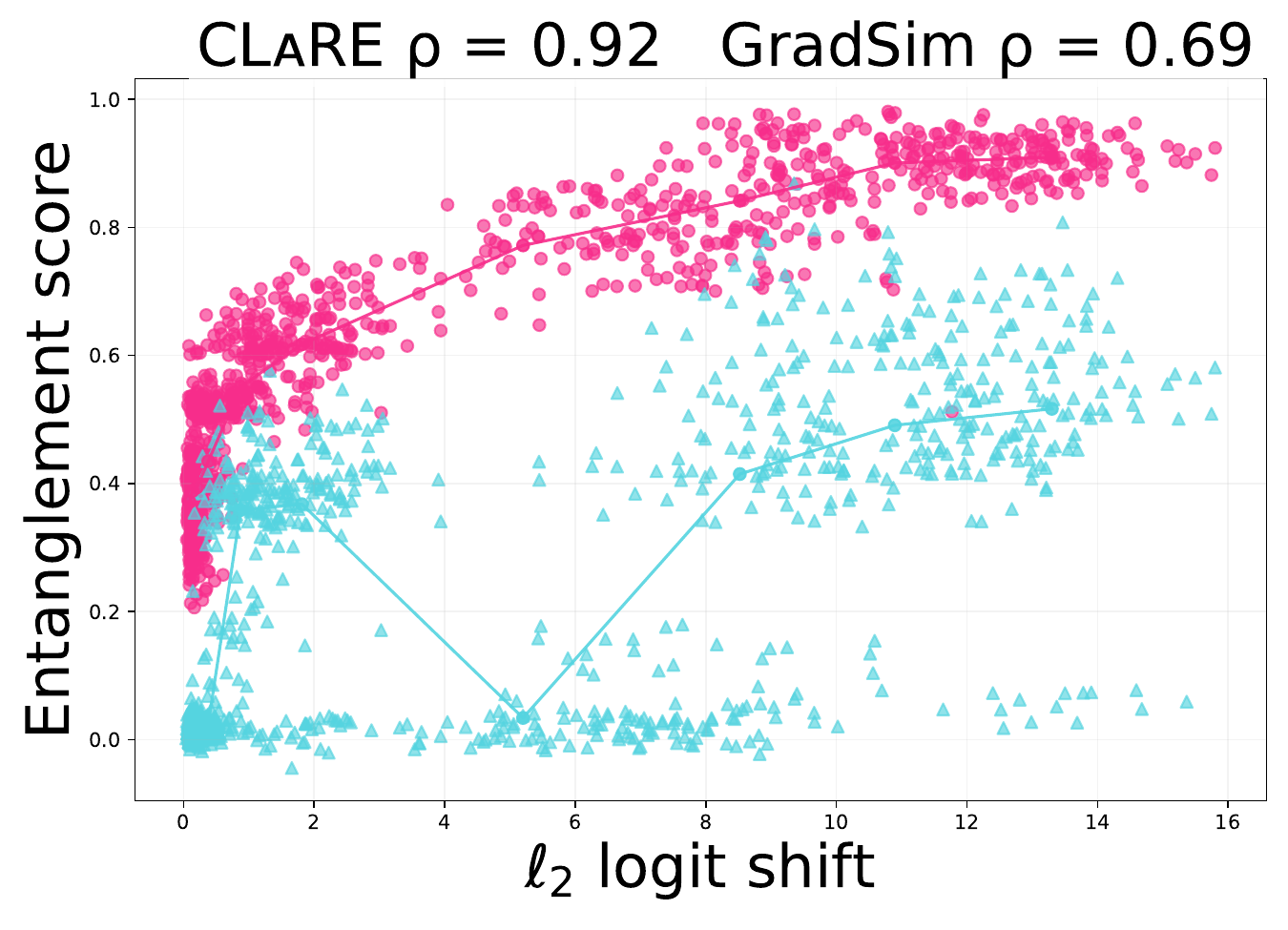}
        \caption{GPT-J}
    \end{subfigure}
    
    \caption{Correlation patterns for PRUNE across different models for entanglement vs $\ell_2$ logit shift.}
    \label{fig:prune_l2}
\end{figure*}

\begin{figure*}[t]
    \centering
    
    \begin{subfigure}[b]{0.32\textwidth}
        \includegraphics[width=\textwidth]{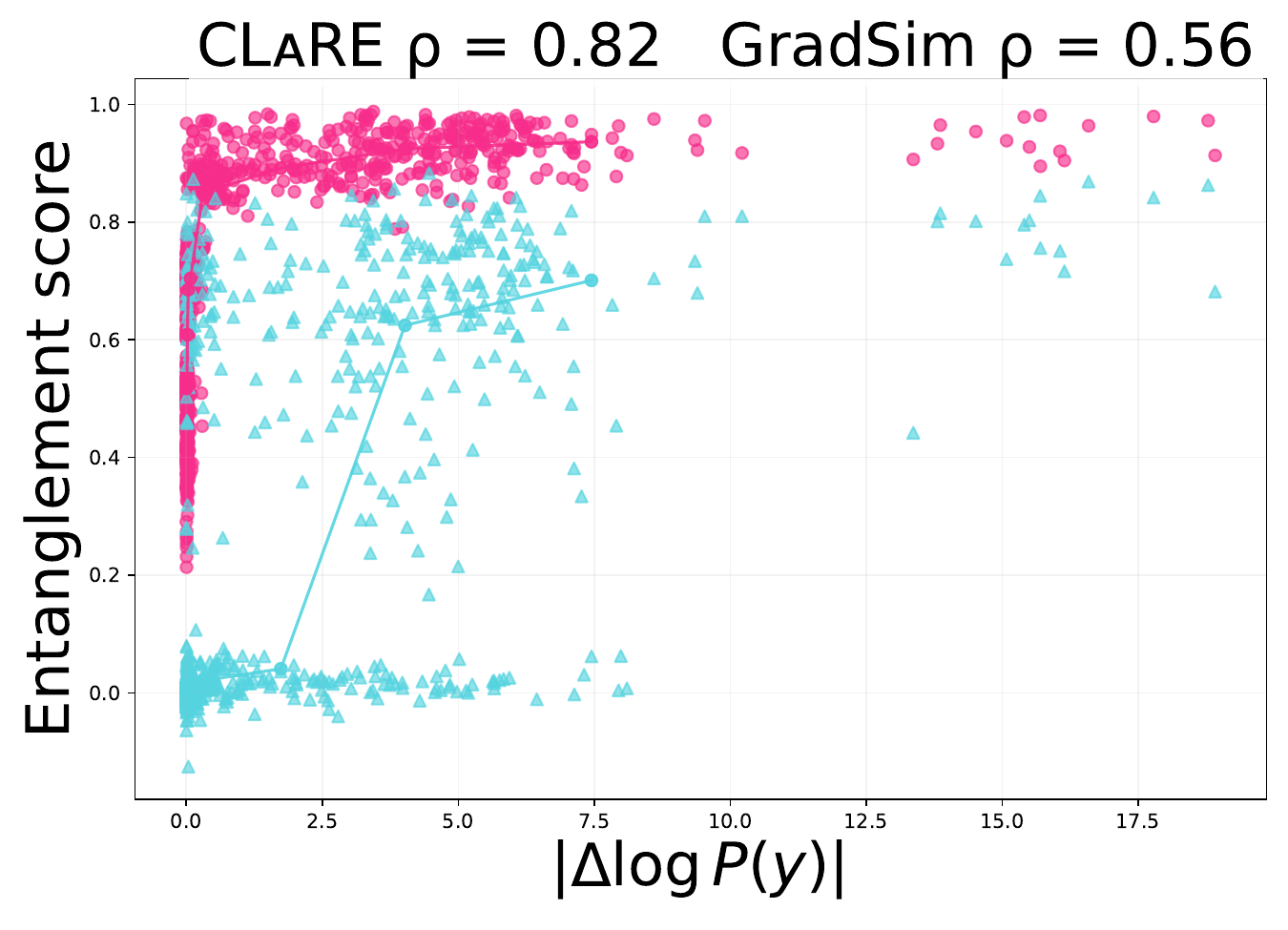}
        \caption{GPT2-XL}
    \end{subfigure}
    \hfill
    \begin{subfigure}[b]{0.32\textwidth}
        \includegraphics[width=\textwidth]{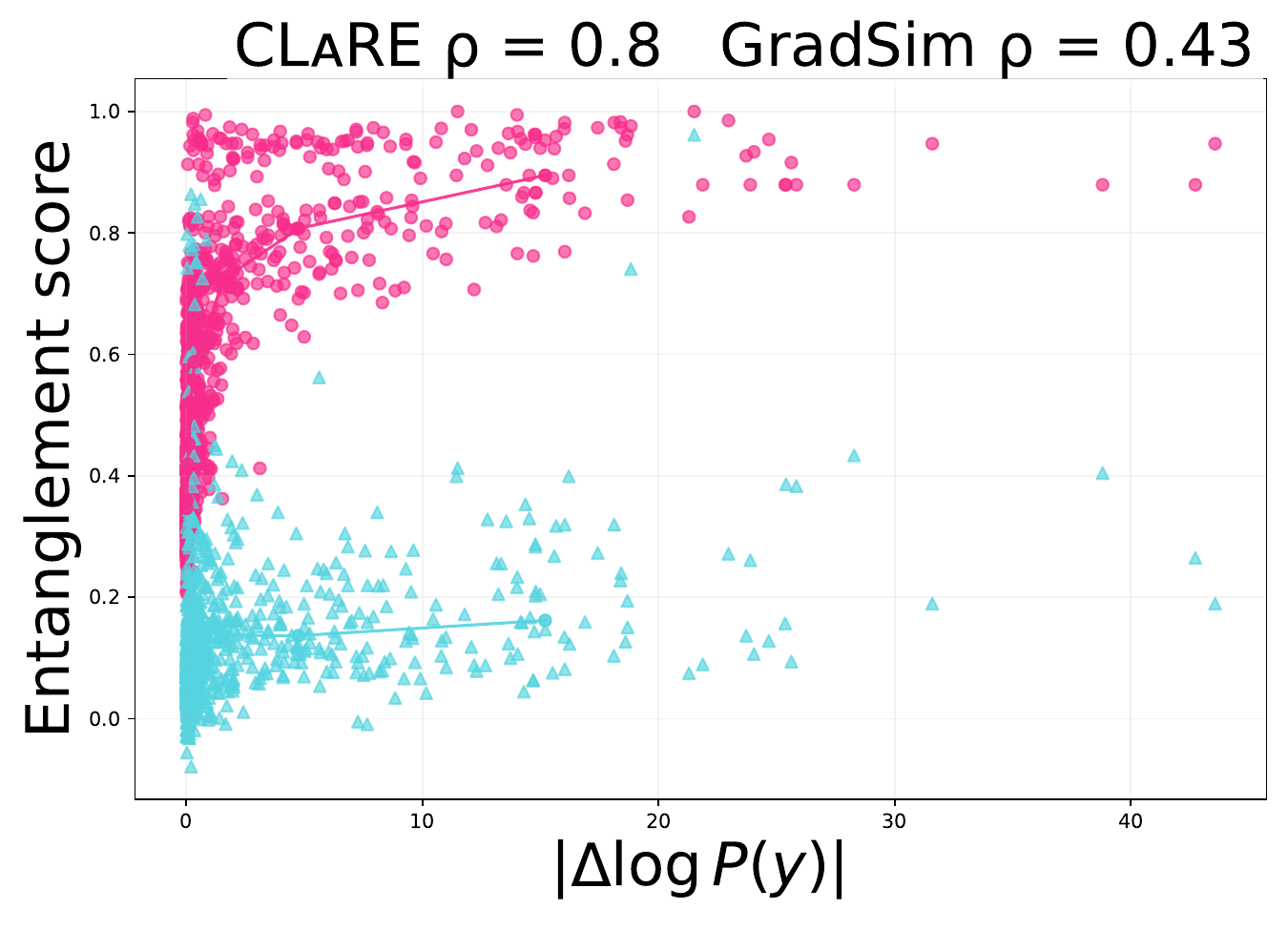}
        \caption{Llama3}
    \end{subfigure}
    \hfill
    \begin{subfigure}[b]{0.32\textwidth}
        \includegraphics[width=\textwidth]{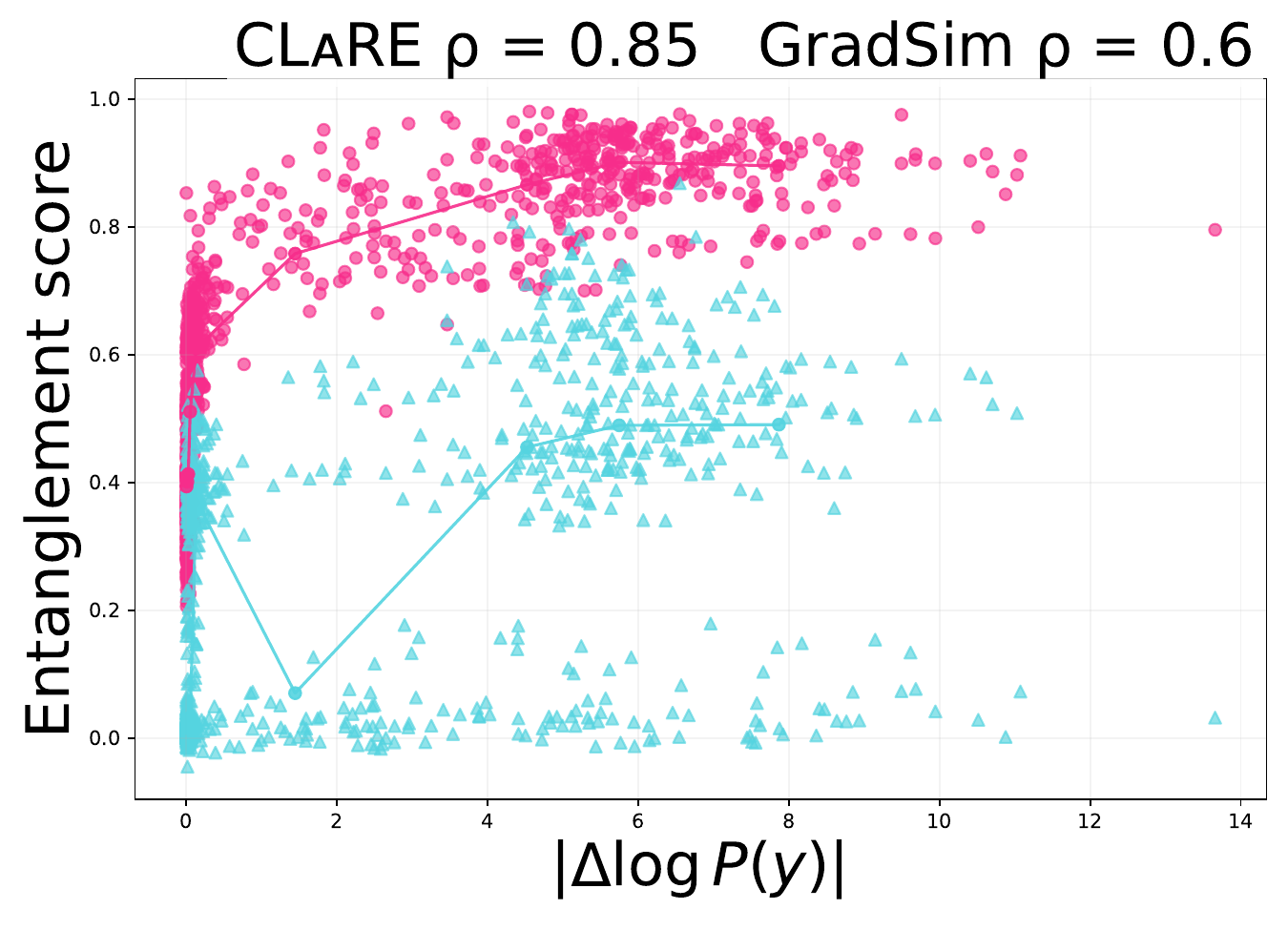}
        \caption{GPT-J}
    \end{subfigure}
    
    \caption{Correlation patterns for PRUNE across different models for entanglement vs $|\Delta \log P(y)|$.}
    \label{fig:prune_delta}
\end{figure*}

\begin{figure*}[t]
    \centering
    
    \begin{subfigure}[b]{0.32\textwidth}
        \includegraphics[width=\textwidth]{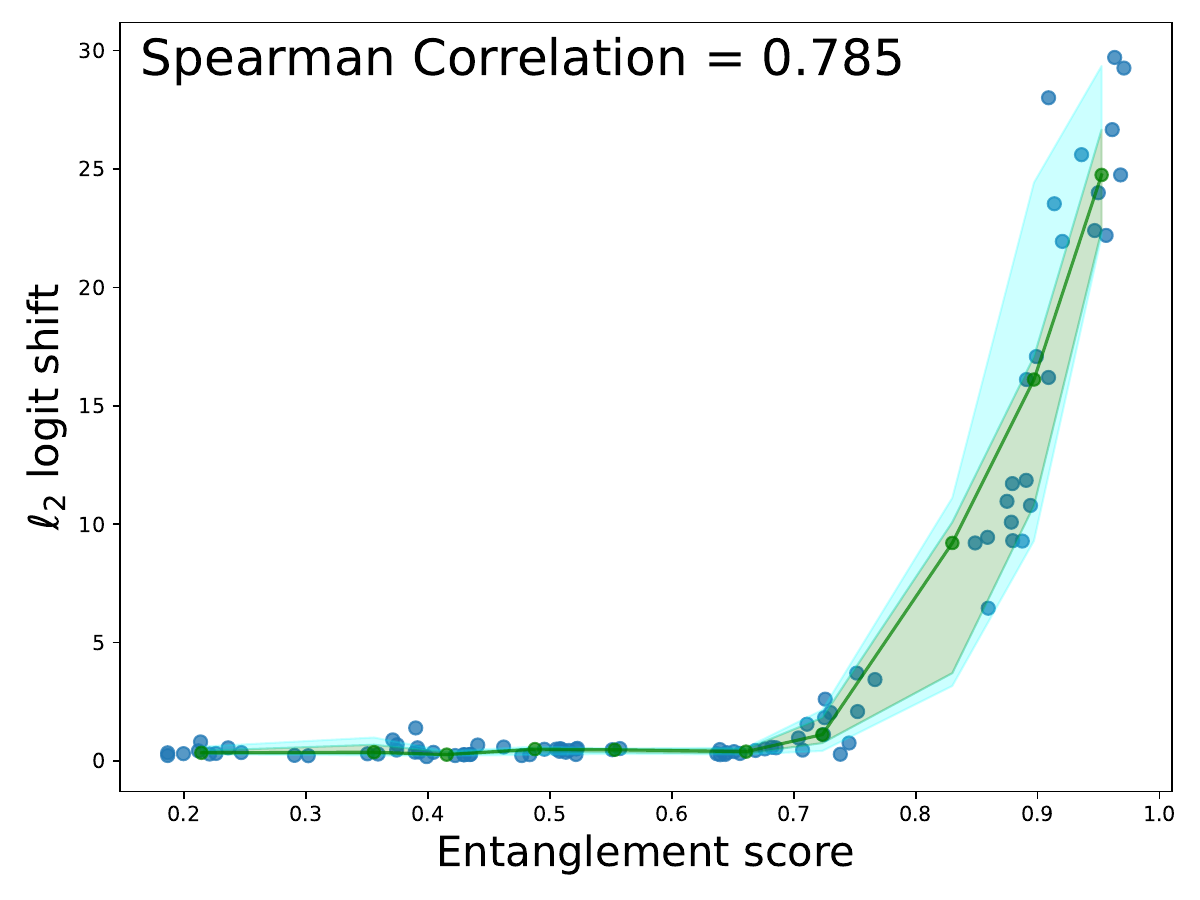}
        \caption{GPT2-XL}
    \end{subfigure}
    \hfill
    \begin{subfigure}[b]{0.32\textwidth}
        \includegraphics[width=\textwidth]{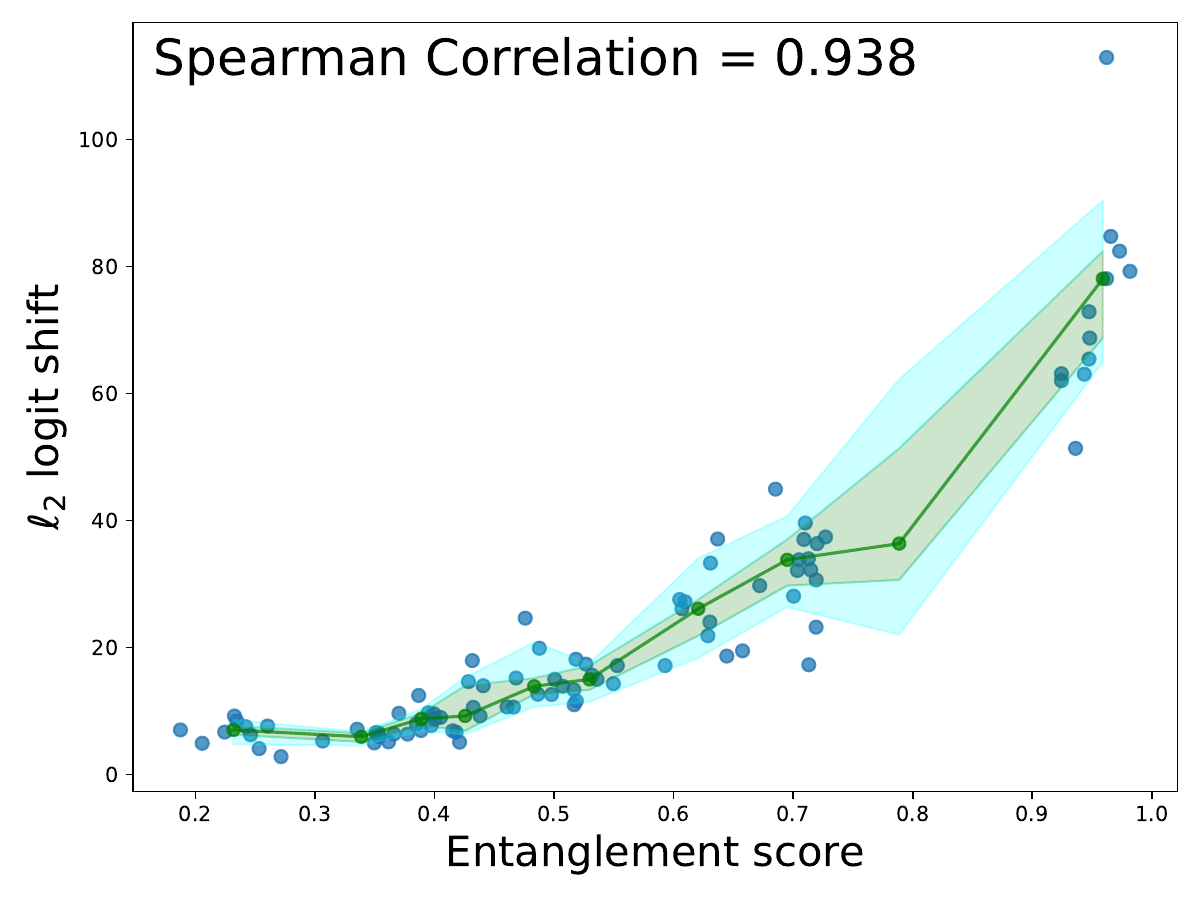}
        \caption{Llama3}
    \end{subfigure}
    \hfill
    \begin{subfigure}[b]{0.32\textwidth}
        \includegraphics[width=\textwidth]{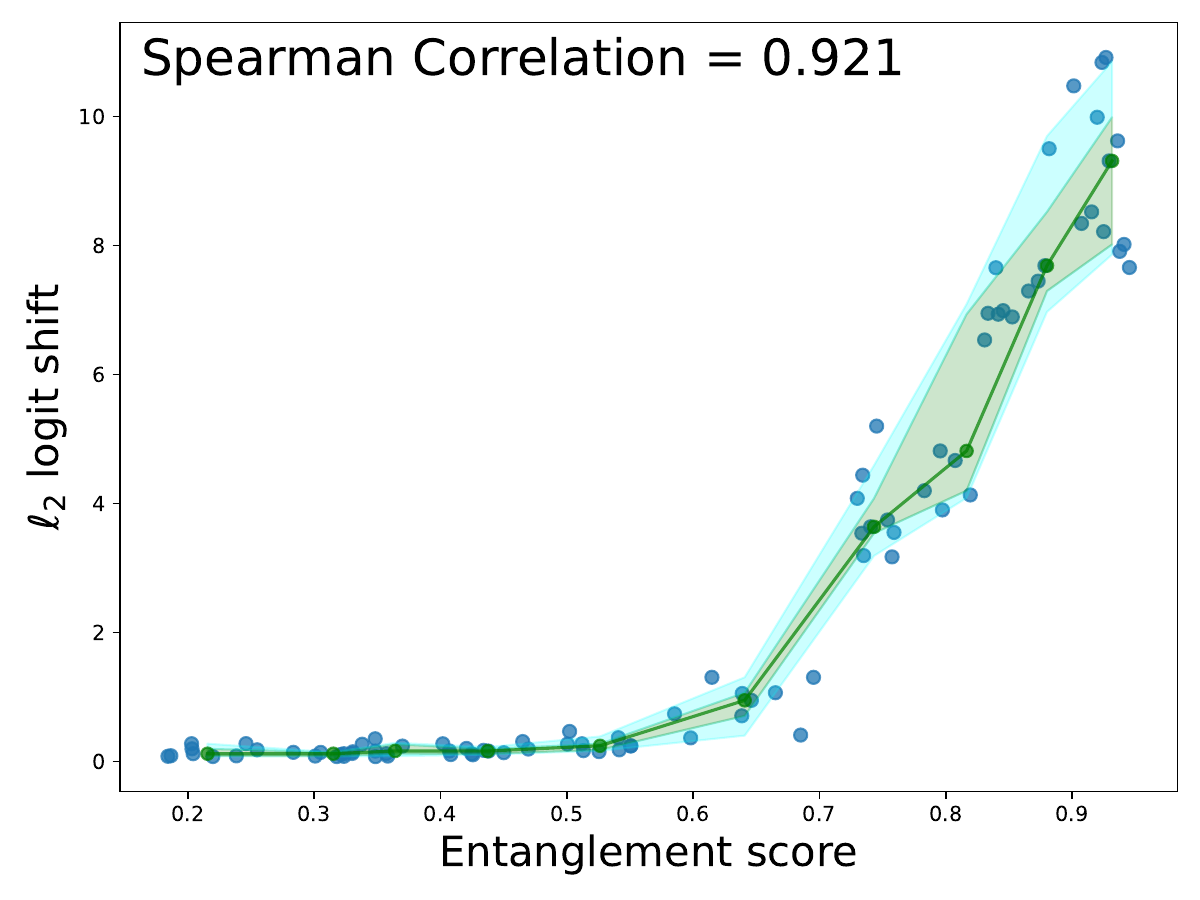}
        \caption{GPT-J}
    \end{subfigure}
    
    \caption{\Clare correlation patterns for AlphaEdit across different models for entanglement vs $\ell_2$ logit shift.}
    \label{fig:alphaedit_l2_clare}
\end{figure*}

\begin{figure*}[t]
    \centering
    
    \begin{subfigure}[b]{0.32\textwidth}
        \includegraphics[width=\textwidth]{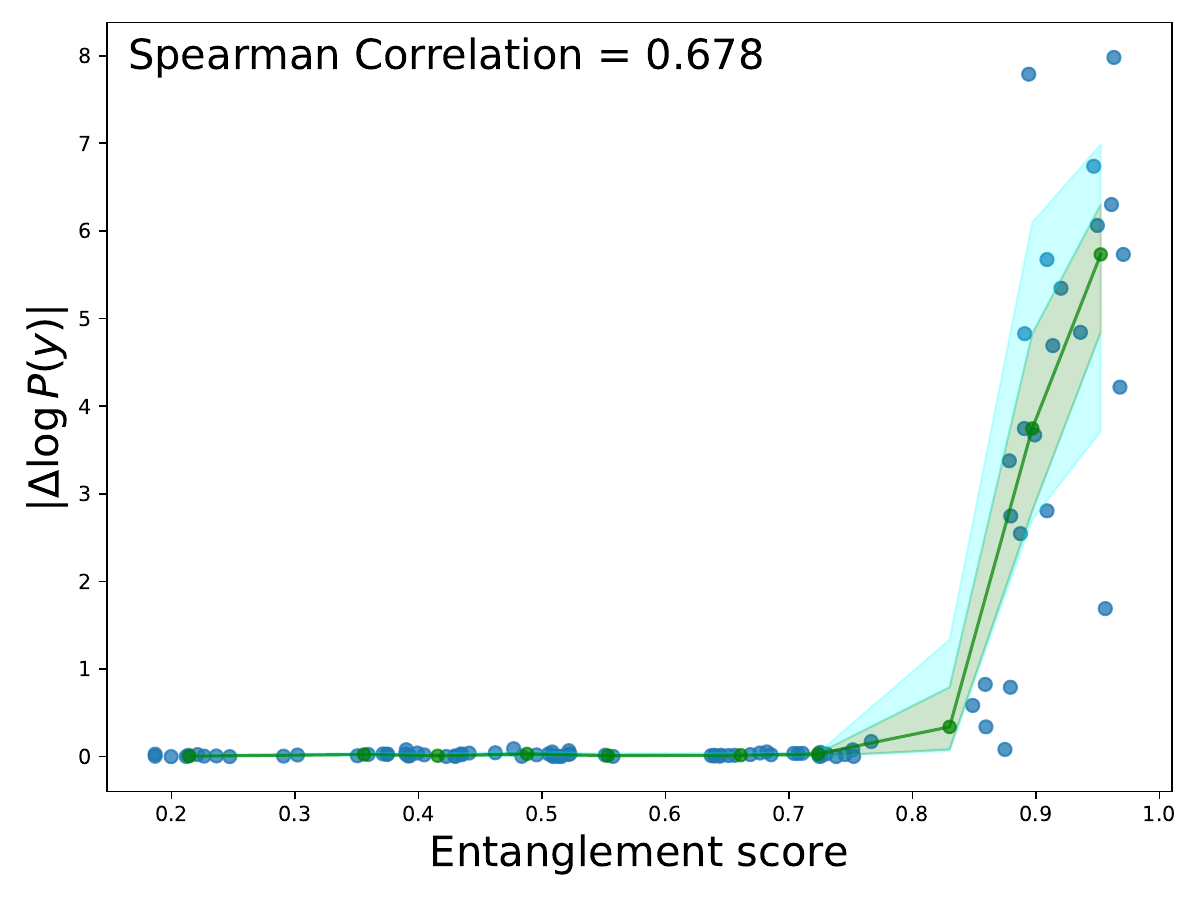}
        \caption{GPT2-XL}
    \end{subfigure}
    \hfill
    \begin{subfigure}[b]{0.32\textwidth}
        \includegraphics[width=\textwidth]{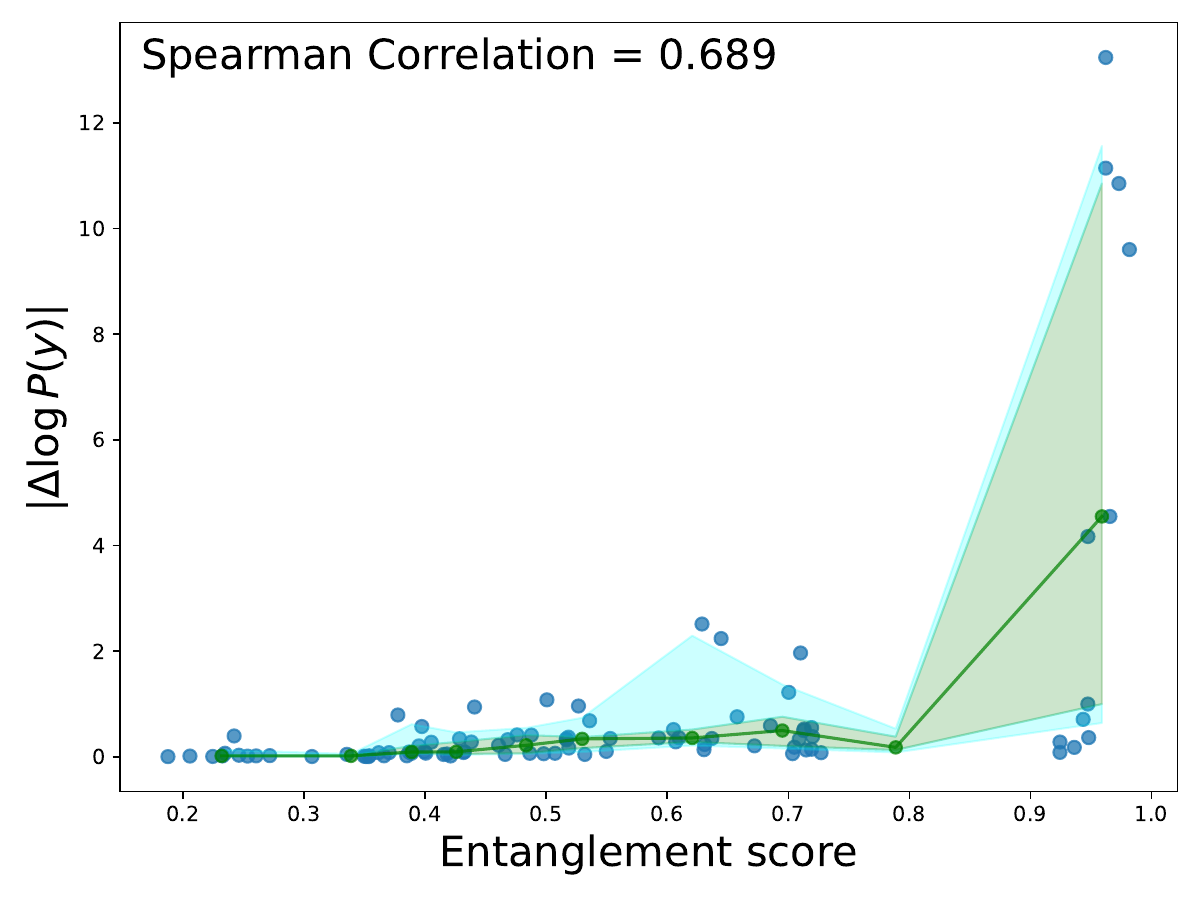}
        \caption{Llama3}
    \end{subfigure}
    \hfill
    \begin{subfigure}[b]{0.32\textwidth}
        \includegraphics[width=\textwidth]{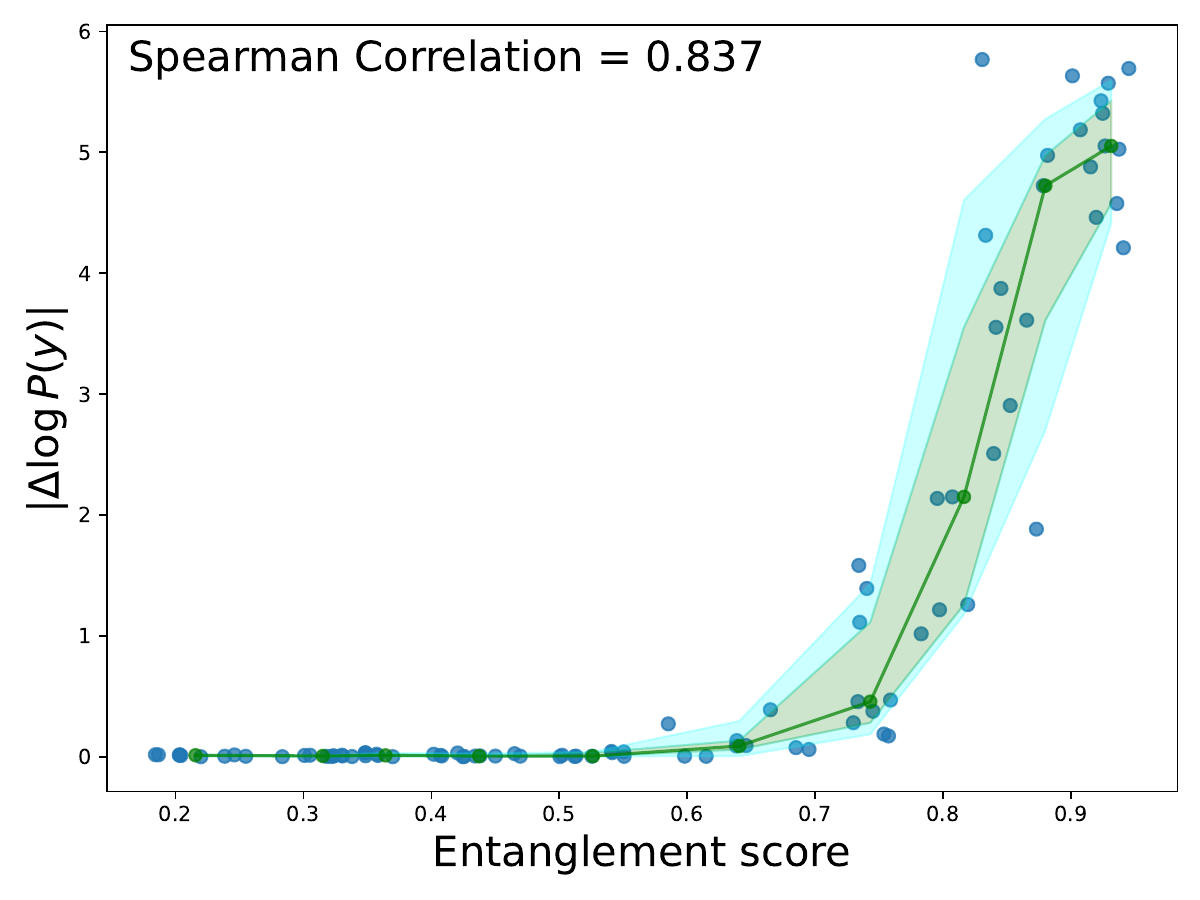}
        \caption{GPT-J}
    \end{subfigure}
    
    \caption{\Clare correlation patterns for AlphaEdit across different models for entanglement vs $|\Delta \log P(y)|$.}
    \label{fig:alphaedit_delta_clare}
\end{figure*}

\begin{figure*}[t]
    \centering
    
    \begin{subfigure}[b]{0.32\textwidth}
        \includegraphics[width=\textwidth]{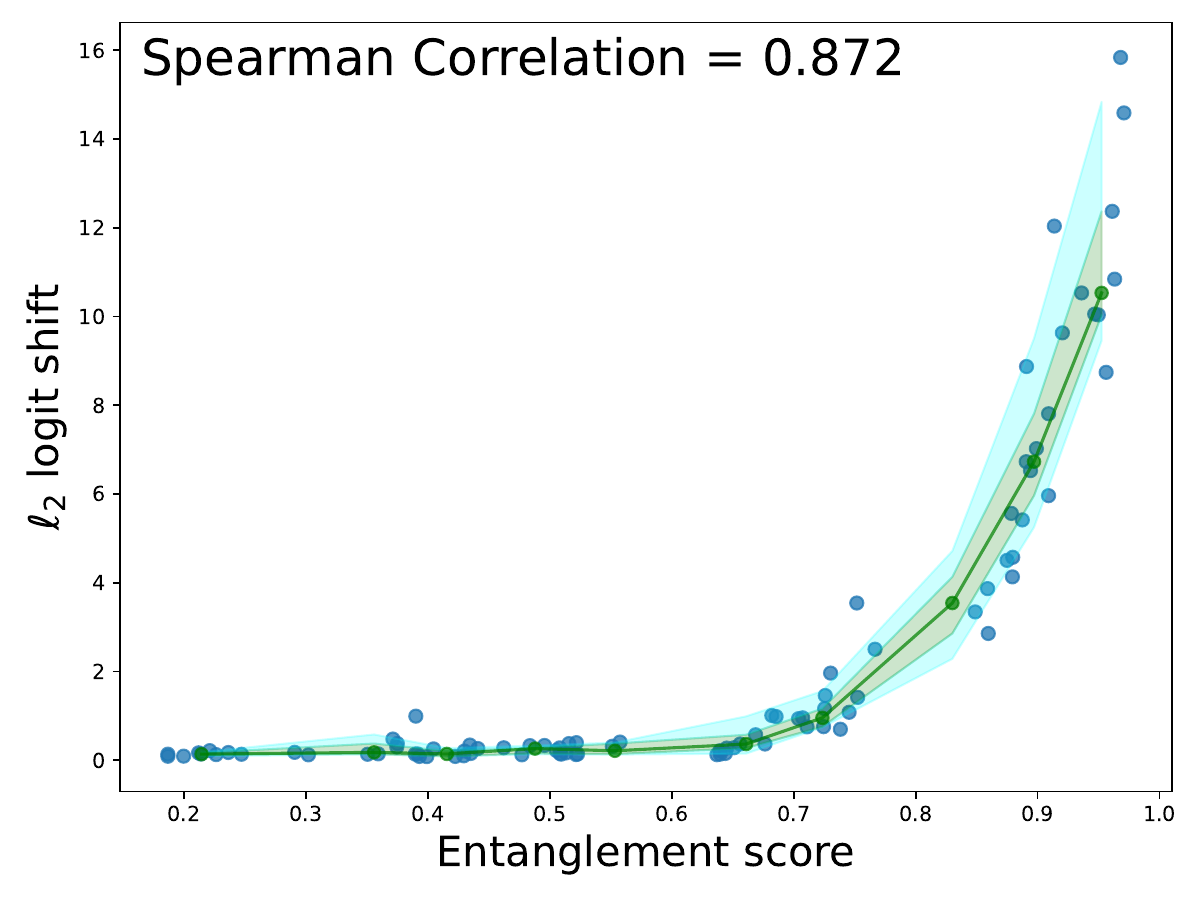}
        \caption{GPT2-XL}
    \end{subfigure}
    \hfill
    \begin{subfigure}[b]{0.32\textwidth}
        \includegraphics[width=\textwidth]{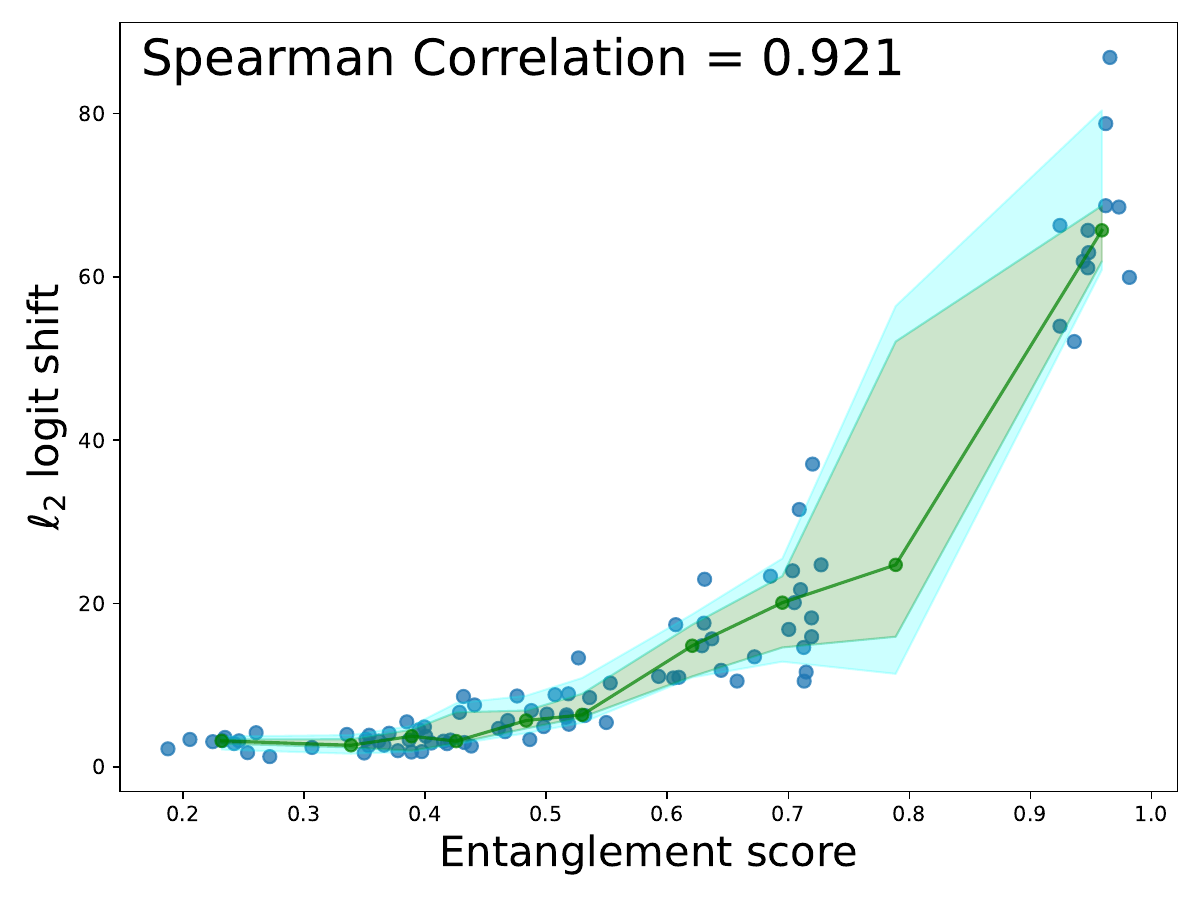}
        \caption{Llama3}
    \end{subfigure}
    \hfill
    \begin{subfigure}[b]{0.32\textwidth}
        \includegraphics[width=\textwidth]{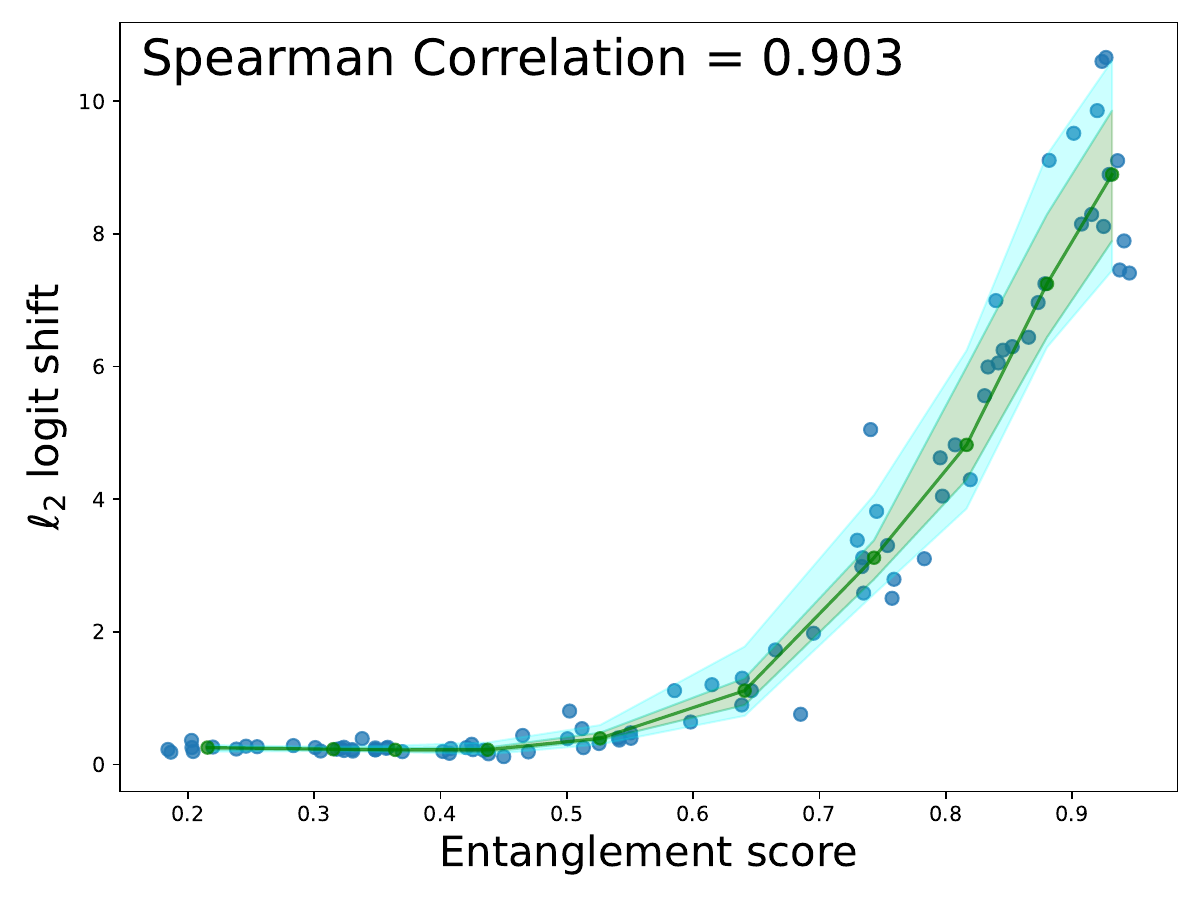}
        \caption{GPT-J}
    \end{subfigure}
    
    \caption{\Clare correlation patterns for RECT across different models for entanglement vs $\ell_2$ logit shift.}
    \label{fig:rect_l2_clare}
\end{figure*}

\begin{figure*}[t]
    \centering
    
    \begin{subfigure}[b]{0.32\textwidth}
        \includegraphics[width=\textwidth]{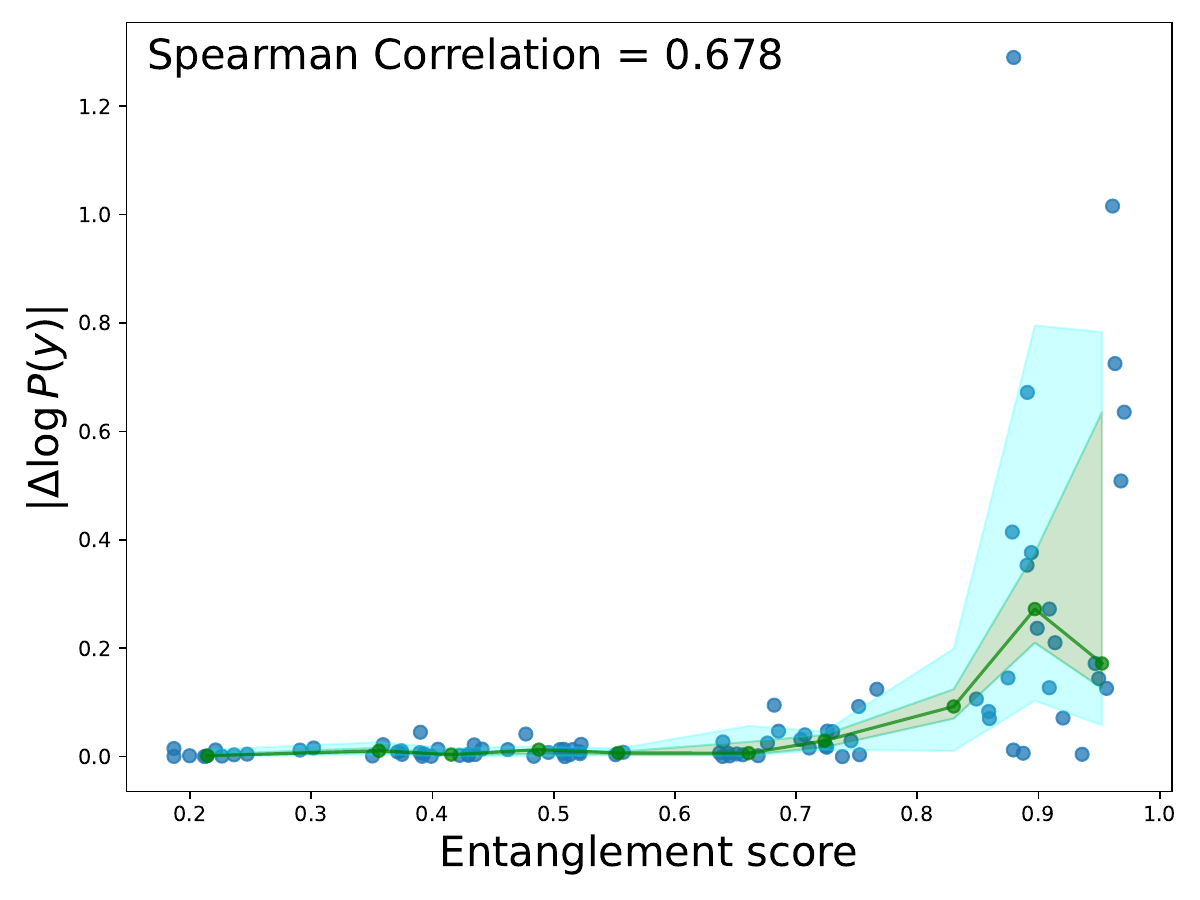}
        \caption{GPT2-XL}
    \end{subfigure}
    \hfill
    \begin{subfigure}[b]{0.32\textwidth}
        \includegraphics[width=\textwidth]{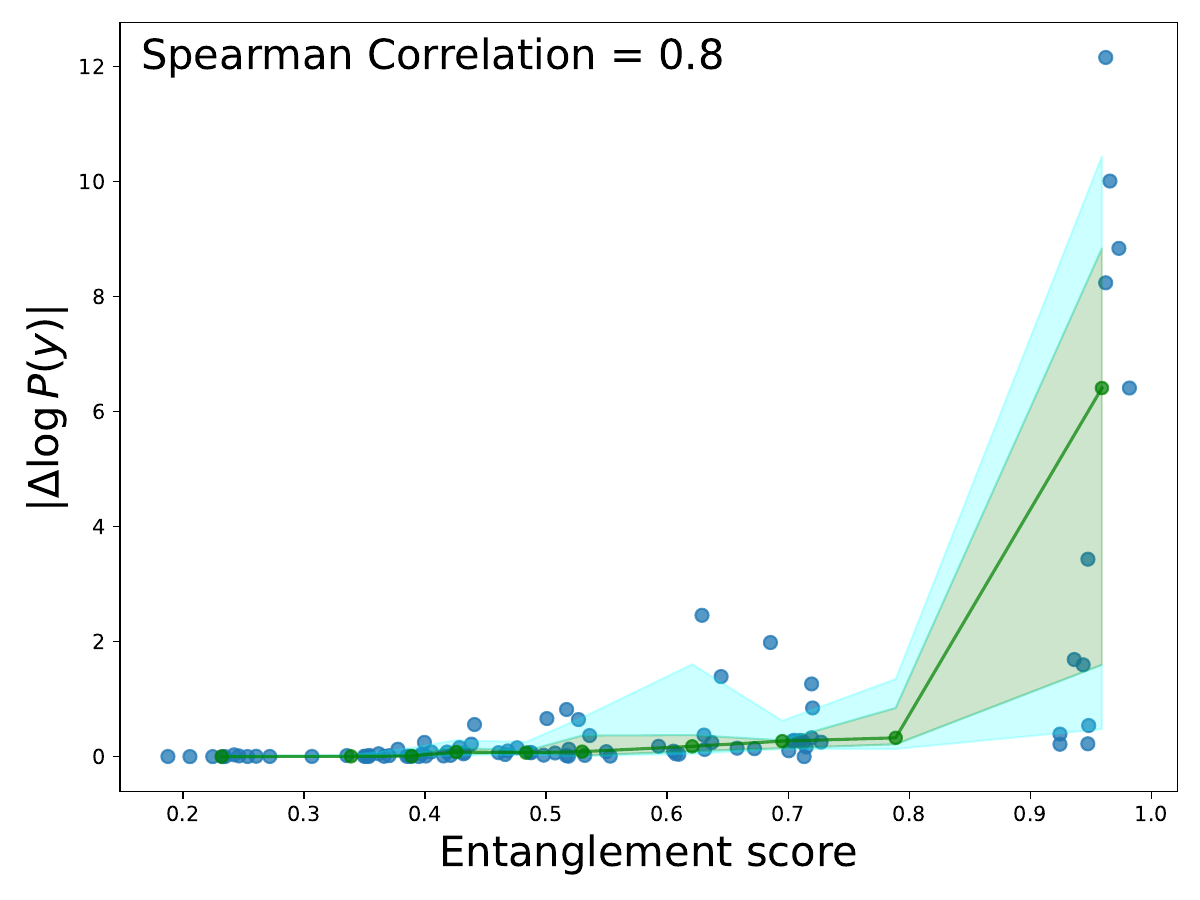}
        \caption{Llama3}
    \end{subfigure}
    \hfill
    \begin{subfigure}[b]{0.32\textwidth}
        \includegraphics[width=\textwidth]{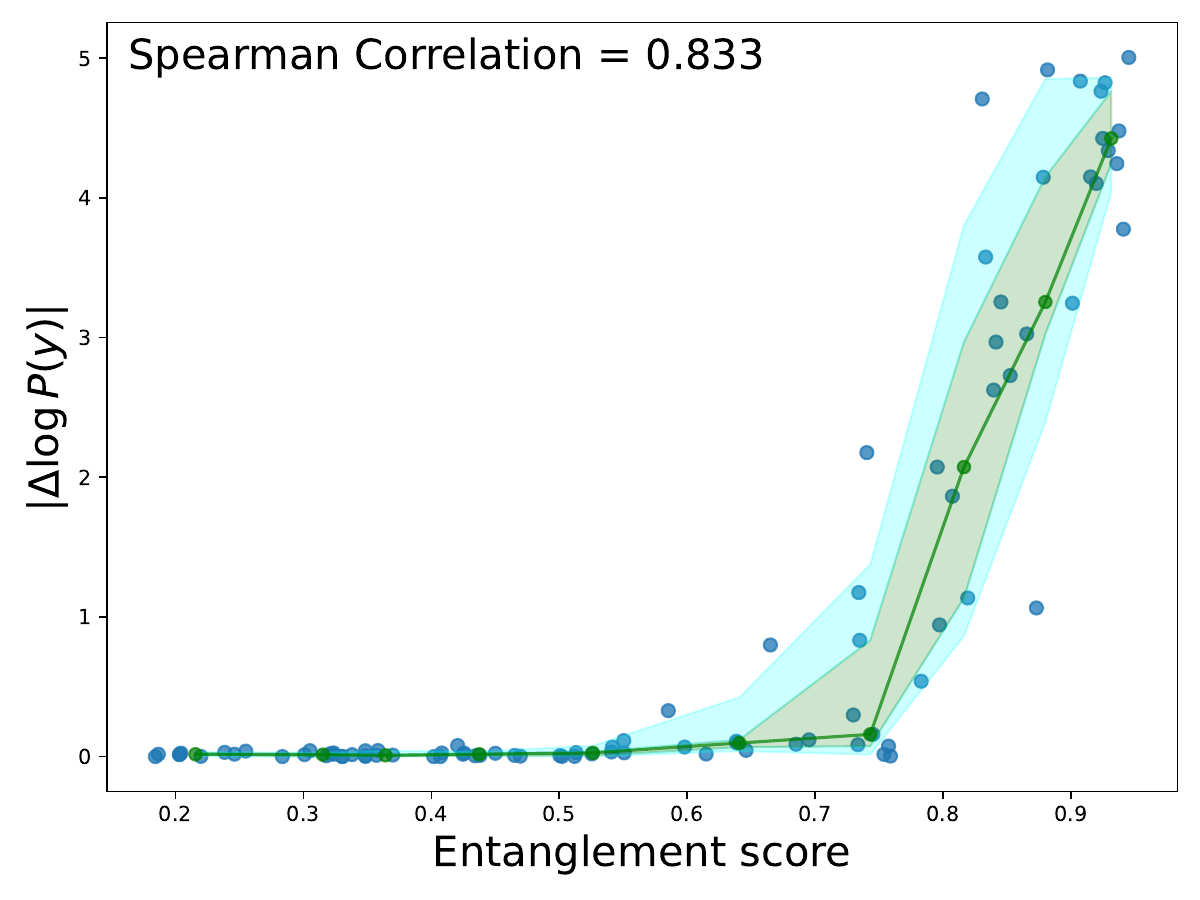}
        \caption{GPT-J}
    \end{subfigure}
    
    \caption{\Clare correlation patterns for RECT across different models for entanglement vs $|\Delta \log P(y)|$.}
    \label{fig:rect_delta_clare}
\end{figure*}

\begin{figure*}[t]
    \centering
    
    \begin{subfigure}[b]{0.32\textwidth}
        \includegraphics[width=\textwidth]{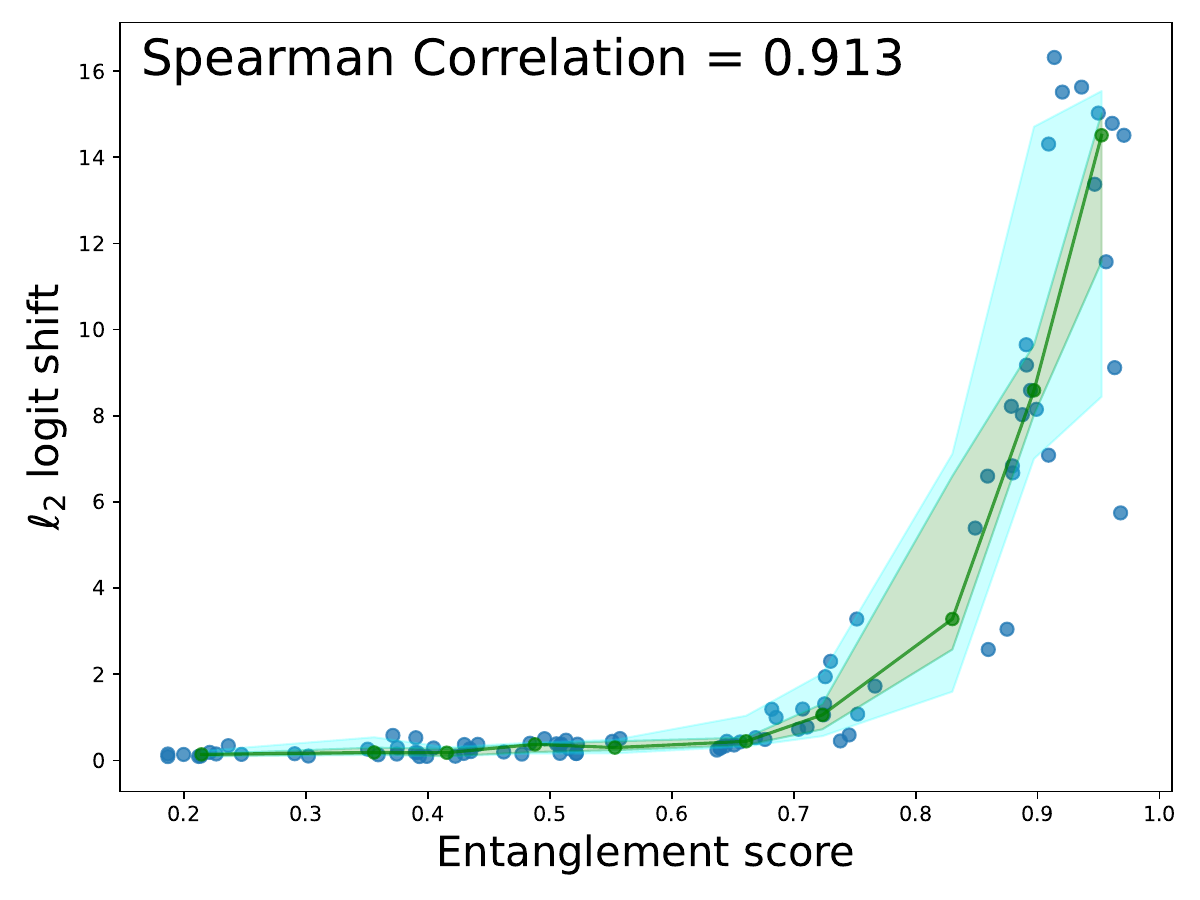}
        \caption{GPT2-XL}
    \end{subfigure}
    \hfill
    \begin{subfigure}[b]{0.32\textwidth}
        \includegraphics[width=\textwidth]{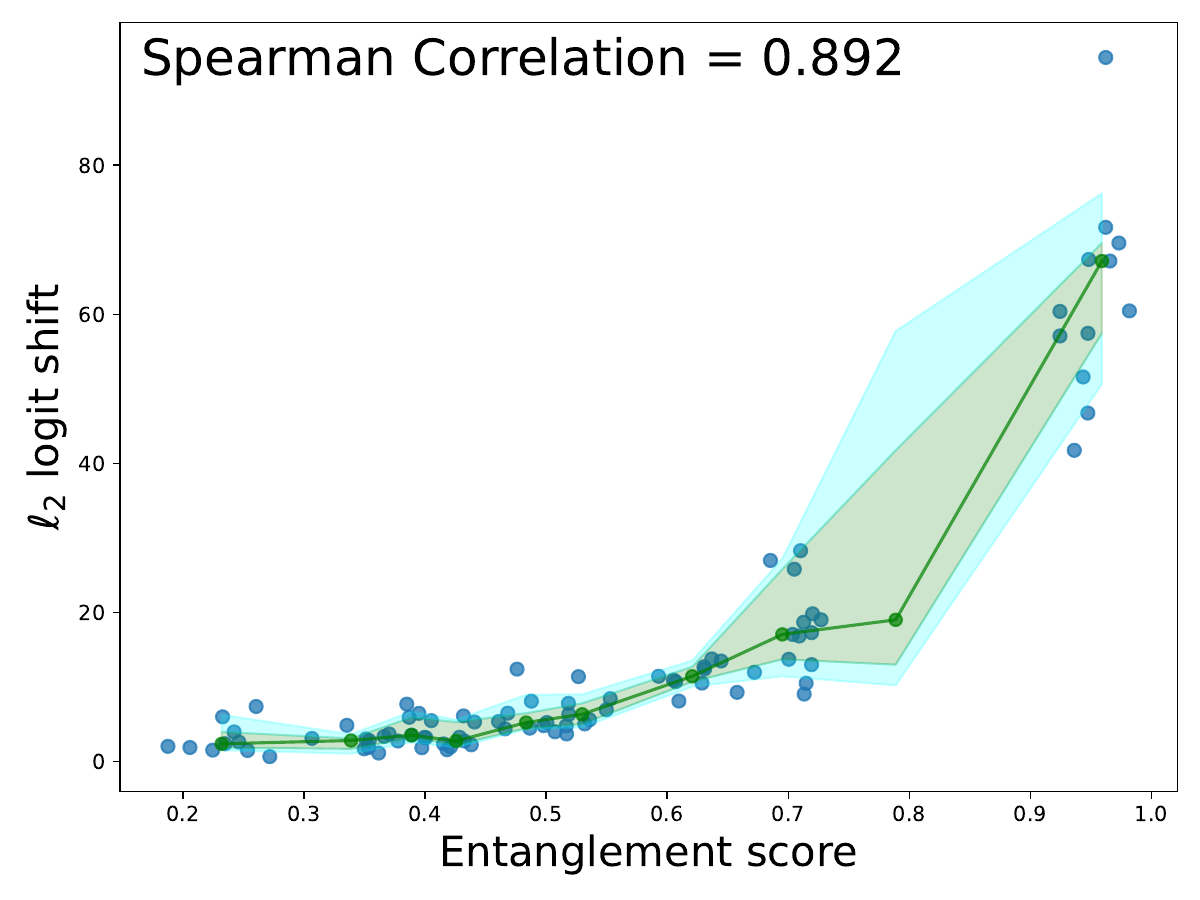}
        \caption{Llama3}
    \end{subfigure}
    \hfill
    \begin{subfigure}[b]{0.32\textwidth}
        \includegraphics[width=\textwidth]{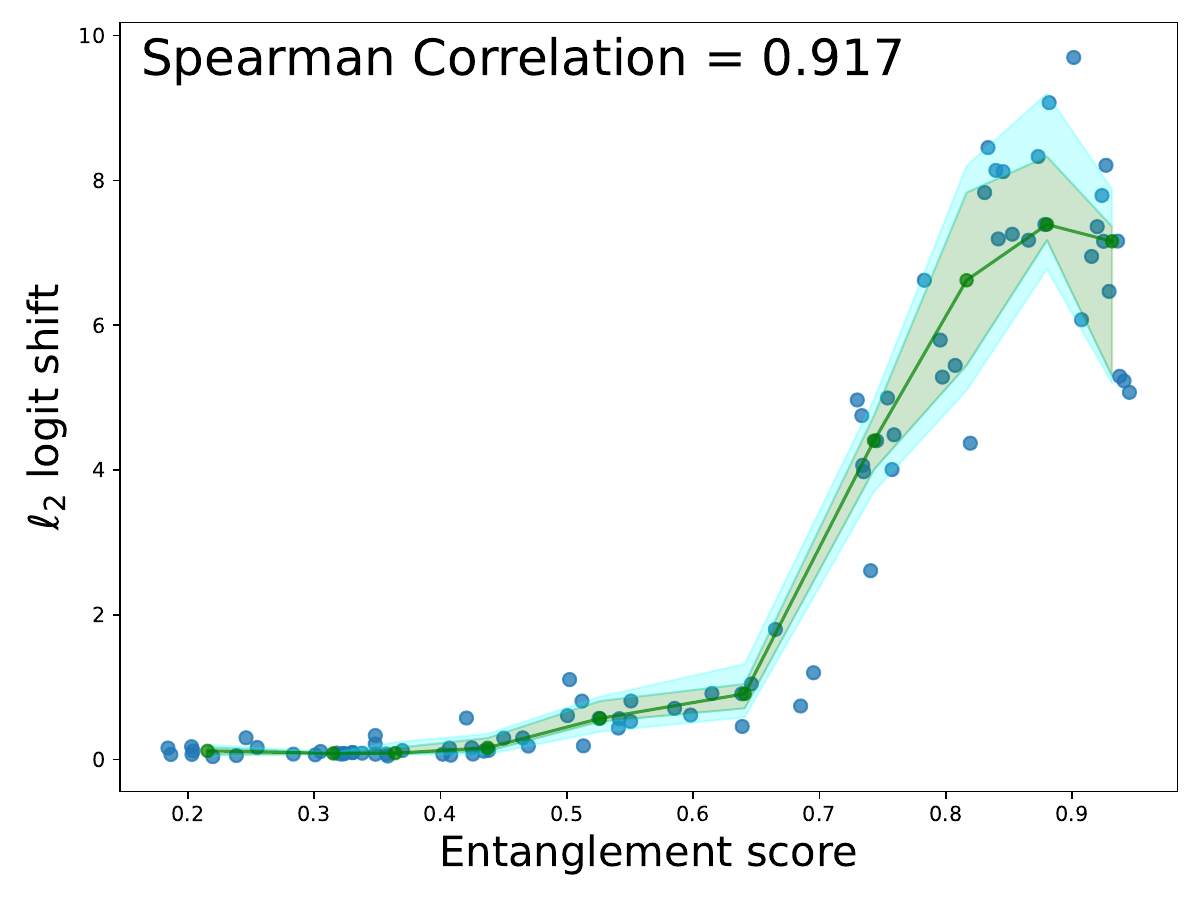}
        \caption{GPT-J}
    \end{subfigure}
    
    \caption{\Clare correlation patterns for MEMIT across different models for entanglement vs $\ell_2$ logit shift.}
    \label{fig:memit_l2_clare}
\end{figure*}

\begin{figure*}[t]
    \centering
    
    \begin{subfigure}[b]{0.32\textwidth}
        \includegraphics[width=\textwidth]{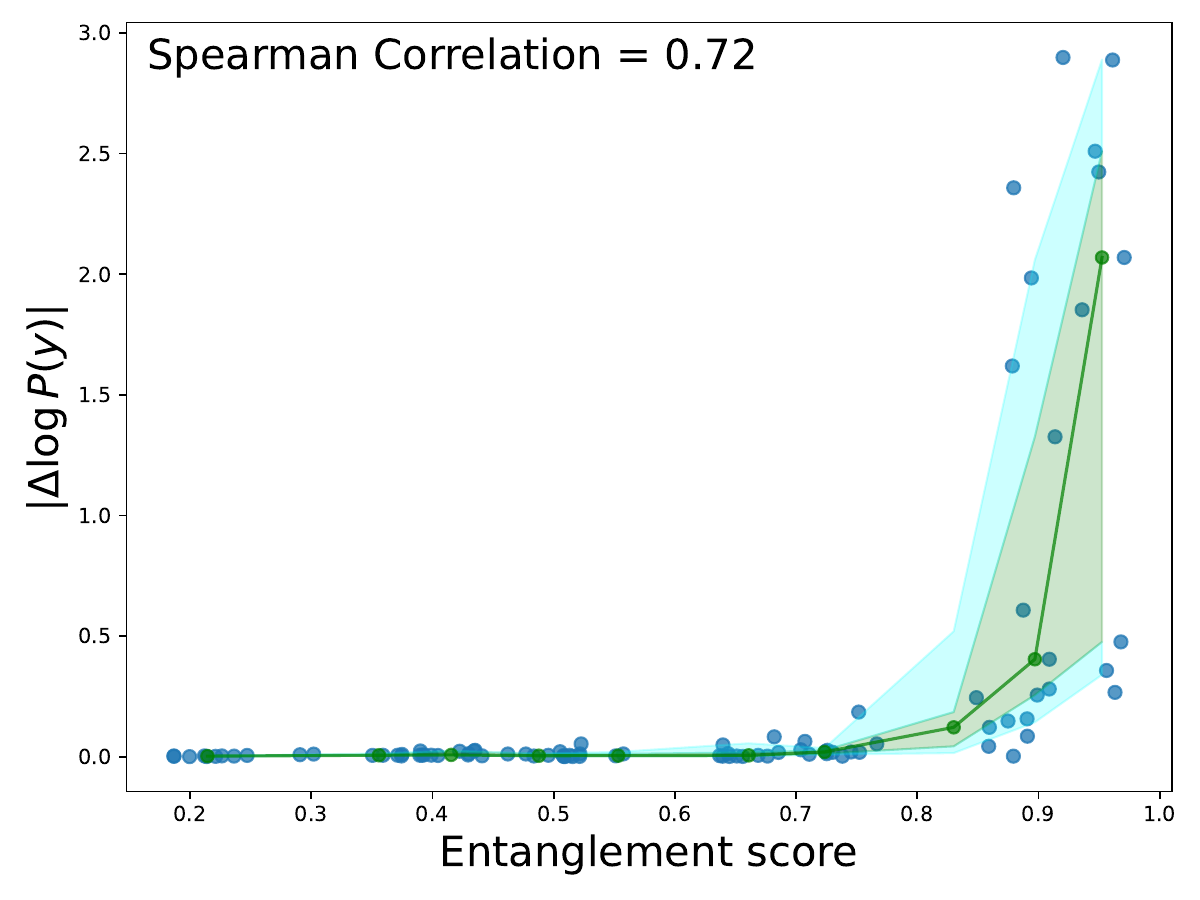}
        \caption{GPT2-XL}
    \end{subfigure}
    \hfill
    \begin{subfigure}[b]{0.32\textwidth}
        \includegraphics[width=\textwidth]{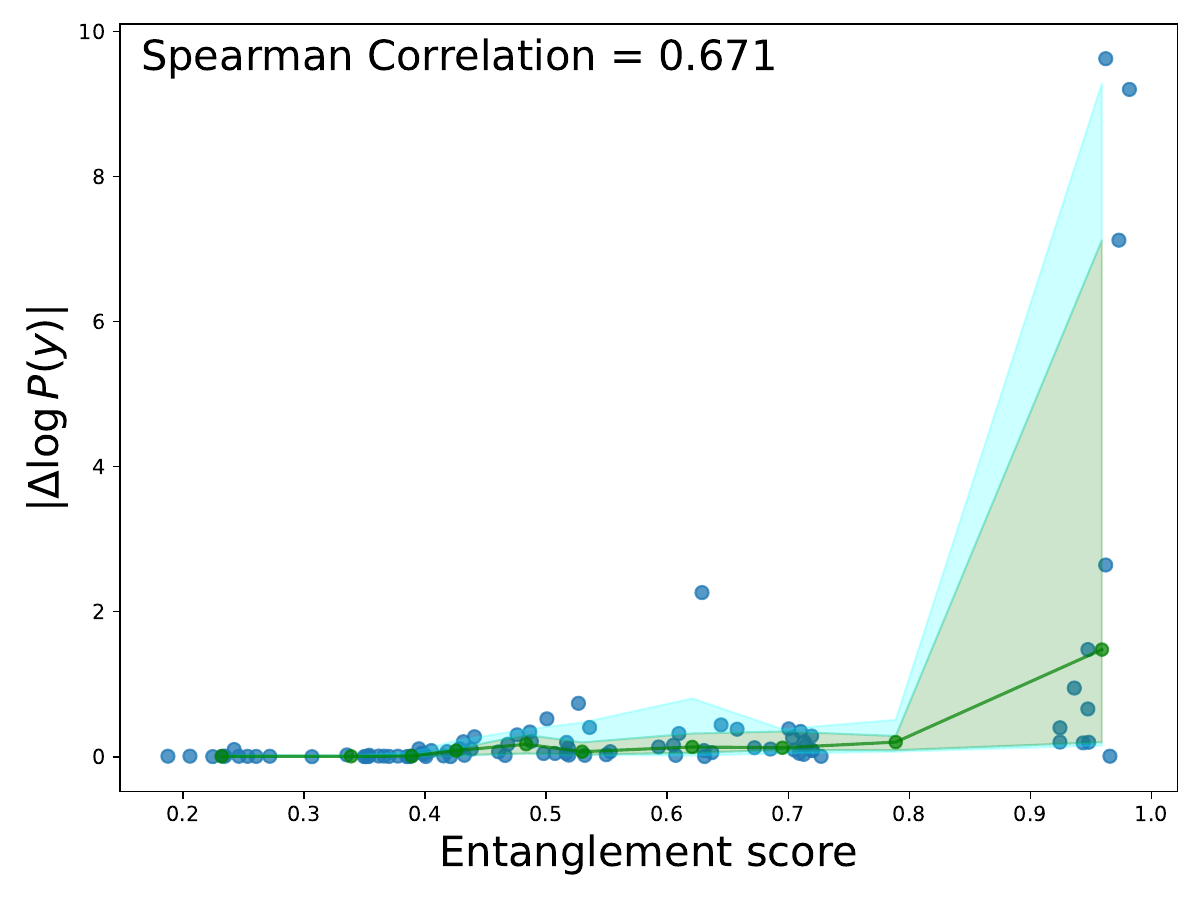}
        \caption{Llama3}
    \end{subfigure}
    \hfill
    \begin{subfigure}[b]{0.32\textwidth}
        \includegraphics[width=\textwidth]{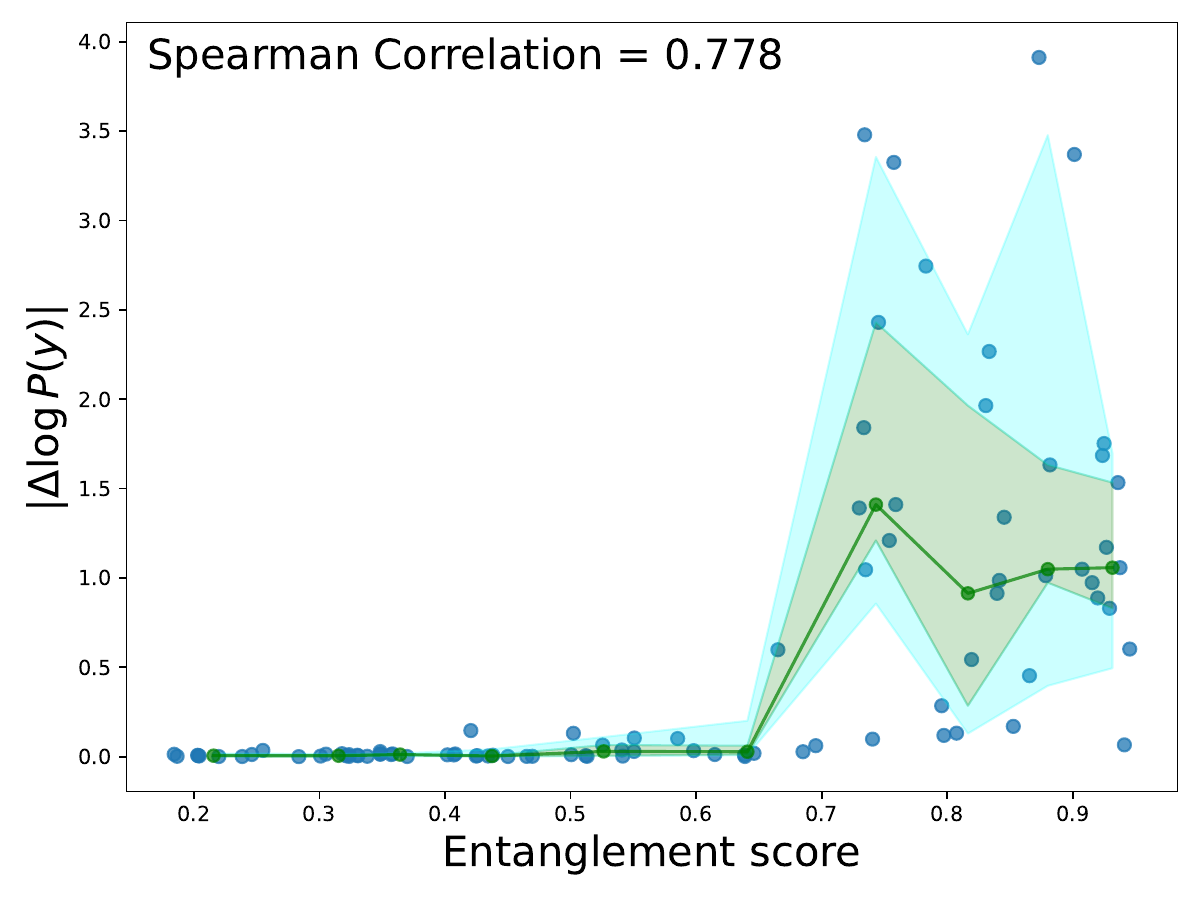}
        \caption{GPT-J}
    \end{subfigure}
    
    \caption{Correlation patterns for MEMIT across different models for entanglement vs $|\Delta \log P(y)|$.}
    \label{fig:memit_delta_clare}
\end{figure*}

\begin{figure*}[t]
    \centering
    
    \begin{subfigure}[b]{0.32\textwidth}
        \includegraphics[width=\textwidth]{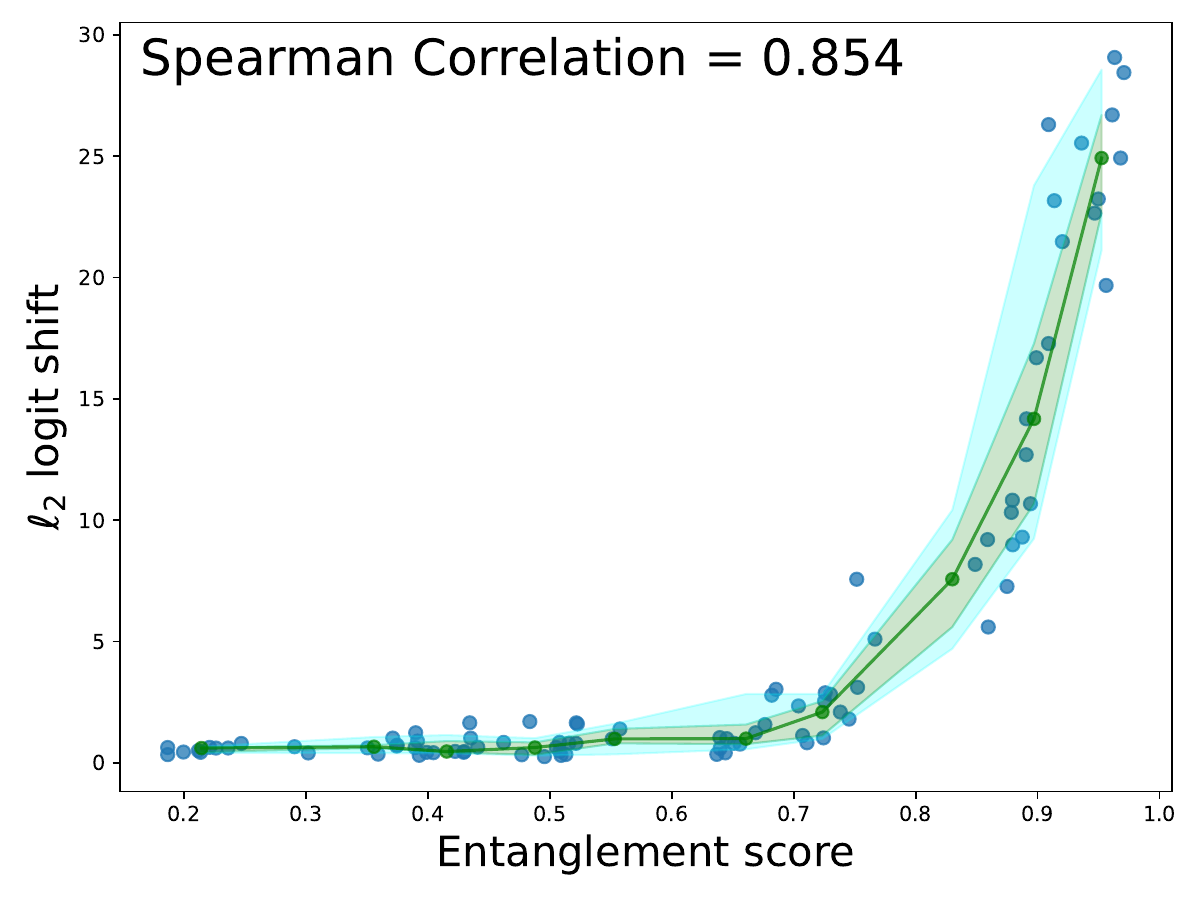}
        \caption{GPT2-XL}
    \end{subfigure}
    \hfill
    \begin{subfigure}[b]{0.32\textwidth}
        \includegraphics[width=\textwidth]{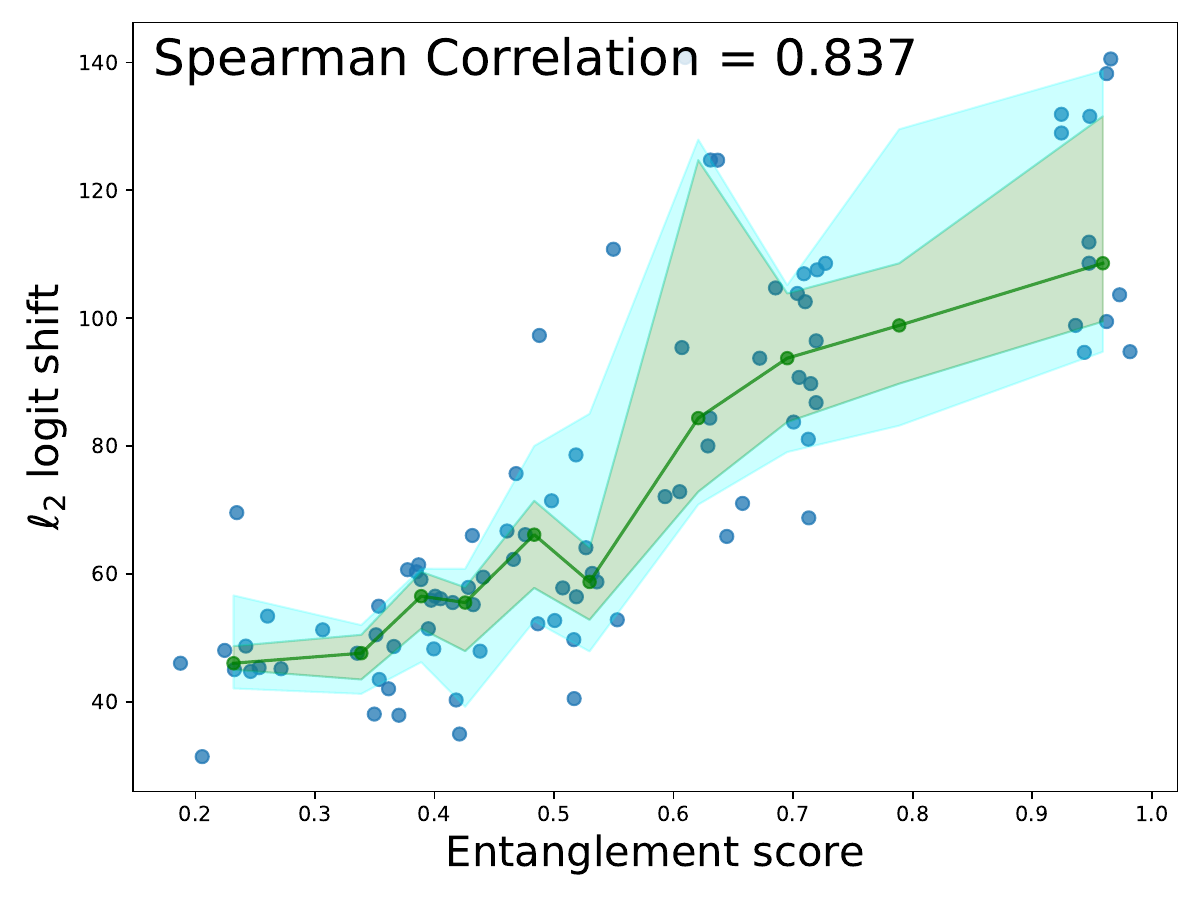}
        \caption{Llama3}
    \end{subfigure}
    \hfill
    \begin{subfigure}[b]{0.32\textwidth}
        \includegraphics[width=\textwidth]{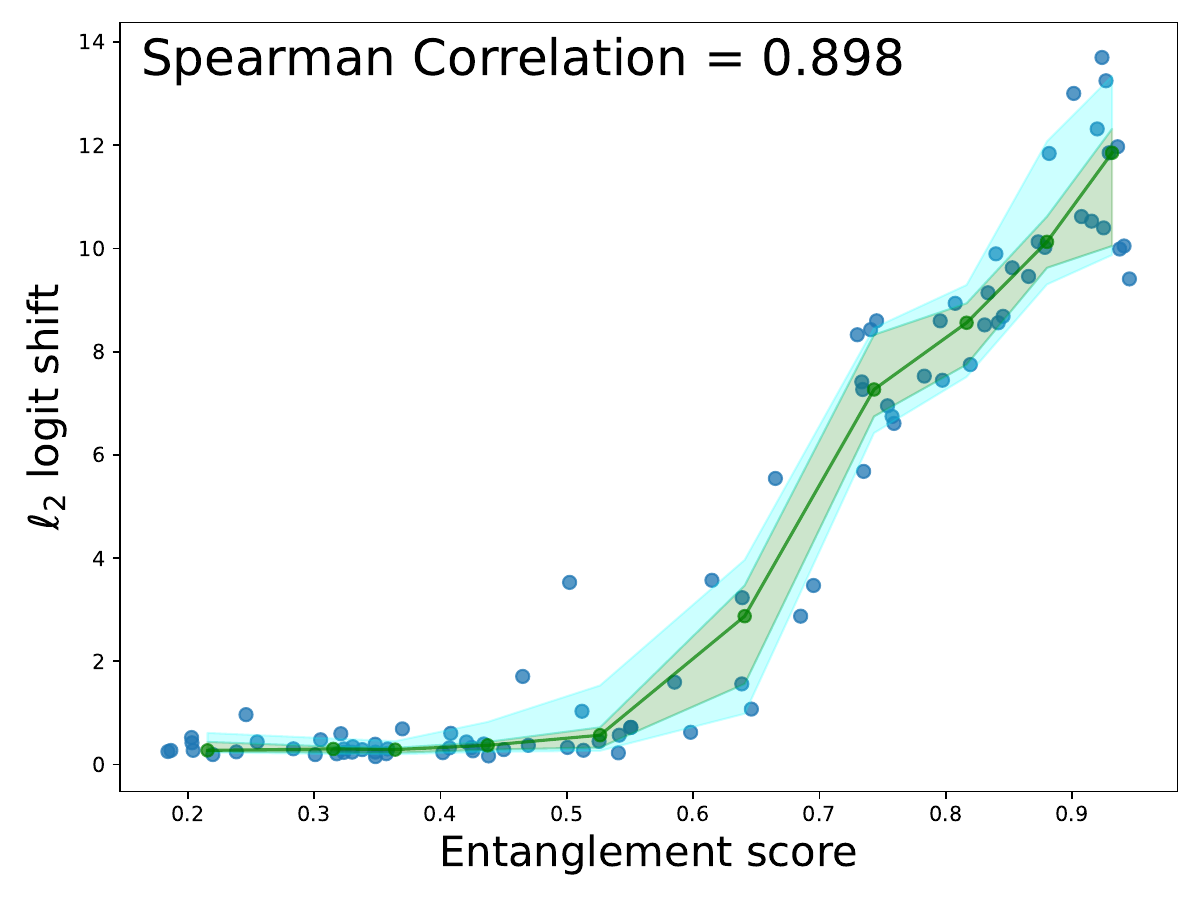}
        \caption{GPT-J}
    \end{subfigure}
    
    \caption{\Clare correlation patterns for ROME across different models for entanglement vs $\ell_2$ logit shift.}
    \label{fig:rome_l2_clare}
\end{figure*}

\begin{figure*}[t]
    \centering
    
    \begin{subfigure}[b]{0.32\textwidth}
        \includegraphics[width=\textwidth]{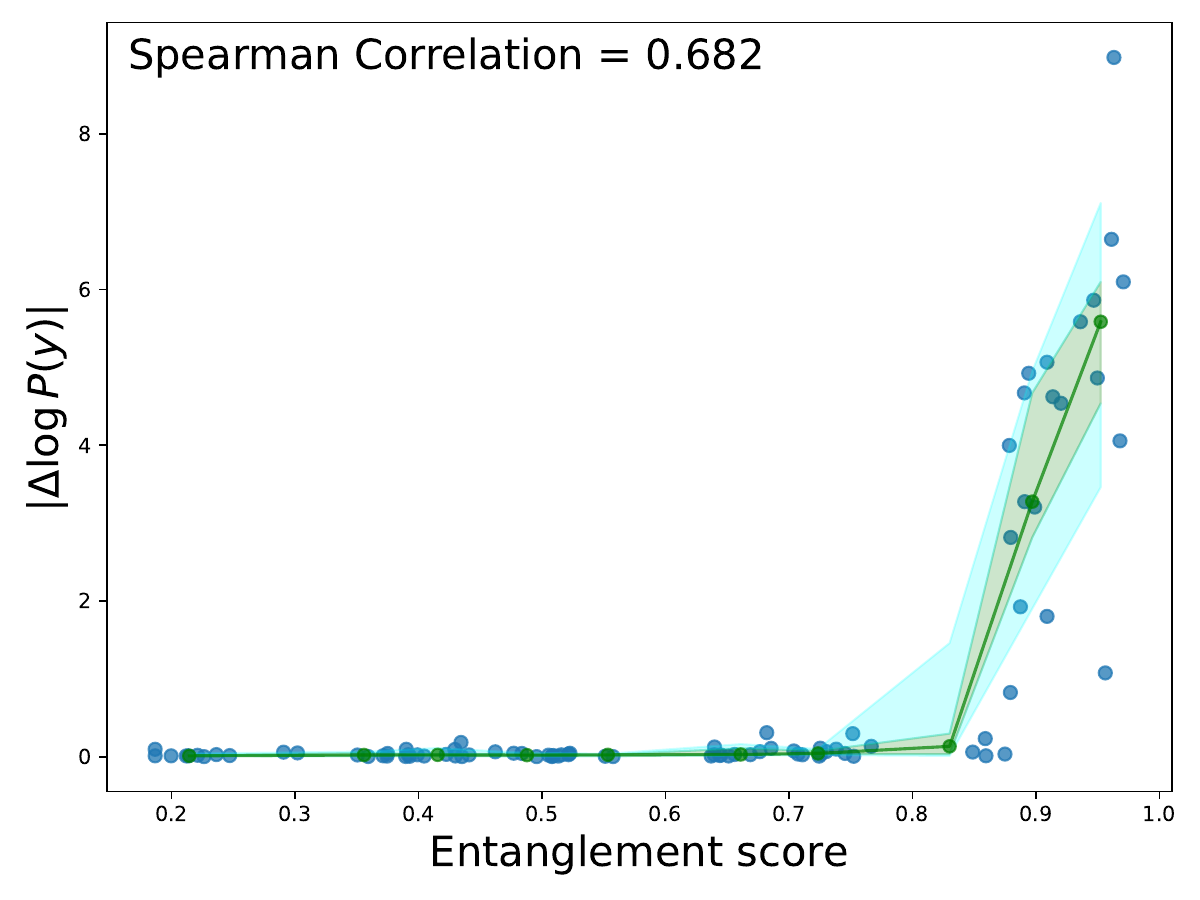}
        \caption{GPT2-XL}
    \end{subfigure}
    \hfill
    \begin{subfigure}[b]{0.32\textwidth}
        \includegraphics[width=\textwidth]{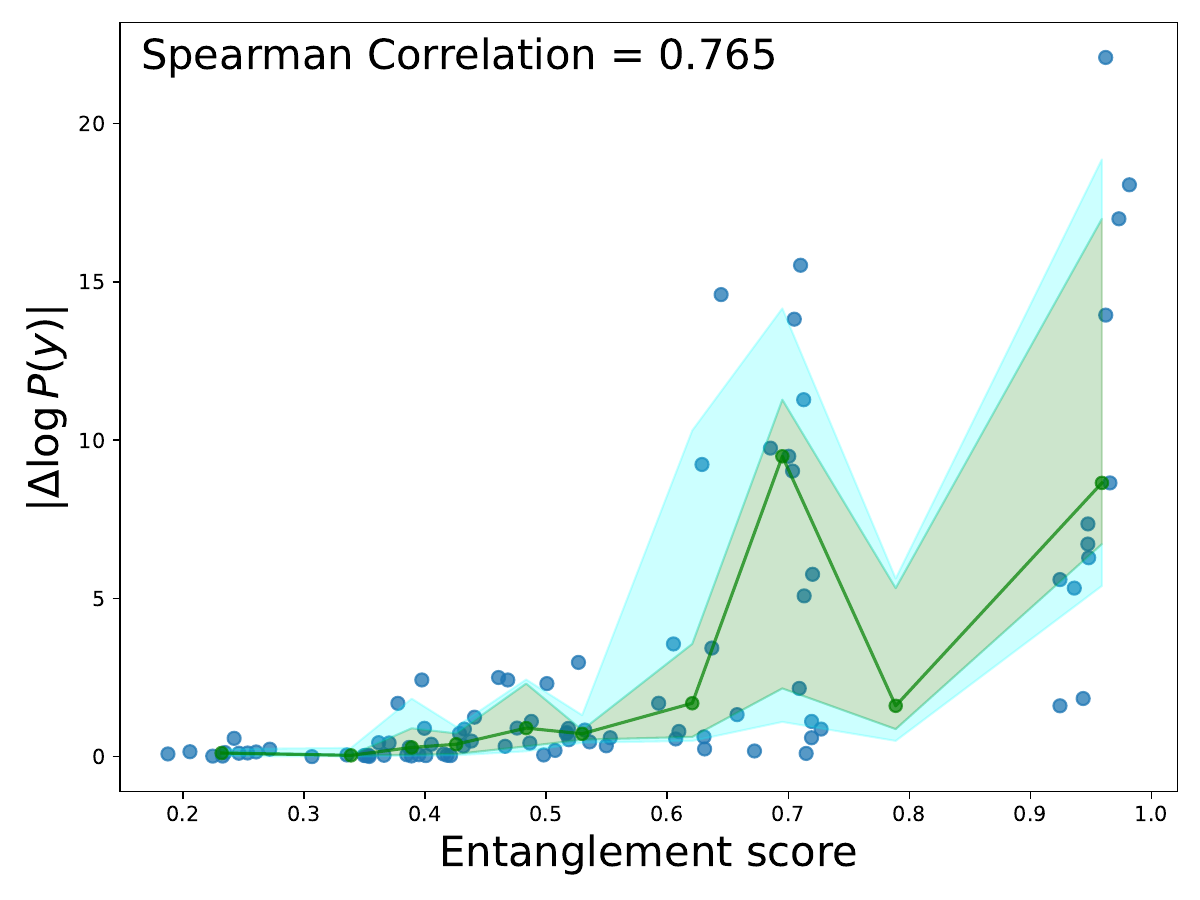}
        \caption{Llama3}
    \end{subfigure}
    \hfill
    \begin{subfigure}[b]{0.32\textwidth}
        \includegraphics[width=\textwidth]{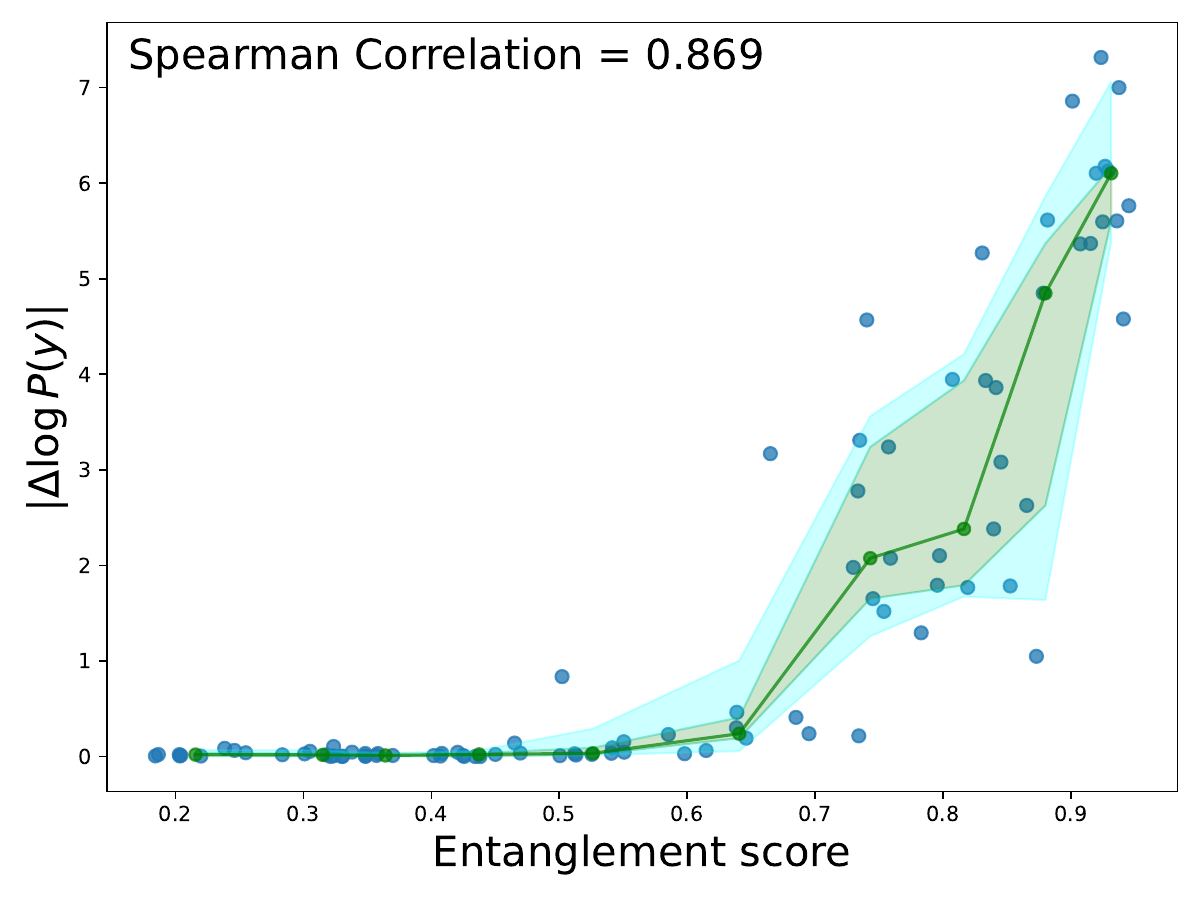}
        \caption{GPT-J}
    \end{subfigure}
    
    \caption{\Clare correlation patterns for ROME across different models for entanglement vs $|\Delta \log P(y)|$.}
    \label{fig:rome_delta_clare}
\end{figure*}

\begin{figure*}[t]
    \centering
    
    \begin{subfigure}[b]{0.32\textwidth}
        \includegraphics[width=\textwidth]{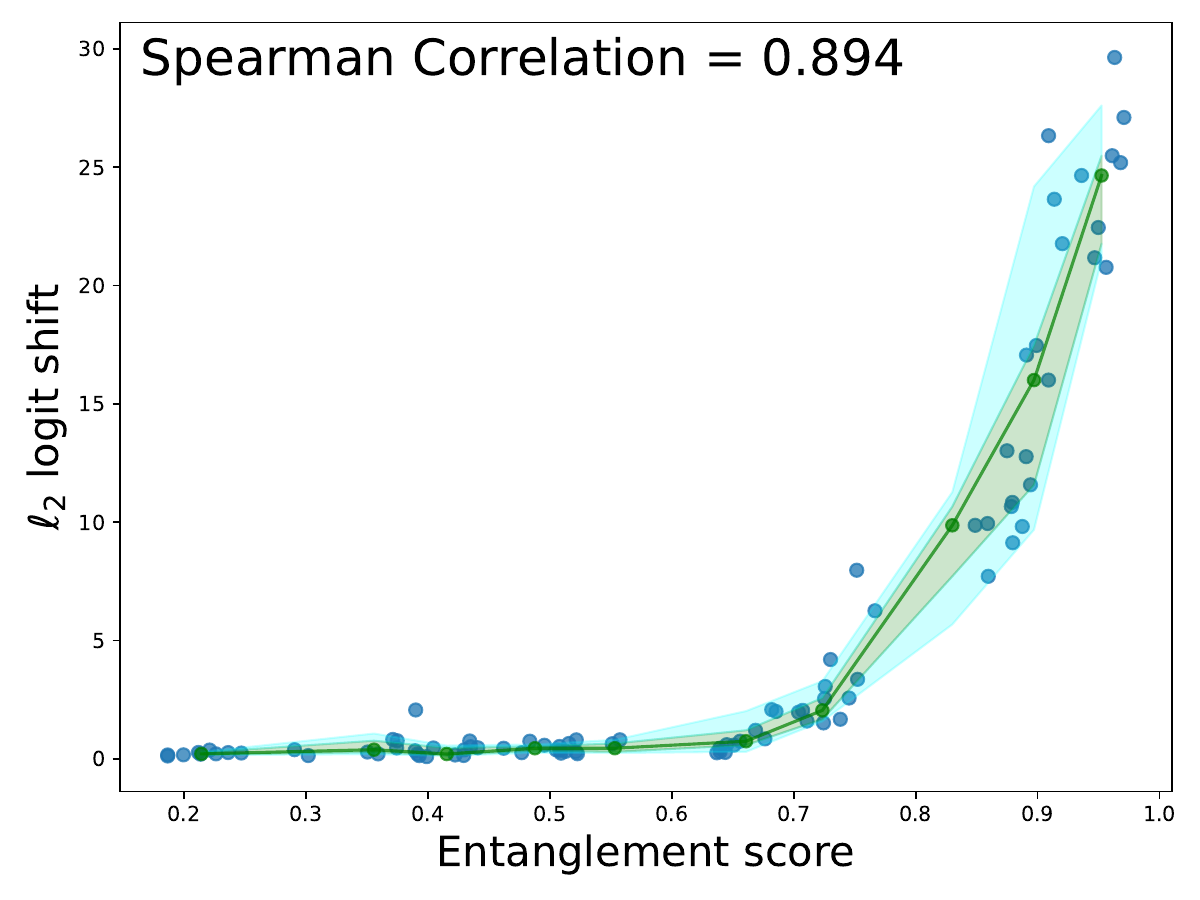}
        \caption{GPT2-XL}
    \end{subfigure}
    \hfill
    \begin{subfigure}[b]{0.32\textwidth}
        \includegraphics[width=\textwidth]{Plots/one_to_many/meta-llama/Meta-Llama-3-8B-Instruct/output_ROME_0_1.pdf}
        \caption{Llama3}
    \end{subfigure}
    \hfill
    \begin{subfigure}[b]{0.32\textwidth}
        \includegraphics[width=\textwidth]{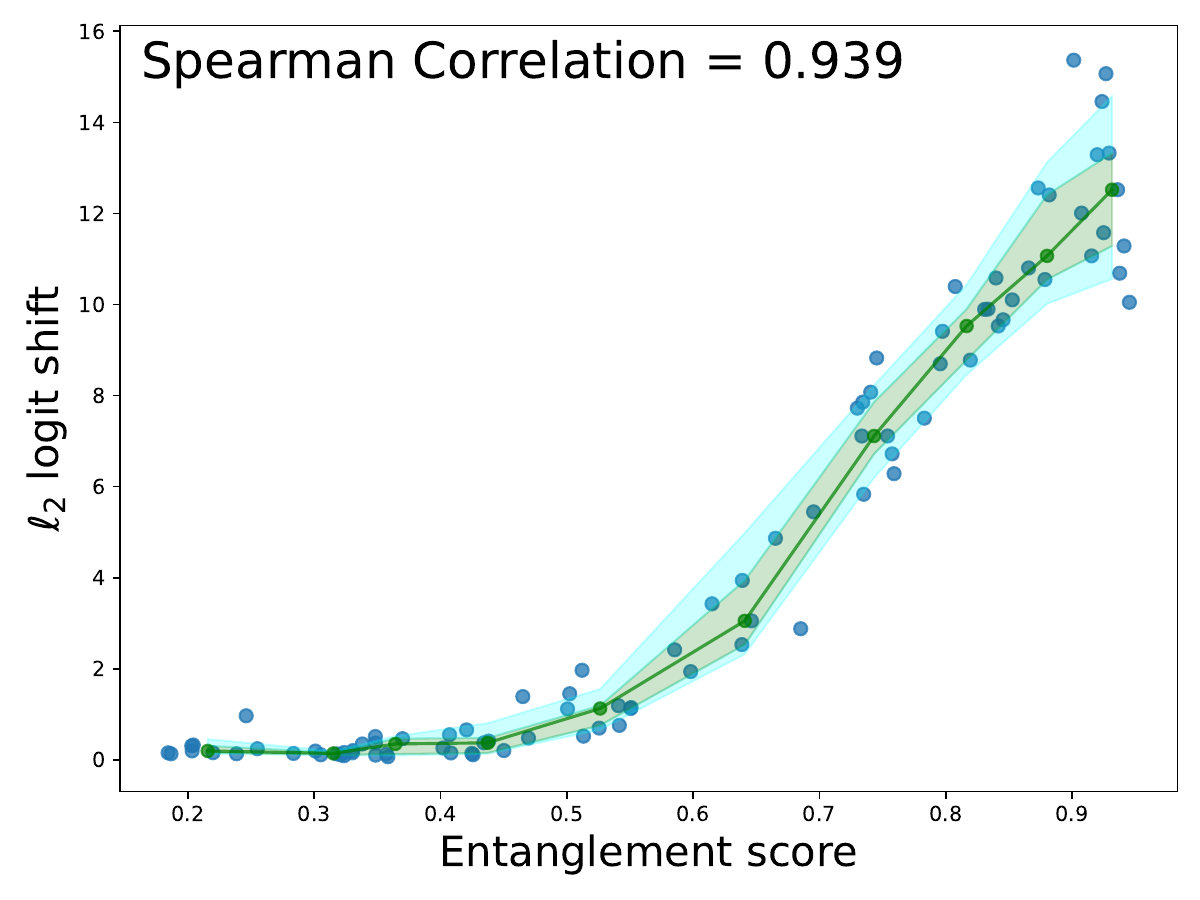}
        \caption{GPT-J}
    \end{subfigure}
    
    \caption{\Clare correlation patterns for PRUNE across different models for entanglement vs $\ell_2$ logit shift.}
    \label{fig:prune_l2_clare}
\end{figure*}

\begin{figure*}[t]
    \centering
    
    \begin{subfigure}[b]{0.32\textwidth}
        \includegraphics[width=\textwidth]{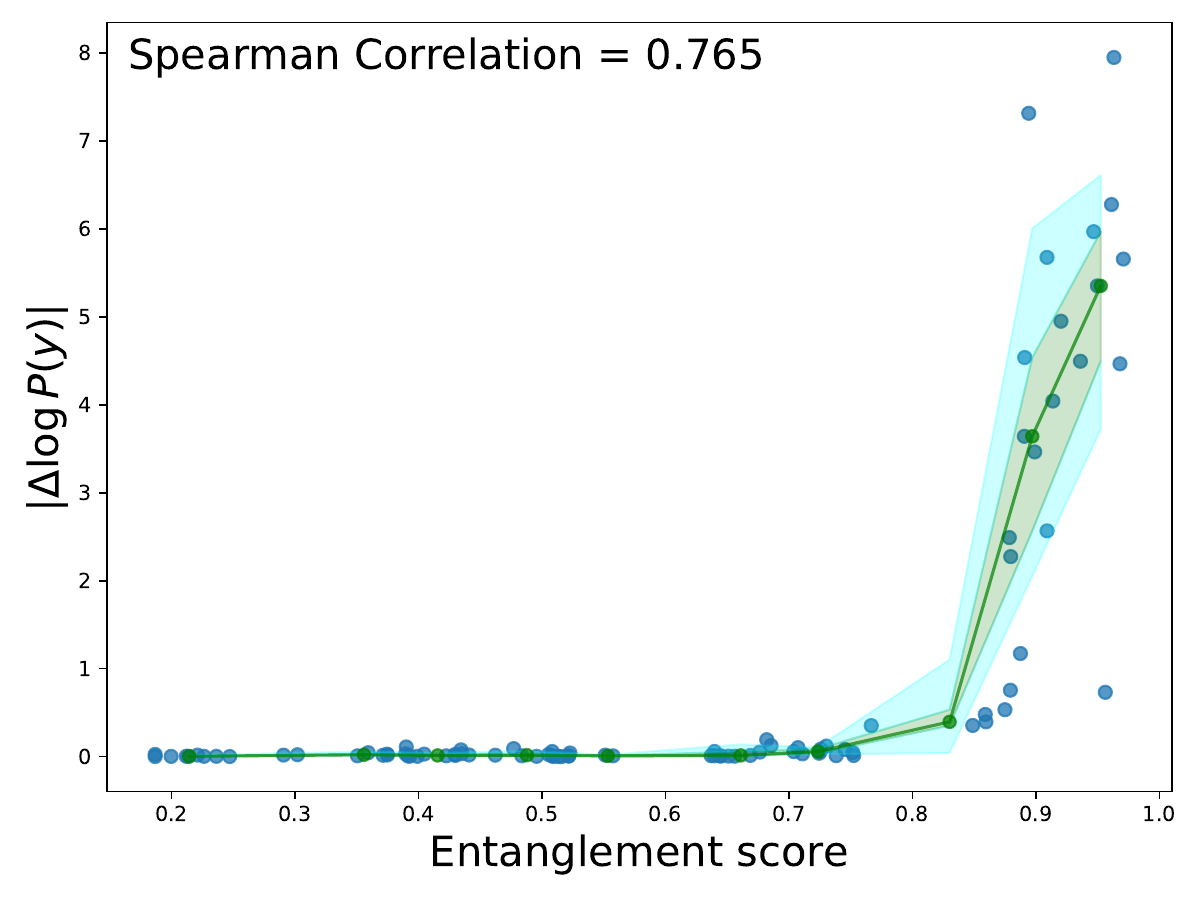}
        \caption{GPT2-XL}
    \end{subfigure}
    \hfill
    \begin{subfigure}[b]{0.32\textwidth}
        \includegraphics[width=\textwidth]{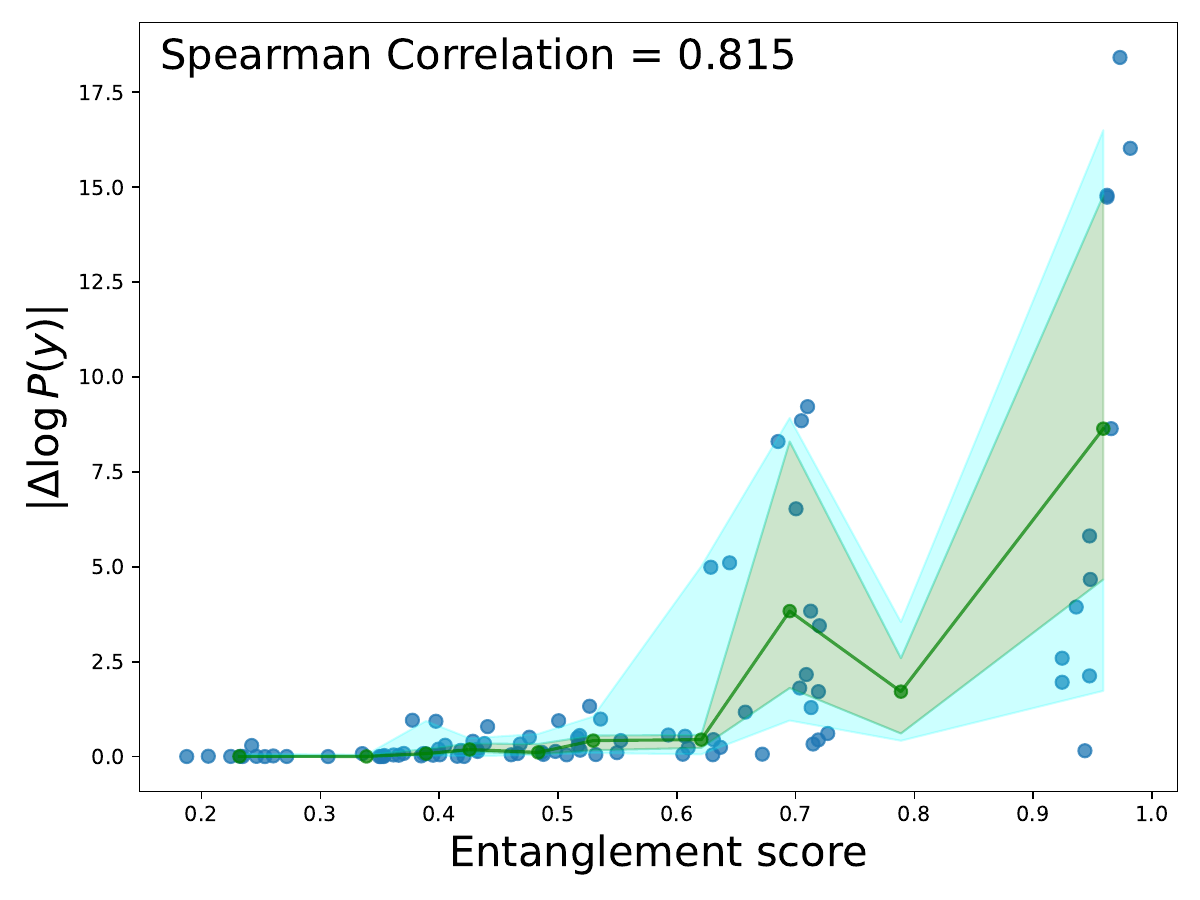}
        \caption{Llama3}
    \end{subfigure}
    \hfill
    \begin{subfigure}[b]{0.32\textwidth}
        \includegraphics[width=\textwidth]{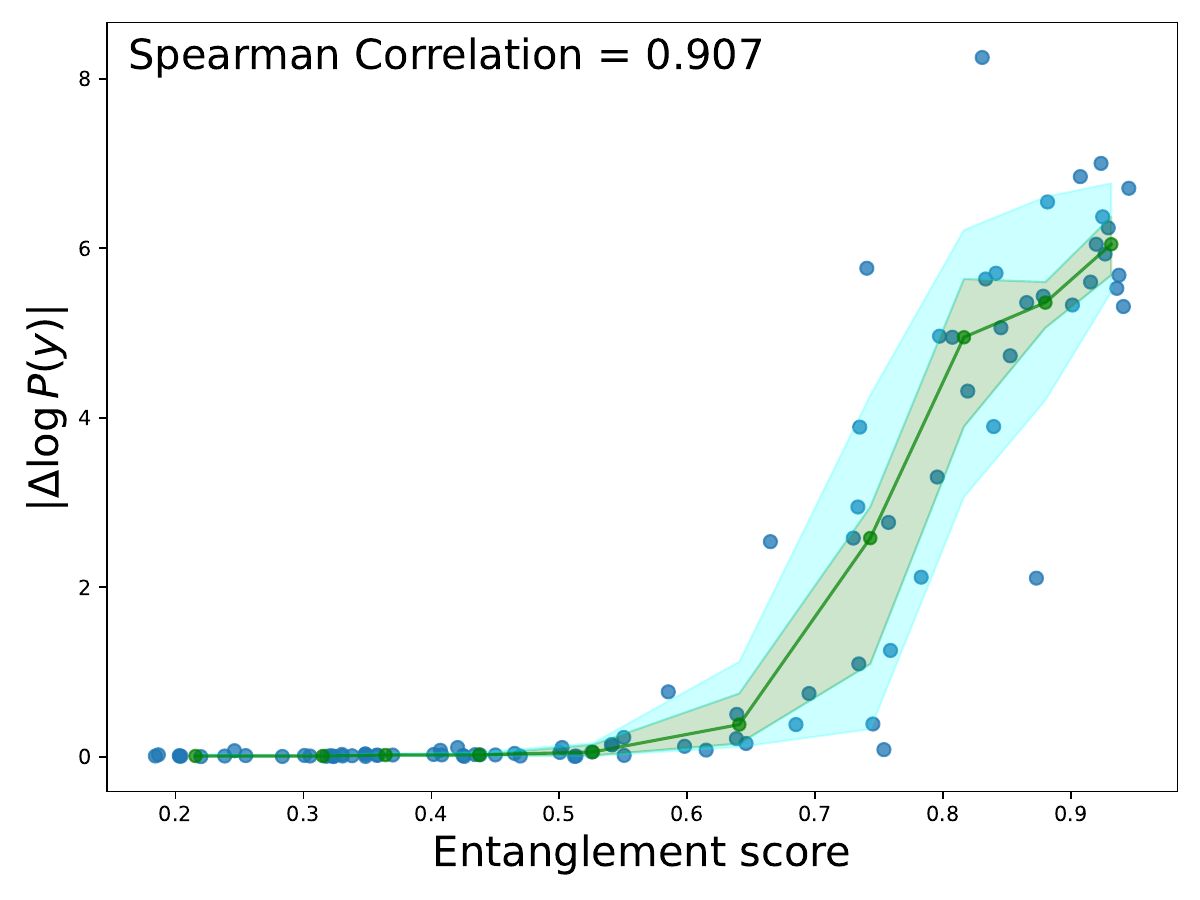}
        \caption{GPT-J}
    \end{subfigure}
    
    \caption{\Clare correlation patterns for PRUNE across different models for entanglement vs $|\Delta \log P(y)|$.}
    \label{fig:prune_delta_clare}
\end{figure*}

Figures \ref{fig:rect_l2}, \ref{fig:rect_delta}, \ref{fig:memit_l2}, \ref{fig:memit_delta}, \ref{fig:rome_l2}, \ref{fig:rome_delta}, \ref{fig:prune_l2} and \ref{fig:prune_delta} show the correlation patterns of GradSim and
\Clare for different models when edited by different edited techniques over the 1,000 samples as discussed in RQ \ding{172}. Figures \ref{fig:alphaedit_l2_clare}, \ref{fig:alphaedit_delta_clare}, \ref{fig:rect_l2_clare}, \ref{fig:rect_delta_clare}, \ref{fig:memit_l2_clare}, \ref{fig:memit_delta_clare}, \ref{fig:rome_l2_clare}, \ref{fig:rome_delta_clare}, \ref{fig:prune_l2_clare}, and \ref{fig:prune_delta_clare} illustrate \Clare's entanglement correlation patterns using a single edited fact probed against multiple control facts, to capture a more fine-grained view of ripple susceptibility. Across all editing techniques, models and metrics ($\ell_2$ logit shift and 
$|\Delta \log P(y)|$), we observe a consistent increasing trend: higher \Clare entanglement scores correspond to larger ripple effects. This monotonic relationship supports the reliability of \Clare as a forward-only proxy for ripple-prone regions, generalizing robustly across diverse settings without the need for gradient computation. Across all settings, we observe a thresholded rising pattern. Entanglement scores below approximately 0.7 correspond to negligible or minimal changes in behavior (flat region), whereas scores above this threshold consistently lead to sharply increasing ripple effects. This pattern consistency across models, techniques, and evaluation metrics assists in separating safe vs. high-risk edits based purely on forward activations.

\begin{table*}
\centering
\caption{Step-by-step peak GPU memory consumption (GB). \Clare steps: Fwd (forward pass) 1 $\rightarrow$ Fwd 2 $\rightarrow$ Sim (Cosine similarity of two vectors). \textbf{GradSim} steps: Gradient 1 (Fwd $\rightarrow$ Loss calculation $\rightarrow$ Bwd (backward pass) $\rightarrow$ Gradient 2 (Fwd $\rightarrow$ Loss $\rightarrow$ Bwd) $\rightarrow$ Sim (Cosine similarity of two gradients).}
\label{tab:gpu_breakdown}
\resizebox{\textwidth}{!}{%
\begin{tabular}{l|ccc|ccc|ccc|c}
\toprule
\multirow{3}{*}{\textbf{Model}} & \multicolumn{3}{c|}{\textbf{Method 1: \Clare}} & \multicolumn{7}{c}{\textbf{Method 2: GradSim}} \\ \cmidrule(lr){2-4} \cmidrule(lr){5-11}
 & \multicolumn{2}{c}{Forward Pass} & \multirow{2}{*}{Sim} & \multicolumn{3}{c|}{Gradient 1 Steps} & \multicolumn{3}{c|}{Gradient 2 Steps} & \multirow{2}{*}{Sim} \\ \cmidrule(lr){2-3} \cmidrule(lr){5-7} \cmidrule(lr){8-10}
 & Fwd 1 & Fwd 2 & & Fwd & Loss & Bwd & Fwd & Loss & Bwd & \\ \midrule
GPT-J & 23.815 & \textbf{23.905} & 22.588 & 33.551 & 33.550 & 45.167 & 56.134 & 56.133 & 67.710 & \textbf{68.474} \\
GPT2-XL & 6.968 & \textbf{7.016} & 6.013 & 9.032 & 9.029 & 12.275 & 15.003 & 15.000 & 18.210 & \textbf{18.212} \\
Llama3.2 & 12.940 & \textbf{12.995} & 12.001 & 18.042 & 18.037 & 25.475 & 30.048 & 30.043 & 37.439 & \textbf{37.445} \\
Llama3 & 31.459 & \textbf{31.539} & 29.946 & 44.012 & 44.007 & 59.898 & 73.967 & 73.961 & 89.813 & \textbf{91.765} \\
Mistral & 28.404 & \textbf{28.478} & 27.009 & 40.331 & 40.330 & 54.021 & 67.348 & 67.346 & 80.999 & \textbf{81.483} \\
Qwen2 & 29.882 & \textbf{29.981} & 28.541 & 41.820 & 41.815 & 57.023 & 70.316 & 70.310 & 85.491 & \textbf{87.513} \\
Qwen2.5 & 29.882 & \textbf{29.981} & 28.541 & 41.820 & 41.815 & 57.023 & 70.316 & 70.310 & 85.491 & \textbf{87.513} \\ \bottomrule
\end{tabular}%
}
\end{table*}

We hypothesize that the observed threshold around 0.7 cosine similarity reflects a critical geometric boundary in the model's representational manifold. In high-dimensional space, a cosine similarity of 0.7 corresponds to an angular separation of approximately $45^{\circ}$, suggesting that facts within this cone may share overlapping subspaces or ``knowledge circuits''~\cite{yao2024knowledgecircuits}. Edits that shift one fact's representation may inadvertently affect other facts within this angular neighborhood, manifesting as ripple effects. Further investigation is needed to validate whether this threshold consistently emerges across different model architectures.

\section{Approximation of the last critical layer}
\label{appendix:critical_layer_approximation}

\begin{table}
\centering
\caption{Ratio of the last critical layer to total number of layers for various models, taken from the EasyEdit repository. Values cluster around one-third depth, supporting the use of a one-third-depth heuristic when the critical layer is unknown.}
\label{tab:critical_layer_ratios}
\begin{tabular}{|l|c|}
\hline
\textbf{Model} & \textbf{Index of $\mathcal{L}$ / Number of layers} \\
\hline
Baichuan-7B & 0.28125 \\
GPT-J-6B & 0.32143 \\
GPT-2-XL & 0.37500 \\
InternLM-7B & 0.28125 \\
Llama-7B & 0.28125 \\
Llama3.2-3B & 0.34615 \\
Mistral-7B & 0.28125 \\
Qwen-7B & 0.28125 \\
Qwen2.5-7B & 0.32143 \\
\hline
\end{tabular}
\end{table}

When causal-tracing data or prior analyses identifying the factual-memory layers are unavailable, we approximate the final critical layer using a simple depth-based heuristic. Prior work and our replication of the EasyEdit repository indicate that critical layers across a range of transformer architectures typically occur within the lower-to-middle portion of the model—specifically, around one-third of the total depth. As shown in Table~\ref{tab:critical_layer_ratios}, the ratio of the last critical layer to total layers spans approximately \textbf{0.28--0.38} across diverse architectures, including GPT-J, GPT-2-XL, Llama, Qwen, Mistral, and Baichuan. This empirical regularity supports the use of the one-third-depth index as a reliable proxy for the last causal layer when such information is unknown.

Thus, we take the layer at one-third of the model's total depth as the \emph{approximated critical layer} for computing \Clare scores. This approximation enables scalable entanglement estimation without requiring causal-tracing interventions or proprietary layer-level annotations. Table~\ref{tab:spearman_depth_third_layers_grouped_peak} reports the Spearman correlation ($\rho_{s}$) between \Clare scores and empirically measured ripple magnitudes when \Clare is computed from this one-third-depth layer. Across all three LLMs (GPT2-XL, GPTJ-6B, and Llama3-8B), the correlation at the approximated layer remains within approximately \textbf{0--1.5 percentage points of the global peak}, indicating that the heuristic provides a robust and practical substitute for exact causal-layer identification. Notably, GPT-J and GPT-2-XL exhibit almost no degradation relative to their optimal layer correlations (differences $<0.5$ p.p. for most methods), while Llama-3-8B shows slightly higher but still marginal deviations (1.0--1.4 p.p.). This suggests that representational entanglement is distributed smoothly around the critical band, and that the one-third-depth heuristic yields near-optimal \Clare estimates.

\begin{table*}
\centering
\caption{Spearman correlation ($\rho_{s}$) between \Clare and ripple effect ($\ell_2$ logit shift) in control facts at the one-third-depth layer (indexed by layer number), compared with the maximum value observed across all layers for each editing method. "Difference from peak" reports the absolute deviation in percentage points (p.p.).}
\label{tab:spearman_depth_third_layers_grouped_peak}
\begin{tabular}{l l c c c c c}
\hline
\textbf{Model} & \textbf{Layer Type} &
\textbf{AlphaEdit} & \textbf{MEMIT} & \textbf{PRUNE} & \textbf{RECT} & \textbf{ROME} \\
\hline
\multirow{3}{*}{Llama-3}
 & 1/3-depth layer = 11      & 64.31\% & 62.74\% & 63.41\% & 61.02\% & 69.53\% \\
 & Maximum $\rho_{s}$ (any layer)    & 65.16\% & 64.16\% & 64.84\% & 61.73\% & 70.54\% \\
 & Difference from peak (p.p.)     & 0.85 & 1.42 & 1.43 & 0.71 & 1.01 \\
\hline
\multirow{3}{*}{GPT-J}
 & 1/3-depth layer = 9       & 89.30\% & 80.13\% & 92.72\% & 90.26\% & 88.46\% \\
 & Maximum $\rho_{s}$ (any layer)    & 89.72\% & 80.13\% & 92.97\% & 90.26\% & 88.58\% \\
 & Difference from peak (p.p.)     & 0.42 & 0.00 & 0.25 & 0.00 & 0.12 \\
\hline
\multirow{3}{*}{GPT2-XL}
 & 1/3-depth layer = 16      & 76.29\% & 87.63\% & 87.99\% & 86.32\% & 85.26\% \\
 & Maximum $\rho_{s}$ (any layer)    & 77.49\% & 87.67\% & 87.99\% & 86.42\% & 85.26\% \\
 & Difference from peak (p.p.)     & 1.20 & 0.04 & 0.00 & 0.10 & 0.00 \\
\hline
\end{tabular}
\end{table*}

\section{Cost-effective red-teaming}
\label{appendix:redteaming}
To systematically identify areas of elevated editing risk, we analyse our corpora to find facts with exceptionally high representational connectivity, i.e., those whose internal representations overlap with a large number of other facts. These high-density knowledge nodes form critical pressure points within the model. Editing them is likely to induce observable widespread ripple effects. Figures~\ref{fig:gpt2xl_hub}, \ref{fig:llama3_hub}, \ref{fig:gptj_hub_1} and \ref{fig:gptj_hub_2} list the 50 most entangled facts in GPT2-XL, Llama3 and GPT-J, showing the number of connections each fact has within the graph with entanglement score more than 0.7. Such facts are not only more vulnerable to unintended interference, but also ideal candidates for red-teaming, or inclusion in post-edit evaluation suites. By spotlighting these highly connected regions, \Clare supports the prioritization of facts for testing under constrained evaluation budgets.

\section{Entanglement clusters}
\label{appendix:clusters}
\begin{figure*}
    \centering
    \begin{subfigure}[b]{0.48\textwidth}
        \centering
        \includegraphics[width=\textwidth]{Plots/test_10_app_diagrams/EleutherAI/gpt-j-6B/ripple_cluster_value_cosine_sim_01.pdf}
        \caption{\protect\shortstack{Cluster 1\\1149 facts \;|\; 397 subjects \;|\; 87.9\% cross-subject edges}}
        \label{fig:gptj_c1}
    \end{subfigure}
    \hfill
    \begin{subfigure}[b]{0.48\textwidth}
        \centering
        \includegraphics[width=\textwidth]{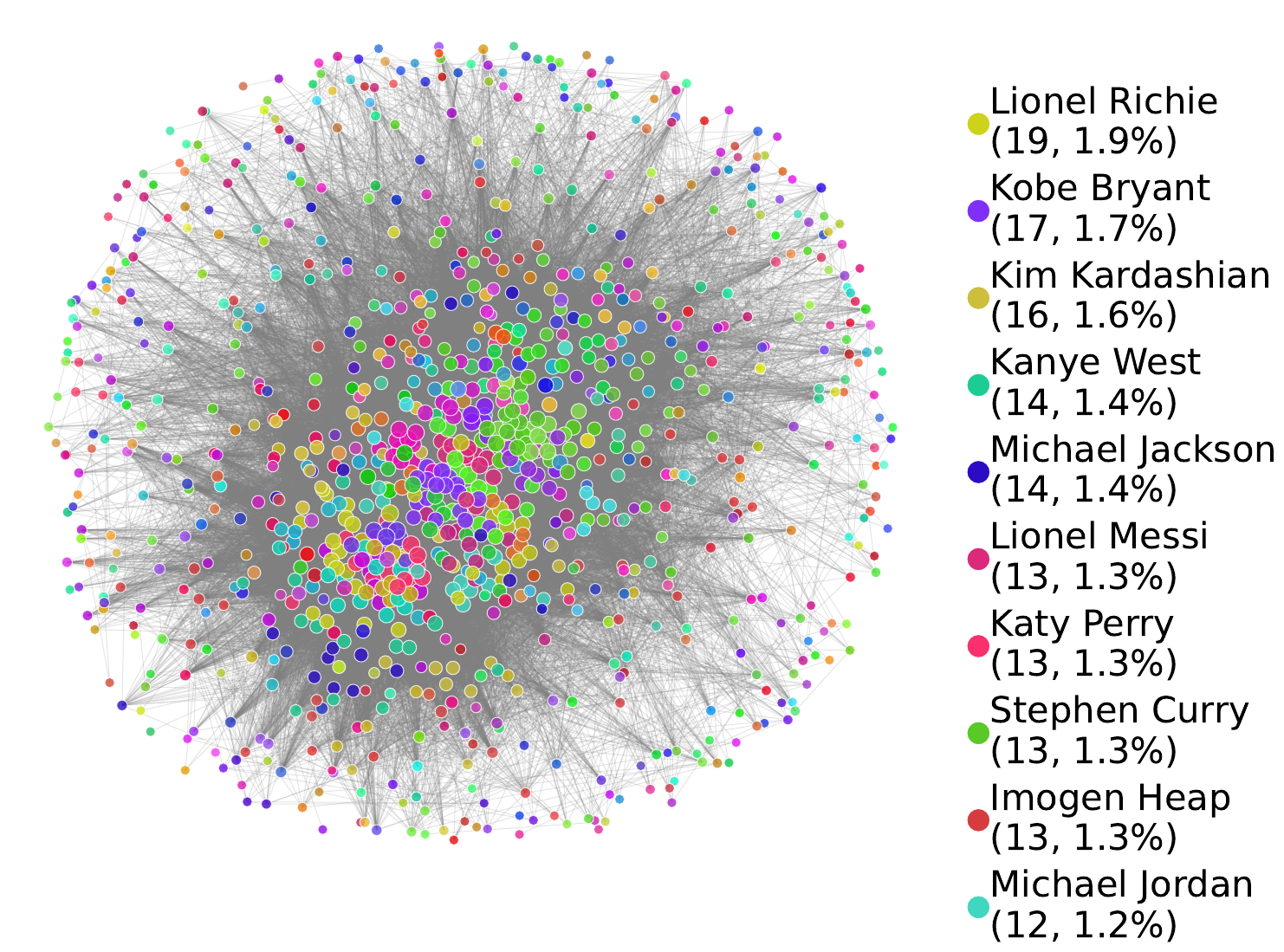}
        \caption{\protect\shortstack{Cluster 2\\1013 facts \;|\; 423 subjects \;|\; 95.8\% cross-subject edges}}
        \label{fig:gptj_c2}
    \end{subfigure}

    \caption{EleutherAI GPT-J-6B: Clusters 01-02}
\end{figure*}

\begin{figure*}
    \centering
    
    \begin{subfigure}[b]{0.48\textwidth}
        \centering
        \includegraphics[width=\textwidth]{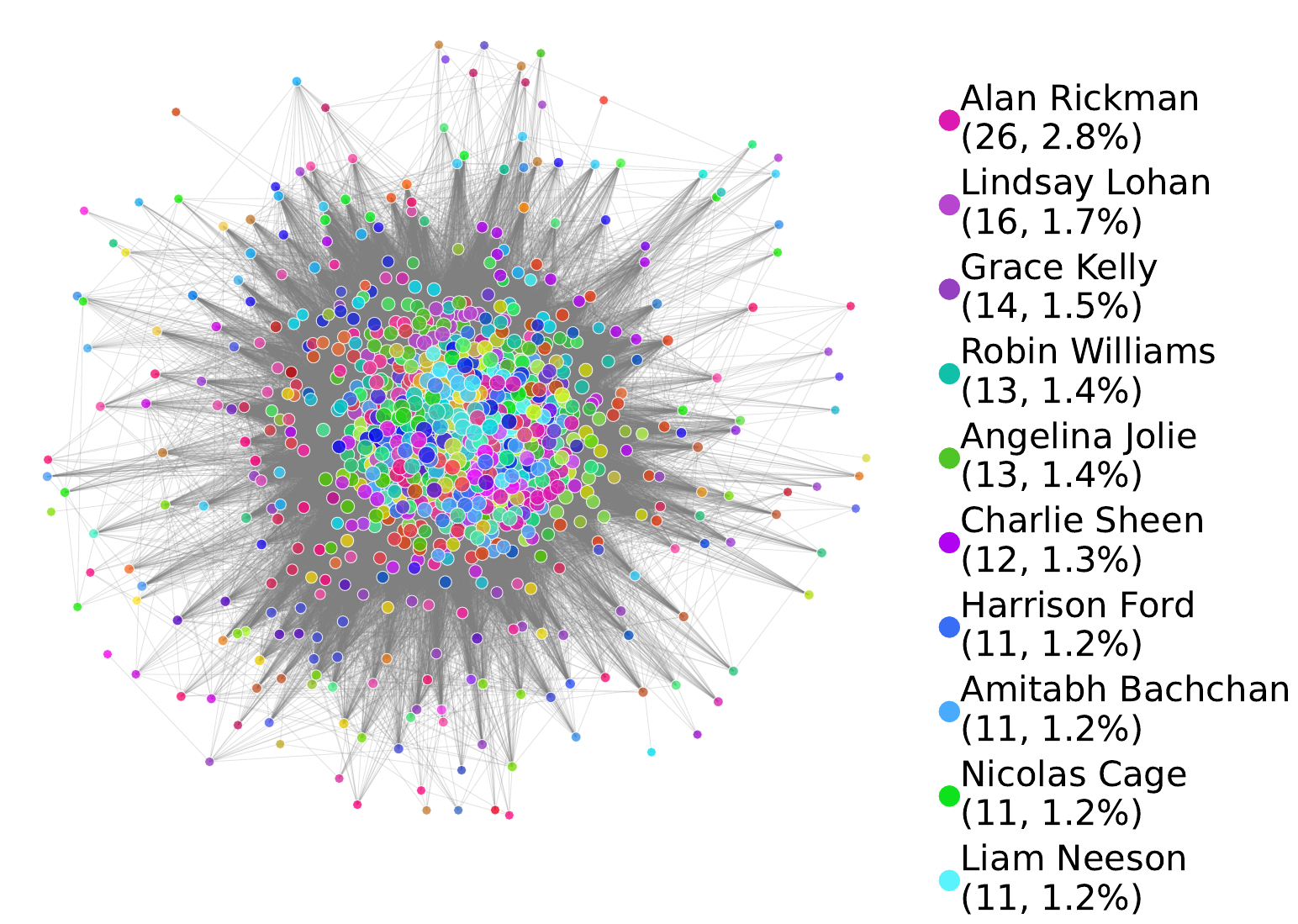}
        \caption{\protect\shortstack{Cluster 3\\929 facts \;|\; 194 subjects \;|\; 98.2\% cross-subject edges}}
        \label{fig:gptj_c3}
    \end{subfigure}
    \hfill
    \begin{subfigure}[b]{0.48\textwidth}
        \centering
        \includegraphics[width=\textwidth]{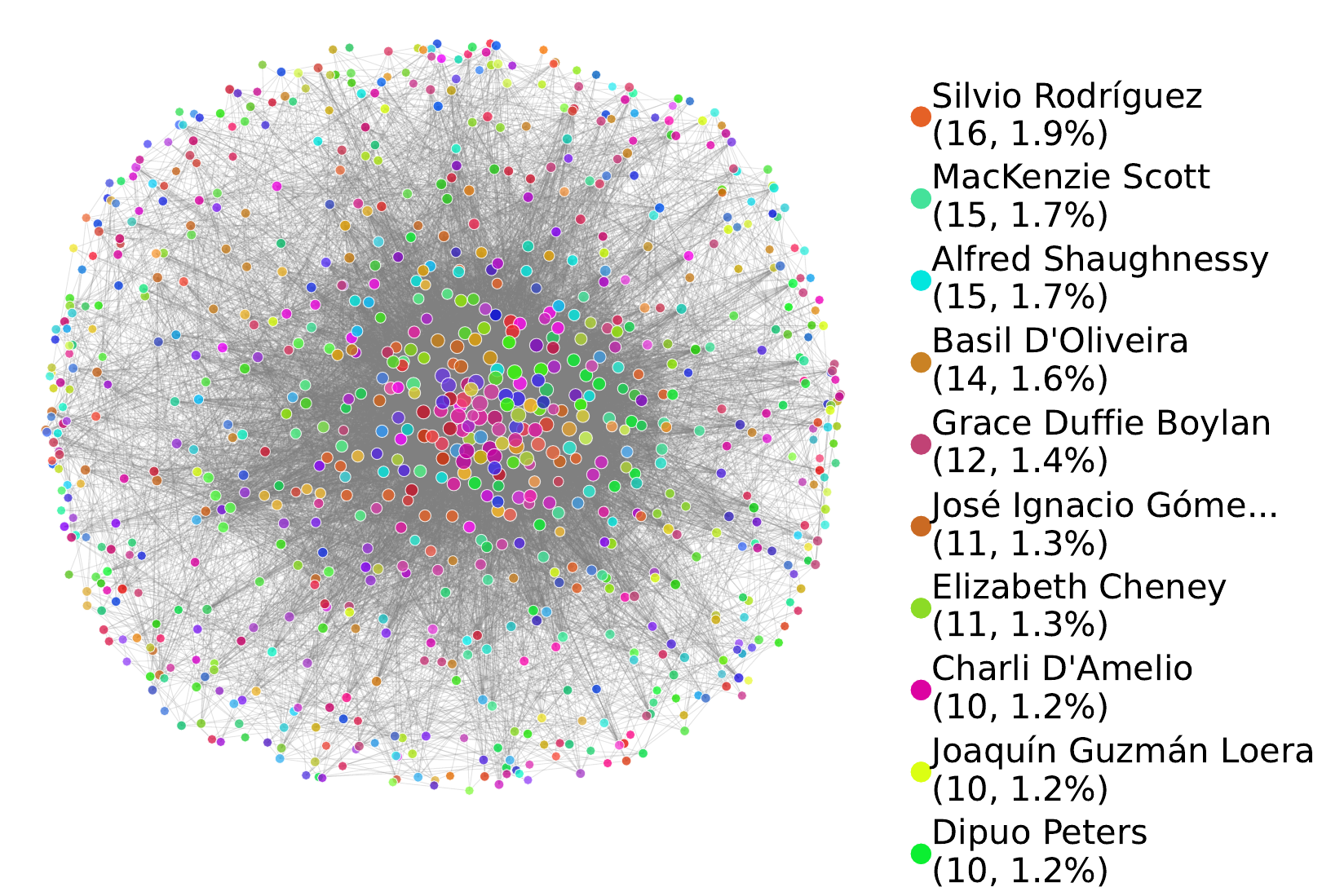}
        \caption{\protect\shortstack{Cluster 4\\861 facts \;|\; 292 subjects \;|\; 88.4\% cross-subject edges}}
        \label{fig:gptj_c4}
    \end{subfigure}

    \begin{subfigure}[b]{0.48\textwidth}
        \centering
        \includegraphics[width=\textwidth]{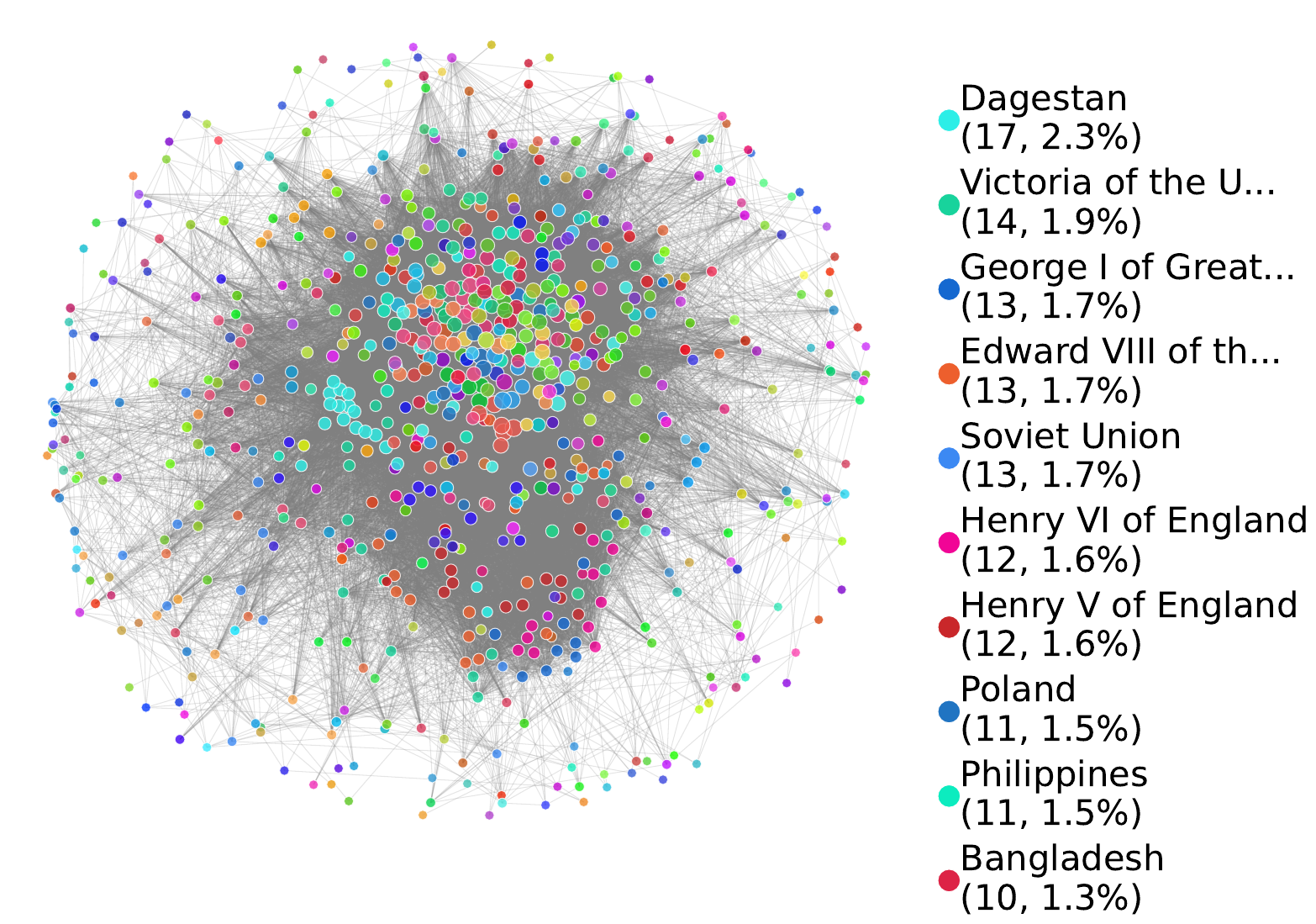}
        \caption{\protect\shortstack{Cluster 5\\744 facts \;|\; 316 subjects \;|\; 95.3\% cross-subject edges}}
        \label{fig:gptj_c5}
    \end{subfigure}
    \hfill
    \begin{subfigure}[b]{0.48\textwidth}
        \centering
        \includegraphics[width=\textwidth]{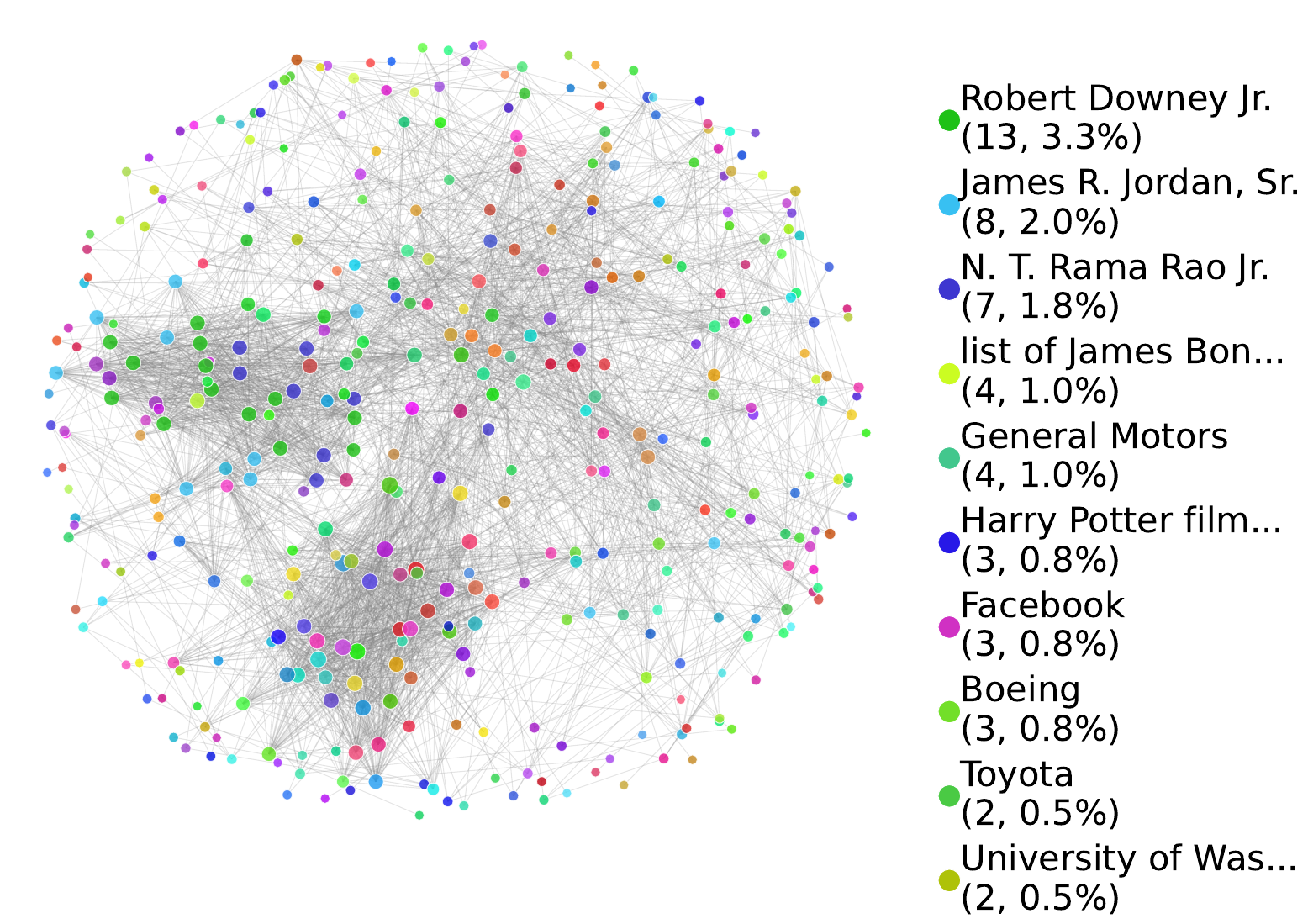}
        \caption{\protect\shortstack{Cluster 6\\394 facts \;|\; 328 subjects \;|\; 94.4\% cross-subject edges}}
        \label{fig:gptj_c6}
    \end{subfigure}

    \begin{subfigure}[b]{0.48\textwidth}
        \centering
        \includegraphics[width=\textwidth]{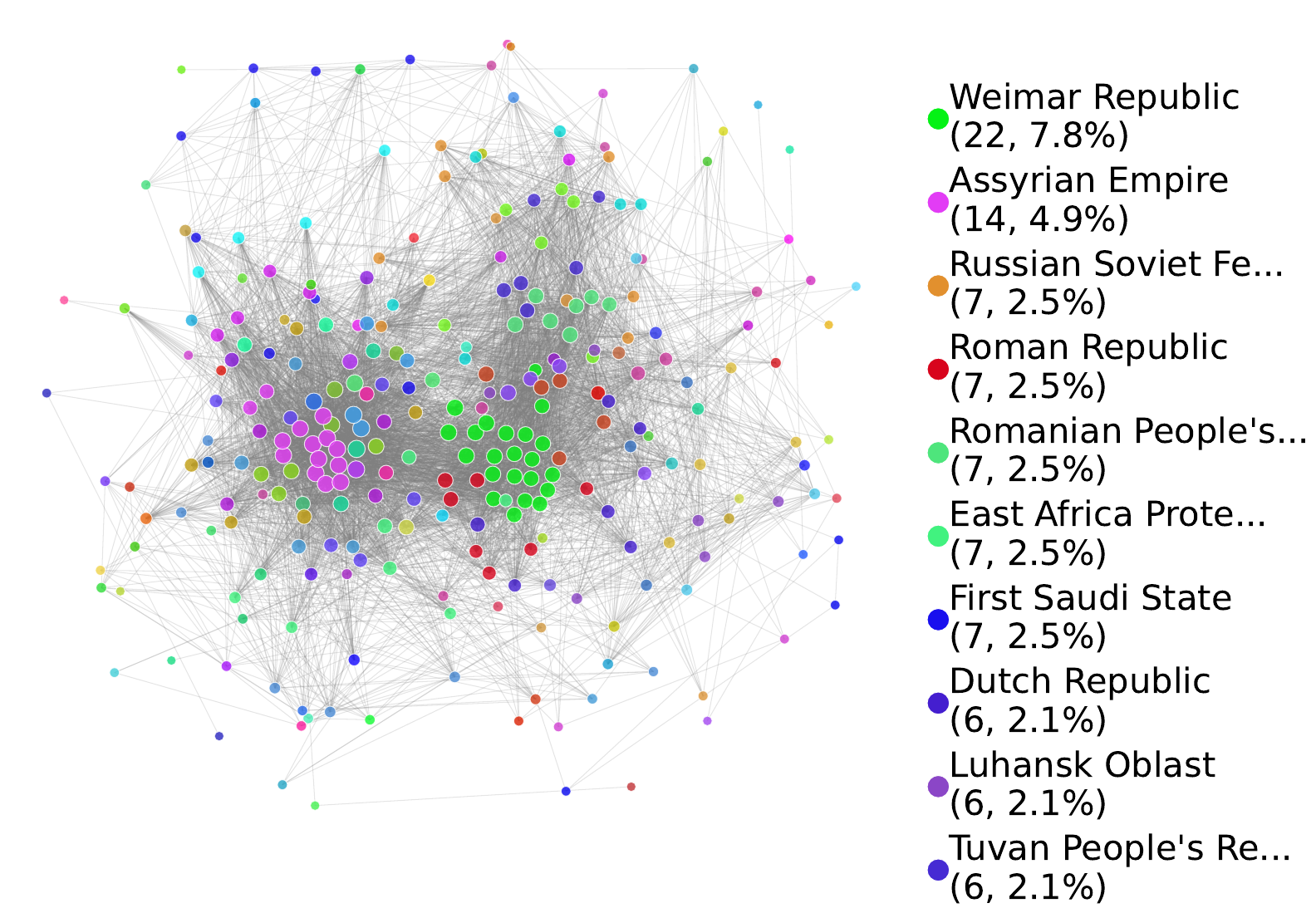}
        \caption{\protect\shortstack{Cluster 7\\283 facts \;|\; 120 subjects \;|\; 90.0\% cross-subject edges}}
        \label{fig:gptj_c7}
    \end{subfigure}
    \hfill
    \begin{subfigure}[b]{0.48\textwidth}
        \centering
        \includegraphics[width=\textwidth]{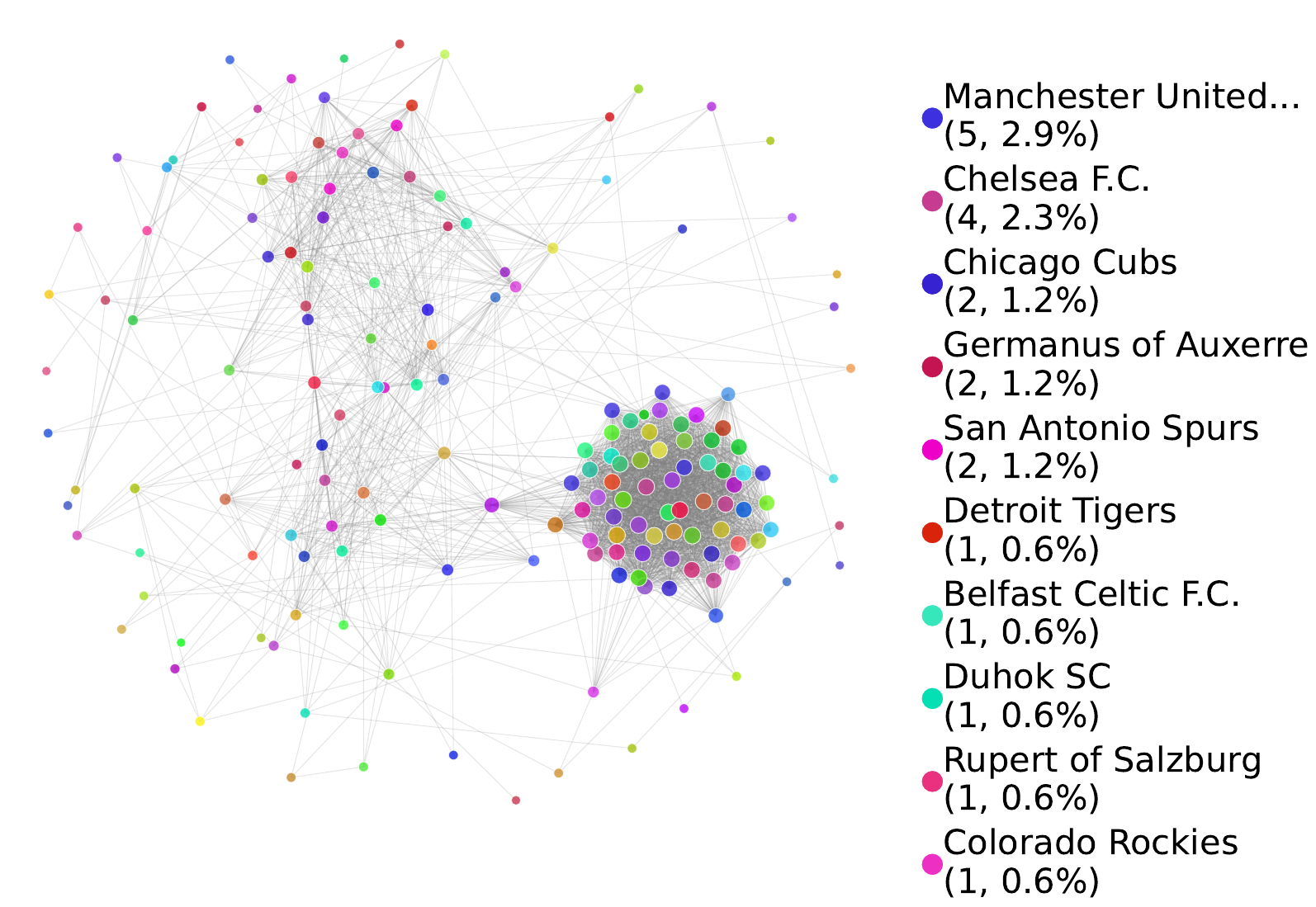}
        \caption{\protect\shortstack{Cluster 8\\172 facts \;|\; 162 subjects \;|\; 99.3\% cross-subject edges}}
        \label{fig:gptj_c8}
    \end{subfigure}

    \caption{EleutherAI GPT-J-6B: Clusters 03-08}
\end{figure*}

\begin{figure*}
    \centering
    
    \begin{subfigure}[b]{0.48\textwidth}
        \centering
        \includegraphics[width=\textwidth]{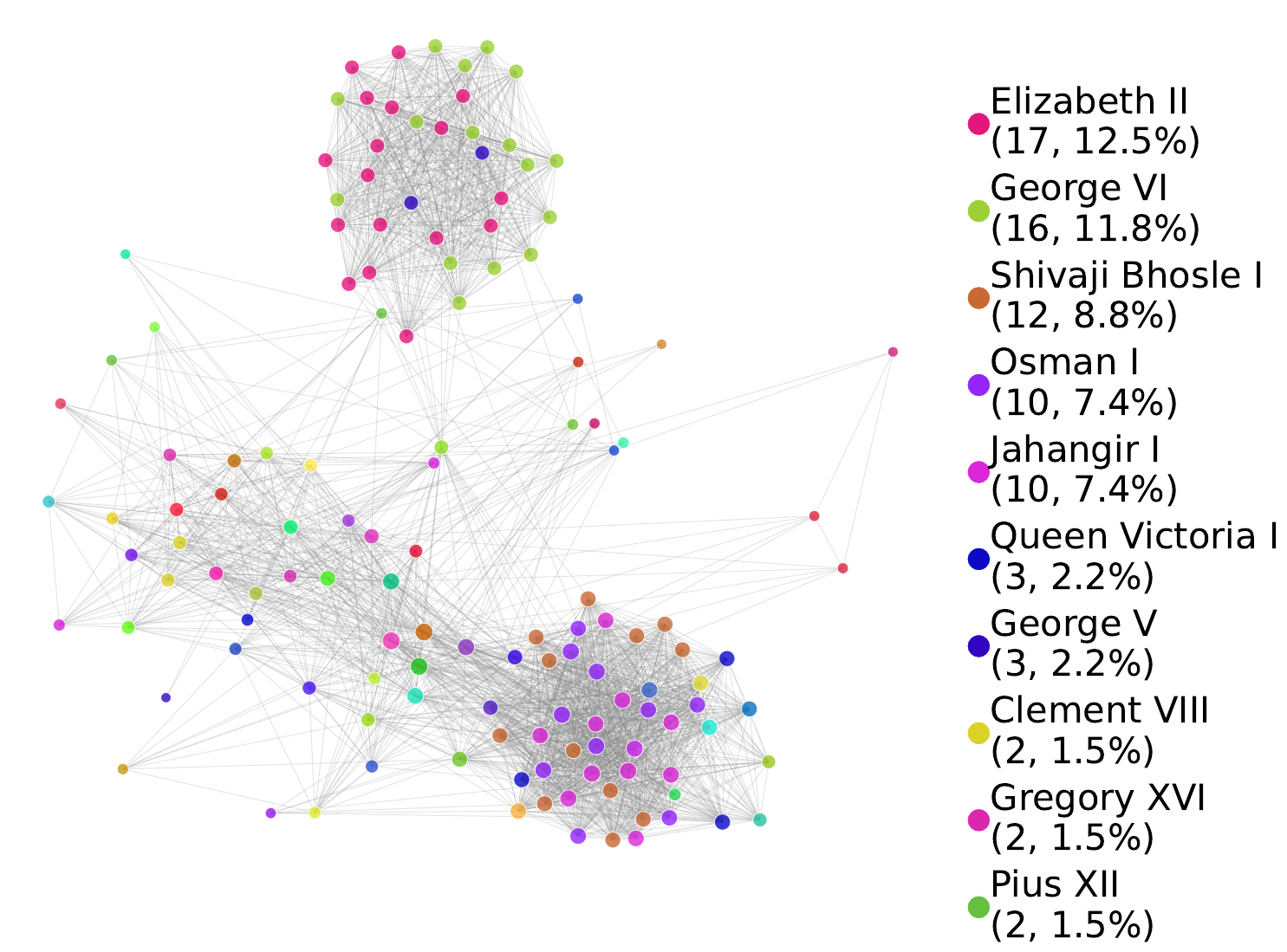}
        \caption{\protect\shortstack{Cluster 9\\136 facts \;|\; 66 subjects \;|\; 81.3\% cross-subject edges}}
        \label{fig:gptj_c9}
    \end{subfigure}
    \hfill
    \begin{subfigure}[b]{0.48\textwidth}
        \centering
        \includegraphics[width=\textwidth]{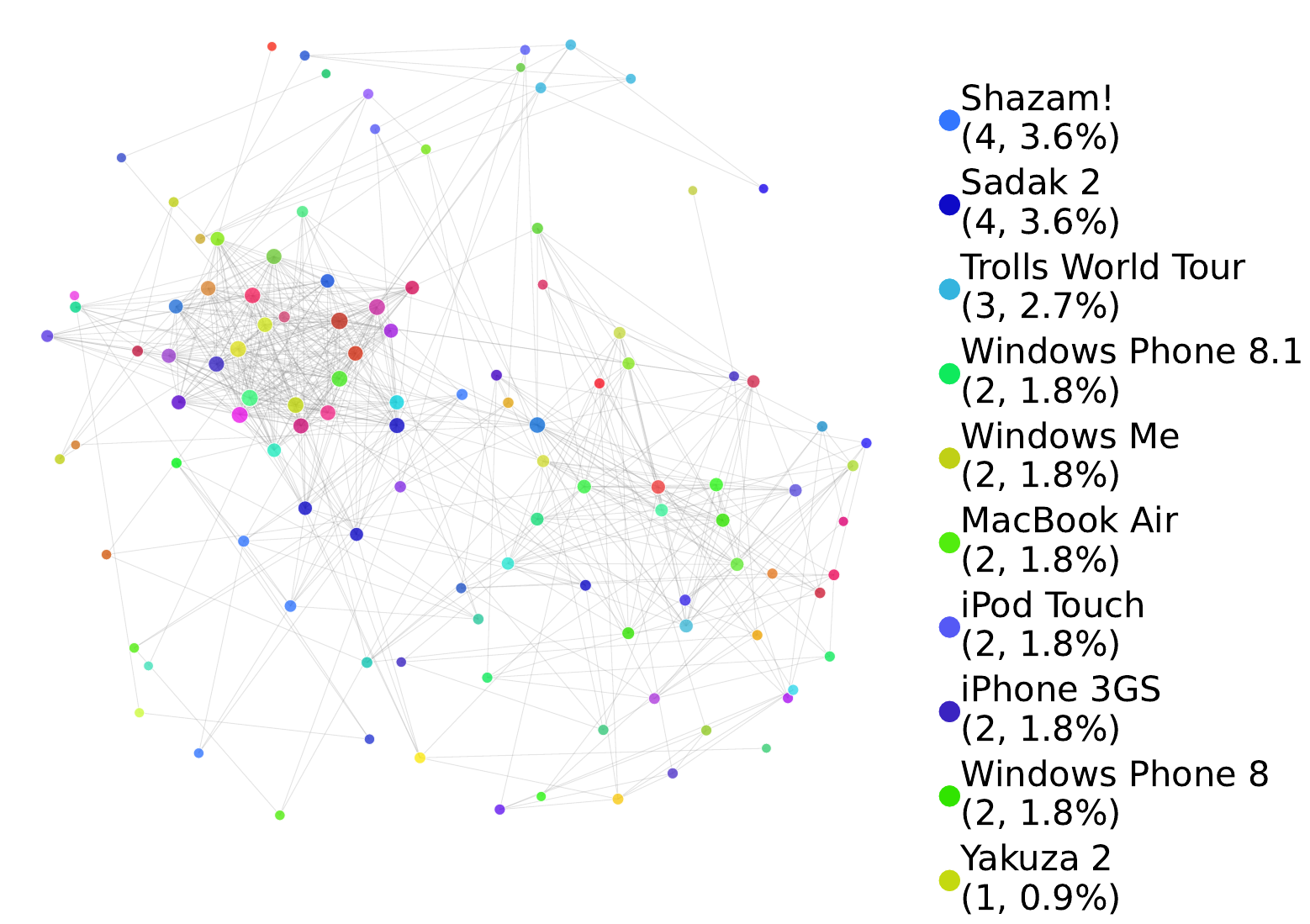}
        \caption{\protect\shortstack{Cluster 10\\112 facts \;|\; 98 subjects \;|\; 96.6\% cross-subject edges}}
        \label{fig:gptj_c10}
    \end{subfigure}
    
    \begin{subfigure}[b]{0.48\textwidth}
        \centering
        \includegraphics[width=\textwidth]{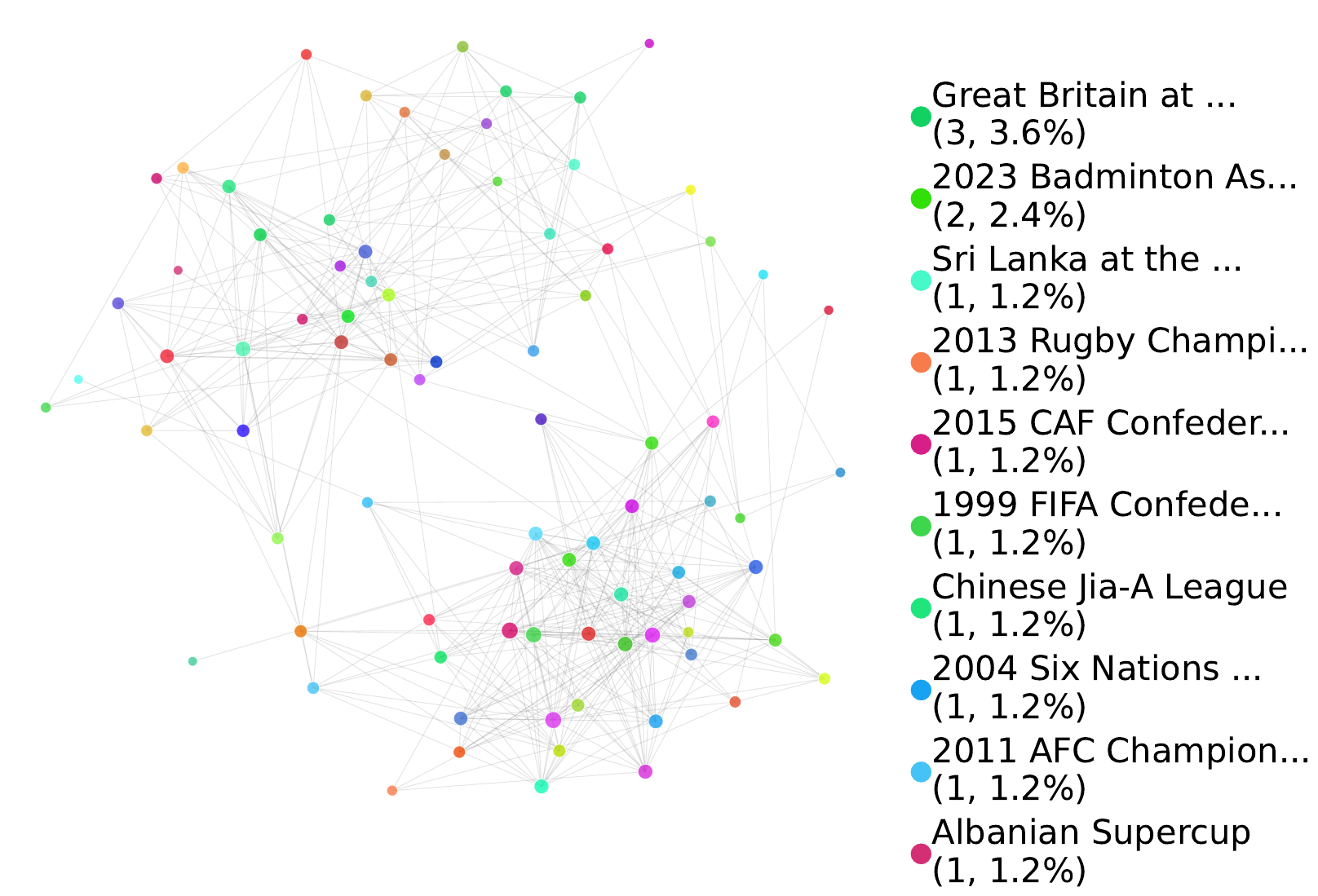}
        \caption{\protect\shortstack{Cluster 11\\83 facts \;|\; 80 subjects \;|\; 99.0\% cross-subject edges}}
        \label{fig:gptj_c11}
    \end{subfigure}
    \begin{subfigure}[b]{0.48\textwidth}
        \centering
        \includegraphics[width=\textwidth]{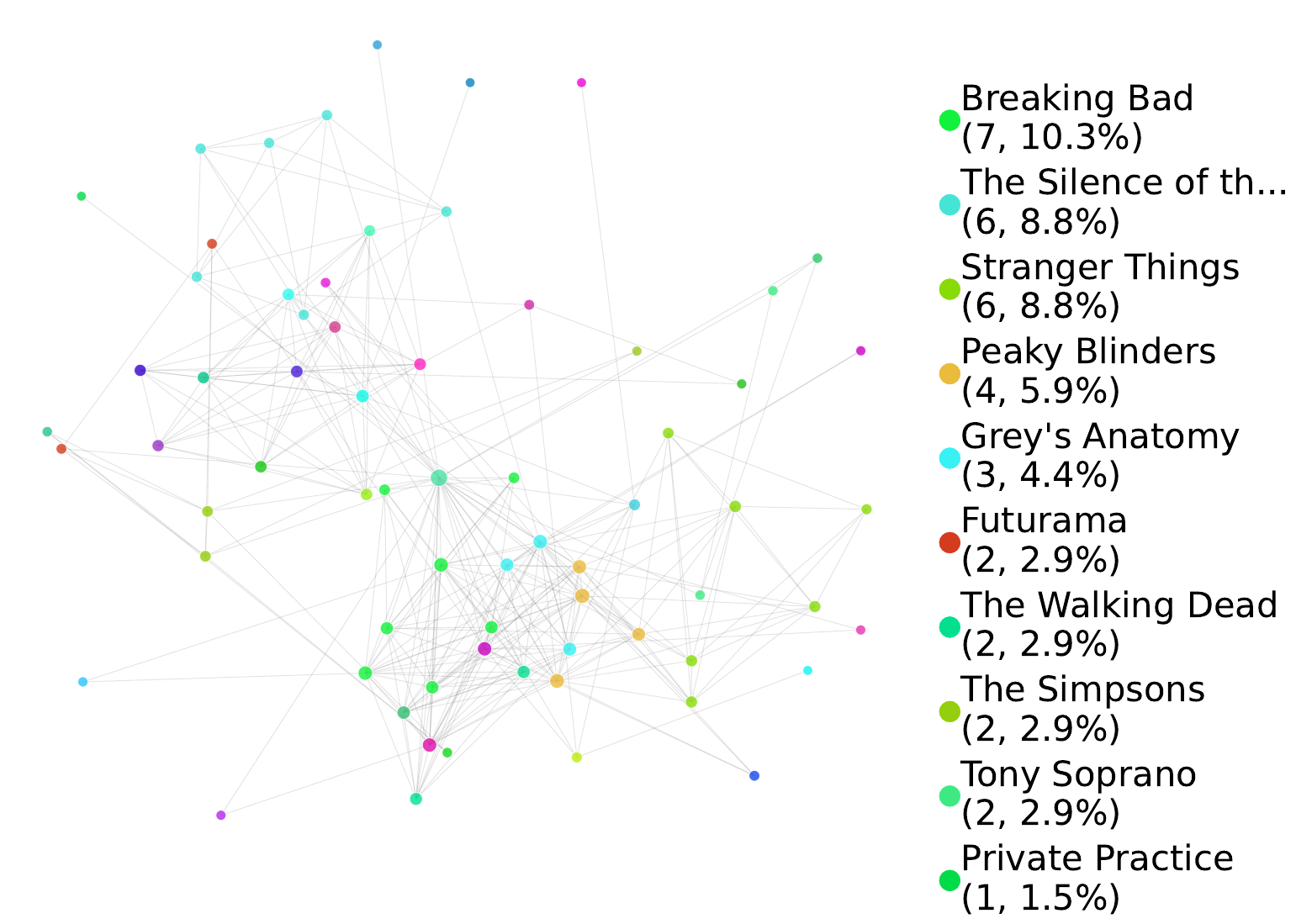}
        \caption{\protect\shortstack{Cluster 12\\68 facts \;|\; 43 subjects \;|\; 78.7\% cross-subject edges}}
        \label{fig:gptj_c12}
    \end{subfigure}

    \begin{subfigure}[b]{0.48\textwidth}
        \centering
        \includegraphics[width=\textwidth]{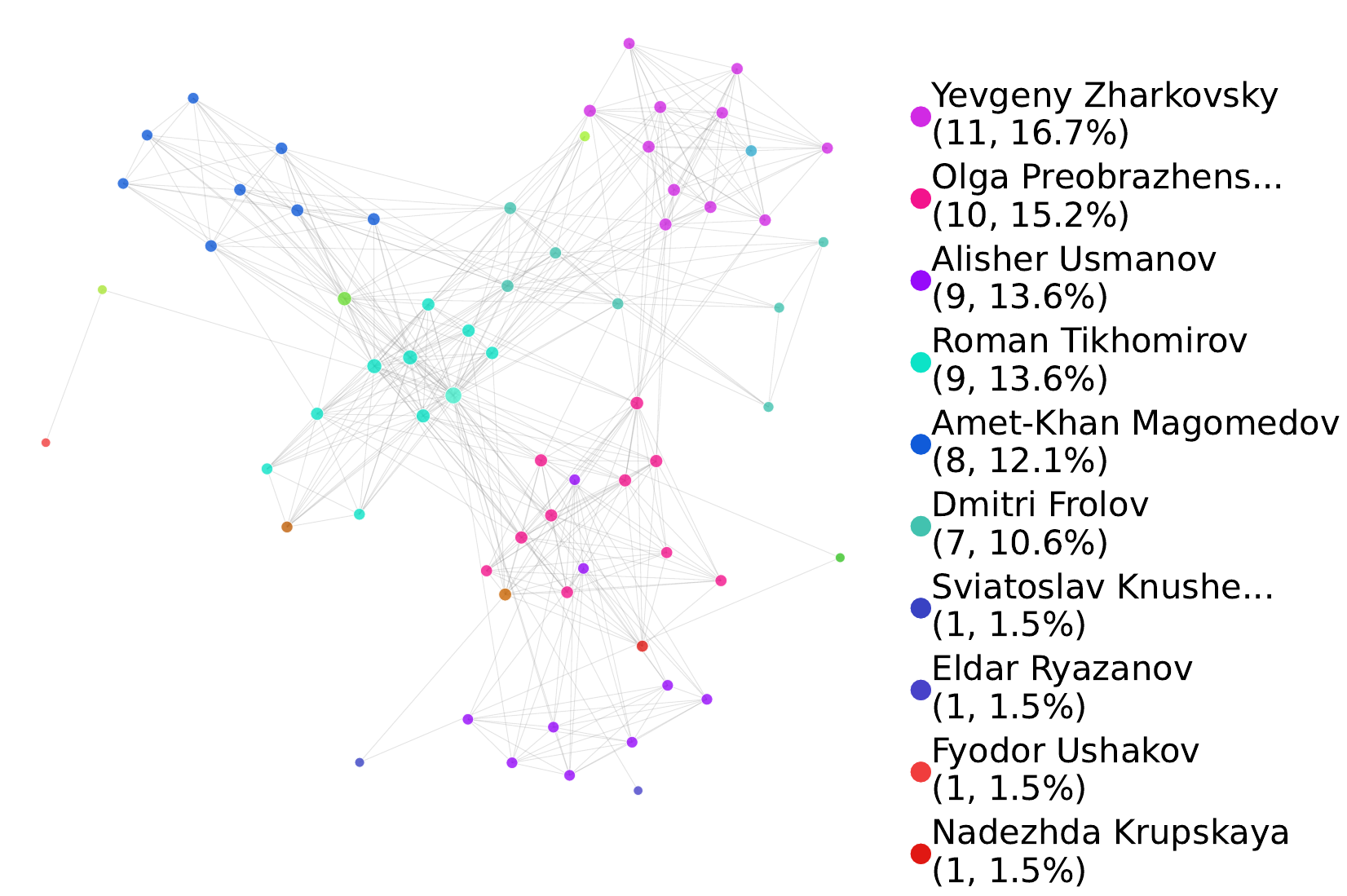}
        \caption{\protect\shortstack{Cluster 13\\66 facts \;|\; 18 subjects \;|\; 46.2\% cross-subject edges}}
        \label{fig:gptj_c13}
    \end{subfigure}
    
    \caption{EleutherAI GPT-J-6B: Clusters 09-13}
\end{figure*}

\begin{figure*}
    \centering
    \begin{subfigure}[b]{0.48\textwidth}
        \centering
        \includegraphics[width=\textwidth]{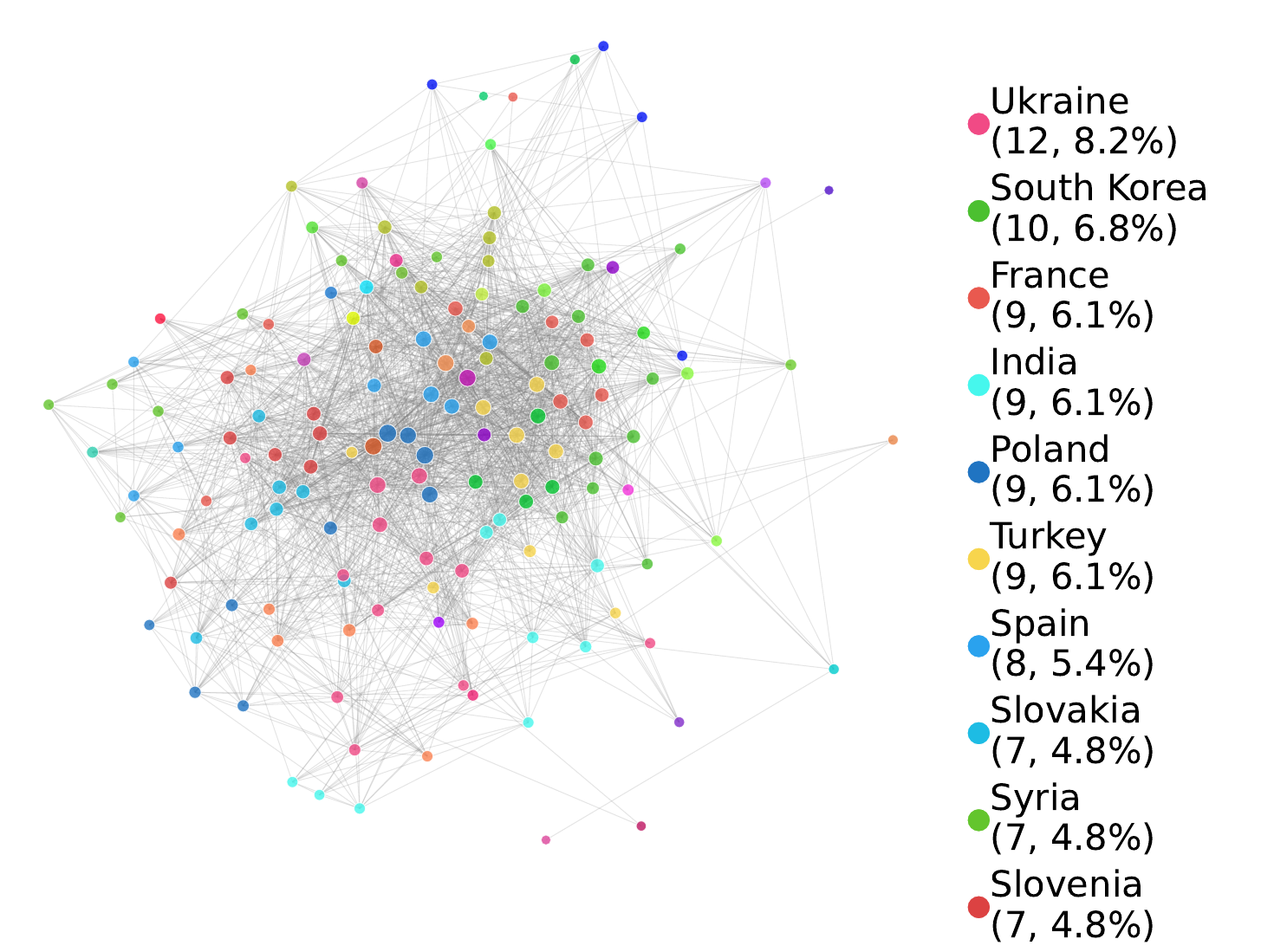}
        \caption{\protect\shortstack{Cluster 1\\147 facts \;|\; 47 subjects \;|\; 81.8\% cross-subject edges}}
        \label{fig:gpt2_c1}
    \end{subfigure}
    \hfill
    \begin{subfigure}[b]{0.48\textwidth}
        \centering
        \includegraphics[width=\textwidth]{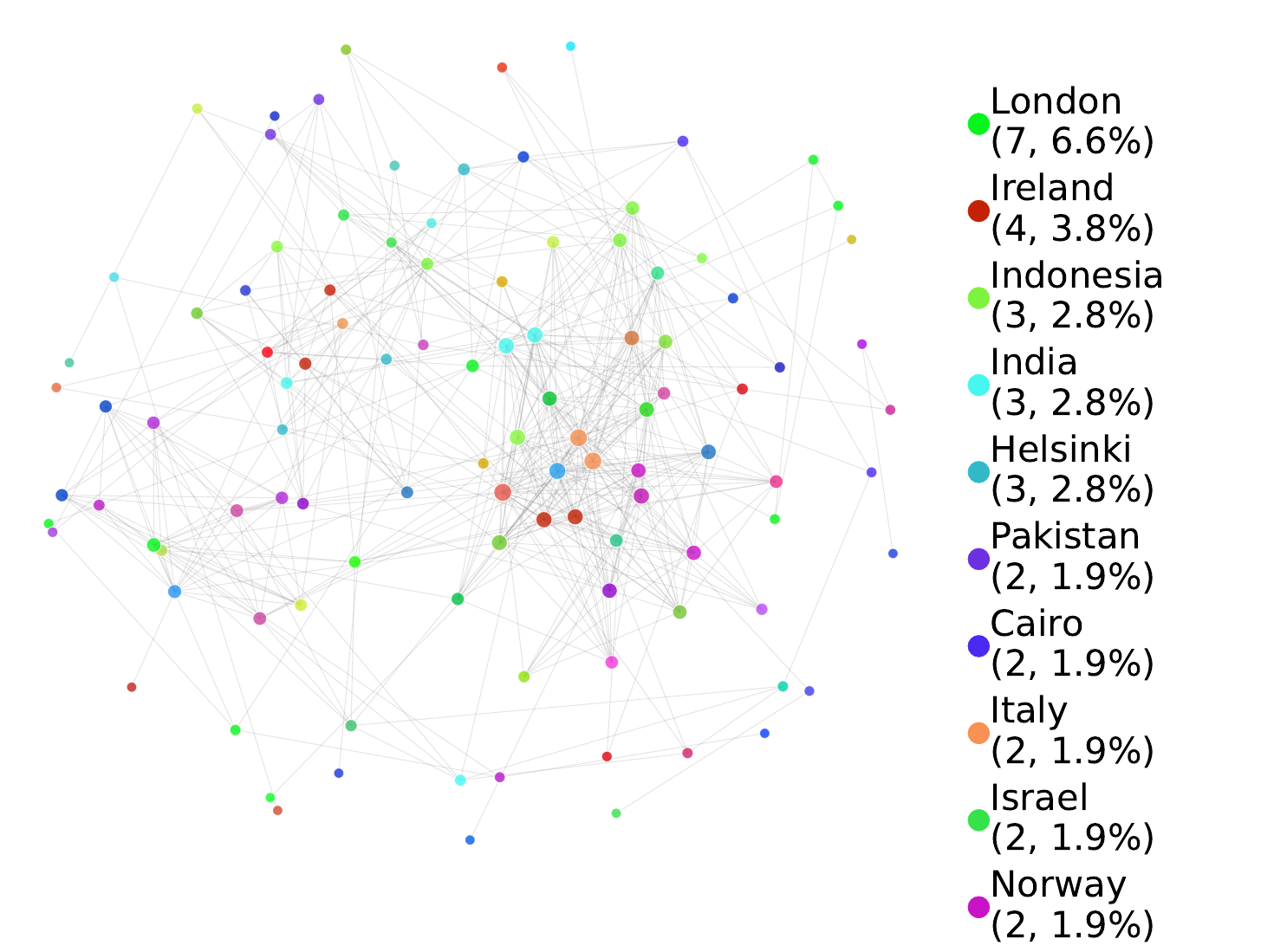}
        \caption{\protect\shortstack{Cluster 2\\106 facts \;|\; 74 subjects \;|\; 91.1\% cross-subject edges}}
        \label{fig:gpt2_c2}
    \end{subfigure}

    \begin{subfigure}[b]{0.48\textwidth}
        \centering
        \includegraphics[width=\textwidth]{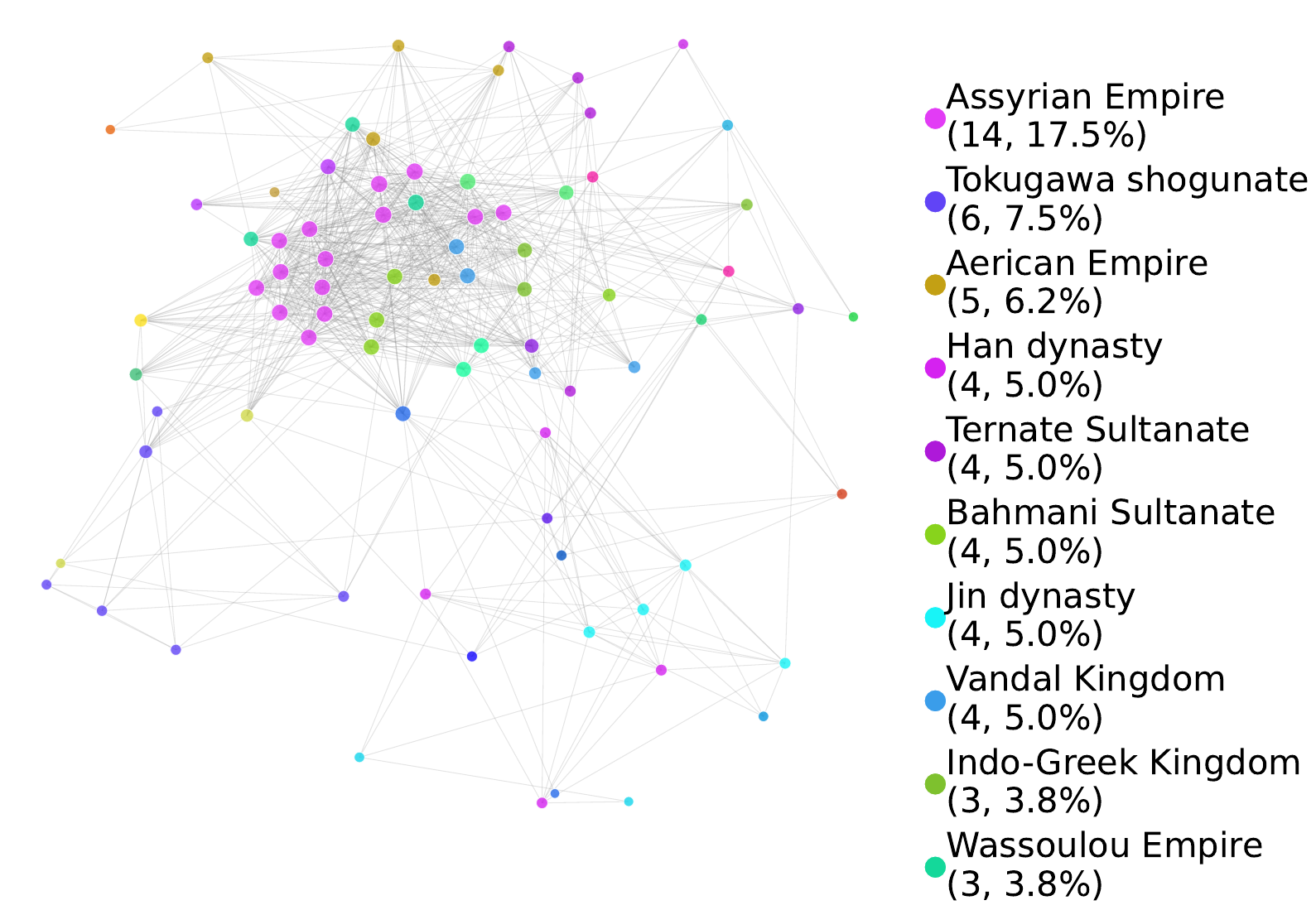}
        \caption{\protect\shortstack{Cluster 3\\80 facts \;|\; 31 subjects \;|\; 78.1\% cross-subject edges}}
        \label{fig:gpt2_c3}
    \end{subfigure}
    \hfill
    \centering
    \begin{subfigure}[b]{0.48\textwidth}
        \centering
        \includegraphics[width=\textwidth]{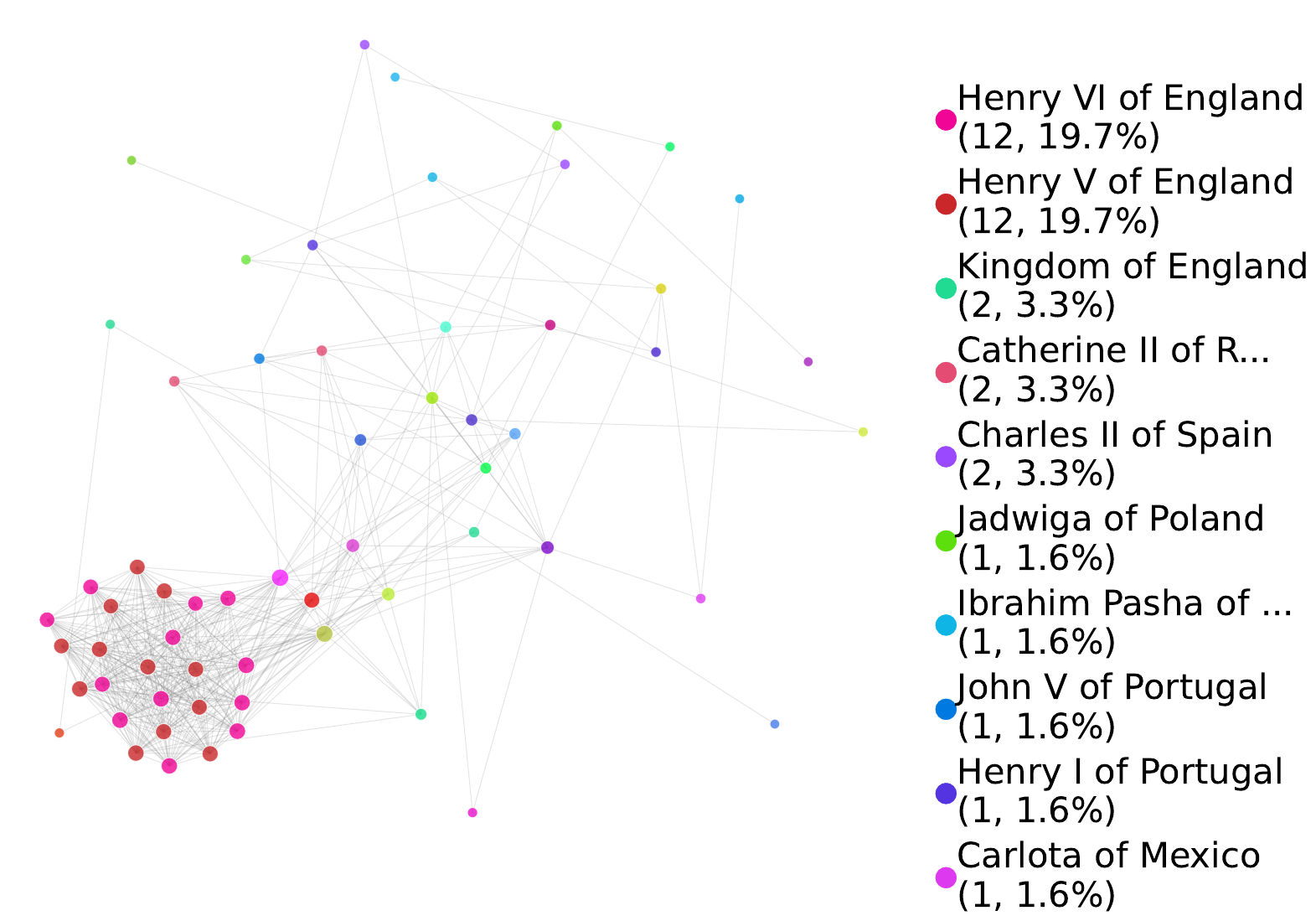}
        \caption{\protect\shortstack{Cluster 4\\61 facts \;|\; 36 subjects \;|\; 69.7\% cross-subject edges}}
        \label{fig:gpt2_c4}
    \end{subfigure}

    \begin{subfigure}[b]{0.48\textwidth}
        \centering
        \includegraphics[width=\textwidth]{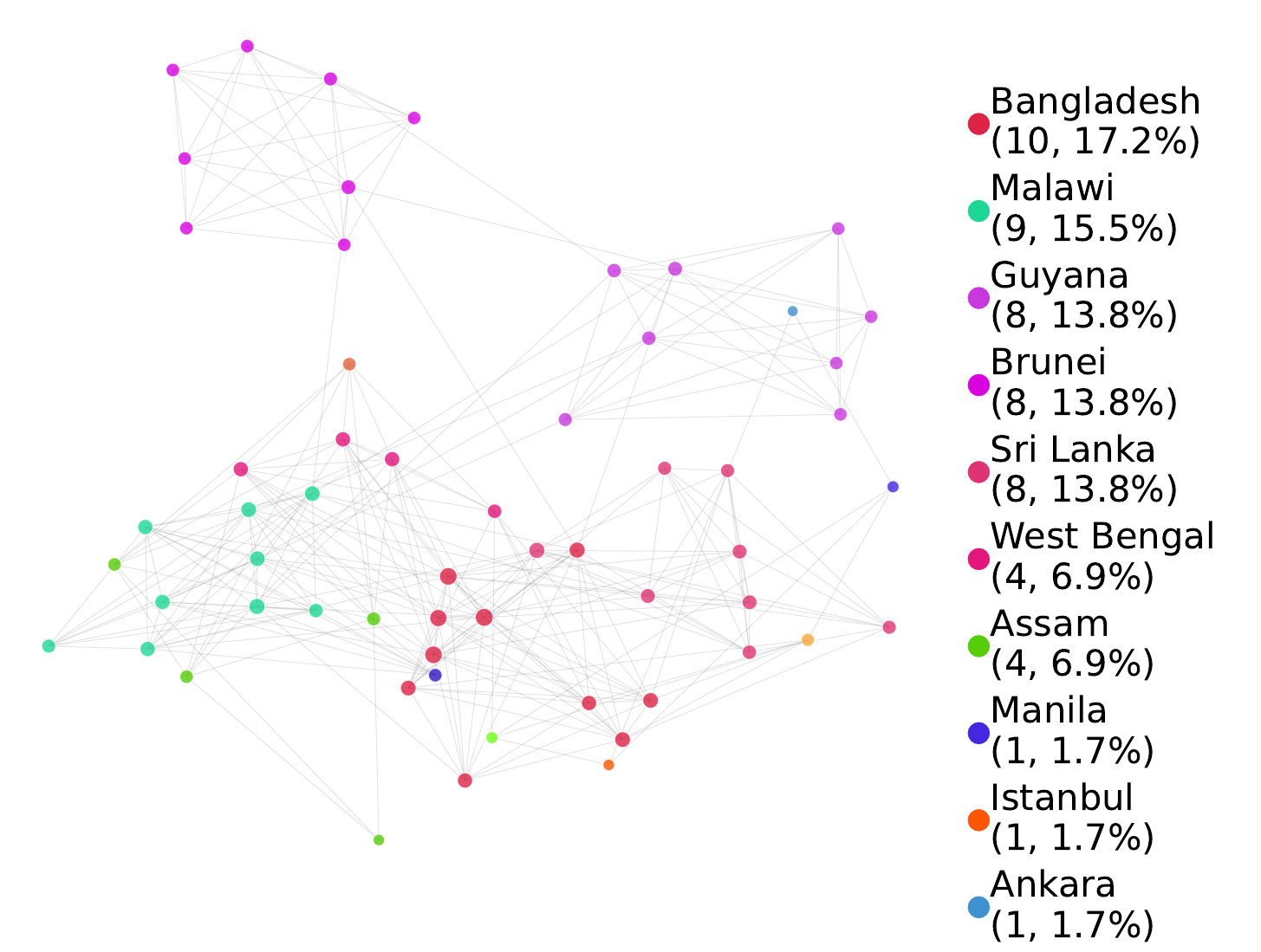}
        \caption{\protect\shortstack{Cluster 5\\58 facts \;|\; 14 subjects \;|\; 34.2\% cross-subject edges}}
        \label{fig:gpt2_c5}
    \end{subfigure}
    \hfill
    \begin{subfigure}[b]{0.48\textwidth}
        \centering
        \includegraphics[width=\textwidth]{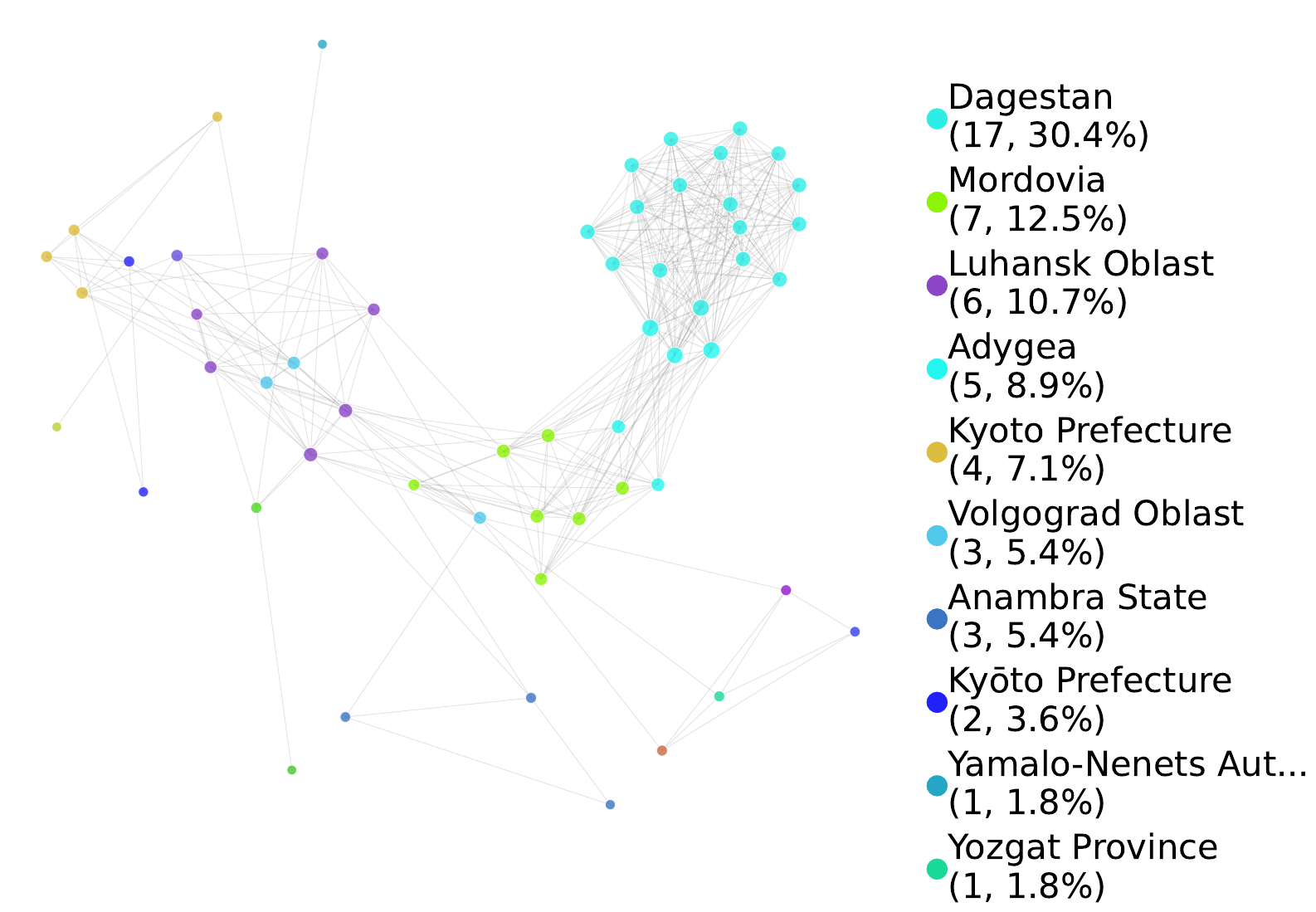}
        \caption{\protect\shortstack{Cluster 6\\56 facts \;|\; 17 subjects \;|\; 44.6\% cross-subject edges}}
        \label{fig:gpt2_c6}
    \end{subfigure}
    
    \caption{GPT2-XL: Clusters 01-06}
\end{figure*}

\begin{figure*}
    \centering

    \begin{subfigure}[b]{0.48\textwidth}
        \centering
        \includegraphics[width=\textwidth]{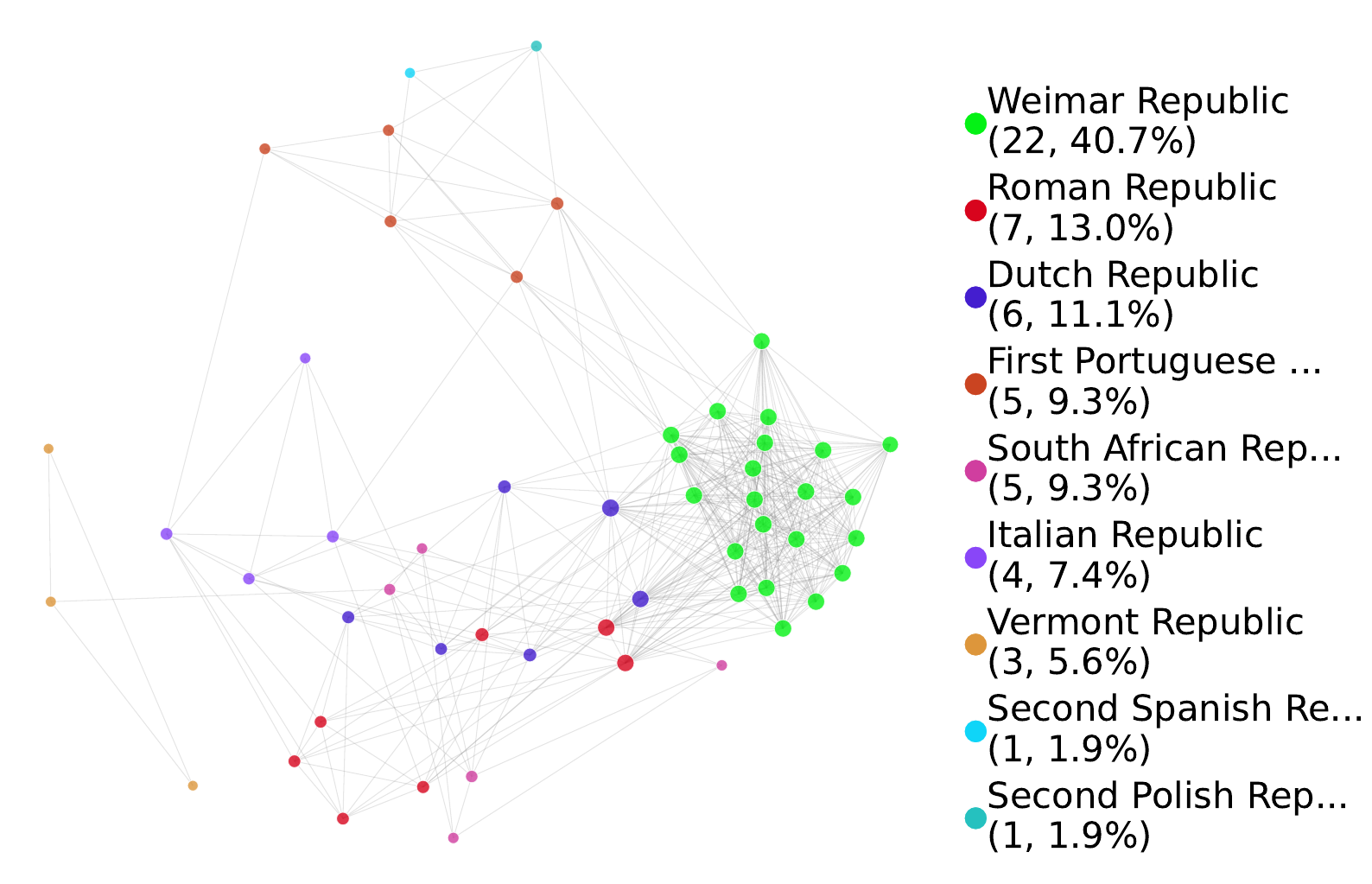}
    \end{subfigure}
    \caption{GPT2-XL: Cluster 07\\\protect\shortstack{54 facts \;|\; 9 subjects \;|\; 29.4\% cross-subject edges}}
    \label{fig:gpt2_c7}
\end{figure*}

\begin{figure*}
    \centering
    \begin{subfigure}[b]{0.48\textwidth}
        \centering
        \includegraphics[width=\textwidth]{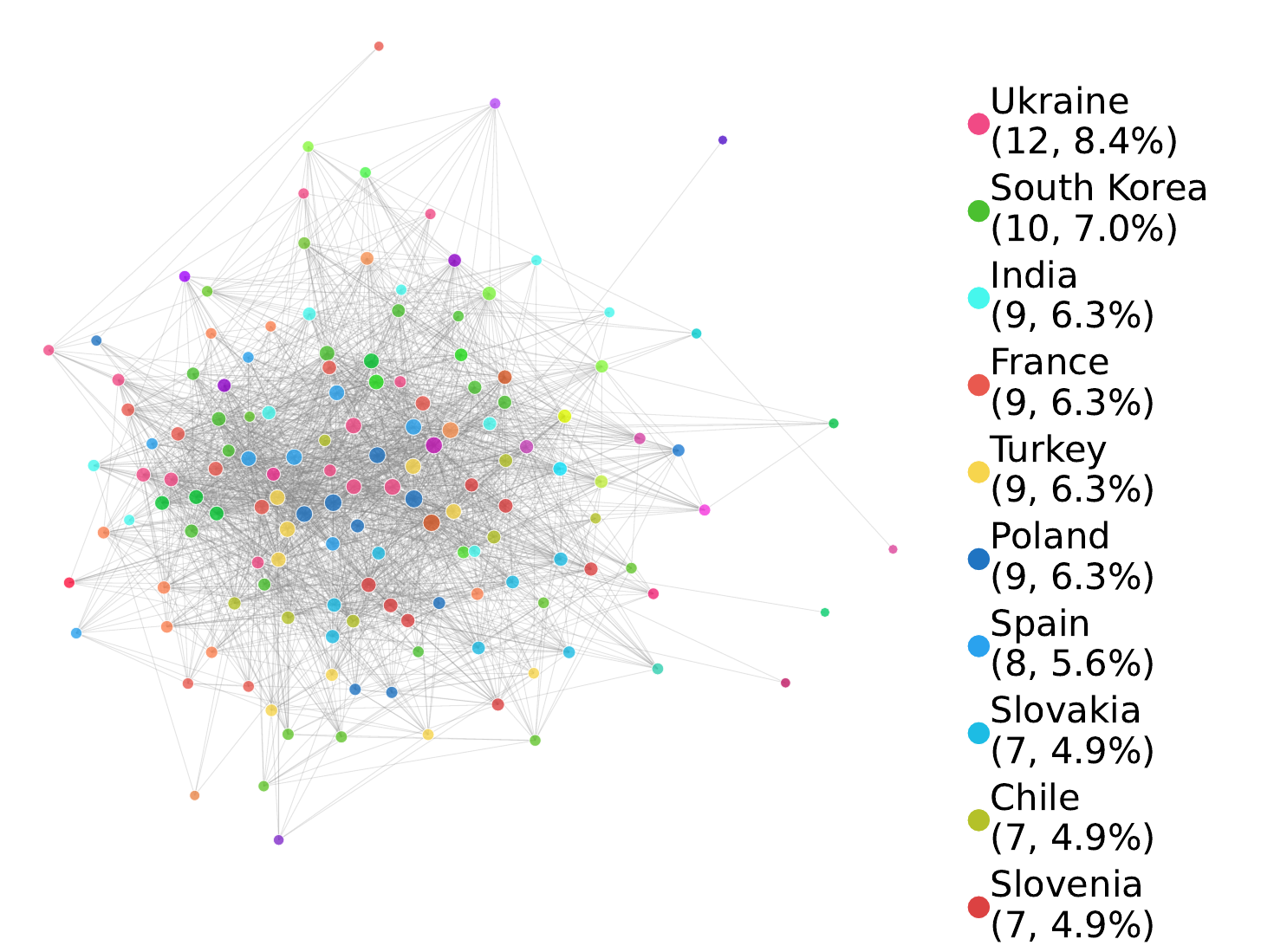}
        \caption{\protect\shortstack{Cluster 1\\143 facts \;|\; 46 subjects \;|\; 81.8\% cross-subject edges}}
        \label{fig:llama_c1}
    \end{subfigure}
    \hfill
    \begin{subfigure}[b]{0.48\textwidth}
        \centering
        \includegraphics[width=\textwidth]{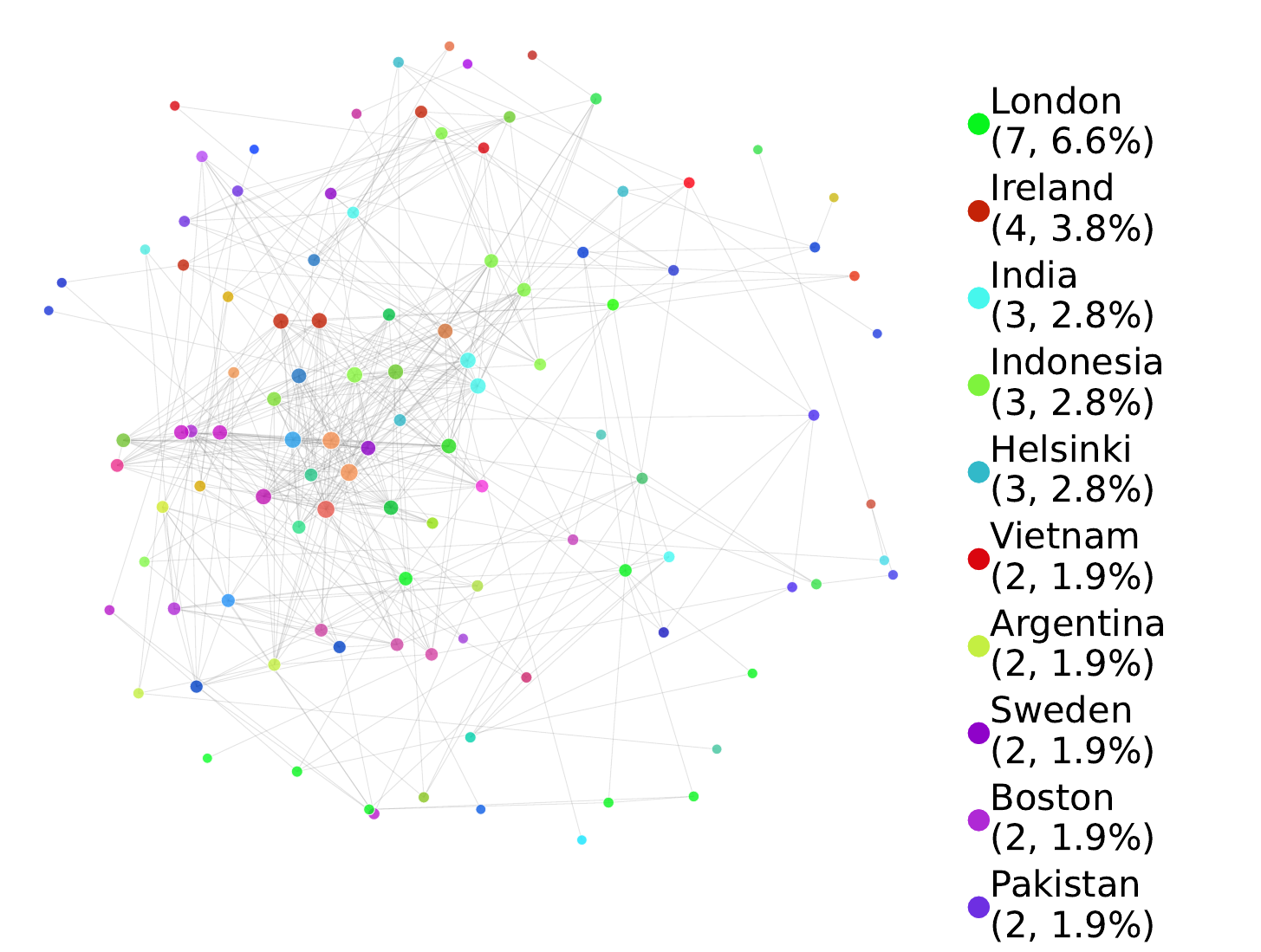}
        \caption{\protect\shortstack{Cluster 2\\106 facts \;|\; 74 subjects \;|\; 91.1\% cross-subject edges}}
        \label{fig:llama_c2}
    \end{subfigure}

    \begin{subfigure}[b]{0.48\textwidth}
        \centering
        \includegraphics[width=\textwidth]{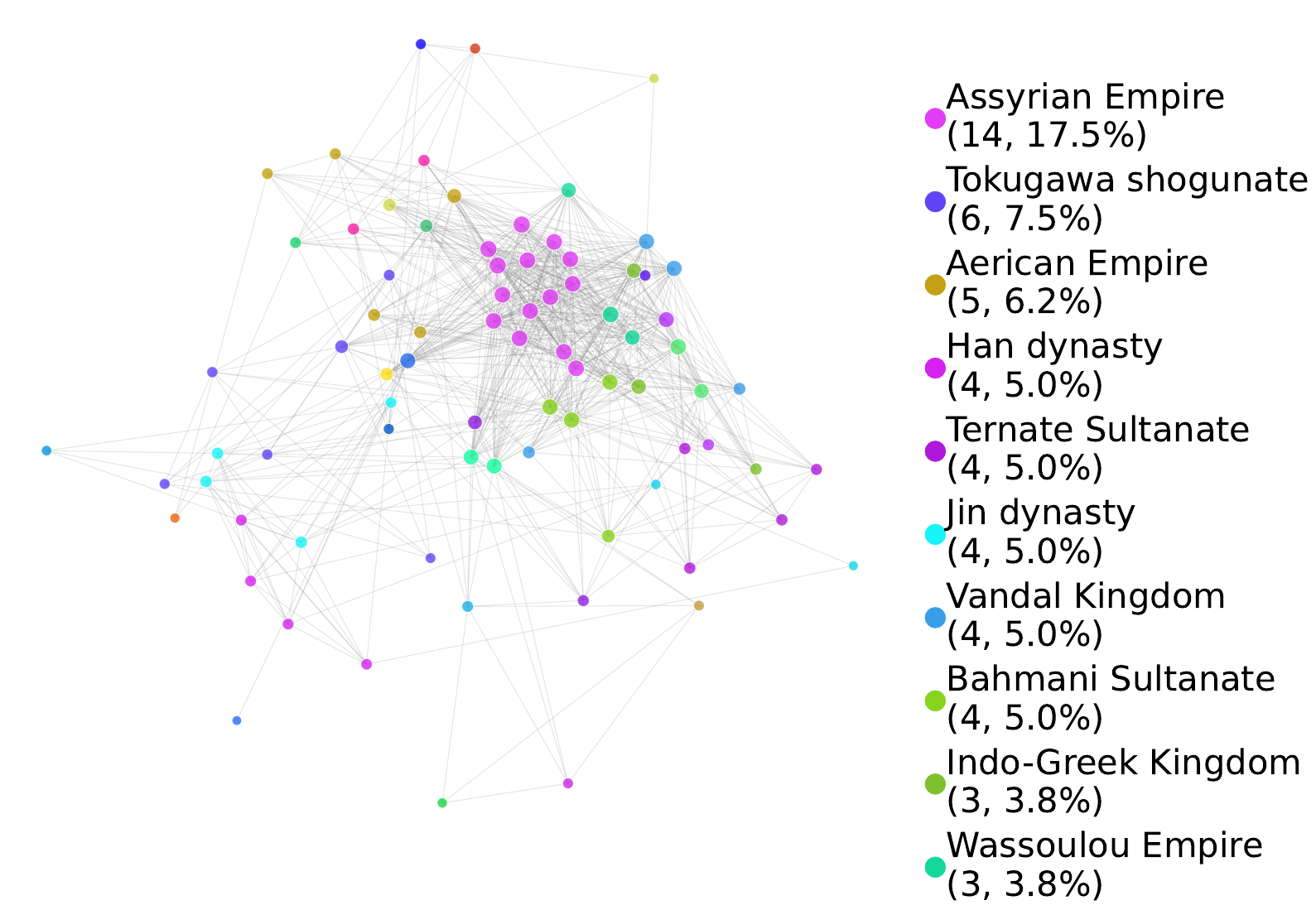}
        \caption{\protect\shortstack{Cluster 3\\80 facts \;|\; 31 subjects \;|\; 78.1\% cross-subject edges}}
        \label{fig:llama_c3}
    \end{subfigure}
    \hfill
    \begin{subfigure}[b]{0.48\textwidth}
        \centering
        \includegraphics[width=\textwidth]{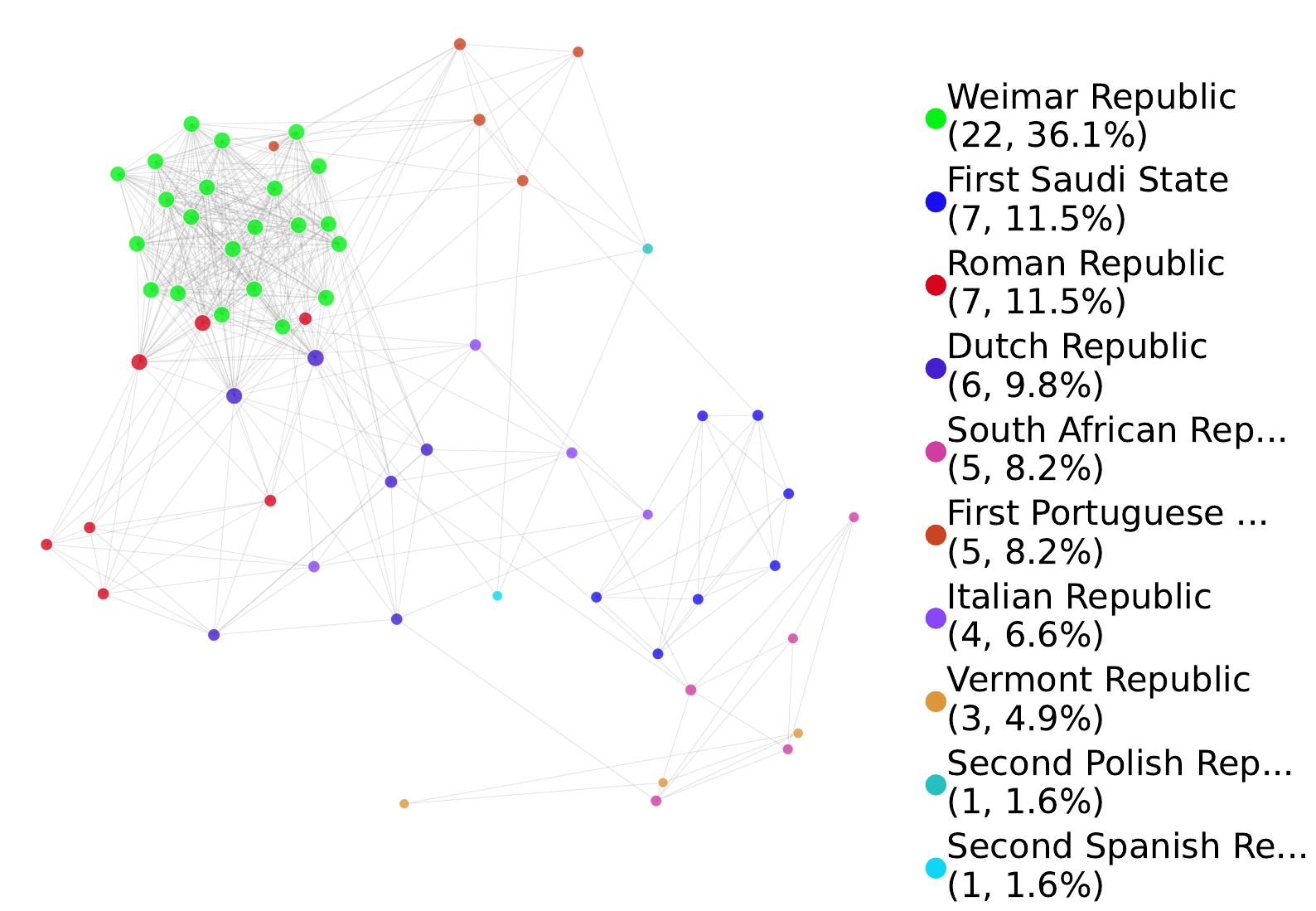}
        \caption{\protect\shortstack{Cluster 4\\61 facts \;|\; 10 subjects \;|\; 28.1\% cross-subject edges}}
        \label{fig:llama_c4}
    \end{subfigure}
    \caption{Llama3-8B: Clusters 01-04}
\end{figure*}

\begin{figure*}
    \centering
    \begin{subfigure}[b]{0.48\textwidth}
        \centering
        \includegraphics[width=\textwidth]{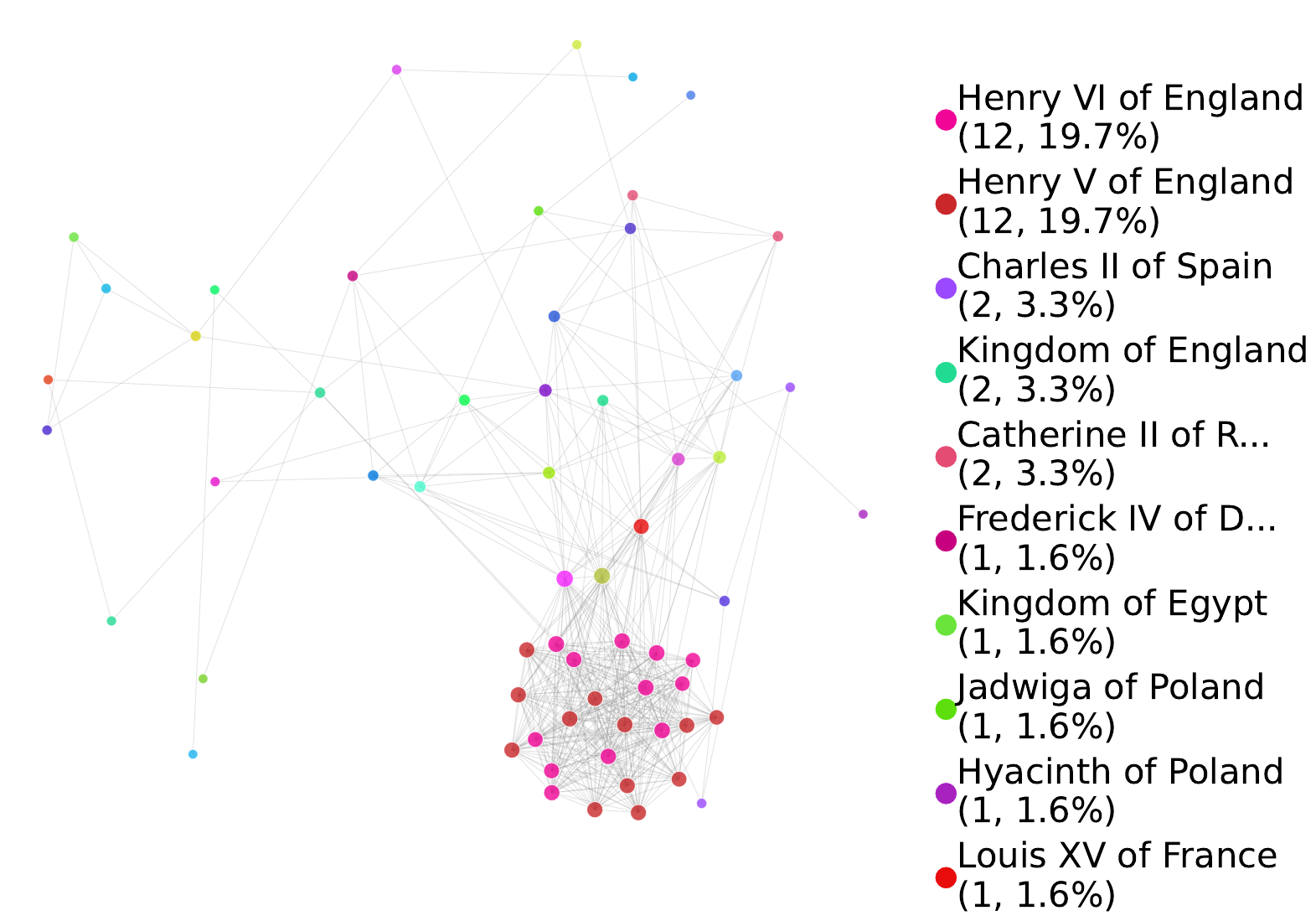}
        \caption{\protect\shortstack{Cluster 5\\61 facts \;|\; 36 subjects \;|\; 69.7\% cross-subject edges}}
        \label{fig:llama_c5}
    \end{subfigure}
    \hfill
    \begin{subfigure}[b]{0.48\textwidth}
        \centering
        \includegraphics[width=\textwidth]{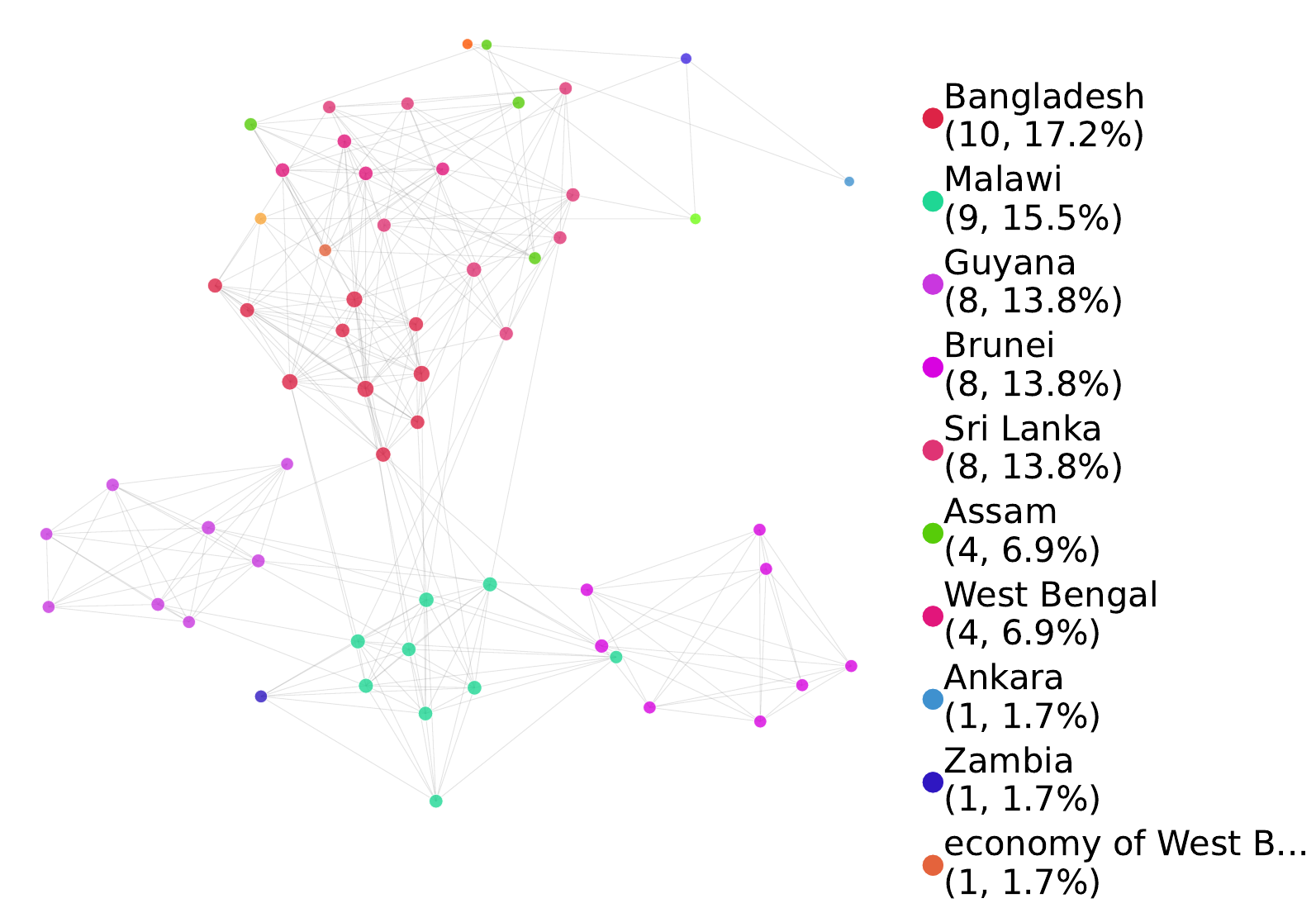}
        \caption{\protect\shortstack{Cluster 6\\58 facts \;|\; 14 subjects \;|\; 34.2\% cross-subject edges}}
        \label{fig:llama_c6}
    \end{subfigure}

    \begin{subfigure}[b]{0.48\textwidth}
        \centering
        \includegraphics[width=\textwidth]{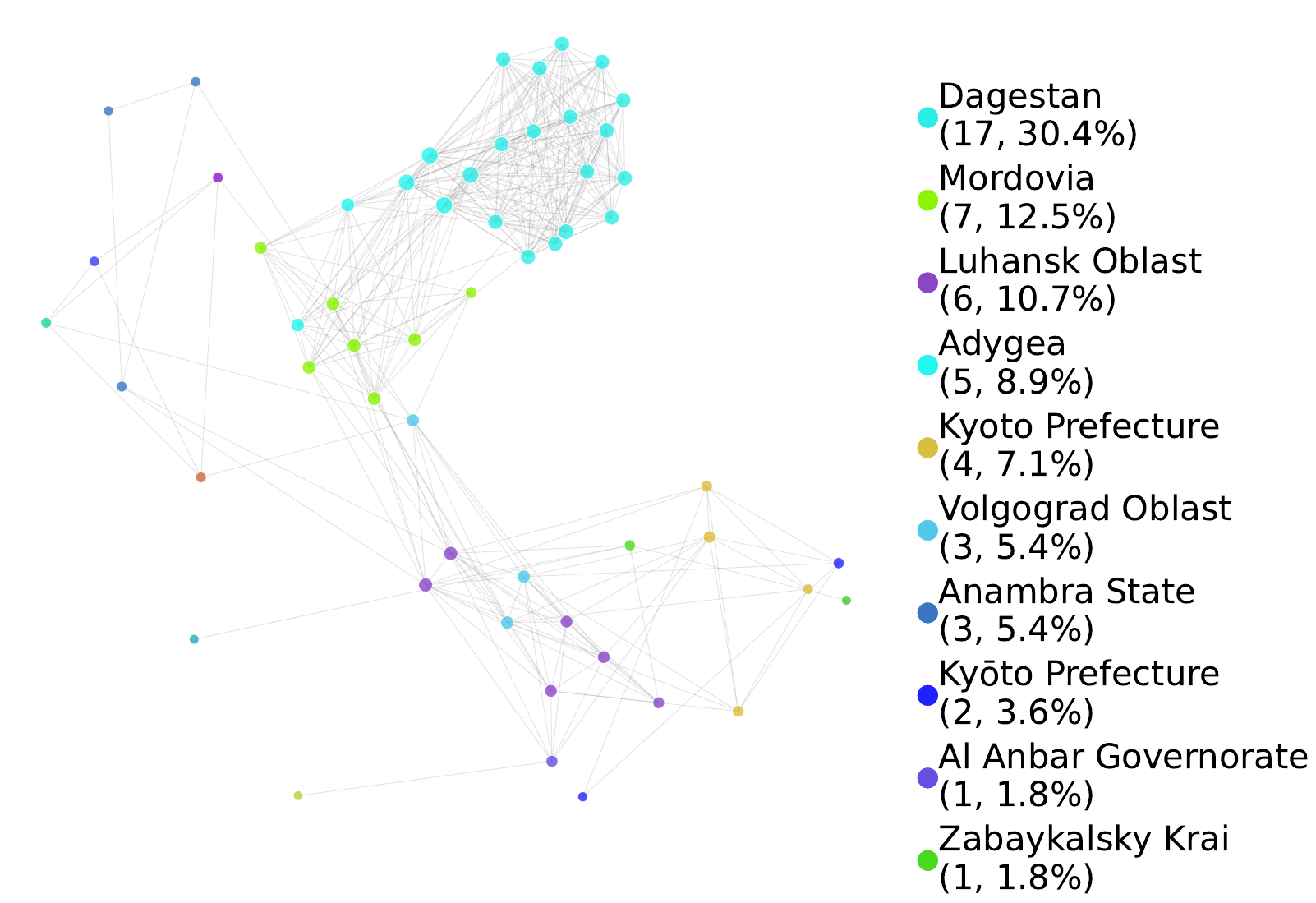}
        \caption{\protect\shortstack{Cluster 7\\56 facts \;|\; 17 subjects \;|\; 44.6\% cross-subject edges}}
        \label{fig:llama_c7}
    \end{subfigure}
    \caption{Llama3-8B: Clusters 05-07}
\end{figure*}

To analyse large-scale entanglement clusters, we construct graphs where nodes represent factual triplets and edges connect pairs with high entanglement scores (cosine similarity > 0.7). As observed in Appendix~\ref{appendix:exp1_diagrams}, they have high tendency of ripple susceptibility. We apply the Louvain community detection algorithm to our corpora, identifying densely connected groups of entanglement clusters of triplets. Figures~\ref{fig:gptj_c1}, \ref{fig:gptj_c2}, \ref{fig:gptj_c3}, \ref{fig:gptj_c4}, \ref{fig:gptj_c5}, \ref{fig:gptj_c6}, \ref{fig:gptj_c7}, \ref{fig:gptj_c8}, \ref{fig:gptj_c9}, \ref{fig:gptj_c10}, \ref{fig:gptj_c11}, \ref{fig:gptj_c12}, \ref{fig:gptj_c13}, \ref{fig:gpt2_c1}, \ref{fig:gpt2_c2}, \ref{fig:gpt2_c3}, \ref{fig:gpt2_c4}, \ref{fig:gpt2_c5}, \ref{fig:gpt2_c6}, \ref{fig:gpt2_c7}, \ref{fig:llama_c1}, \ref{fig:llama_c2}, \ref{fig:llama_c3}, \ref{fig:llama_c4}, \ref{fig:llama_c5}, \ref{fig:llama_c6} and \ref{fig:llama_c7} illustrate these clusters for GPT-J, GPT2-XL and Llama3. All of these contain 10 subjects that have strongest ripple potential because changing that fact influences the largest number of other facts. These clusters reflect internal representational overlap, and thus pinpoint regions where edits to one fact are likely to induce widespread ripple effects. We perform this clustering for Llama3, GPT2-XL, and GPT-J, and visualize all clusters containing more than 50 facts. Nodes are colored by subject, allowing visual identification of dominant subjects and high-risk hubs. This analysis reveals latent knowledge bottlenecks where facts are densely entangled, guiding safer and more targeted editing.

\begin{table*}
\centering
\caption{Comparison between \Clare, ripple-effect measurements, and standard editing evaluation metrics.}
\label{tab:metric_comparison}
\resizebox{0.99\textwidth}{!}{
\begin{tabular}{clc}
\toprule
\textbf{Metric} & \textbf{Evaluation Focus} & \textbf{Temporal Application} \\
\midrule

\Clare
& Measures shared subspaces between facts using cosine similarity at the last critical layer. 
& Pre-edit \\

Efficacy
& Measures whether the model correctly outputs the new target fact after editing. 
& Post-edit \\

Locality
& Measures whether the model's predictions for unrelated facts remain unchanged. 
& Post-edit \\

Generalization
& Measures whether the edit holds across different semantic phrasings of the same fact.
& Post-edit \\

$|\Delta \log P(y)|$ 
& Measures the change in log-probability of the correct answer before and after editing. 
& Post-edit \\

$\ell_2$ logit shift 
& Measures the global geometric displacement in the output logit distribution. 
& Post-edit \\

\bottomrule
\end{tabular}
}
\end{table*}

\begin{table*}
\centering
\caption{Representative examples where fact pairs exhibit high \Clare similarity but minimal observed ripple effects. These cases illustrate that \Clare measures susceptibility rather than guaranteeing interference.}
\label{tab:counterexamples}
\small
\begin{tabular}{l p{6cm} p{3.8cm} c c}
\toprule
\textbf{Model} & \textbf{Edited Fact} & \textbf{Entangled Fact} & \Clare & $|\Delta \log P(y)|$ \\
\midrule

Llama3
& The official language of Germany is (German $\rightarrow$ English) 
& The official language of Poland is: Polish 
& 0.84 & 1.43 \\

Llama3
& Queen's Park F.C. is associated with the sport of (association football $\rightarrow$ basketball) 
& St Patrick's Athletic F.C. is headquartered in: Dublin 
& 0.87 & 1.17 \\

GPT2-XL
& The official language of Italian Republic is (Italian $\rightarrow$ French) 
& The official language of Dutch Republic is: Dutch 
& 0.93 & 2.12 \\

GPT2-XL
& The capital of Denmark is (Copenhagen $\rightarrow$ London) 
& The capital of Sweden is: Stockholm 
& 0.88 & 0.07 \\

GPT-J
& The name of the capital city of France is (Paris $\rightarrow$ London) 
& The name of the capital city of Russia is: Moscow 
& 0.87 & 1.03 \\

GPT-J
& The name of the award Norman Mailer won is (Pulitzer Prize for Fiction $\rightarrow$ Academy Award) 
& Ernest Hemingway is a citizen of: United States of America 
& 0.79 & 1.16 \\

\bottomrule
\end{tabular}
\end{table*}

\section{Comparison of model-editing evaluation metrics}

\review{To further clarify the relationship between \Clare and existing model-editing evaluation frameworks, we summarize the key differences between \Clare and commonly used model-editing evaluation metrics in Table~\ref{tab:metric_comparison}.}

\review{Standard editing metrics are primarily post-edit behavioral assessments. They measure the success or failure of an edit after model weights have already been modified. For example, locality evaluates whether an edit remains confined to its target, but it can only detect interference after the damage has occurred. In contrast, \Clare is a pre-edit diagnostic tool. It estimates the susceptibility to ripple effects by quantifying the representational entanglement in the model's hidden space prior to any parameter updates. By identifying ``high-risk'' facts before editing, \Clare allows for the construction of targeted preservation sets, moving the field toward a preventive rather than reactive approach to model safety.}

\section{\Clare as a probabilistic risk indicator}

\review{While \Clare provides a strong signal for estimating factual entanglement, it is a probabilistic risk indicator rather than a deterministic predictor of ripple effects. In practice, we observe cases where two facts have high \Clare similarity but exhibit only minimal ripple effects after editing.}

\review{Through our experiments, we show that the variance of ripple effects increases with higher \Clare similarity scores. This indicates that highly entangled facts are more susceptible to interference when one fact is edited. However, high entanglement does not guarantee a large ripple effect for every fact pair. Instead, it reflects elevated risk. In contrast, fact pairs with low \Clare similarity consistently exhibit negligible ripple effects with very few outliers.}

\review{Table~\ref{tab:counterexamples} presents representative examples across three models where facts with high \Clare similarity experience minimal ripple effects after editing. In these cases, the observed ripple-effect metric $|\Delta \log P(y)|$ remains very small despite strong representational similarity between the edited fact and the control fact.}

\section{Comparison with Low-Rank Gradient Projection Approaches}
\review{Recent works have explored low-rank gradient projection as a strategy to reduce the memory footprint of gradient-based attribution methods. LOGRA~\cite{choe2024your} exploits the Kronecker product structure of per-layer gradients to project them into a compact subspace without materializing the full gradient, achieving substantial reductions in both memory and compute. TrackStar~\cite{chang2024scalable} similarly applies two-sided random projection across layer blocks to make gradient indexing tractable at pretraining scale.
While these methods reduce the cost of storing gradients, they still require a full backward pass for each example, with associated GPU memory overhead. \Clare, by contrast, extracts a single hidden-state vector from one forward pass up to the last critical layer, requiring no gradient computation. This yields substantially smaller per-example representations and enables the corpus-wide entanglement analysis described in Section~\ref{rq3}.}

\section{Computational Cost of Critical Layers Identification}
\review{
Several model editing techniques, including ROME~\cite{meng2022locating}, rely on identifying a critical transformer layer through causal tracing. While ROME demonstrates that factual associations localize to middle-layer MLP modules, the computational cost of this identification step was not explicitly analyzed. We attempt to address this gap here. This cost is incurred once per model architecture rather than per edit. Techniques without layer selection (e.g., gradient-based approaches) avoid this overhead.}

\review{\textbf{Causal Tracing Procedure.} Causal tracing estimates hidden state influence via three runs:
\begin{enumerate}
    \item \textbf{Clean run}: Normal forward pass,
    \item \textbf{Corrupted run}: Noise injected into subject token embeddings,
    \item \textbf{Restoration runs}: Corrupted execution with specific hidden states restored at layer $\hat{\ell}$, token $\hat{i}$.
\end{enumerate}
Average Indirect Effect (AIE)~\cite{meng2022locating} identifies layers with largest causal impact.}

\review{\textbf{Complexity Analysis.} For a transformer with $L$ layers, $T$ tokens, $H$ hidden dimension, $D$ MLP dimension (typically $D = 4H$), and $N$ factual prompts, a single forward pass costs:
\begin{equation}
    O\big(L(T^2 H + THD)\big)
\end{equation}}

\review{Causal tracing evaluates $T \times L$ restoration passes per prompt, yielding the dominant cost:
\begin{equation}
    O\big(N \cdot T^2 \cdot L^2 \cdot (TH + HD)\big)
\end{equation}}
\review{When $D = 4H$ and $H \gg T$, this simplifies to $O(N \cdot T^2 \cdot L^2 \cdot H^2)$. Table~\ref{tab:causal_tracing_complexity} details the phase-wise breakdown.}

\begin{table}
\centering
\caption{Phase-wise complexity analysis of causal tracing for critical layer identification.}
\label{tab:causal_tracing_complexity}
\small
\begin{tabular}{lc}
\toprule
\textbf{Phase} & \textbf{Complexity} \\
\midrule
Clean run & $O(N \cdot T \cdot L \cdot (T \cdot H + H \cdot D))$ \\
Corrupted run & $O(N \cdot T \cdot L \cdot (T \cdot H + H \cdot D))$ \\
Restoration passes & $O(N \cdot T^2 \cdot L^2 \cdot (T \cdot H + H \cdot D))$ \\
\midrule
\textbf{Total} & $\mathbf{O(N \cdot T^2 \cdot L^2 \cdot (T \cdot H + H \cdot D))}$ \\
\bottomrule
\end{tabular}
\end{table}

\begin{figure*}
\centering
\begin{Verbatim}[breaklines=true,breakanywhere=true,fontsize=\small]
1. The name of the head of government of Spain is -> Pedro Sánchez → affects 296 other facts
2. The name of the head of government of France is -> Élisabeth Borne → affects 289 other facts
3. The name of the head of state of France is -> Emmanuel Macron → affects 273 other facts
4. The name of the current head of state in France is -> Emmanuel Macron → affects 271 other facts
5. The name of the head of state of Spain is -> Felipe VI of Spain → affects 266 other facts
6. The name of the capital city of France is -> Paris → affects 258 other facts
7. The name of the anthem of France is -> La Marseillaise → affects 252 other facts
8. The name of the capital city of Spain is -> Madrid → affects 251 other facts
9. The official language of Italy is -> Italian → affects 250 other facts
10. The name of the currency in Poland is -> Złoty → affects 250 other facts
11. The official language of Spain is -> Spanish → affects 247 other facts
12. The official language of France is -> French → affects 243 other facts
13. The name of the head of government of Poland is -> Mateusz Morawiecki → affects 242 other facts
14. The official language of Poland is -> Polish → affects 242 other facts
15. The official language of Germany is -> German → affects 240 other facts
16. The name of the capital city of Russia is -> Moscow → affects 239 other facts
17. The name of the currency in Sri Lanka is -> Sri Lankan rupee → affects 238 other facts
18. The name of the head of state of Poland is -> Andrzej Duda → affects 236 other facts
19. The name of the currency in Spain is -> euro → affects 233 other facts
20. The official language of Sri Lanka is -> Sinhala → affects 230 other facts
21. The name of the head of government of India is -> Narendra Modi → affects 230 other facts
22. The name of the head of state of Sri Lanka is -> Ranil Wickremesinghe → affects 229 other facts
23. The name of the capital city of occupation of Japan is -> Tokyo → affects 224 other facts
24. The name of the capital city of Poland is -> Warsaw → affects 224 other facts
25. The name of the head of government of Sri Lanka is -> Ranil Wickremesinghe → affects 222 other facts
26. The official language of Australia is -> English → affects 220 other facts
27. The official language of Slovakia is -> Slovak → affects 216 other facts
28. The official language of Romania is -> Romanian → affects 215 other facts
29. The name of the current head of state in Portugal is -> Marcelo Rebelo de Sousa → affects 215 other facts
30. The name of the head of state of Slovakia is -> Zuzana Čaputová → affects 214 other facts
31. The name of the anthem of Sri Lanka is -> Sri Lanka Matha → affects 213 other facts
32. The capital of Lithuania is -> Vilnius → affects 210 other facts
33. The capital of Indonesia is -> Jakarta → affects 208 other facts
34. The name of the currency in India is -> Indian rupee → affects 207 other facts
35. The name of the currency in Slovakia is -> euro → affects 206 other facts
36. The official language of Argentina is -> Spanish → affects 206 other facts
37. The name of the head of government of Slovakia is -> Eduard Heger → affects 205 other facts
38. The name of the head of state of India is -> Droupadi Murmu → affects 205 other facts
39. The capital of Romania is -> Bucharest → affects 205 other facts
40. The name of the head of state of Slovenia is -> Nataša Pirc Musar → affects 204 other facts
41. The official language of India is -> Hindi → affects 204 other facts
42. The name of the currency in Ukraine is -> Hryvnia → affects 203 other facts
43. The name of the head of government of Slovenia is -> Robert Golob → affects 202 other facts
44. Louis XVII of France died in the city of -> Paris → affects 202 other facts
45. The name of the anthem of Slovakia is -> Nad Tatrou sa blýska → affects 201 other facts
46. The official language of Peru is -> Spanish → affects 201 other facts
47. The official language of Slovenia is -> Slovene → affects 201 other facts
48. Louis XV of France is affiliated with the religion of -> Catholic Church → affects 199 other facts
49. The official language of Japan is -> Japanese → affects 198 other facts
50. The name of the capital city of Slovakia is -> Bratislava → affects 196 other facts
\end{Verbatim}
\caption{Most entangled facts ranked by representational connectivity in GPT2-XL.}
\label{fig:gpt2xl_hub}
\end{figure*}

\begin{figure*}
\centering
\begin{Verbatim}[breaklines=true,breakanywhere=true,fontsize=\small]
1. The name of the head of government of Poland is -> Mateusz Morawiecki → affects 89 other facts
2. The official language of Poland is -> Polish → affects 85 other facts
3. The official language of Romania is -> Romanian → affects 78 other facts
4. The name of the head of state of Poland is -> Andrzej Duda → affects 77 other facts
5. The official language of Germany is -> German → affects 73 other facts
6. The name of the capital city of Poland is -> Warsaw → affects 71 other facts
7. The official language of Italy is -> Italian → affects 68 other facts
8. The official language of Ukraine is -> Ukrainian → affects 67 other facts
9. The name of the head of government of Spain is -> Pedro Sánchez → affects 66 other facts
10. The name of the head of government of Turkey is -> Recep Tayyip Erdoğan → affects 66 other facts
11. The official language of Spain is -> Spanish → affects 63 other facts
12. The name of the capital city of Ukraine is -> Kyiv → affects 62 other facts
13. The name of the anthem of Turkey is -> İstiklâl Marşı → affects 58 other facts
14. The name of the currency in Ukraine is -> Hryvnia → affects 58 other facts
15. The official language of Japan is -> Japanese → affects 57 other facts
16. The name of the capital city of Turkey is -> Ankara → affects 57 other facts
17. The name of the capital city of Russia is -> Moscow → affects 56 other facts
18. The name of the capital city of Spain is -> Madrid → affects 56 other facts
19. The name of the capital city of South Korea is -> Seoul → affects 55 other facts
20. The official language of Turkey is -> Turkish → affects 55 other facts
21. The name of the head of government of France is -> Élisabeth Borne → affects 55 other facts
22. The name of the head of state of Spain is -> Felipe VI of Spain → affects 54 other facts
23. The name of the head of state of France is -> Emmanuel Macron → affects 53 other facts
24. The name of the head of state of Turkey is -> Recep Tayyip Erdoğan → affects 52 other facts
25. The official language of France is -> French → affects 52 other facts
26. The name of the head of state of Russia is -> Vladimir Putin → affects 49 other facts
27. The name of the head of government of Russia is -> Mikhail Mishustin → affects 49 other facts
28. The name of the spouse of George I of Great Britain is -> Sophia Dorothea of Celle → affects 48 other facts
29. The name of the head of government of Slovenia is -> Robert Golob → affects 48 other facts
30. The name of the head of state of Slovakia is -> Zuzana Čaputová → affects 48 other facts
31. The name of the head of government of South Korea is -> Yoon Suk Yeol → affects 47 other facts
32. The name of the field of work of George I of Great Britain is -> politics → affects 47 other facts
33. The capital of Indonesia is -> Jakarta → affects 47 other facts
34. The name of the child of George I of Great Britain is -> George II of Great Britain → affects 46 other facts
35. The name of the anthem of Russia is -> State Anthem of the Russian Federation → affects 46 other facts
36. The name of the head of government of Ukraine is -> Denys Shmyhal → affects 45 other facts
37. The name of the head of state of India is -> Droupadi Murmu → affects 45 other facts
38. The name of the religion which George I of Great Britain is associated with is -> Lutheranism → affects 45 other facts
39. The official language of Peru is -> Spanish → affects 45 other facts
40. The name of the head of state of Ukraine is -> Volodymyr Zelenskyy → affects 45 other facts
41. The capital of Serbia is -> Belgrade → affects 44 other facts
42. The name of the currency in Spain is -> euro → affects 43 other facts
43. The name of the current head of state in France is -> Emmanuel Macron → affects 43 other facts
44. The name of the position held by George I of Great Britain is -> Prince-Elector → affects 43 other facts
45. The name of the capital city of Slovenia is -> Ljubljana → affects 43 other facts
46. The capital of Romania is -> Bucharest → affects 43 other facts
47. The name of the head of state of Slovenia is -> Nataša Pirc Musar → affects 43 other facts
48. The name of the capital city of France is -> Paris → affects 43 other facts
49. The place of death of George I of Great Britain is -> Osnabrück → affects 42 other facts
50. The official language of Slovakia is -> Slovak → affects 42 other facts
\end{Verbatim}
\caption{Most entangled facts ranked by representational connectivity in Llama3.}
\label{fig:llama3_hub}
\end{figure*}

\begin{figure*}
\centering
\begin{Verbatim}[breaklines=true,breakanywhere=true,fontsize=\small]
1. The name of the country of citizenship of Kate Winslet is -> United Kingdom → affects 1257 other facts
2. The place of burial of Audrey Hepburn is -> Cemetery of Tolochenaz → affects 1233 other facts
3. The place of birth of Audrey Hepburn is -> Rue Keyenveld - Keienveldstraat → affects 1222 other facts
4. The place of death of Audrey Hepburn is -> Tolochenaz → affects 1218 other facts
5. The name of the country of citizenship of Mark Wahlberg is -> United States of America → affects 1204 other facts
6. The place of birth of Pierce Brosnan is -> Drogheda → affects 1184 other facts
7. The name of the country of citizenship of Leonardo DiCaprio is -> United States of America → affects 1178 other facts
8. The place of birth of Kate Winslet is -> Reading → affects 1178 other facts
9. The name of the employer of Audrey Hepburn is -> UNICEF → affects 1167 other facts
10. The name of the country of citizenship of Avril Lavigne is -> Canada → affects 1165 other facts
11. The name of the spouse of Pierce Brosnan is -> Keely Shaye Smith → affects 1162 other facts
12. The name of the field of work of Audrey Hepburn is -> acting → affects 1159 other facts
13. The name of the country of citizenship of Pierce Brosnan is -> Republic of Ireland → affects 1148 other facts
14. The place of birth of Avril Lavigne is -> Belleville → affects 1147 other facts
15. The name of the spouse of Mark Wahlberg is -> Rhea Durham → affects 1146 other facts
16. The name of the alma mater of Mark Wahlberg is -> Snowden International School → affects 1143 other facts
17. The names of the siblings of Kate Winslet are -> Beth Winslet → affects 1141 other facts
18. The name of the award Kate Winslet won is -> Commander of the Order of the British Empire → affects 1141 other facts
19. The name of the alma mater of Avril Lavigne is -> Napanee District Secondary School → affects 1140 other facts
20. The name of the position held by Pierce Brosnan is -> UNICEF Goodwill Ambassador → affects 1137 other facts
21. The name of the spouse of Kate Winslet is -> Ned Rocknroll → affects 1134 other facts
22. The place of birth of Katy Perry is -> Santa Barbara → affects 1129 other facts
23. The name of the country of citizenship of Priyanka Chopra is -> India → affects 1126 other facts
24. The place of birth of Priyanka Chopra is -> Jamshedpur → affects 1123 other facts
25. The name of the position held by Audrey Hepburn is -> UNICEF Goodwill Ambassador → affects 1121 other facts
26. The name of the spouse of Avril Lavigne is -> Deryck Whibley → affects 1113 other facts
27. The name of the alma mater of Priyanka Chopra is -> La Martiniere Lucknow → affects 1108 other facts
28. The name of the country of citizenship of Sandra Bullock is -> United States of America → affects 1108 other facts
29. The names of the siblings of Mark Wahlberg are -> Robert Wahlberg → affects 1106 other facts
30. The place of birth of Jennifer Aniston is -> Sherman Oaks → affects 1103 other facts
31. The name of the child of Kate Winslet is -> Joe Mendes → affects 1101 other facts
32. The names of the siblings of Audrey Hepburn are -> Arnoud Quarles van Ufford → affects 1100 other facts
33. The name of the country of citizenship of Jennifer Connelly is -> United States of America → affects 1097 other facts
34. The name of the country of citizenship of Matthew McConaughey is -> United States of America → affects 1097 other facts
35. The name of the alma mater of Kate Winslet is -> Redroofs Theatre School → affects 1096 other facts
36. The name of the country of citizenship of Ben Affleck is -> United States of America → affects 1094 other facts
37. The place of birth of Matthew McConaughey is -> Uvalde → affects 1082 other facts
38. The name of the country of citizenship of Jennifer Aniston is -> United States of America → affects 1082 other facts
39. The name of the mother of Mark Wahlberg is -> Alma Elaine Donnelly → affects 1078 other facts
40. The name of the country of citizenship of Katy Perry is -> United States of America → affects 1078 other facts
...
\end{Verbatim}
\caption{Most entangled facts ranked by representational connectivity in GPT-J.}
\label{fig:gptj_hub_1}
\end{figure*}

\begin{figure*}
\centering
\begin{Verbatim}[breaklines=true,breakanywhere=true,fontsize=\small]
41. The name of the country of citizenship of Adam Sandler is -> United States of America → affects 1076 other facts
42. The name of the alma mater of Pierce Brosnan is -> Saint Martin's School of Art → affects 1075 other facts
43. The name of the position held by Priyanka Chopra is -> UNICEF Goodwill Ambassador → affects 1074 other facts
44. The name of the employer of Pierce Brosnan is -> UNICEF → affects 1069 other facts
45. The name of the ethnic group which Audrey Hepburn is associated with is -> British people → affects 1065 other facts
46. The name of the child of Mark Wahlberg is -> Brendan Wahlberg → affects 1062 other facts
47. The place of birth of Adam Sandler is -> Brooklyn → affects 1061 other facts
48. The name of the country of citizenship of Drew Barrymore is -> United States of America → affects 1061 other facts
49. The name of the award Sandra Bullock won is -> Academy Award for Best Actress → affects 1060 other facts
50. The name of the mother of Avril Lavigne is -> Judith Rosanne Loshaw → affects 1057 other facts
\end{Verbatim}
\caption{Most entangled facts ranked by representational connectivity in GPT-J (contd.).}
\label{fig:gptj_hub_2}
\end{figure*}




\end{document}